\documentclass[lettersize,journal]{IEEEtran}
\usepackage{amsmath,amsfonts}
\usepackage{algorithmicx}
\usepackage[ruled]{algorithm2e}
\usepackage{array}
\usepackage[caption=false,font=normalsize,labelfont=sf,textfont=sf]{subfig}
\usepackage{textcomp}
\usepackage{stfloats}
\usepackage{url}
\usepackage{verbatim}
\usepackage{graphicx}
\usepackage{cite}
\usepackage{cases}
\usepackage{bm}
\usepackage{amssymb}
\usepackage{diagbox}
\usepackage{multirow}
\usepackage{epstopdf}
\usepackage{threeparttable}
\usepackage{epsfig}
 \usepackage{amsthm}
 \usepackage{color}

\newtheorem{thm}{\bf Theorem}

\newtheorem{rmk}{\bf Remark}

\begin{document}
\title{MetaUE: Model-based Meta-learning for Underwater Image Enhancement}

\author{Zhenwei Zhang,
        Haorui Yan,
        Ke Tang,
        and Yuping Duan$^*$
\thanks{The work was partially supported by the National Natural Science Foundation of China (NSFC 12071345, 11701418). \emph{Asterisk indicates the corresponding author.}}
\thanks{Z. Zhang and H. Yan are with Center for Applied Mathematics, Tianjin University, Tianjin 300072, China.}
\thanks{K. Tang is with the Guangdong Key Laboratory of Brain-Inspired Intelligent Computation, Department of Computer Science and Engineering, Southern University of
Science and Technology, Shenzhen 518055, China, and also with the Research Institute of Trustworthy Autonomous Systems, Southern University of Science and Technology, Shenzhen 518055, China. E-mail: tangk3@sustech.edu.cn.}
\thanks{Y. Duan is with Center for Applied Mathematics, Tianjin University, Tianjin 300072, China. E-mail: doveduan@gmail.com.}
}
\markboth{Journal of \LaTeX\ Class Files,~Vol.~14, No.~8, August~2021}%
{Shell \MakeLowercase{\textit{et al.}}: A Sample Article Using IEEEtran.cls for IEEE Journals}


\maketitle

\begin{abstract}
The challenges in recovering underwater images are the presence of diverse degradation factors and the lack of ground truth images.
Although synthetic underwater image pairs can be used to overcome the problem of inadequately observing data,
it may result in over-fitting and enhancement degradation.
This paper proposes a model-based deep learning method for restoring clean images under various underwater scenarios, which exhibits good interpretability and generalization ability.
More specifically, we build up a multi-variable convolutional neural network model to estimate the clean image, background light and transmission map, respectively. An efficient loss function is also designed to closely integrate the variables based on the underwater image model.
The meta-learning strategy is used to obtain a pre-trained model on the synthetic underwater dataset, which contains different types of degradation to cover the various underwater environments.
The pre-trained model is then fine-tuned on real underwater datasets to obtain a reliable underwater image enhancement model, called MetaUE.
Numerical experiments demonstrate that the pre-trained model has good generalization ability,
allowing it to remove the color degradation for various underwater attenuation images such as blue, green and yellow, etc.
The fine-tuning makes the model able to adapt to different underwater datasets, the enhancement results of which outperform the state-of-the-art underwater image restoration methods. All our codes and data are available at \url{https://github.com/Duanlab123/MetaUE}.
\end{abstract}


\section{Introduction}

\begin{figure}[htbp]
\centering
\subfloat[Blue image]{ \includegraphics[width=0.23\textwidth]{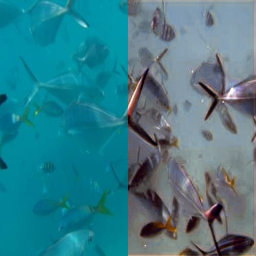}}
\subfloat[Haze image]{\includegraphics[width=0.23\textwidth]{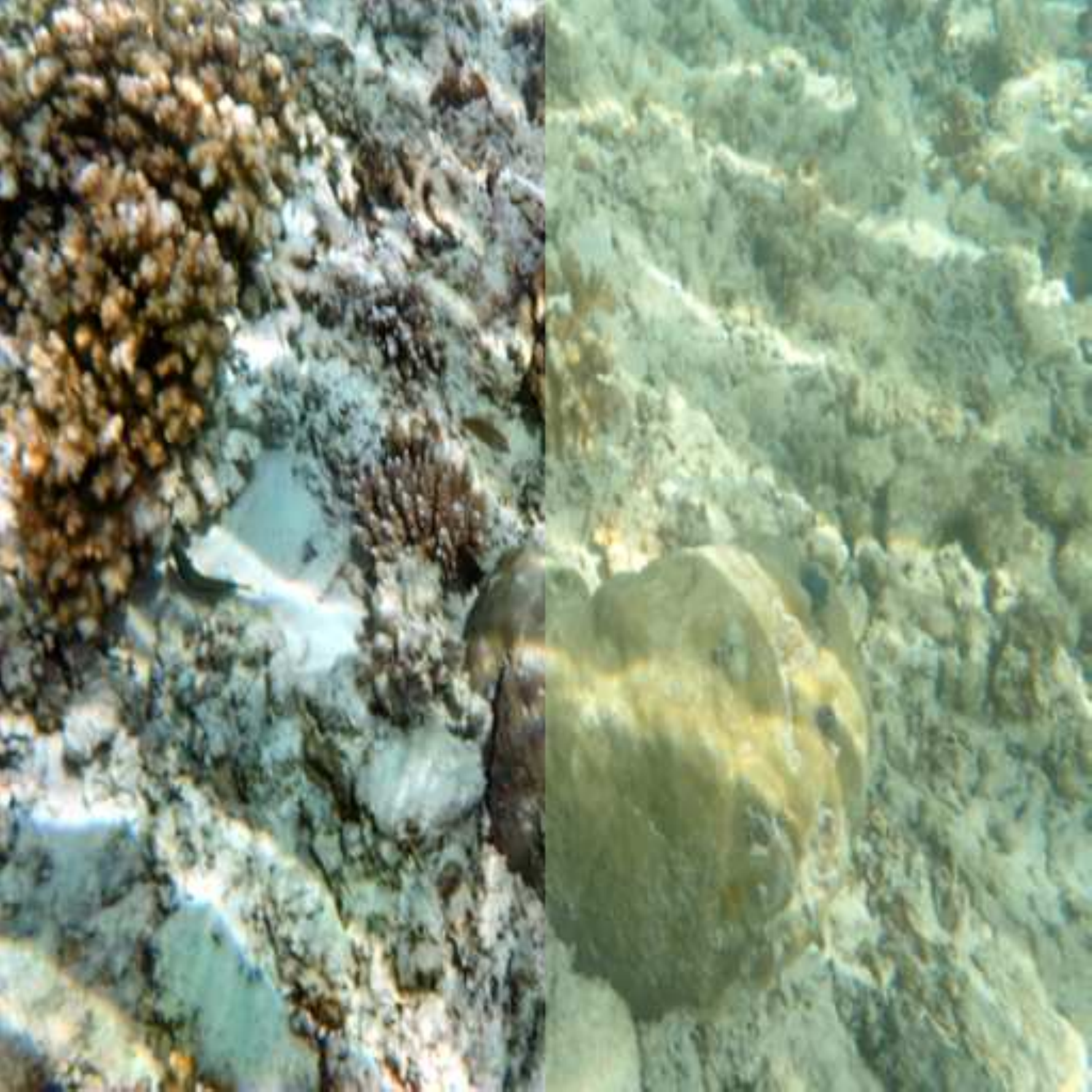}}\\
\subfloat[Green image]{ \includegraphics[width=0.23\textwidth]{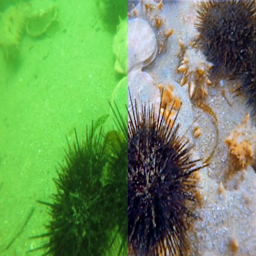}}
\subfloat[Low luminance image]{\includegraphics[width=0.23\textwidth]{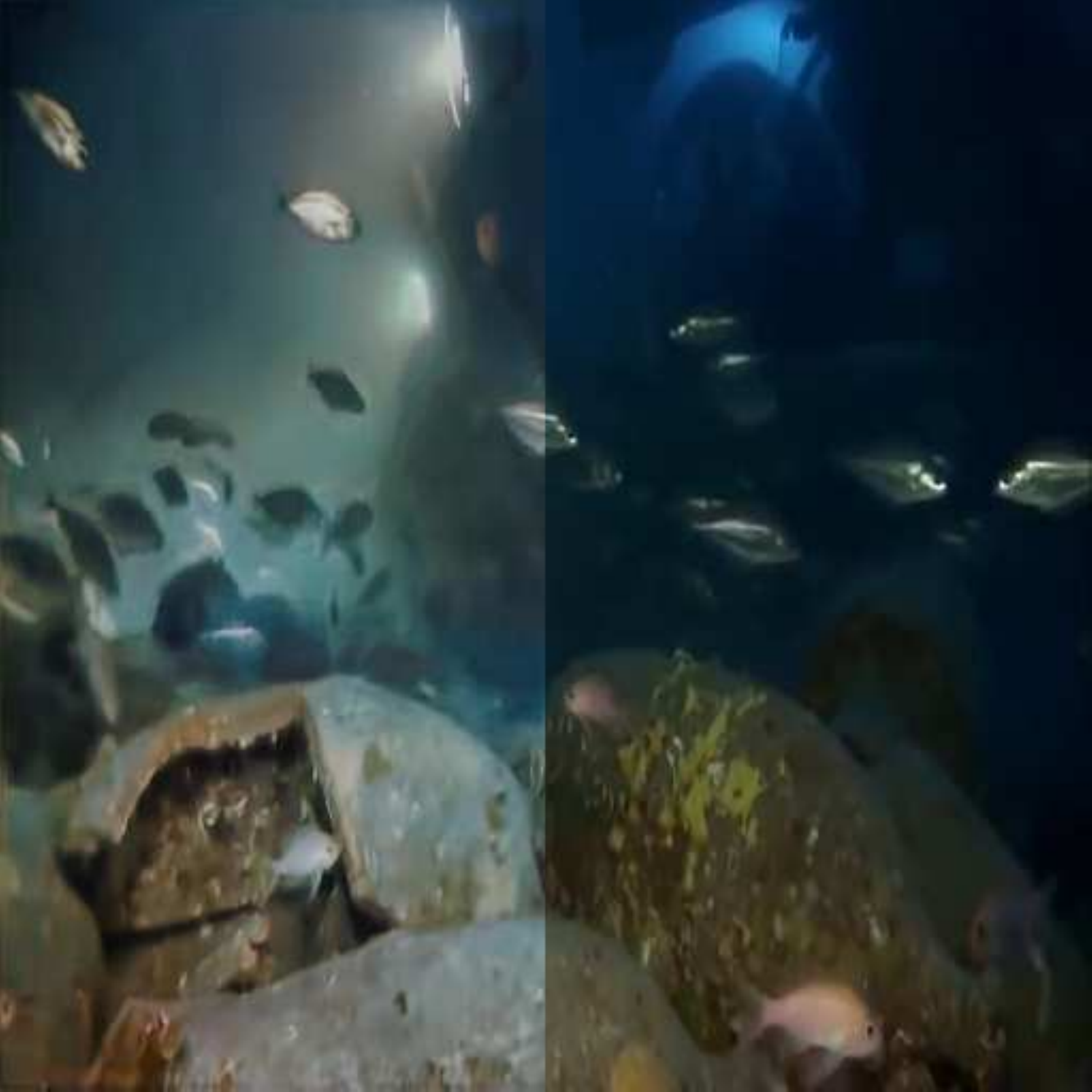}}\\			
\caption{Illustration of typical underwater images and our enhancement results.}
\label{4 underwater}
\end{figure}

\IEEEPARstart{U}{nderwater}  imaging has played an important role in deep-sea exploration, underwater robot technology and marine ecological monitoring.
However, underwater optical images are vulnerable to underwater turbulence, diffusion, severe absorption, scattering of
water bodies, various noises, low contrast, uniform illumination, monotonous color, and complex underwater scenes \cite{jian2021underwater}. Four typical underwater images are displayed in Fig. \ref{4 underwater}, which makes the high-level image processing tasks based on underwater images become very challenging.

Underwater image degradation is a complex process,  mainly influenced by light absorption and scattering.
Various methods have been developed to recover clean images from underwater images including traditional methods and learning-based methods.
Akkaynak \emph{et al.} \cite{akkaynak2017space,akkaynak2019sea} proposed an accurate underwater image formation model and used mathematical methods to estimate the coefficients, which can consistently remove water from underwater images.
By a rank-one transmission prior, Liu \emph{et al.} \cite{liu2021rank,liu2022rank}  presented a real-time light correction method to recover the degraded scenes of underwater, sandstorms and haze, etc.
The regularization methods were also used to recover underwater images,
such as depth-aware total variation regularization \cite{ding2021depth}, curvature variation regularization \cite{hou2020underwater}, and adaptive non-convex non-smooth variation regularization \cite{jiao2021underwater}, etc.
In recent years, the deep learning methods have provided new solutions for underwater image enhancement, which can be roughly divided into Generative Adversarial Network (GAN)-based methods and Convolutional Neural Network (CNN)-based methods.
Since the discriminator of GAN can eliminate unreliable data, Li \emph{et al.} \cite{li2018watergan}  generated realistic underwater images from the in-air images for color correction of underwater images.
Later, GAN-based methods have been intensively studied for underwater image enhancement \cite{li2018emerging,ye2019deep,yang2020underwater,liu2022twin,fabbi2018enhancing,islam2020fast}.
On the other hand, Li \emph{et al.} \cite{li2020underwater} presented a CNN model using the scene prior for underwater image enhancement on the synthesized underwater datasets.
Since then, various CNN-based models have been developed such as the Water-Net \cite{UIEB}, JLCL-Net \cite{xue2021joint}, Ucolor \cite{li2021underwater}, etc. However, the existing deep learning-based underwater image enhancement methods may lose their effects on images of unknown water, which limits their usages in real applications.
Indeed, model-driven deep learning methods are well-known for their high interpretability and abilities in dealing with different image processing tasks.
The plug-and-play strategy as a typical model-driven method has been used for image super resolution \cite{zhang2019deep}, image denoising \cite{zhang2021plug}, image deblurring \cite{fang2022robust}, etc.
Most recently, Jonathan, Jain and Abbeel \cite{ho2020denoising} presented the diffusion probabilistic models inspired by considerations from nonequilibrium thermodynamics, which have emerged as a powerful new family of deep generative models with a record-breaking performance in many applications, including image synthesis \cite{dhariwal2021diffusion,rombach2022high}, denoising \cite{wyatt2022anoddpm,chung2022mr}, and segmentation \cite{hoogeboom2021argmax}, etc.

Deep meta-learning \cite{santoro2016meta,wang2019meta,hospedales2021meta} is a knowledge-driven machine learning framework that attempts to solve the problem of how to learn. Meta-learning has achieved remarkable results on no or fewer reference image processing problems.
For instance, Zhu \emph{et al.} \cite{zhu2020metaiqa} presented a no-reference image quality assessment  metric based on deep meta-learning to learn the meta-knowledge shared by human when evaluating the quality of images with various distortions. Zhao \emph{et al.} \cite{zhao2022federated} proposed a novel federated meta-learning enhanced acoustic radio cooperative framework to do the transfer.
Wang \emph{et al.} \cite{wang2022remember} proposed a meta memory bank to improve the generalization of segmentation networks by bridging the domain
gap between source and target domains.
The model-agnostic nature of meta-learning can facilitate underwater image enhancement tasks, to produce good generalization performance in diverse underwater environments.

In this paper, we propose a task-tailored deep learning method based on the physical model of underwater images to recover clean underwater images, which can accurately describe the underwater image distortions. Our network consists of three sub-networks to estimate the clean image, background light, and transmission map, respectively, which are combined according to the underwater image model to obtain the supervision loss.
Meta-learning is used to obtain a pre-trained model based on synthetic images, which is then fine-tuned on real underwater images.
The pre-trained model is made to capture the shared prior knowledge of the degraded underwater images, while the fine-tuned model can make the pre-trained model easy to adapt to unknown distortions.
To sum up, our major contributions are concluded as follows
\begin{enumerate}
\item We propose a model-driven multi-variable convolution neural network based on the underwater image degradation model, which improves the interpretability by estimating different variables through sub-networks.
\item We implement the meta-learning strategy to learn the shared prior knowledge among different types of distortions on a sophisticated underwater image dataset generated from in-air images to cover a wide range of degradation images, which greatly promotes the generalization ability on diversified distortions.
\item We evaluate our meta-learning underwater image enhancement model (shorted by MetaUE) on different underwater image datasets, which outperforms the state-of-the-art methods in terms of enhancement qualities and generalization ability, particularly for images with strong light scattering and insufficient lighting.
\end{enumerate}
The rest of the paper is organized as follows. Section \ref{sec2} reviews the most important supervised and unsupervised learning-based methods for underwater image enhancement tasks.
In Section \ref{sec3},  we propose the underwater image enhancement method based on the underwater image physical model.
The meta-learning strategy is presented in Section \ref{sec4}, where a synthetic underwater image dataset is developed for training the pre-trained model.
Numerical experiments are provided in Section \ref{sec5} by comparing with the state-of-the-art underwater enhancement methods.
We conclude the work in Section \ref{sec6} and discuss its possible applications for other low-quality image enhancement problems.

\section{Related works}\label{sec2}

\subsection{Supervised learning-based methods}

With the development of deep learning methods, it is possible to restore clean images from complex underwater environments.
Supervised learning-based methods can effectively extract feature expressions from degraded underwater images to high-quality
images.
As a pioneering work, Li \emph{et al.}\cite{li2018emerging} proposed a weakly supervised color transfer (WSCT) method by learning a cross-domain mapping function between ground-truth images and underwater images.
Hong \emph{et al.}\cite{hong2021wsuie} proposed a weakly supervised underwater
enhancement model, where a GAN-based architecture is designed to enhance underwater images by unpaired image-to-image transformation.
Similarly, weakly supervised methods are proposed to deal with underwater tasks, such as underwater fish segmentation \cite{laradji2021weakly,saleh2022transformer}, underwater object detection \cite{cai2022underwater}.

However, these weakly supervised methods still have much room for improvement to obtain satisfactory underwater images.
Using the paired underwater/clean images to train the supervised methods is a good strategy to improve the performance of underwater problems.
Li  \emph{et al.} \cite{li2020underwater} presented an underwater image enhancement convolutional neural network (UWCNN) based on the synthesized underwater dataset,
which was built up based on the physical underwater image model and Jelov water type \cite{wang2022remember}.
The training dataset in \cite{li2020underwater} was also used in \cite{uplavikar2019all}, where the all-in-one network adversatively learned the domain-agnostic features to generate enhanced underwater images from the degraded images of ten different water types.
Liu \emph{et al.} \cite{info13040187} derived a new physical underwater image model to generate a wide range of degradation images from the clean underwater images, where the proposed simulation images were selected by color deviations and the Sobel edge map to generate reliable dataset.
Wang \emph{et al.} \cite{wang2023domain} noticed that only training on simulation data makes many learning methods fail in real underwater images. Thus, they proposed a novel two-phase underwater domain adaptation network to simultaneously minimize the synthetic data and real underwater data gap.

Since there exist significant differences between synthetic and real underwater datasets, which may cause poor generalization in real underwater environments.
Li \emph{et al.} \cite{UIEB} constructed an underwater enhancement benchmark (UIEB) for real underwater images, which obtained real underwater images by using the existing enhancement methods.
Based on the UIEB \cite{UIEB}, the following effective deep-learning networks are proposed to recover the underwater images.
Xue \emph{et al.} \cite{xue2021joint} proposed the joint luminance and chrominance learning network (JLCL-Net).
Wang \emph{et al.} \cite{wang2021uiec} developed the UIEC$^2$-Net using two color spaces to improve luminance and saturation, respectively.
Li \emph{et al.} \cite{li2021underwater} proposed a network containing the multi-color spaces guided by the medium transmission to cope with color casts and low contrast problems.
Liu \emph{et al.} \cite{liu2022adaptive} developed an adaptive learning attention network for underwater image enhancement based on supervised learning. 

\subsection{Unsupervised learning-based methods}
On the other hand, unsupervised methods were also intensively studied for underwater image enhancement.
Fabbri \emph{et al.} \cite{fabbi2018enhancing} used the cycle-consistent adversarial network (CycleGAN) to directly learn to generate degraded underwater images based on two separate groups of clean and degraded real-world images, which are then used to train a UNet for underwater image enhancement.
Islam \emph{et al.}\cite{islam2020fast} utilized the same CycleGAN-based method to build up a large dataset called EUVP and proposed a
lightweight conditional GAN to recover the underwater images (FUnIE-GAN).
Yang \emph{et al.} \cite{yang2020underwater}  proposed a conditional generative adversarial network to improve the quality of the underwater images.
Zhou \emph{et al.} \cite{zhou2022deep} proposed an unsupervised underwater loop enhancement network (ULENet) to improve the turbid underwater images.
Based on the underwater image formation model, Chai \emph{et al.}\cite{chai2022unsupervised} proposed an unsupervised method to estimate the components of the physical image model, such as scene, backscatter, etc.
Guo \emph{et al.} \cite{guo2022unsupervised} used the transformer module to capture the global information to train the unsupervised network.
Liu \emph{et al.} \cite{liu2022twin} proposed an unsupervised method called twin adversarial contrastive learning-based underwater enhancement method (TACL) to enhance the quality of the underwater images.
Jiang \emph{et al.} \cite{jiang2022two} proposed a novel domain adaptation framework for real-world underwater image enhancement inspired by transfer learning, which transfers in-air image dehazing to real-world underwater image enhancement.
Alejandro \emph{et al.} \cite{espinosa2023efficient} presented a novel state-of-the-art end-to-end deep learning architecture for underwater image enhancement focused on solving key image degradations related to blur, haze, and color casts and inference efficiency.

\section{Our approach}\label{sec3}
\subsection{Underwater image formation}
\begin{figure}[htbp]
\centering
\includegraphics[width=0.45\textwidth]{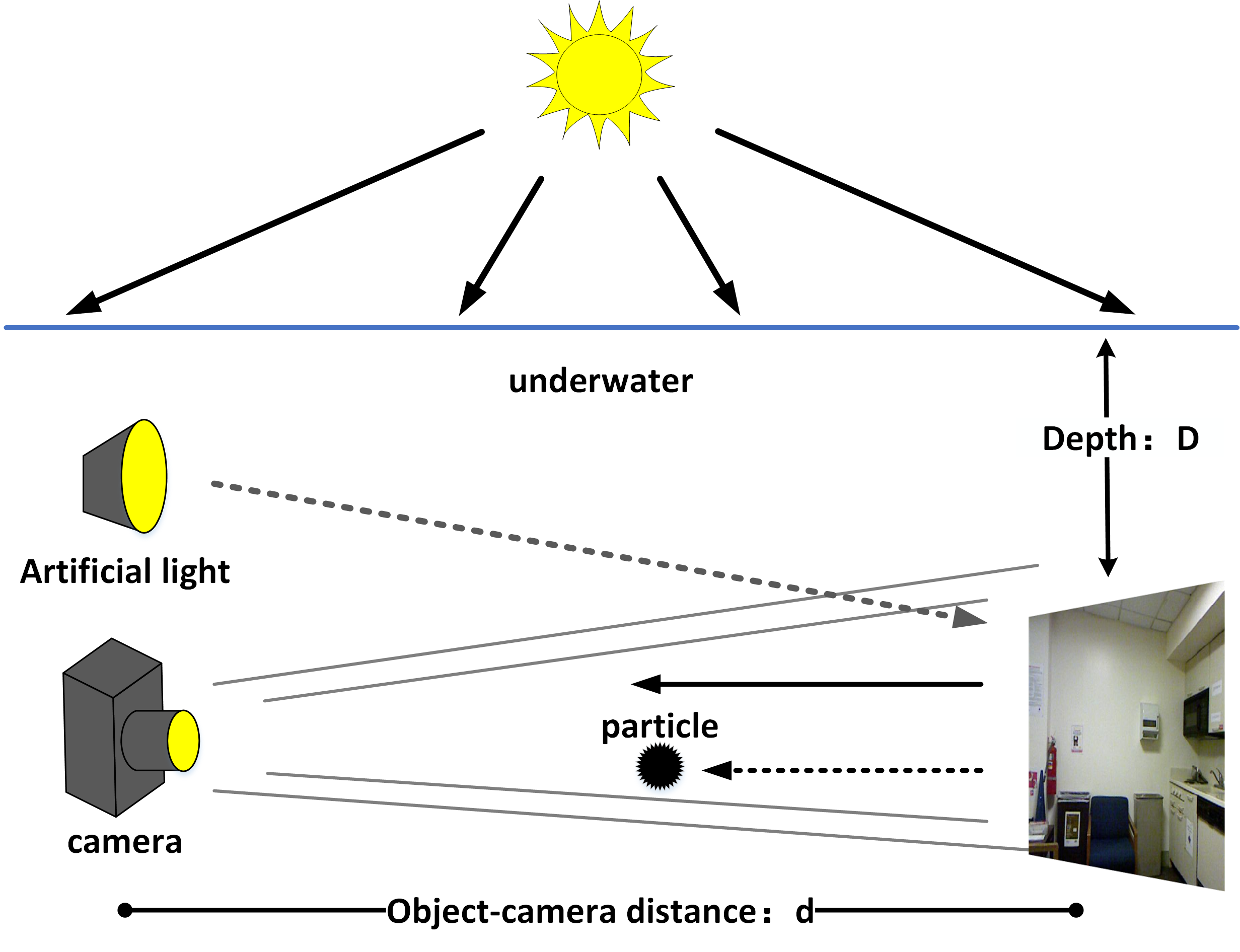}
\caption{The radiance perceived by the camera $I_\lambda$ is the sum of
 direct signal and background scattering.}
\label{illustration}
\end{figure}

One crucial challenge of underwater image enhancement is the endless diversity versus the limited diversity of the training samples, which harms the
generalization ability of the conventional underwater methods trained on the synthesized dataset. As shown in Fig. \ref{illustration}, the radiance perceived by the camera $I_\lambda$ is the sum of direct signal and background scattering, which is  caused by light reflected by particles suspended in the water column.
By ignoring artificial light, the underwater images can be modeled as follows
\begin{equation}\label{uw model}
I_\lambda(x)=J_\lambda(x)t_\lambda(x)+B_\lambda(1-t_\lambda(x)),~~\lambda\in \{R,G,B\},
\end{equation}
where $J_\lambda(x)$ is the clean image,
$B_\lambda$ is the background light and  $t_\lambda(x)$ is the transmission map defined by $t_\lambda(x)=e^{-c_\lambda d(x)}$ with $c_\lambda$ being the attenuation coefficients and $d(x)$ being  the transmission distance.
We propose a multi-variable CNN model to simultaneously estimate the clean image, background light, and transmission map according to the underwater image model \eqref{uw model}, which can ideally  adapt to complex and diversified underwater environments.

\begin{figure*}[t]
\centering
\includegraphics[width=0.85\textwidth]{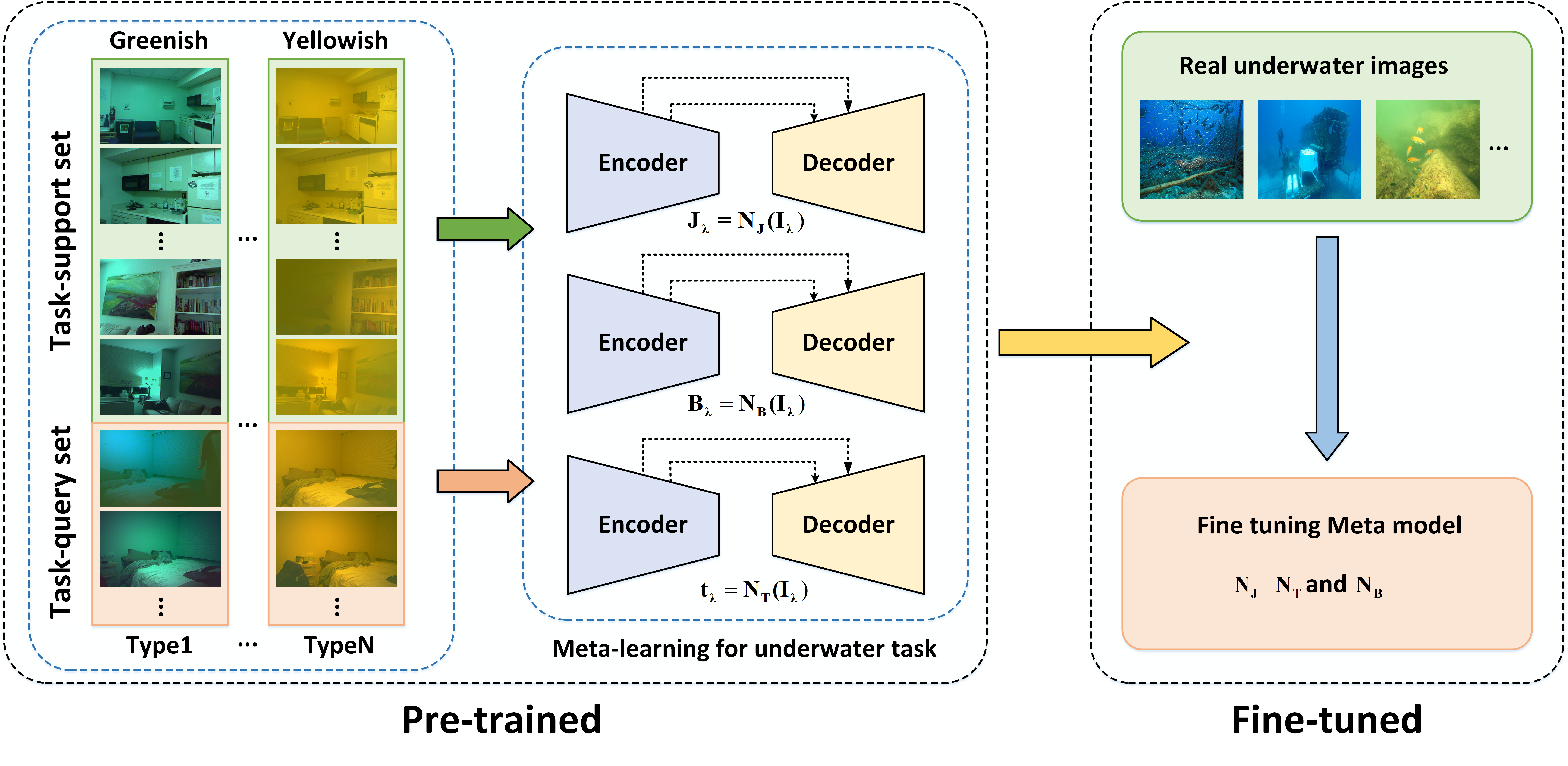}
\caption{The overview framework of our deep meta-learning approach for underwater images enhancement, where three variables are jointly pre-trained and tine-tuned on synthetic and real underwater image dataset, respectively.}
\label{meta learning}
\end{figure*}

\subsection{Network architecture}
\begin{figure}[htbp]
\centering
\includegraphics[width=0.5\textwidth]{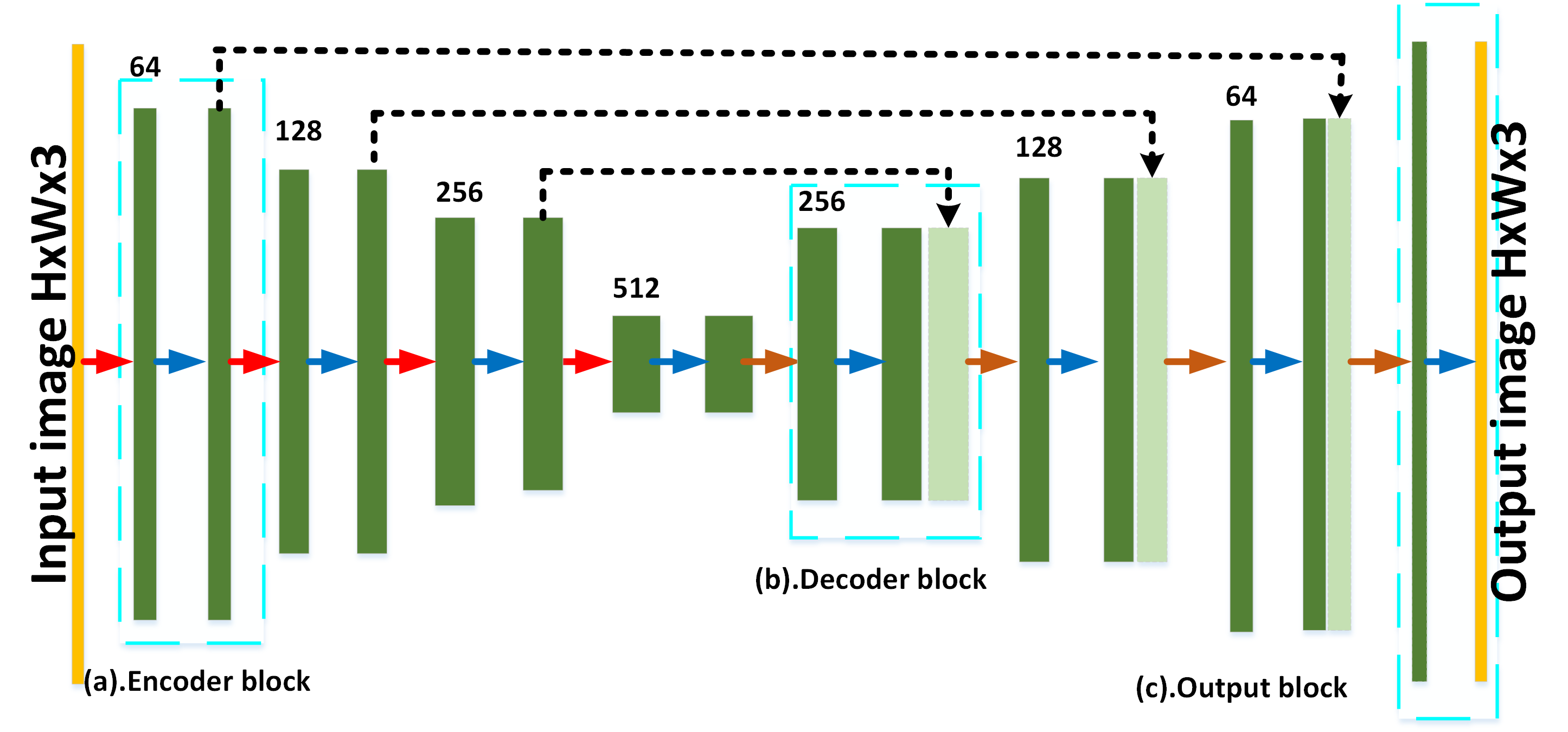}
\caption{The architecture of the proposed underwater image enhancement network, where (a) is an encoder block,
(b) is the decoder block and (c) is the output block.  }
\label{Unet}
\end{figure}

The overall framework of our model-based meta-learning model is illustrated in Fig. \ref{meta learning}, which contains three sub-networks, i.e.,
the clean image estimation sub-network $N_{J}$, background light estimation sub-network $N_B$, and the transmission map estimation sub-network $N_{T}$.
More specifically, the \textbf{clean image estimation sub-network} is used to estimate the clean image $J_\lambda$ from the observed underwater image as follows
\[ J_\lambda= N_{J}(I_\lambda;\theta_{J}),  \]
where $I_\lambda$ is the input image and $\theta_{J}$ represents the parameters of network $N_{J}$. The \textbf{background light estimation sub-network} is used to evaluate the background light $B_\lambda$
\[B_\lambda= N_{B}(I_\lambda;\theta_{B}), \]
where $\theta_{B}$ represents the parameters of network $N_{B}$. Finally, the \textbf{transmission map estimation sub-network} aims to estimate the transmission map $t_\lambda(x)$ from an underwater image as below
\[ t_\lambda = N_{T}(I_\lambda;\theta_T),\]
where $\theta_T$ is the parameters of network $N_{T}$. All the sub-networks are developed based on an encoder-decoder architecture with skip connections and shortcut connections \cite{he2016deep} to perform the translations from degraded images to target images. The detailed network architectures are provided in Fig. \ref{Unet}.
As can be seen, the network consists of four encoder blocks and four decoder blocks with the skip connections from each encoder block to the corresponding decoder block.
To ensure that useful information is well preserved in the block outputs, we also build an open path by shortcut connection in each block. In the encoder, the image is finally downsampled into 512 feature maps, while we add a bilinear interpolation operation to up-sample the input feature maps in the decoder part.

\section{Meta-training strategy}\label{sec4}

\begin{table*}[t]
  \centering
  \caption{ Attenuation coefficients $e^{-c_\lambda}$ are used for synthesizing nine kinds of underwater images.} \label{attenuation coefficients}
    \begin{tabular}{c|c|c|c|c|c|c|c|c|c}
    \hline
\hline
       Types   & I       & II    & III &  B & 3 &  G &5  &7 &Y \\
      \hline
         blue & 0.982& 0.94& 0.89& 0.88 & 0.8& 0.75&0.67 &0.5& 0.40  \\
         green &0.961& 0.925& 0.885& 0.80& 0.82& 0.79& 0.73 & 0.61&0.60  \\
         red  & 0.805& 0.8&  0.75 &0.70 & 0.71& 0.69&0.67 &0.62&0.61 \\
          \hline
\hline
    \end{tabular}%
\end{table*}%

\subsection{Meta-learning method}

As shown in Fig. \ref{meta learning}, the meta-learning strategy is used to build up our underwater image enhancement model. More specifically, we generate the meta training set $D_{meta}^{p(\tau)}=\{\mathcal{D}_s^{\tau_n},\mathcal{D}_q^{\tau_n}\}_{n=1}^N$, where $\mathcal{D}_s^{\tau_n}$ and $\mathcal{D}_q^{\tau_n}$ are the support set and query set of each task, and $N$ is the total number of tasks.
To learn a generalized model from different tasks, the sample $k$ tasks are randomly chosen as a mini-batch from the meta training set.
When the model is applied to a new task $\tau_i$, the parameters of our model are updated according to the task requirements, with the loss function of the $i$-th support set  $\mathcal{D}_s^{\tau_i}$ denoted by $L_{\tau_i}$ for $i\in \{1,2,\cdots,k\}$.
To quickly obtain such ability, we use a two-level gradient descent method to update the model parameters on the support set and query set accordingly as follows
\begin{equation*}
\left\{
\begin{split}
\theta_{J,i}'&\leftarrow \theta_{J}-\alpha \nabla L_{\tau_i} (N_{J}(\theta_{J}));\qquad\mbox{on the support set}\\
\theta_{J,i} &\leftarrow\theta_{J,i}'-\alpha \nabla L_{\tau_i} (N_{J}(\theta_{J,i}') );~\quad\mbox{on the query set}
\end{split}\right.
\end{equation*}
where $\alpha$ is the step size of meta-learning.
The gradients of all tasks are integrated  to obtain the final update for the model parameters such as
\[\theta_{J} \leftarrow \theta_{J}-\beta \frac 1k \sum_{i=1}^k (\theta_{J}- \theta_{J,i}), \]
where $\beta$ is the outer learning rate.
Similarly, the parameters $\theta_B$ and $\theta_T$ are also updated by two-level gradient descent method.

To adapt to real underwater environments, we fine-tune the pre-trained model on real underwater images with unknown distortions.
Then, we leverage the Adam optimizer to update the enhancement network $N_{J}$ on real underwater images as follows
\[\theta_{J}\leftarrow \theta_{J}-\alpha_f \nabla L(N_{J}(\theta_{J})), \]
where $\alpha_f$ is the learning rate of fine-tuning and the other network parameters $\theta_B$ and $\theta_T$ are updated in the similar way.
Then, the fine-tuned model can be obtained for enhancing the underwater images with unknown distortions.
The implementation details of MetaUE are summarized in Algorithm \ref{alg}.

\begin{algorithm}[htbp]
\caption{Model-based Meta-learning for Underwater Enhancement (MetaUE)}
\LinesNumbered
\label{alg}
\KwIn{Meta-trained set $\mathcal{D}_{meta}^{p(\tau)}=\{\mathcal{D}_s^{\tau_n}, \mathcal{D}_q^{\tau_n} \}_{n=1}^N$}
\KwOut{$J_\lambda$, $B_\lambda$ and $t_\lambda$}
\tcc{Pre-trained on synthetic image dataset}
\For {$epoch=1,2,...$}
{
Sample a mini-batch of $k$ tasks in $\mathcal{D}_{meta}^{\tau_i}$ with $i\in \{1,2,\cdots,k\}$\;
\For {$i=1,2,\cdots,k$}
{
Compute $\theta_{J,i}\leftarrow\theta_{J}-\alpha \nabla L_{\tau_i} (N_{J}(\theta_{J}) )$\;
Compute $\theta_{B,i}\leftarrow\theta_{B}-\alpha \nabla L_{\tau_i} (N_{B}(\theta_{B}) )$\;
Compute $\theta_{T,i}\leftarrow\theta_{T}-\alpha \nabla L_{\tau_i} (N_{T}(\theta_{T}) )$\;
}
update $\theta_{J} \leftarrow \theta_{J}-\beta \frac 1k \sum_{i=1}^k (\theta_{J}- \theta_{J,i})$\;
update $\theta_{B} \leftarrow \theta_{B}-\beta \frac 1k \sum_{i=1}^k (\theta_{B}- \theta_{B,i})$\;
update $\theta_{T} \leftarrow \theta_{T}-\beta \frac 1k \sum_{i=1}^k (\theta_{T}- \theta_{T,i})$\;
}

\tcc{Fine-tuned on real image dataset}
Fine-tuning the model $\theta_{J}$, $\theta_{B}$, and $\theta_{T}$ as
\begin{equation*}
\left\{
\begin{split}
\theta_{J}&\leftarrow \theta_{J}-\alpha_f \nabla L(N_{J}(\theta_{J}));\\
\theta_{B}&\leftarrow \theta_{B}-\alpha_f \nabla L(N_{B}(\theta_{B}));\\
\theta_{T}&\leftarrow \theta_{T}-\alpha_f \nabla L(N_{T}(\theta_{T})).
\end{split}\right.
\end{equation*}
\end{algorithm}

\subsection{Synthetic underwater image dataset}
In the following, we build up an synthetic underwater image dataset to cover different underwater degradation images for implementing Algorithm \ref{alg}.
As shown in Fig.\ref{illustration}, the incident light passing through the surface of water is the major source of illumination in an underwater environment. Here, we use the NYU-V1 dataset \cite{silberman2011indoor} as a benchmark dataset to generate various underwater images. Due to beam attenuation, the incident light passing through the water may make the underwater environment appearing blue, green or yellow.
On the other hand, the influence of artificial light should be also considered to overcome the disadvantage of insufficient underwater illumination.
Therefore, the underwater illumination can be simplified as
\begin{equation}\label{illumination}
E_\lambda(x) = \omega_a E_\lambda^S e^{-c_\lambda D} +  \omega_b E_\lambda^{A} e^{-c^\lambda d(x)},
\end{equation}
where $\omega_a$ and $\omega_b$ are two weights, $E_\lambda^S$ is light on the water surface, $E_\lambda^{A}$ is artificial light, $d(x)$ is the distance form object to the camera and $D$ is the water depth.
The backscattering component mainly contributes to the hazy look of underwater images.
An efficient model was proposed in \cite{zhao2015deriving} to calculate the background light $B_\lambda$ by
\[B_\lambda(x) =\kappa E_\lambda(x) /c_\lambda,\]
where $\kappa$ is a scalar defined by the camera system.
Then the simulation formula from the scene $J^{gt}$ without attenuation to the underwater image $I_\lambda$ can be expressed as follows
\begin{equation}\label{Liu uw model}
I_\lambda(x)=t_\lambda(x)J_\lambda^{gt}(x) \frac{E_\lambda(x)}{E_\lambda^S } + \frac{\kappa E_\lambda(x)}{c_\lambda} (1-t_\lambda(x)).
\end{equation}
Here, depending on the camera principle, $J_\lambda^{gt}(x)\frac{E_\lambda(x)}{E_\lambda^S }$ is to remove the light $E_\lambda^S$ at $x$ of clean image $J_\lambda^{gt}$ and use $E_\lambda(x)$ as the new lighting condition.

\textbf{Simulation of lighting parameters $E_\lambda(x)$.}
To effectively simulate various illumination conditions, the range of air light $E_\lambda^S$ is set to [0.7,1] and the depth variable $D$ is randomly selected from [5m,20m].
The artificial illumination is the only light source to provide uneven brightness and limited visibility.
Here, we use two-dimensional Gaussian distribution to simulate the artificial light in the water, where beam pattern $E_\lambda^A=\mathcal{P}(\widetilde{x}|E_\lambda^{art}, \sigma)$ with a  peak value  $E_\lambda^{art}\in [0.7,1]$ and standard deviation $\sigma$ proportional to the width of the image by a random rate in $[0.2,1.1]$. The light source $\widetilde{x}$ is randomly picked from the image.

\textbf{Simulation of channel-wise attenuation coefficients $c_\lambda$.}
The attenuation coefficients of Jerlov water types \cite{li2020underwater,mobley1994light} are used to generate different water types.
To better adapt to complex underwater situations, we remove the clearest water quality Type 1 and the most turbid Type 9 and introduce three kinds of water quality for the three common underwater colors (i.e., Blue, Green, and Yellow), named Type B, G, Y. The corresponding values of $c_\lambda$ for all nine types of underwater images are presented in Table \ref{attenuation coefficients}.

\textbf{Simulation of transmission distance $d(x)$}.
We use the NYU-V1 dataset \cite{silberman2011indoor} to generate the synthetic underwater image dataset, which contains a total of 3733 RGB images and corresponding depth maps.

To sum up, we use the parameters in Table \ref{notation} to build up a synthetic underwater image dataset based on the RGB-D NYU-V1 dataset. In particular,
 we randomly selected  parameters three times for each water type to obtain a total of $27$ specific underwater distortion types and 100791 images.

\begin{table}[t]
  \centering
  \caption{The parameters used to generate the synthetic underwater image dataset, where $\lambda\in{\{R,G,B\}}$.}\label{notation}%
    \begin{tabular}{c|c|c}
 \hline
 \hline
 Notation & Description & Range\\
 \hline
 $D$ & Water depth & $[5m,20m]$\\
 $d$ &Transmission distance & NYU-V1 dataset \cite{silberman2011indoor}\\
 $c_\lambda$ & Attenuation coefficients & Table \ref{attenuation coefficients} \\
 $E_\lambda^S$ &Air light&   $[0.7,1]$\\
 $E_\lambda^{art}$ & Peak value of artificial light &  $[0.7,1]$\\
 $\widetilde{x}$ & Location of $E_\lambda^{art}$ & A random point in image\\
 $\sigma$ & Coverage of  artificial light & Random rate  $[0.2,1.1]$\\
 $\omega_a, \omega_b$ & Weights of lighting & $\omega_a\in[0,1] ~\mbox{and}~\omega_a+\omega_b=1$\\
 $\kappa$ &Camera system parameter & $[0.7,1.1]$\\
 \hline
 $E_\lambda^A$ &Artificial light& $E_\lambda^A=\mathcal{P}(\widetilde{x}|E_\lambda^{art}, \sigma)$ \\
 $E_\lambda$ & Underwater illumination & estimated by \eqref{illumination}\\
 $t_\lambda$ & Transmission map & $t_\lambda=e^{-c_\lambda d}$ \\
 $B_\lambda$ & Background light & $B_\lambda=\kappa E_\lambda /c_\lambda$\\
 \hline
 \hline
    \end{tabular}%
\end{table}%

\subsection{Loss function}
To effectively train our multi-variable CNN model, we introduce a novel task-specific loss function as below.

\textbf{Clean image supervision loss.}
To remove water, we compensate the disparities of wavelength attenuation for traversing the water depth to the surface of water.
To be specific,  we use $L_1$ loss function to measure supervision loss of clean image
\[L_J=  \| J_\lambda -J_\lambda^{gt} \|_1, \]
where $J_\lambda^{gt}$ denotes the clean image without attenuation.

\textbf{Background light supervision loss.} Similarly, the background light is supervised by the $L_1$ loss function as follows
\[L_B =\|B_\lambda  -B_\lambda^{gt}\|_1, \]
where $B_\lambda^{gt}=\kappa E_\lambda/c_\lambda$ can be regarded as the ground truth of the background light.

\textbf{Transmission map supervision loss.} We penalize the difference between
$t_\lambda$ and the reference transmission map $t^{gt}_\lambda$ by
\[L_T =\|t_\lambda  -t^{gt}_\lambda\|_1, \]
where the reference transmission map is given by $t^{gt}_\lambda= e^{-c_\lambda d}$.

\textbf{Underwater supervision loss.}
Based on the physical model \eqref{uw model}, the underwater image can be calculated by
\[\widetilde{I}_\lambda(x)= J_\lambda t_\lambda + B_\lambda(1-t_\lambda).\]
Thus, the loss based on the physical process is defined
to minimize the difference between the underwater $I_\lambda$ and the physics-based estimation solution as follows
\[L_{I} =\|I_\lambda  -\widetilde{I}_\lambda\|_1. \]
Finally, our multi-variable CNN model is jointly optimized by a weighted combination of $L_J$, $L_B$,$L_{T}$ , and $L_{I}$ as follows
\begin{equation}\label{loss function}
L=c_JL_J+c_BL_B+c_TL_T+ c_{I} L_{I},
\end{equation}
where $c_J,c_B,c_T,c_{I}$ are the weights used to balance different terms.
During the meta-learning implementation, the loss function $L_{\tau_i}$ of $i-$th support set is defined as \eqref{loss function}.
The weights in our loss function \eqref{loss function} are empirically set as $c_J=c_B=c_T=1$ and $c_{I}=0.5$ for the pre-trained model.
Due to the lack of ground truth of transmission map and background lights for underwater images, we set the weights as $c_J=1$, $c_B=c_T=0$ and $c_{I}=1$ when we fine-tune the model.

\section{Experimental results}\label{sec5}

\subsection{Implementation details}
In the proposed model, all training images are cropped to $256\times 256$ pixel patches  to feed  into the proposed model.
We train our model using bi-level gradient optimization with an inner learning rate $\alpha$ of $1e-4$ and outer learning rate $\beta$ of $5e-5$, which is implemented based on Pytorch \cite{paszke2017automatic}.
The total epoch is set to $40$, and the batch size of training data is fixed to 8.
We set the fine-tuned learning rate $\alpha_f$ to $1e-5$ and the total epoch is 30.
These learning rates drop to a factor of $0.8$ after every five epochs.
The mini-batch size is set as $k=5$.
The learning rates drop to a factor of $0.8$ after every two epochs.

\begin{figure}[t]
\centering
\subfloat[\small{Validation Loss}]{\includegraphics[width=0.245\textwidth]{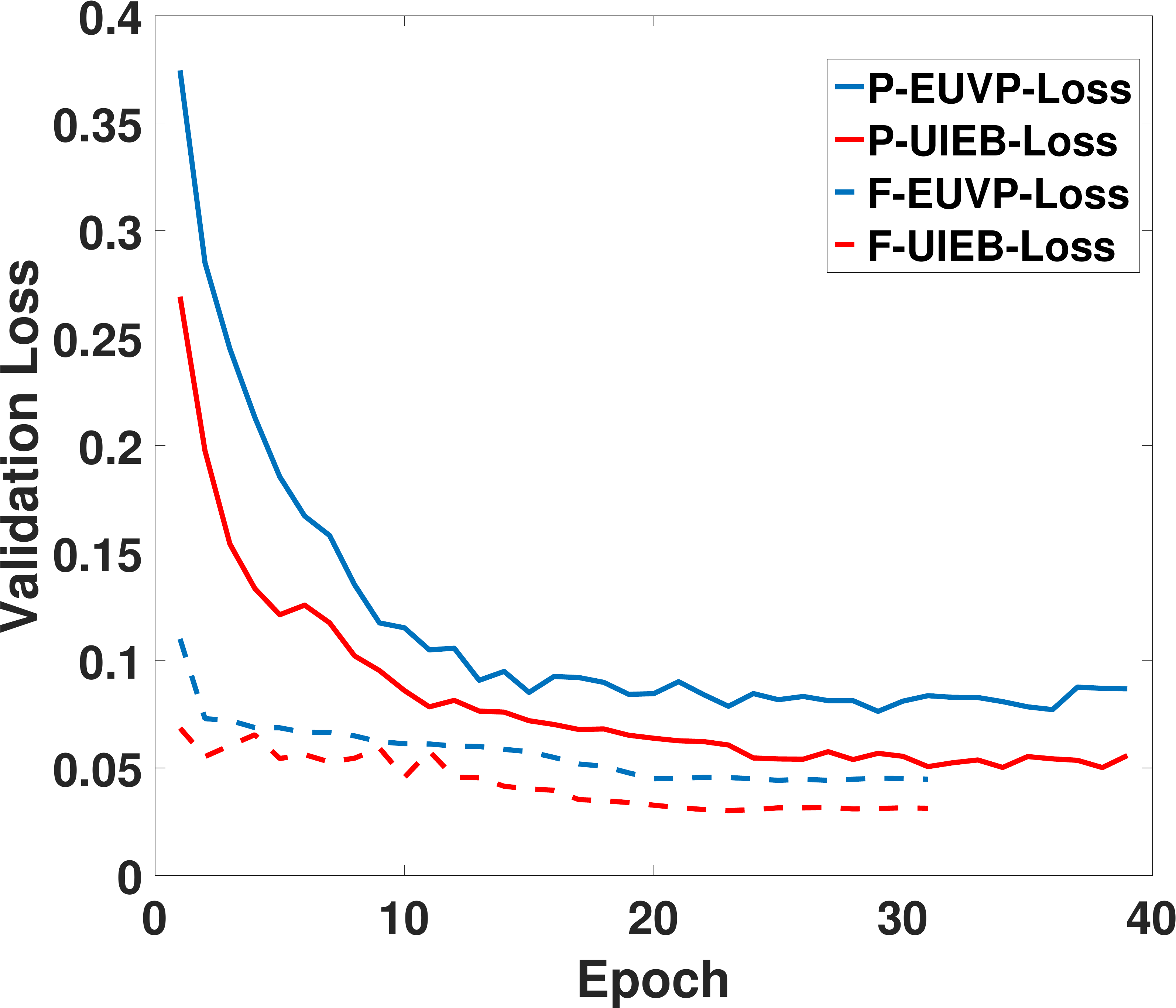}}
\hspace{-0.7mm}
\subfloat[\small{PSNR}]{\includegraphics[width=0.24\textwidth]{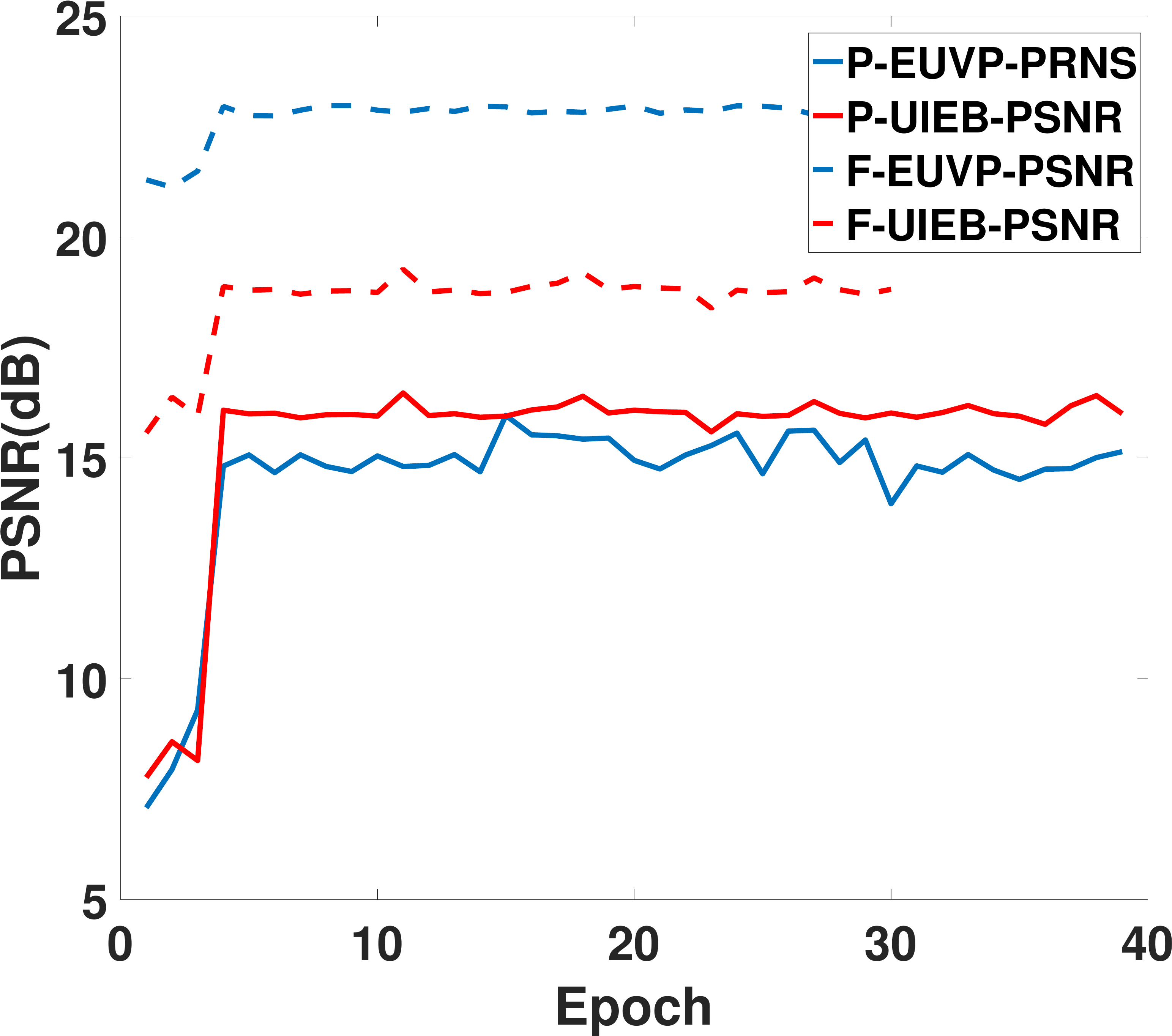}}
\caption{There are the performances of validation loss and averaged PSNR for the pre-trained and fine-tuned model on the UIEB dataset and EUVP dataset during the training process.
}
\label{loss-psnr}
\end{figure}

\begin{figure*}[htbp]
\centering
\setcounter{subfigure}{0}
\hspace{-3.5mm}
\subfloat{\rotatebox{90}{\footnotesize{~~~~~Input}}
\includegraphics[width=0.095\textwidth]{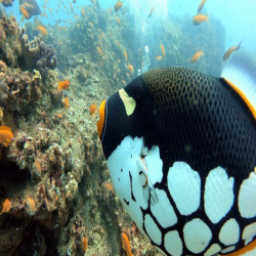}
\hspace{-2mm}
\includegraphics[width=0.095\textwidth]{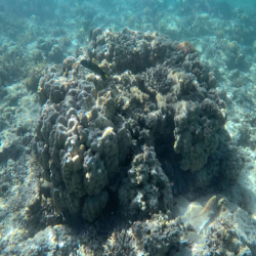}}
\hspace{-0.7mm}
\subfloat{\includegraphics[width=0.095\textwidth]{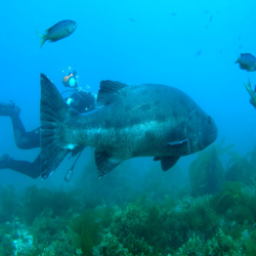}}
\hspace{-0.7mm}
\subfloat{\includegraphics[width=0.095\textwidth]{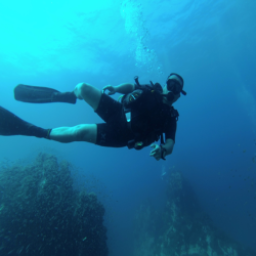}}
\hspace{-0.7mm}
\subfloat{\includegraphics[width=0.095\textwidth]{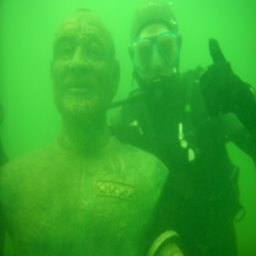}}
\hspace{-0.7mm}
\subfloat{\includegraphics[width=0.095\textwidth]{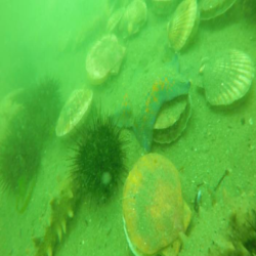}}
\hspace{-0.7mm}
\subfloat{\includegraphics[width=0.095\textwidth]{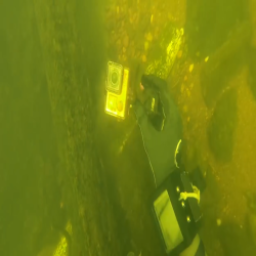}}
\hspace{-0.7mm}
\subfloat{\includegraphics[width=0.095\textwidth]{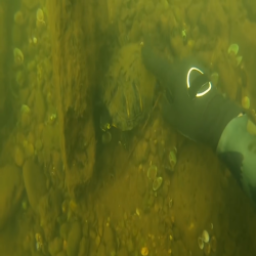}}
\hspace{-0.7mm}
\subfloat{\includegraphics[width=0.095\textwidth]{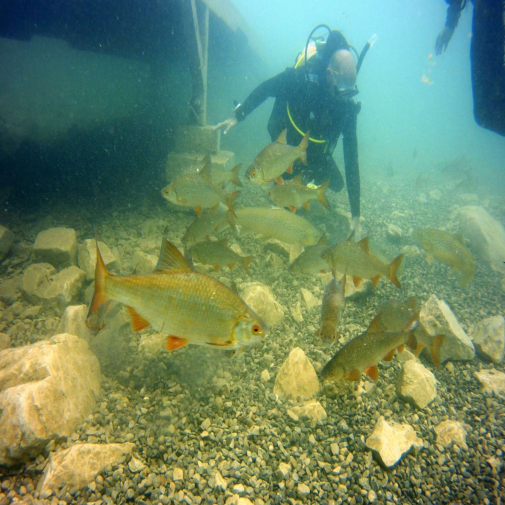}}
\hspace{-0.7mm}
\subfloat{\includegraphics[width=0.095\textwidth]{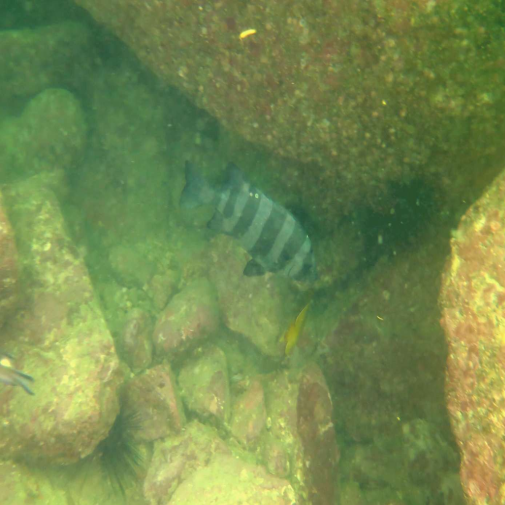}}\\

\vspace{0.5mm}
\hspace{-3mm}
\subfloat{\rotatebox{90}{\footnotesize{~~Pre-trained}} \includegraphics[width=0.095\textwidth]{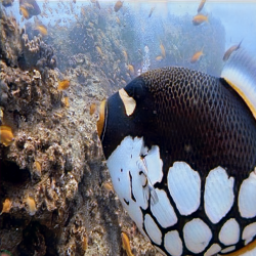}}
\hspace{-0.7mm}
\subfloat{\includegraphics[width=0.095\textwidth]{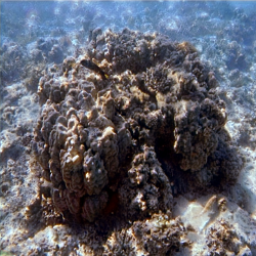}}
\hspace{-0.7mm}
\subfloat{\includegraphics[width=0.095\textwidth]{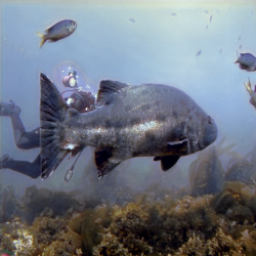}}
\hspace{-0.7mm}
\subfloat{\includegraphics[width=0.095\textwidth]{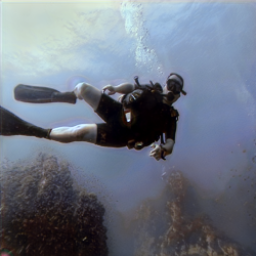}}
\hspace{-0.7mm}
\subfloat{\includegraphics[width=0.095\textwidth]{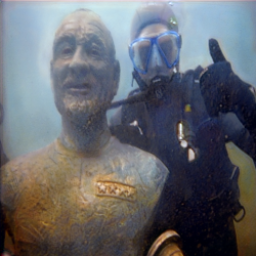}}
\hspace{-0.7mm}
\subfloat{\includegraphics[width=0.095\textwidth]{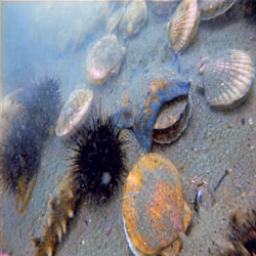}}
\hspace{-0.7mm}
\subfloat{\includegraphics[width=0.095\textwidth]{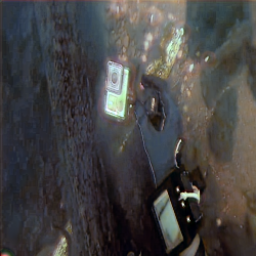}}
\hspace{-0.7mm}
\subfloat{\includegraphics[width=0.095\textwidth]{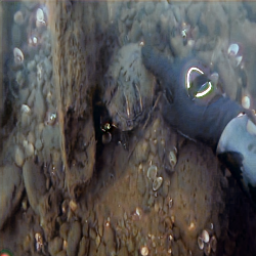}}
\hspace{-0.7mm}
\subfloat{\includegraphics[width=0.095\textwidth]{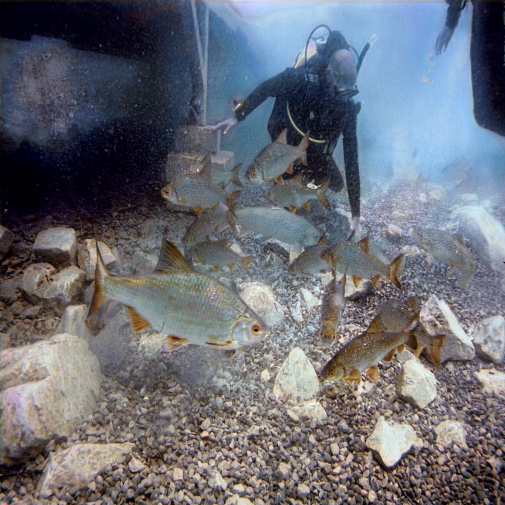}}
\hspace{-0.7mm}
\subfloat{\includegraphics[width=0.095\textwidth]{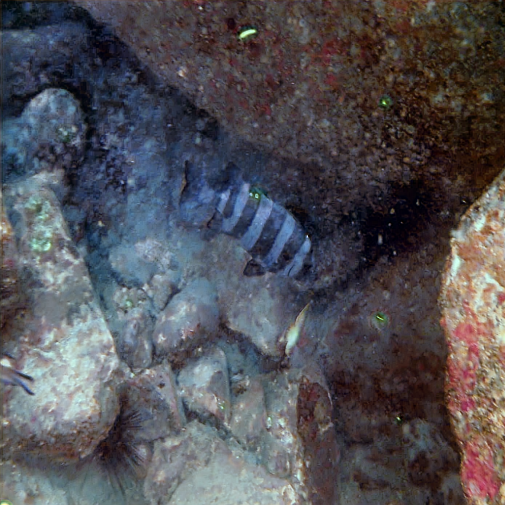}}

\setcounter{subfigure}{0}
\vspace{0.5mm}
\hspace{-3mm}
\subfloat{\rotatebox{90}{\footnotesize{~~~F-EUVP}}
\includegraphics[width=0.095\textwidth]{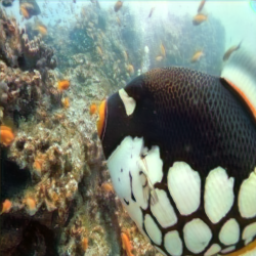}
\hspace{-1.9mm}
\includegraphics[width=0.095\textwidth]{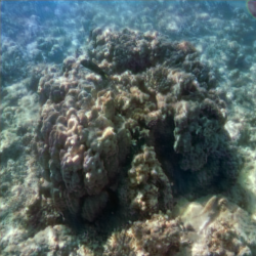}
}
\hspace{-3.1mm}
\subfloat{
\includegraphics[width=0.095\textwidth]{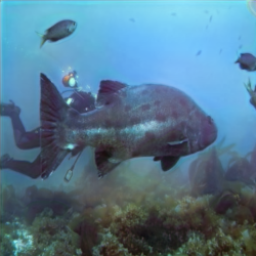}
\hspace{-1.9mm}
\includegraphics[width=0.095\textwidth]{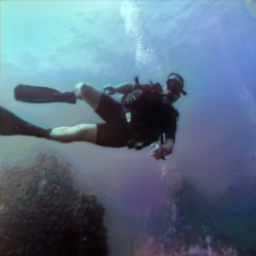}
}
\hspace{-3.1mm}
\subfloat{
\includegraphics[width=0.095\textwidth]{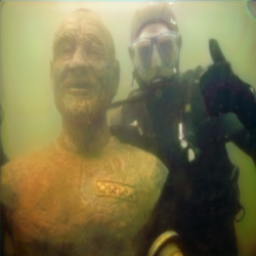}
\hspace{-1.9mm}
\includegraphics[width=0.095\textwidth]{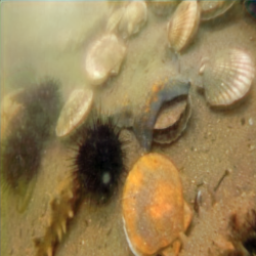}}
\hspace{-1.9mm}
\subfloat{
\includegraphics[width=0.095\textwidth]{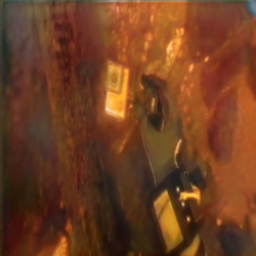}
\hspace{-1.9mm}
\includegraphics[width=0.095\textwidth]{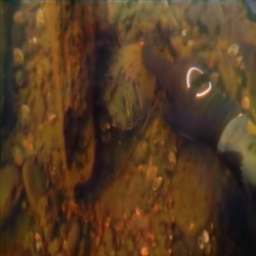}}
\hspace{-1.9mm}
\subfloat{
\includegraphics[width=0.095\textwidth]{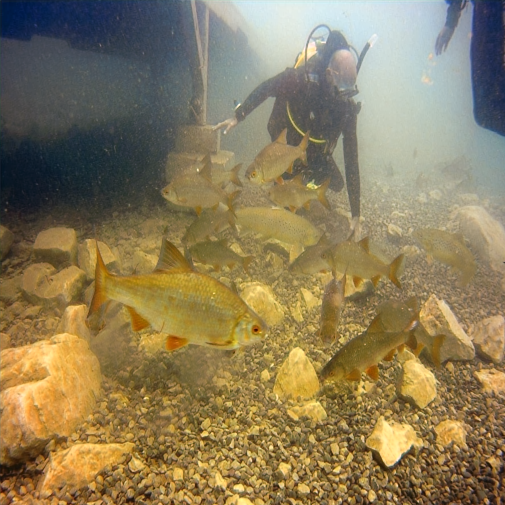}
\hspace{-1.9mm}
\includegraphics[width=0.095\textwidth]{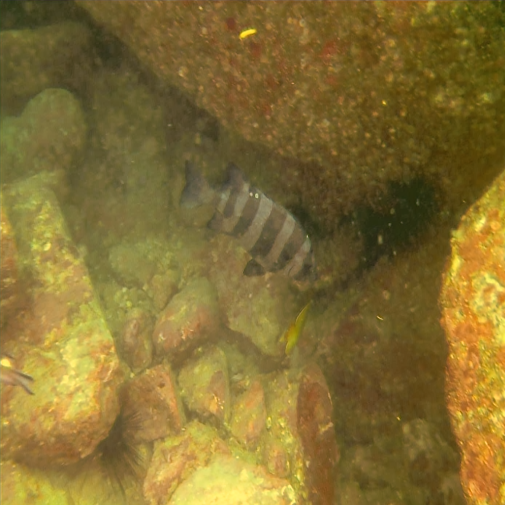}}

\setcounter{subfigure}{0}
\vspace{0.5mm}
\hspace{-3mm}
\subfloat[\small{Haze}]{\rotatebox{90}{\footnotesize{~~~F-UIEB}}
\includegraphics[width=0.095\textwidth]{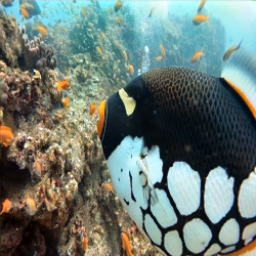}
\hspace{-1.9mm}
\includegraphics[width=0.095\textwidth]{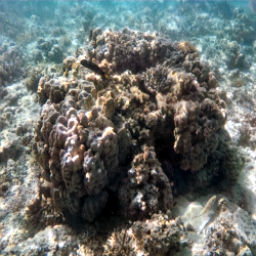}}
\hspace{-1.9mm}
\subfloat[\small{Blue}]{
\includegraphics[width=0.095\textwidth]{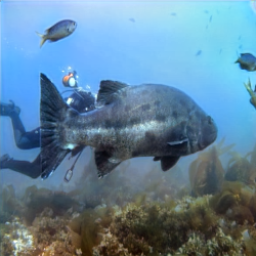}
\hspace{-1.9mm}
\includegraphics[width=0.095\textwidth]{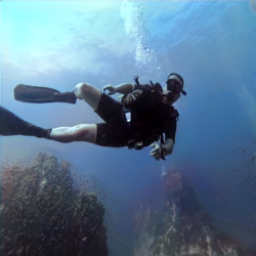}}
\hspace{-1.9mm}
\subfloat[\small{Green}]{
\includegraphics[width=0.095\textwidth]{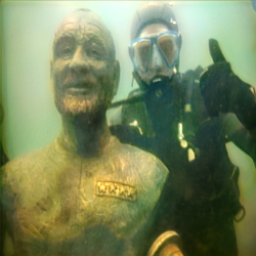}
\hspace{-1.9mm}
\includegraphics[width=0.095\textwidth]{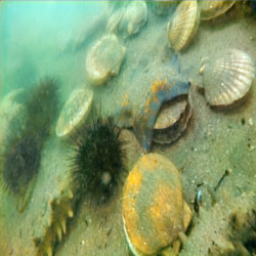}}
\hspace{-1.9mm}
\subfloat[\small{Yellow}]{
\includegraphics[width=0.095\textwidth]{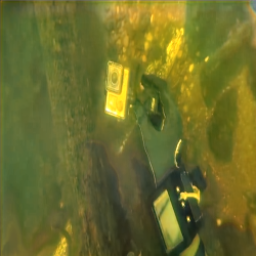}
\hspace{-1.9mm}
\includegraphics[width=0.095\textwidth]{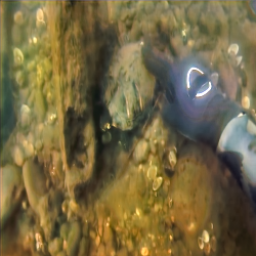}}
\hspace{-1.9mm}
\subfloat[\small{Turbid}]{
\includegraphics[width=0.095\textwidth]{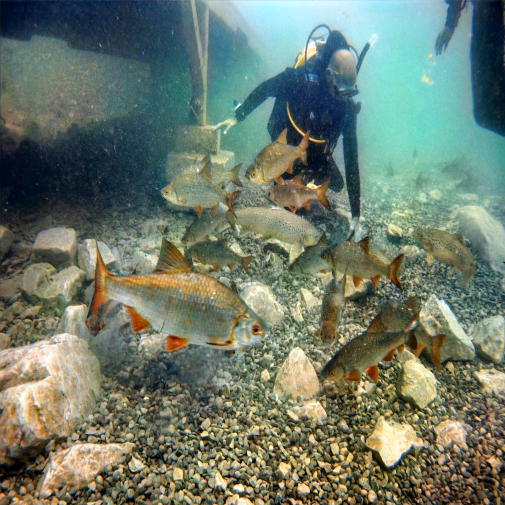}
\hspace{-1.9mm}
\includegraphics[width=0.095\textwidth]{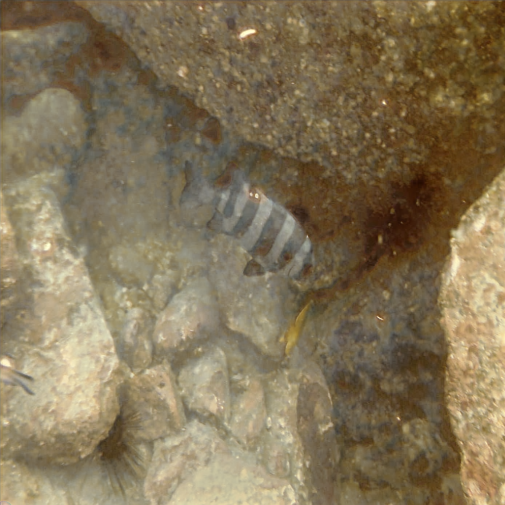}}
\caption{The enhanced results of pre-trained model for common underwater degradation types, where F-EUVP and F-UEIB are the fine-tuned models on EUVP and UIEB, respectively.}
\label{enhance}
\end{figure*}

\subsection{Comparison methods}
The performance of our MetaUE is compared against eight state-of-the-art underwater image enhancement methods, including three model-based methods, i.e., ACDC \cite{zhang2022underwaterACDC}, Rank1 \cite{liu2021rank} and  MMLE \cite{zhang2022underwater}, and
five learning-based methods, i.e., UWCNN \cite{li2020underwater},  Water-Net \cite{UIEB},  FUnIE-GAN \cite{islam2020fast}, Ucolor \cite{li2021underwater}, and TACL \cite{liu2022twin}.
Note that both Water-Net \cite{UIEB} and Ucolor \cite{li2021underwater} were trained on the UIEB dataset, while FUnIE-GAN \cite{islam2020fast} was trained on the EUVP dataset. The UWCNN\cite{li2020underwater} was trained in simulation of multiple water types,
while TACL \cite{liu2022twin} was trained on UIEB and BSD500 to train the twin adversarial enhancement network.

\textbf{Training samples.} The pre-trained model is fine-tuned on two underwater image datasets with reference images, i.e., UIEB \cite{UIEB} and  EUVP \cite{islam2020fast}, respectively.
For a fair comparison, we also retrained the Water-Net and Ucolor on the EUVP dataset and retrained FUnIE-GAN on the UIEB dataset.
Since the trainable codes of both UWCNN and TACL are not available, we use the public models directly in comparison.

\textbf{Testing samples.}
For testing the learning ability and generalization ability of our MetaUE, six underwater image datasets are used to evaluate the performance, which are UIEB \cite{UIEB}, EUVP \cite{islam2020fast}, U45 \cite{li2019fusion}, UIEB-60 \cite{UIEB}, EUVPUN \cite{islam2020fast}, UCCS \cite{duarte2016dataset}.
In particular, it contains 90 paired images of UIEB and 200 paired images of EUVP, respectively.  We randomly selected 150 unpaired images from EUVPUN, 20 images from UCCS, and all real underwater images of U45 and UIEB-60 for comparison.

\subsection{Quantitative evaluation}

\textbf{Full Reference Evaluation:}  For the test datasets with reference images, we  conduct a full reference evaluation using three commonly-used metrics, PSNR, SSIM, and MSE.
These results of full-reference image quality evaluation by using the reference images can provide
realistic feedback on the performance of different methods to some extent.
A higher PSNR and SSIM or lower MSE indicates that the result is close to the reference image.

\textbf{Non-reference Evaluation:}  For the test datasets without references, we  employ two non-reference metrics, i.e., UCIQE \cite{yang2015underwater} and UIQM \cite{panetta2015human}. A higher UCIQE or UIQM score indicates  a better human visual perception.
The UCIQE is a linear combination of chroma $\sigma_c$, contrast $con_l$, and saturation $\mu_s$ defined as follows
\[\mbox{UCIQE}= c_1\times \sigma_c+c_2\times con_l+c_3\times \mu_s\]
with $c_1=0.4680$, $c_2=0.2745$ and $c_3=0.2575$ as in \cite{yang2015underwater}. A higher UCIQE gives a better tradeoff among chroma, contrast and saturation.
The UIQM is a linear composition of UICM, UISM, and UIConM, which represent the metrics of colorfulness, sharpness and contrast, respectively. The formula of UIQM is given as follows
\[\mbox{UIQM}=c_1\times \mbox{UICM}+c_2\times \mbox{UISM}+ c_3\times \mbox{UIconM}\]
with $c_1=0.0282$, $c_2=0.2953$, $c_3=3.5753$ as suggest in \cite{panetta2015human}.
 A higher UIQM indicates a better tradeoff among colorfulness, sharpness and contrast.
\begin{table}[t]
  \centering
  \caption{The averaged unsupervised scores for the test images in Fig. \ref{enhance}.}
    \begin{tabular}{c|c|ccc}
    \hline\hline
      Index    & \multicolumn{1}{c|}{UCIQE} & \multicolumn{1}{c}{$\sigma_c$} & \multicolumn{1}{c}{$con_l$} & \multicolumn{1}{c}{$\mu_s$} \\
          \hline
    Pre-trained & \textbf{0.50} & \textbf{0.5483} & \textbf{0.8732} & 0.001 \\
    F-UIEB & 0.37 & 0.2819 & 0.8581 & 0.0014  \\
    F-EUVP & 0.37 & 0.3302 & 0.782 & \textbf{0.0015}\\
    \hline
  Index   & \multicolumn{1}{c|}{UIQM} & \multicolumn{1}{c}{UICM} & \multicolumn{1}{c}{UISM} & \multicolumn{1}{c}{UIConM} \\
    \hline
    Pre-trained & \textbf{1.66} & 4.5596 & \textbf{1.7938} & 0.2763 \\
    F-UIEB & 1.59  & \textbf{4.9446} & 1.3277 & 0.299 \\
    F-EUVP & 1.55  & 4.7811 & 0.953 & \textbf{0.3114} \\
    \hline\hline
    \end{tabular}%
  \label{Pre-training-uciqe}%
\end{table}%

\begin{figure*}[htbp]
\subfloat{\includegraphics[width=0.095\textwidth]{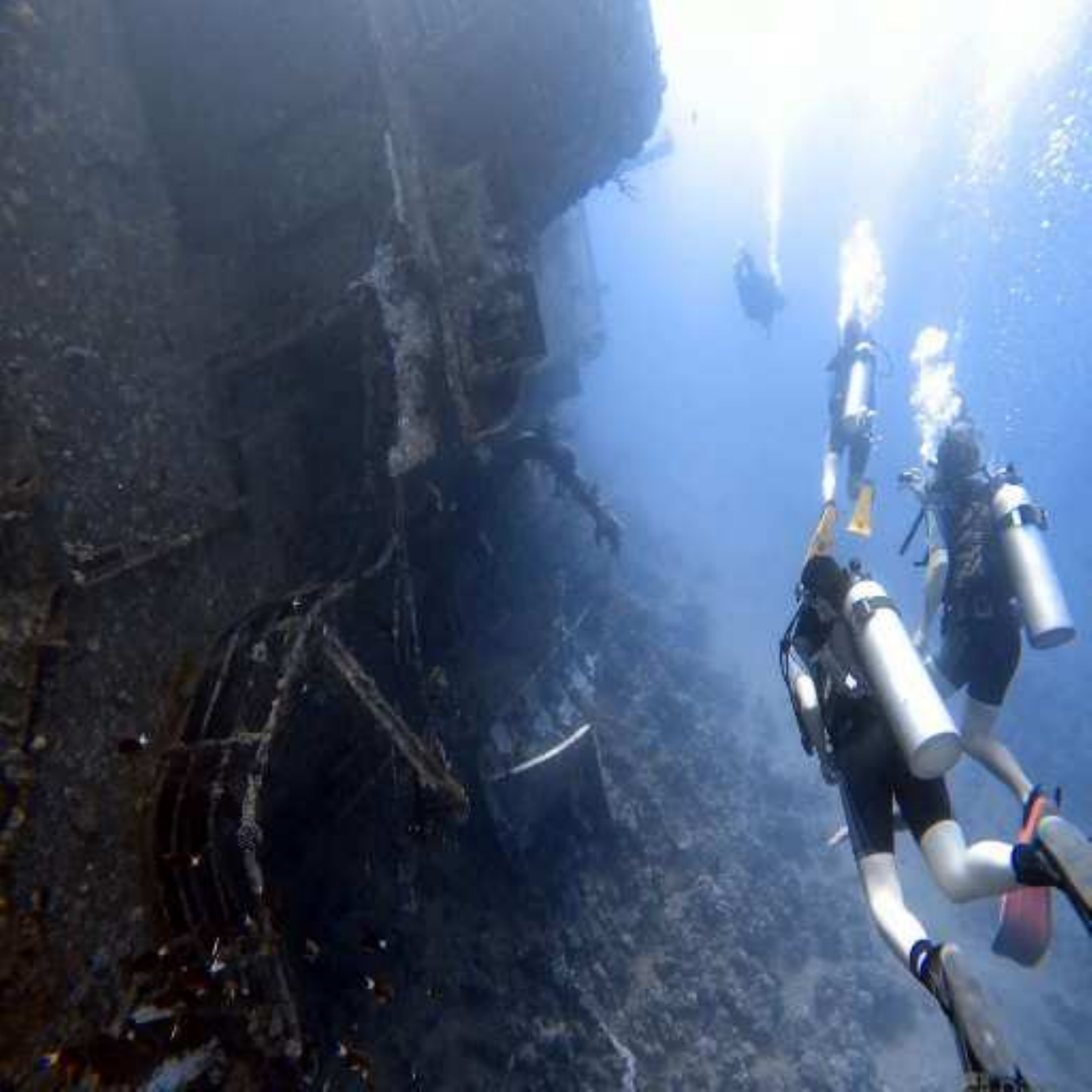}}
\hspace{-0.7mm}
\subfloat{\includegraphics[width=0.095\textwidth]{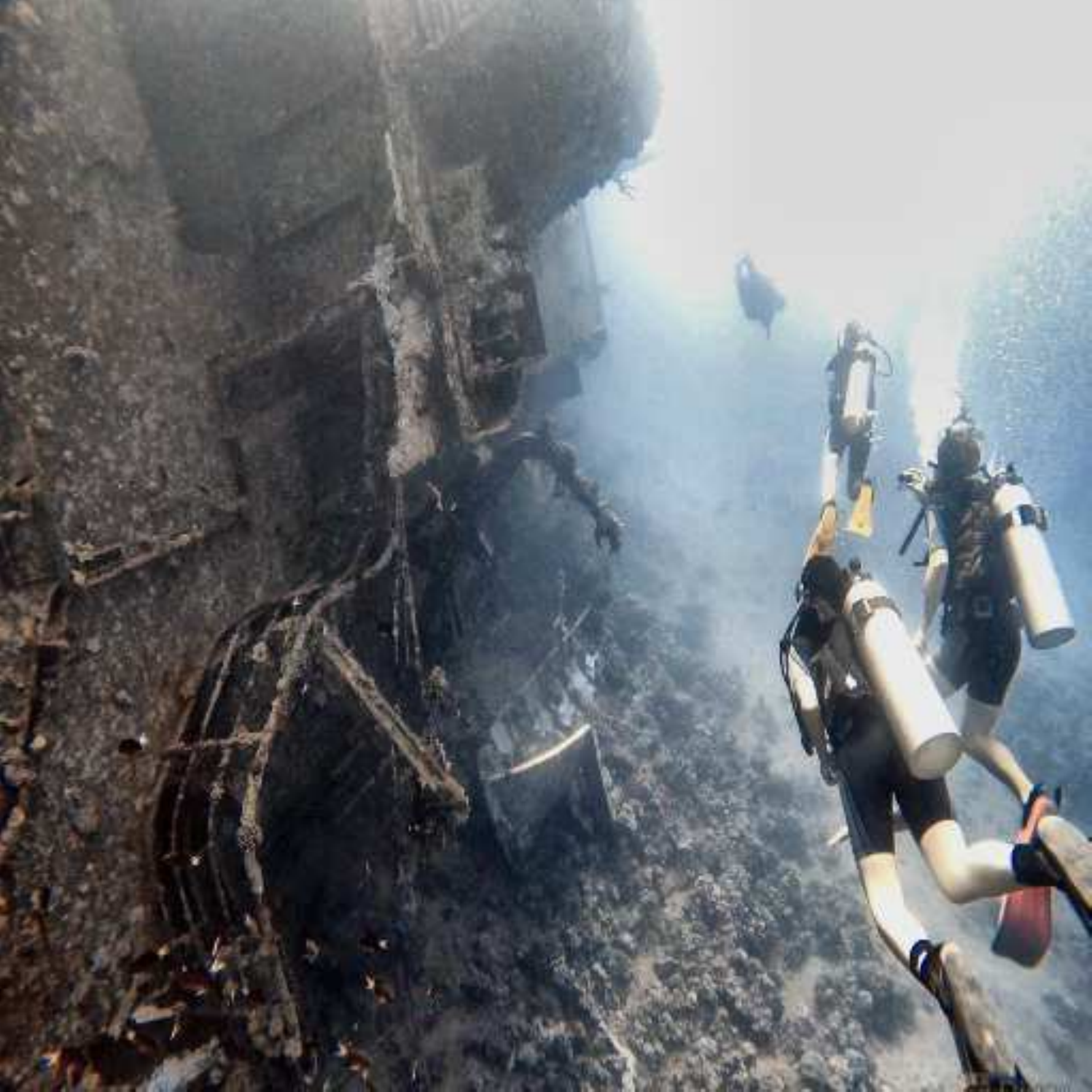}}
\hspace{-0.7mm}
\subfloat{\includegraphics[width=0.095\textwidth]{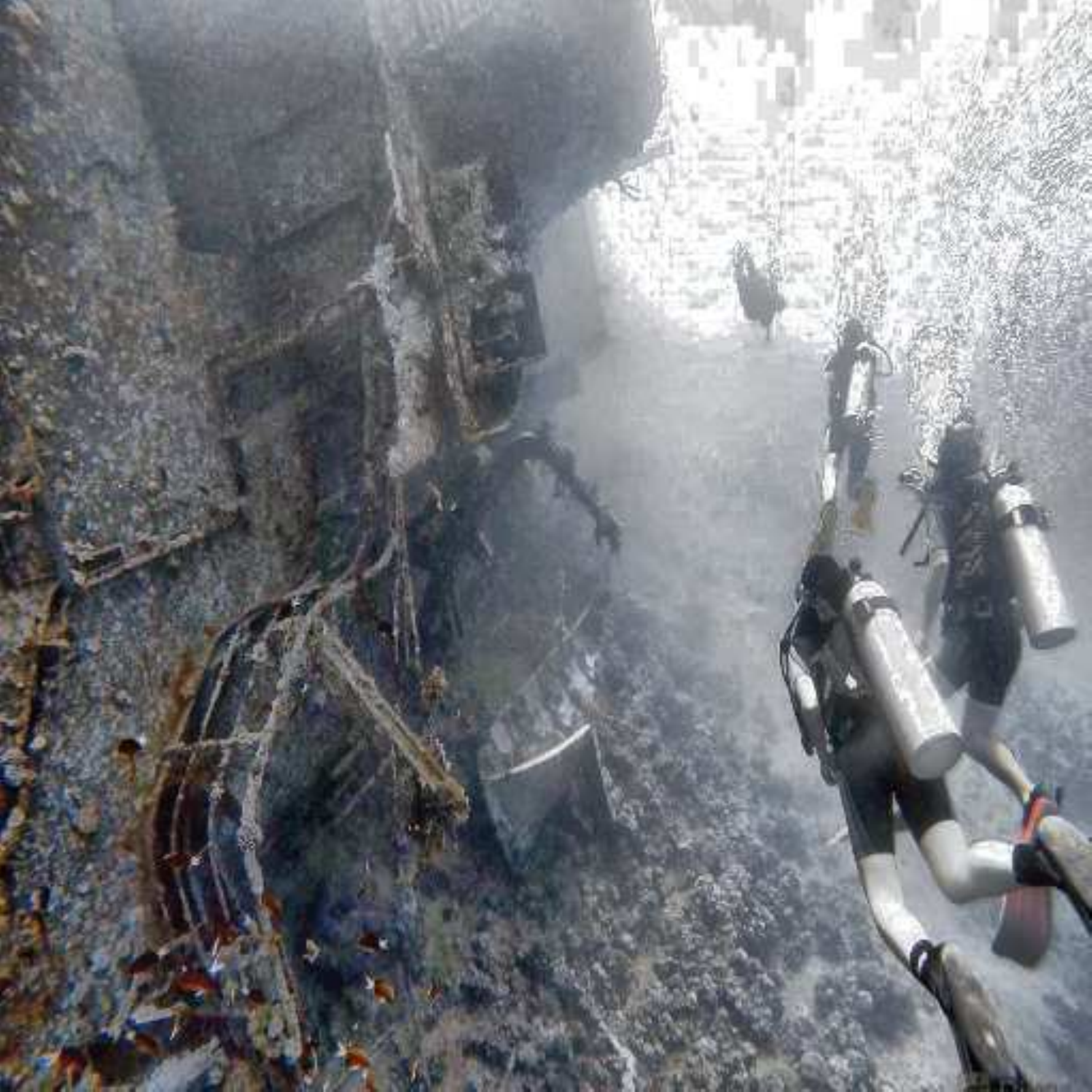}}
\hspace{-0.7mm}
\subfloat{\includegraphics[width=0.095\textwidth]{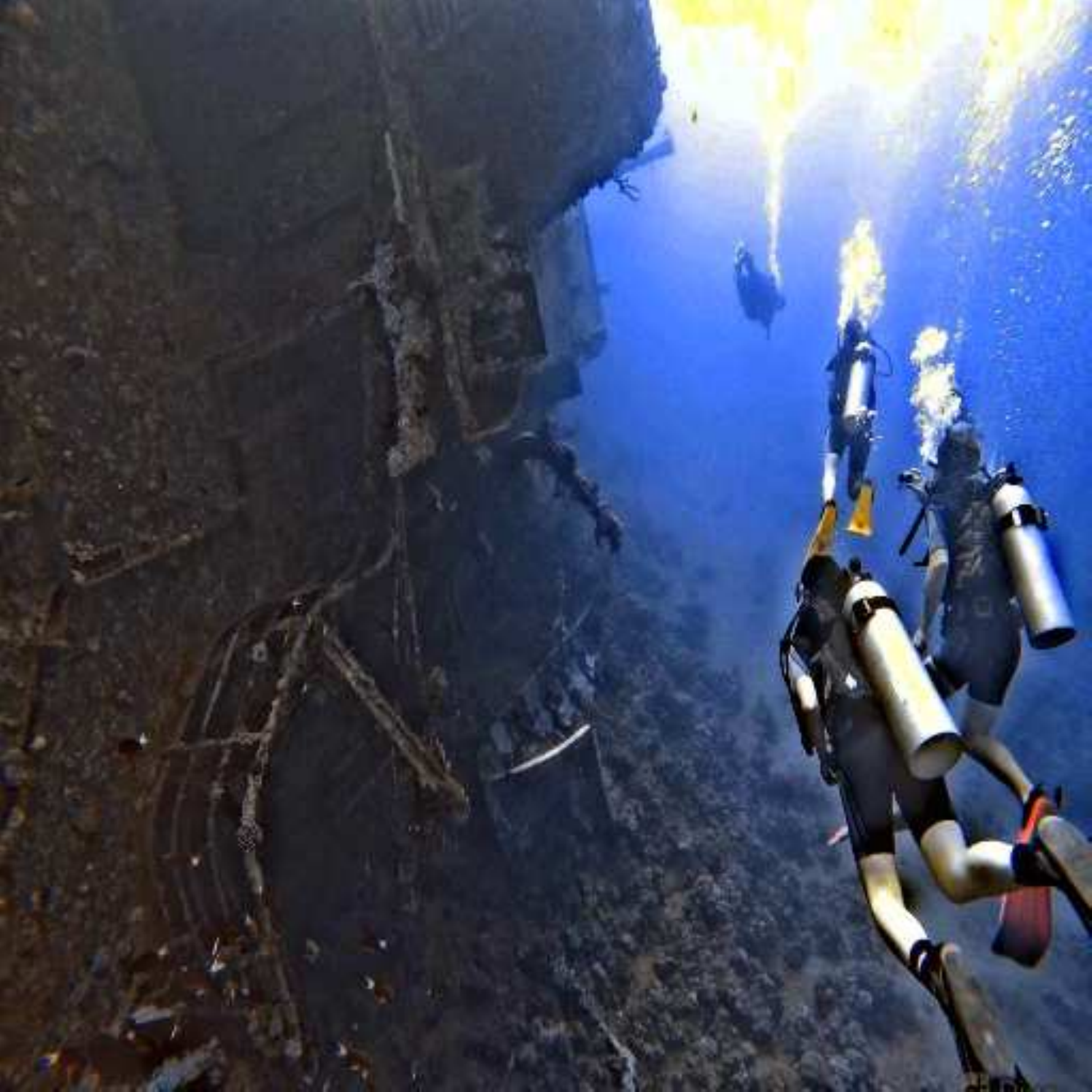}}
\hspace{-0.7mm}
\subfloat{\includegraphics[width=0.095\textwidth]{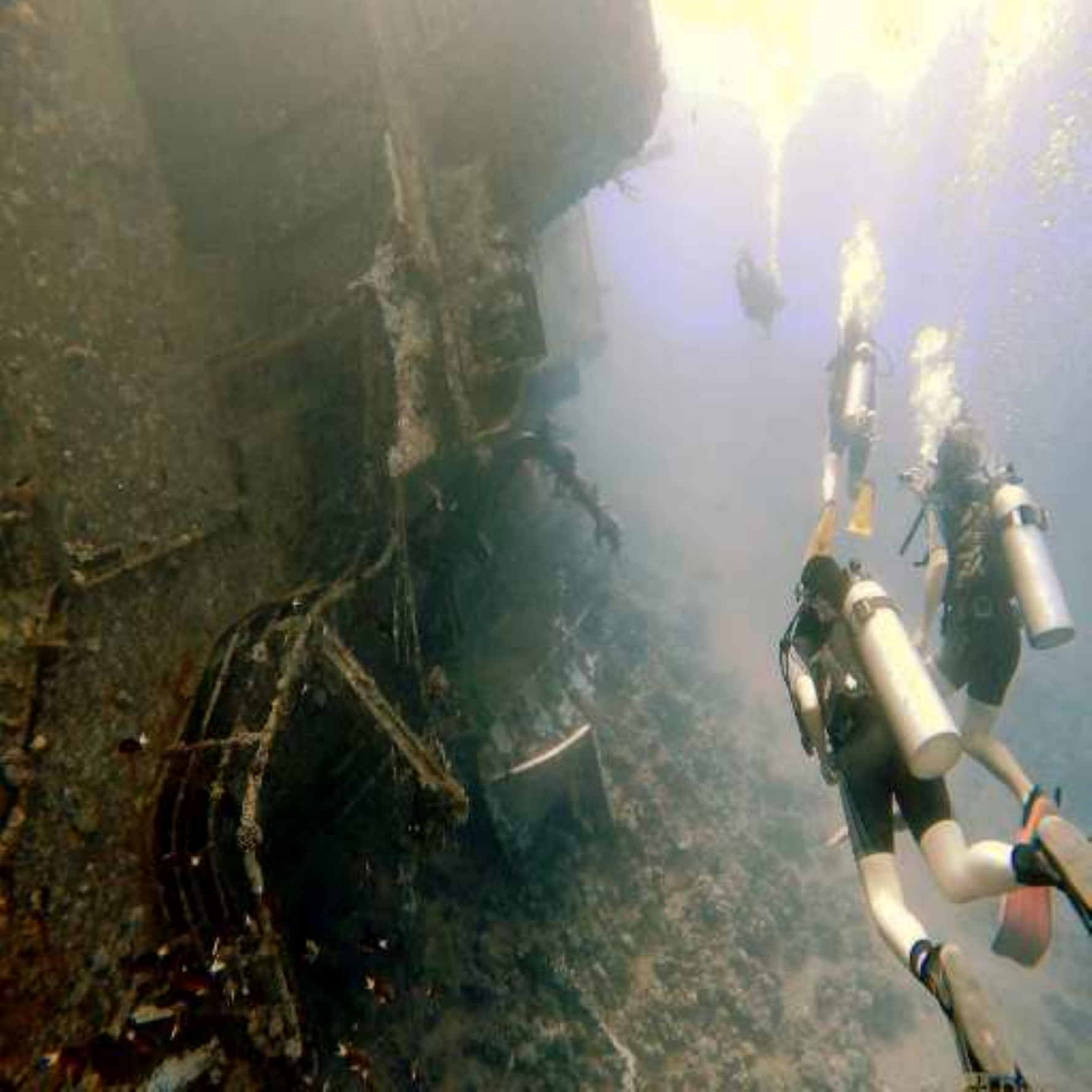}}
\hspace{-0.7mm}
\subfloat{\includegraphics[width=0.095\textwidth]{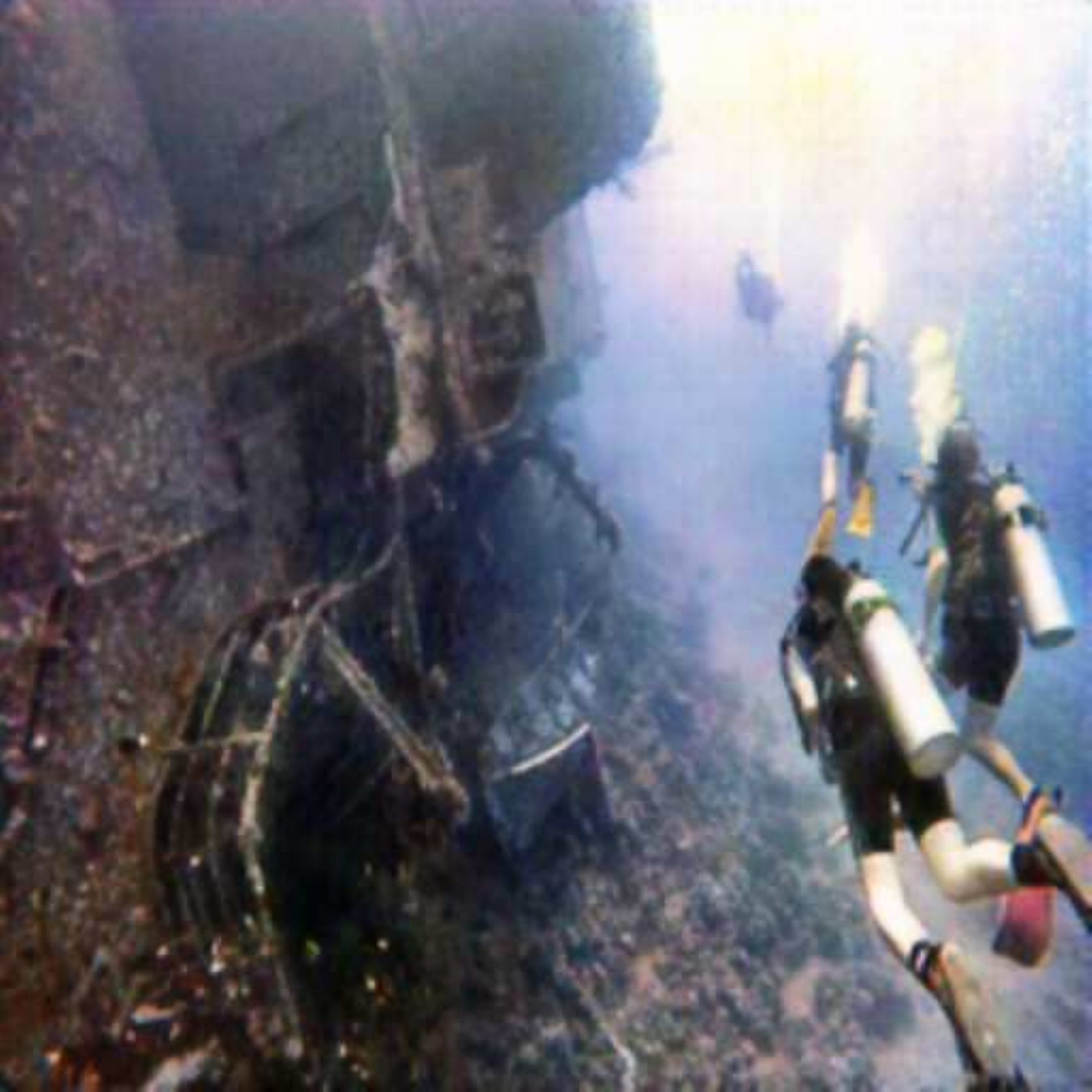}}
\hspace{-0.7mm}
\subfloat{\includegraphics[width=0.095\textwidth]{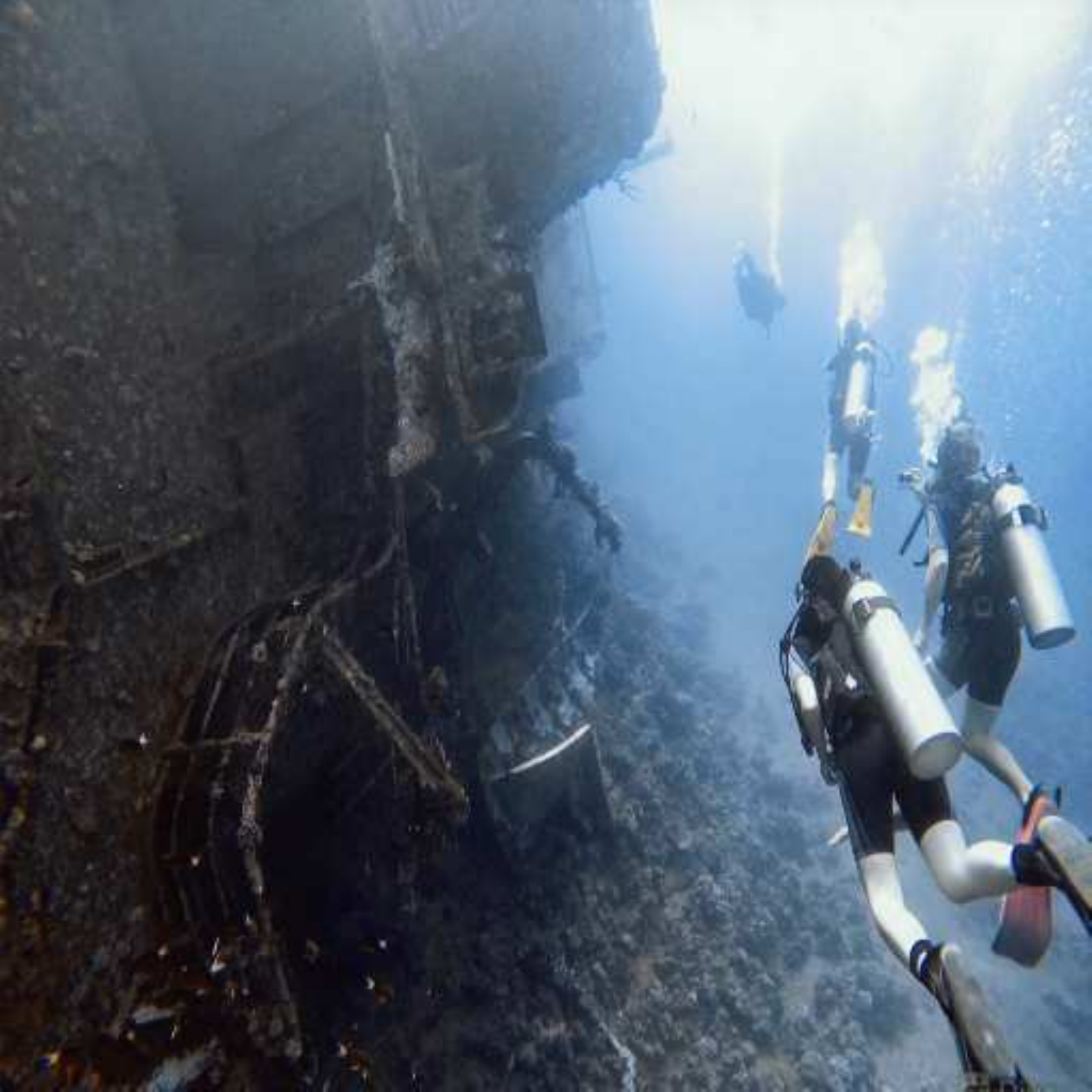}}
\hspace{-0.7mm}
\subfloat{\includegraphics[width=0.095\textwidth]{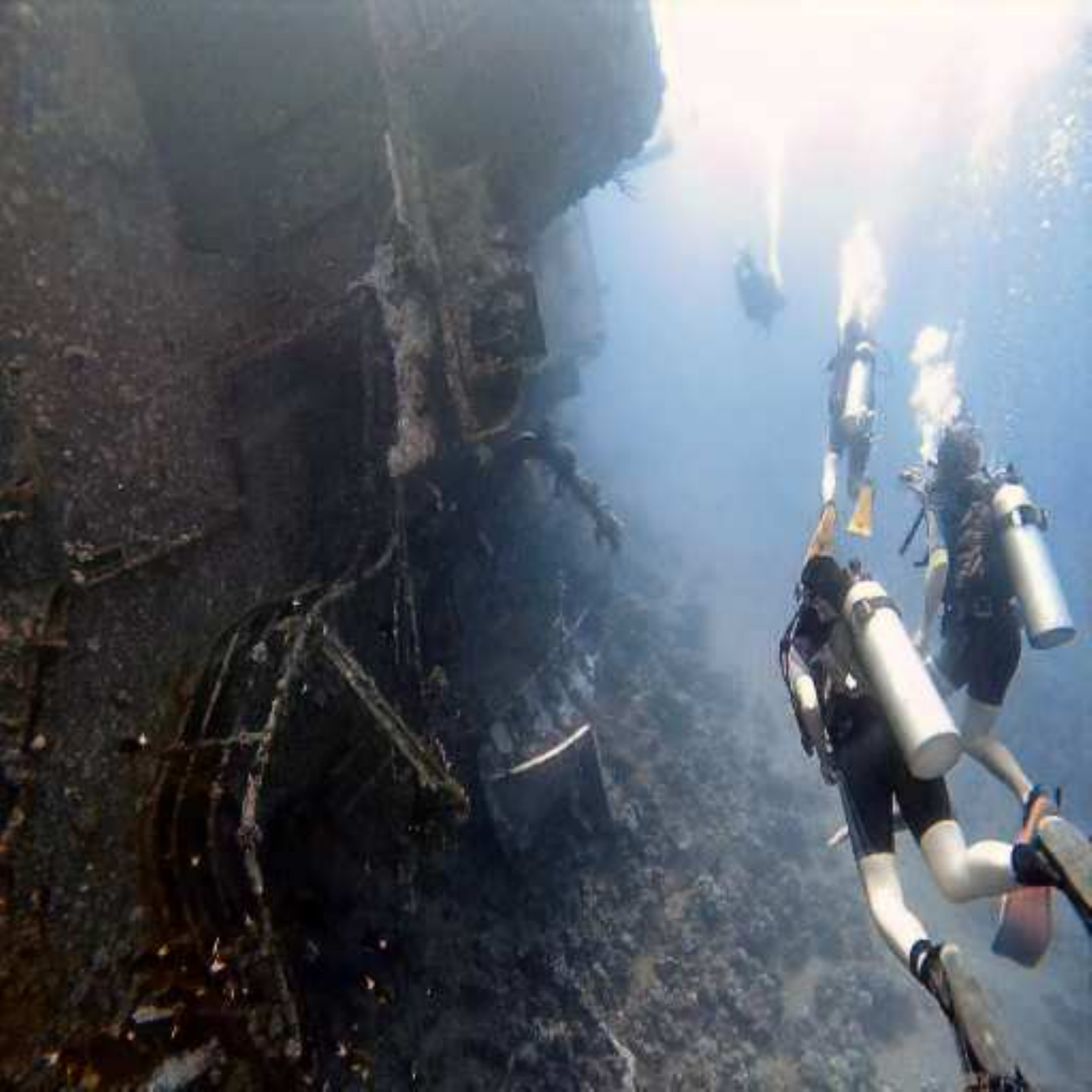}}
\hspace{-0.7mm}
\subfloat{\includegraphics[width=0.095\textwidth]{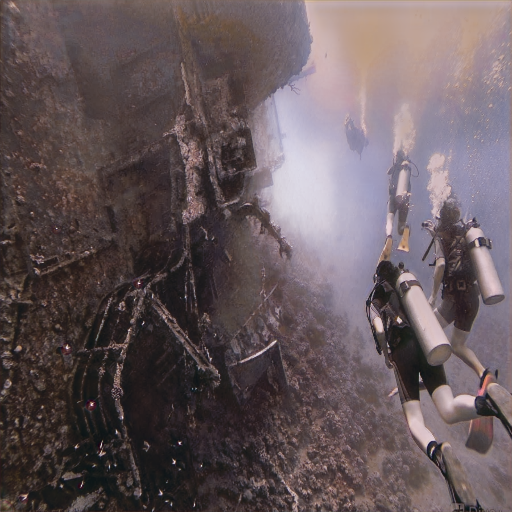}}
\hspace{-0.7mm}
\subfloat{\includegraphics[width=0.095\textwidth]{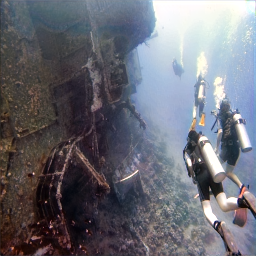}}

\vspace{-3mm}
\subfloat{\includegraphics[width=0.095\textwidth]{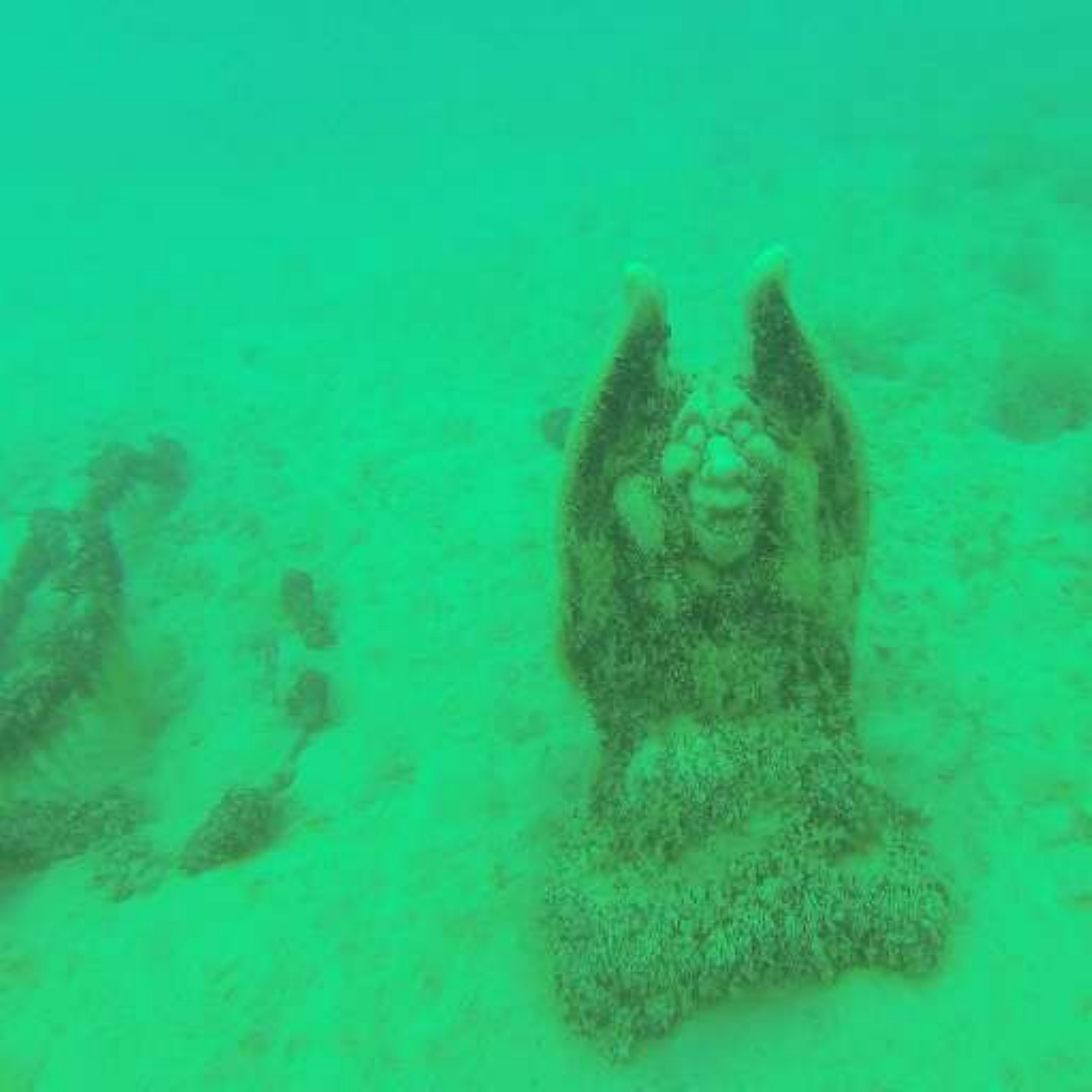}}
\hspace{-0.7mm}
\subfloat{\includegraphics[width=0.095\textwidth]{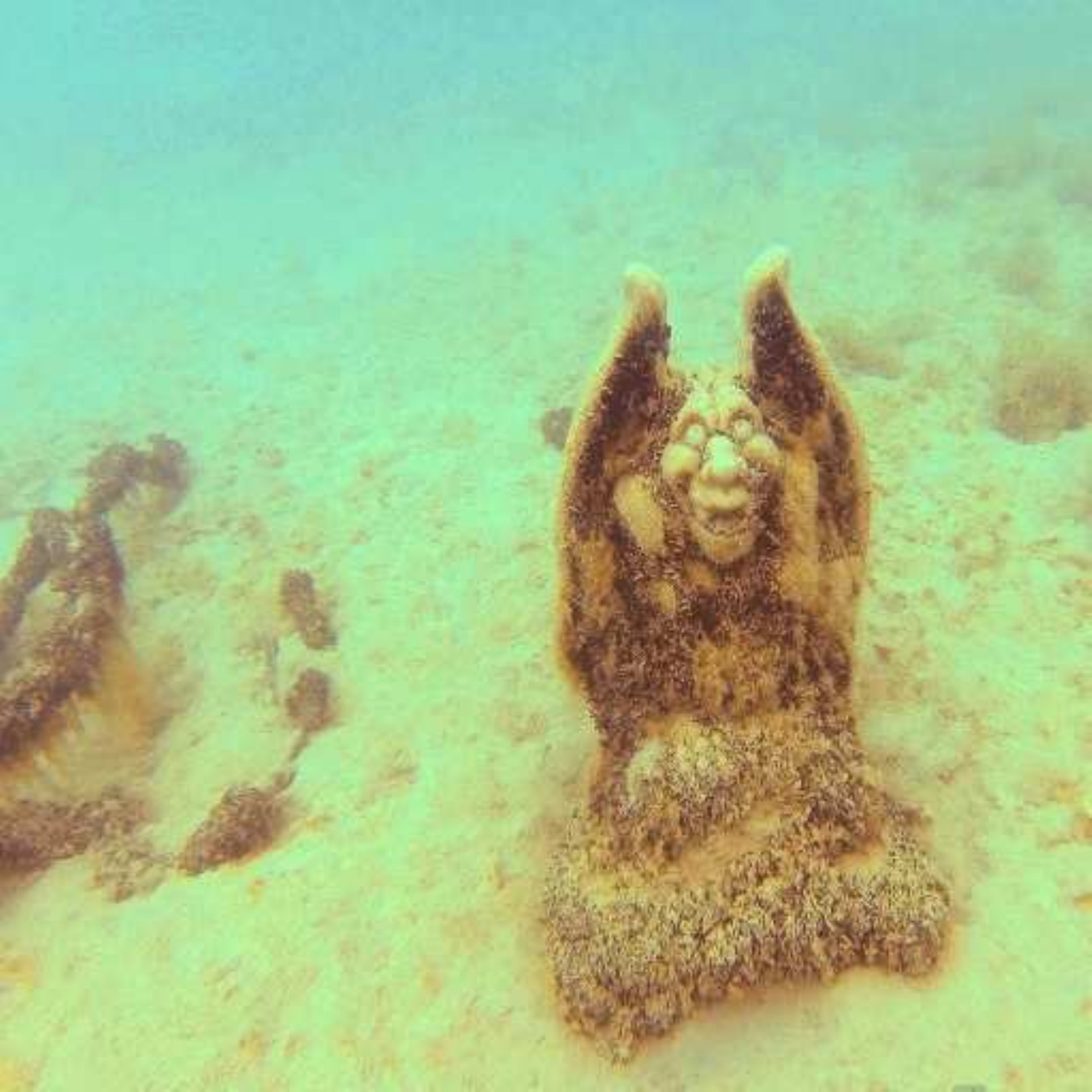}}
\hspace{-0.7mm}
\subfloat{\includegraphics[width=0.095\textwidth]{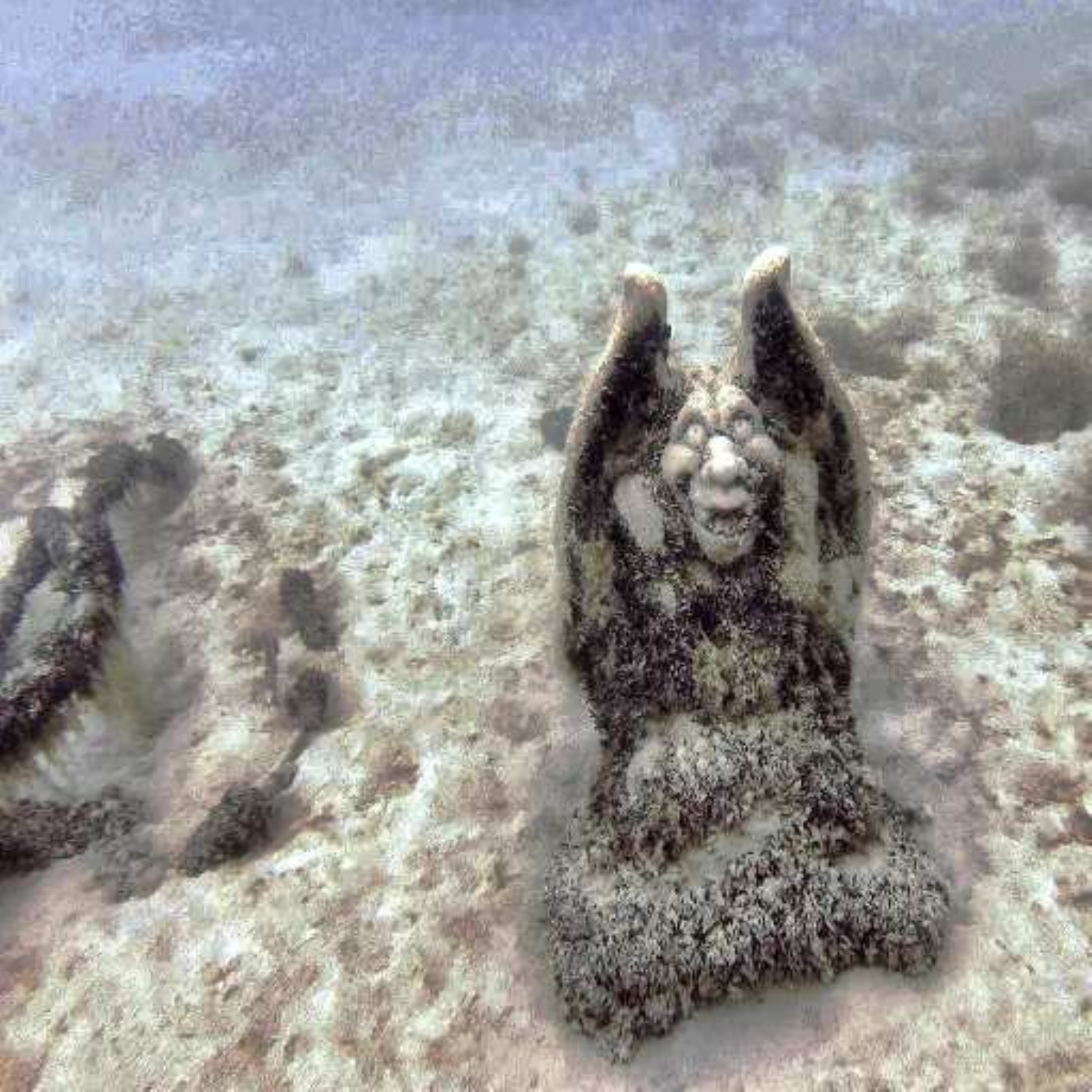}}
\hspace{-0.7mm}
\subfloat{\includegraphics[width=0.095\textwidth]{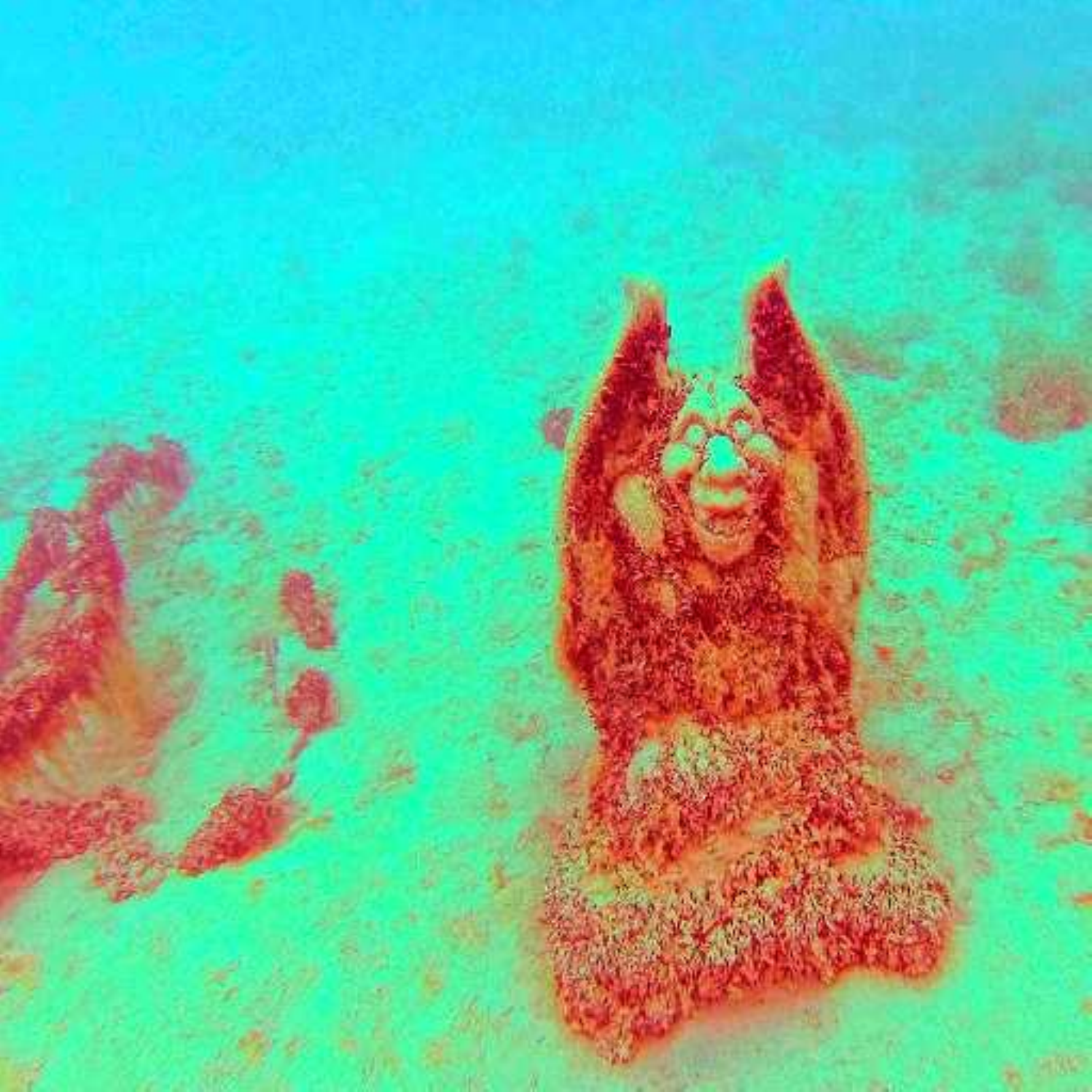}}
\hspace{-0.7mm}
\subfloat{\includegraphics[width=0.095\textwidth]{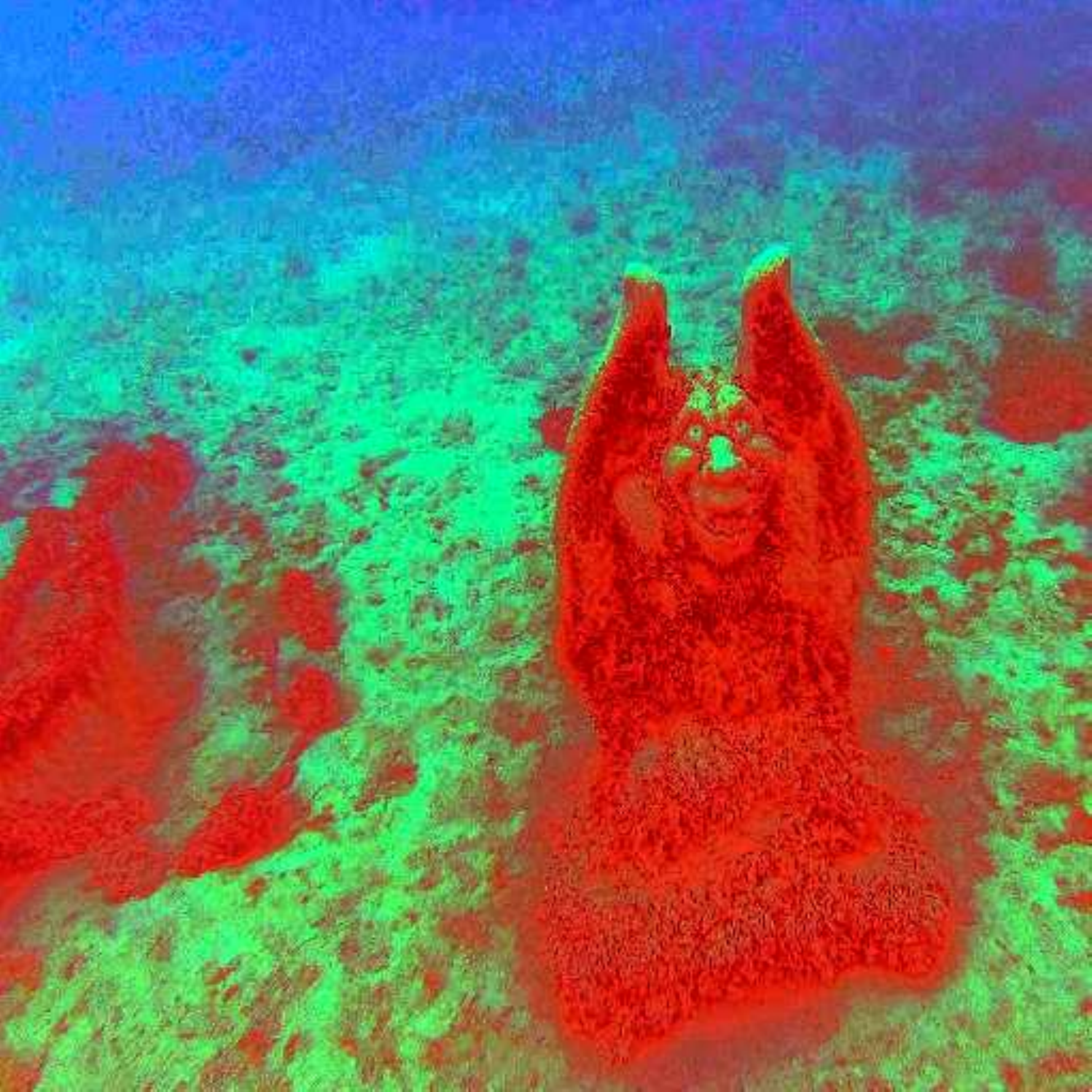}}
\hspace{-0.7mm}
\subfloat{\includegraphics[width=0.095\textwidth]{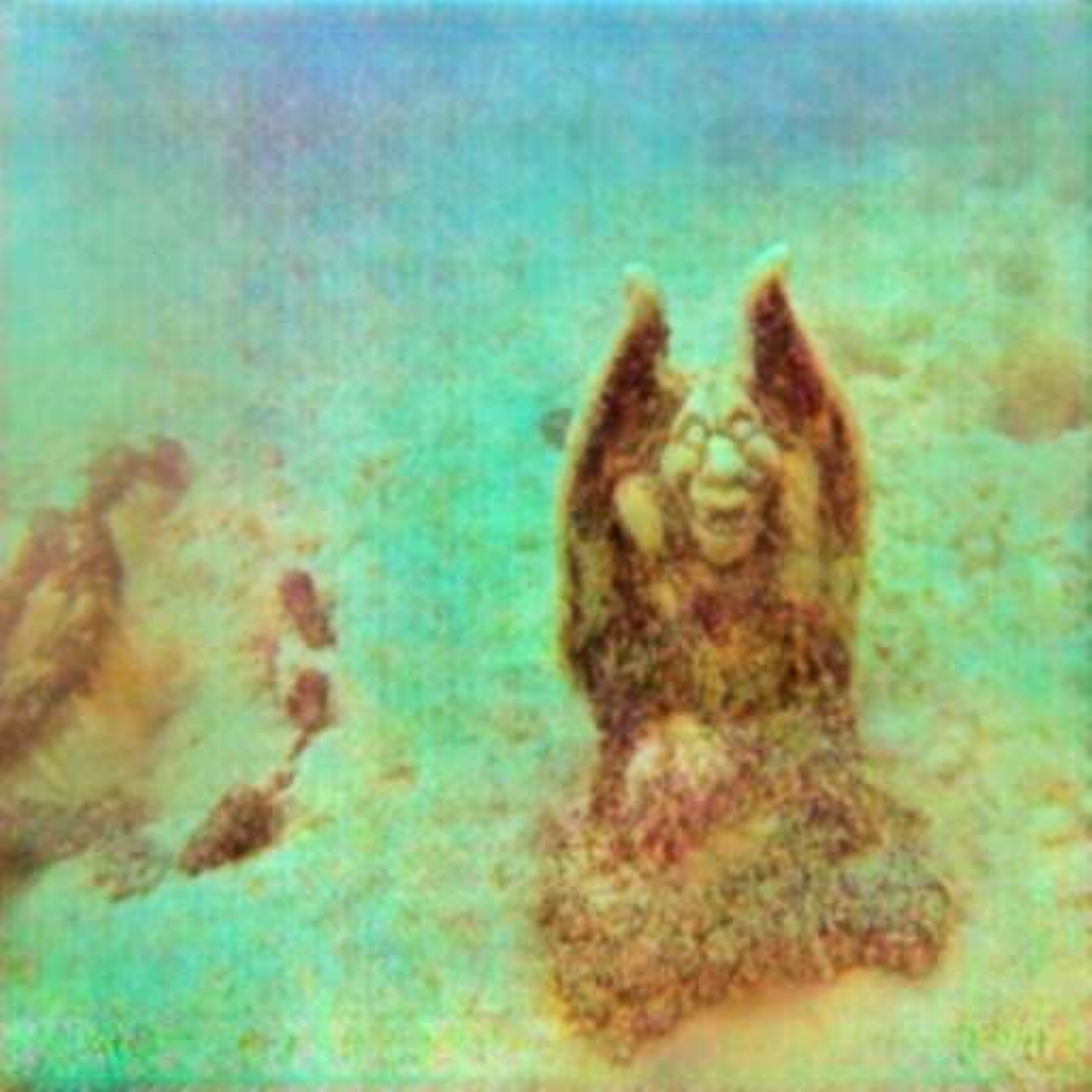}}
\hspace{-0.7mm}
\subfloat{\includegraphics[width=0.095\textwidth]{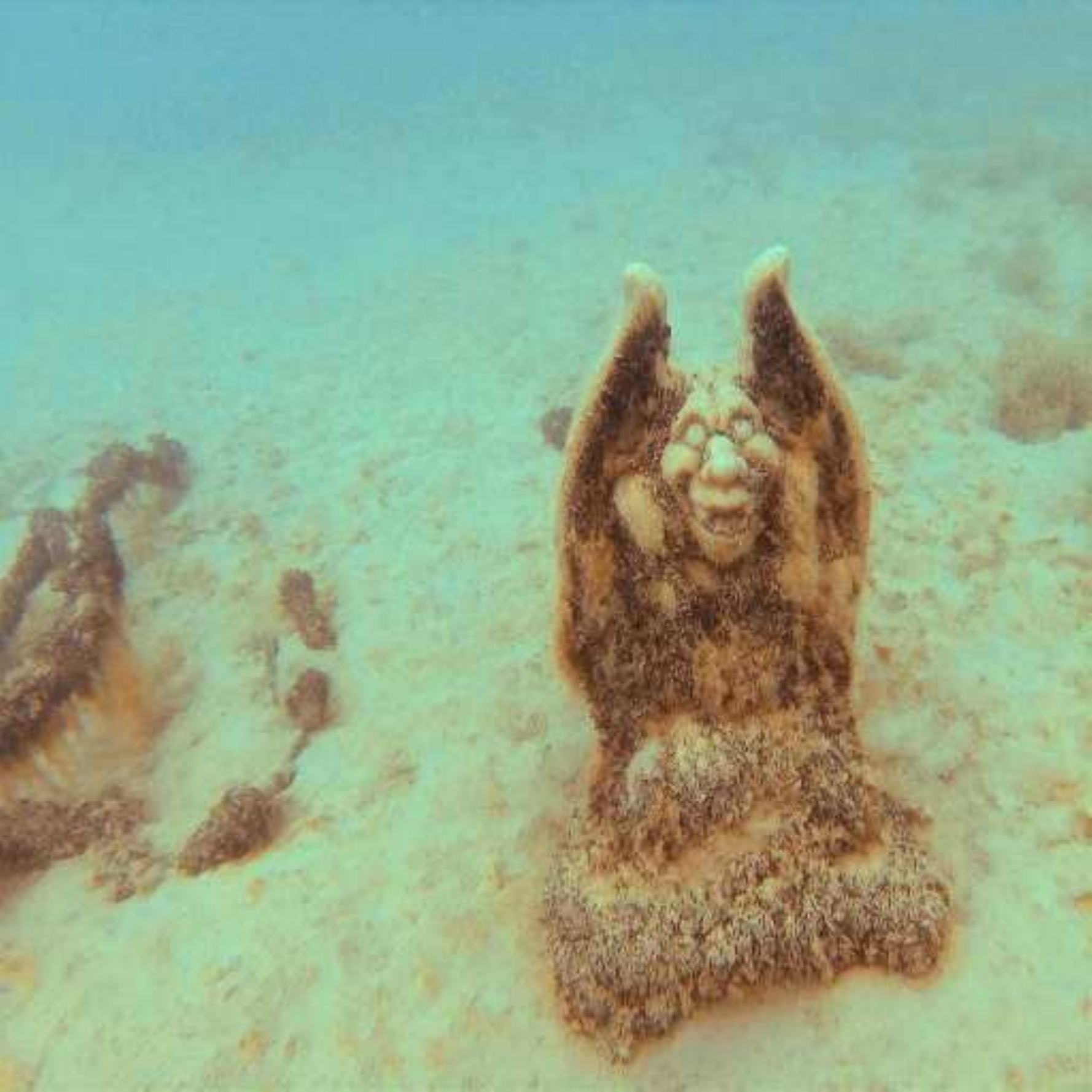}}
\hspace{-0.7mm}
\subfloat{\includegraphics[width=0.095\textwidth]{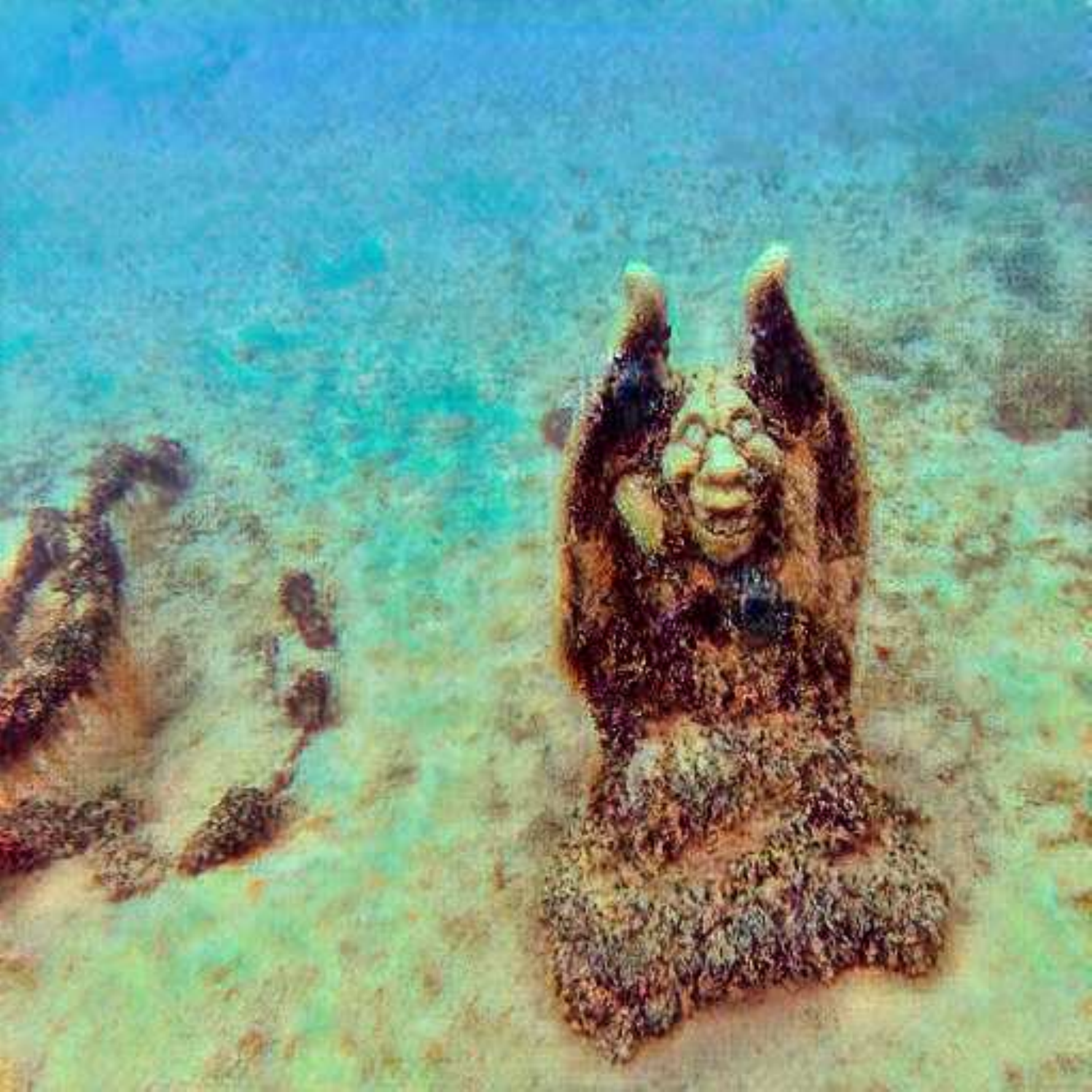}}
\hspace{-0.7mm}
\subfloat{\includegraphics[width=0.095\textwidth]{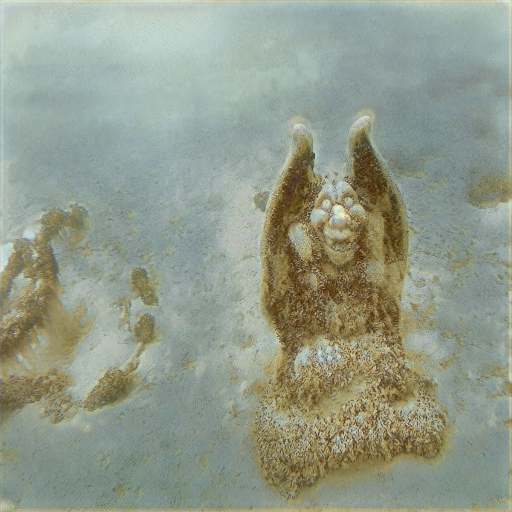}}
\hspace{-0.7mm}
\subfloat{\includegraphics[width=0.095\textwidth]{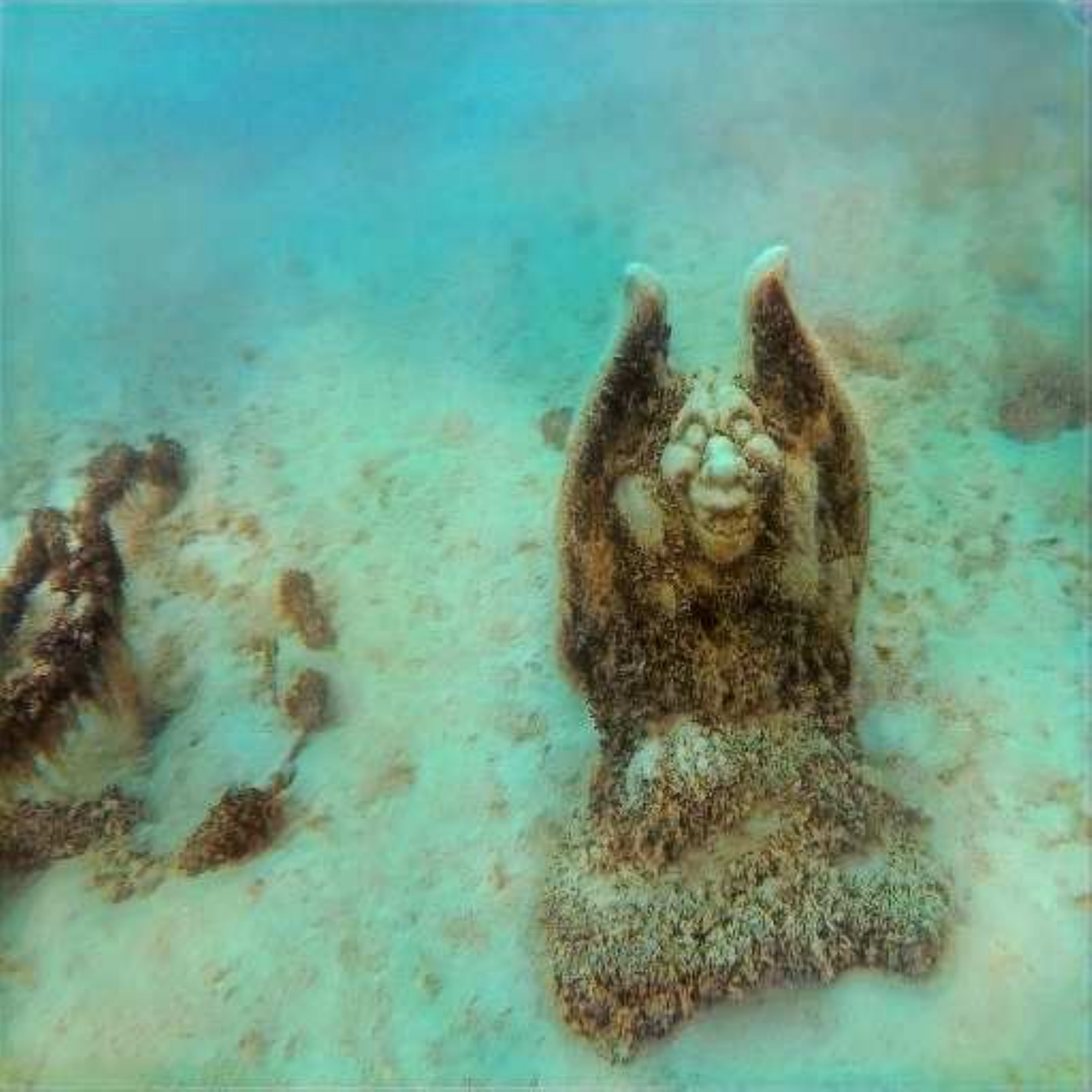}}

\vspace{-3mm}
\subfloat{\includegraphics[width=0.095\textwidth]{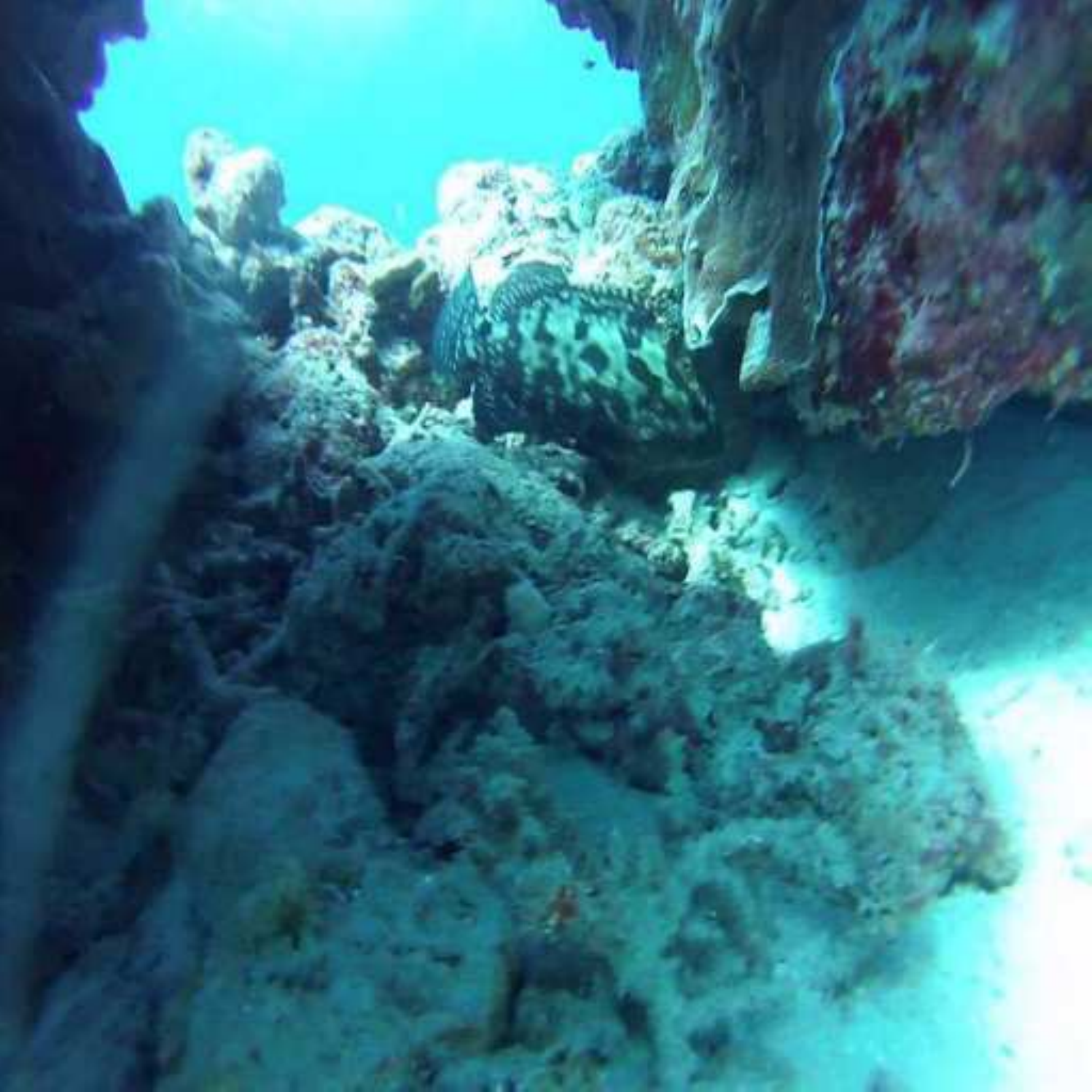}}
\hspace{-0.7mm}
\subfloat{\includegraphics[width=0.095\textwidth]{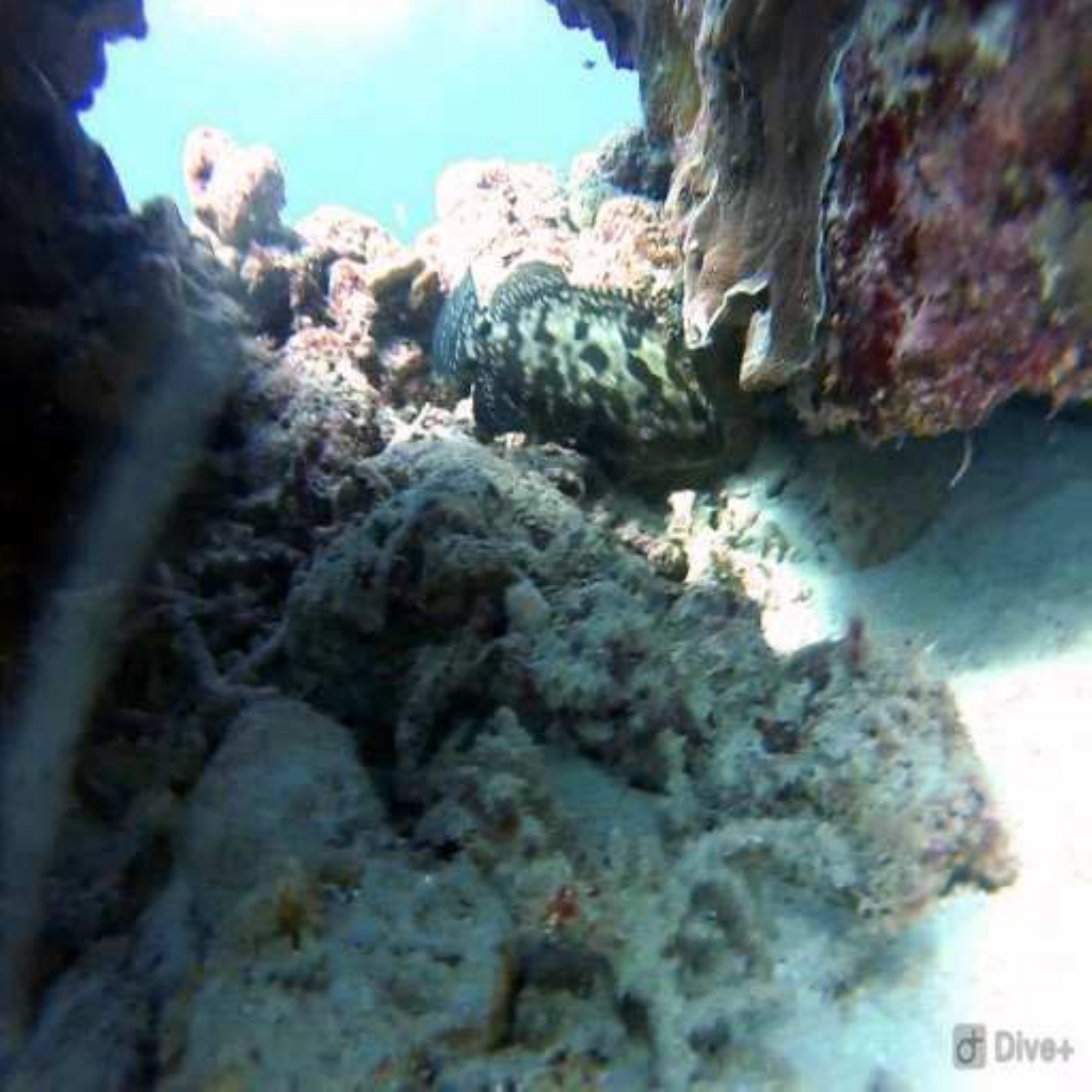}}
\hspace{-0.7mm}
\subfloat{\includegraphics[width=0.095\textwidth]{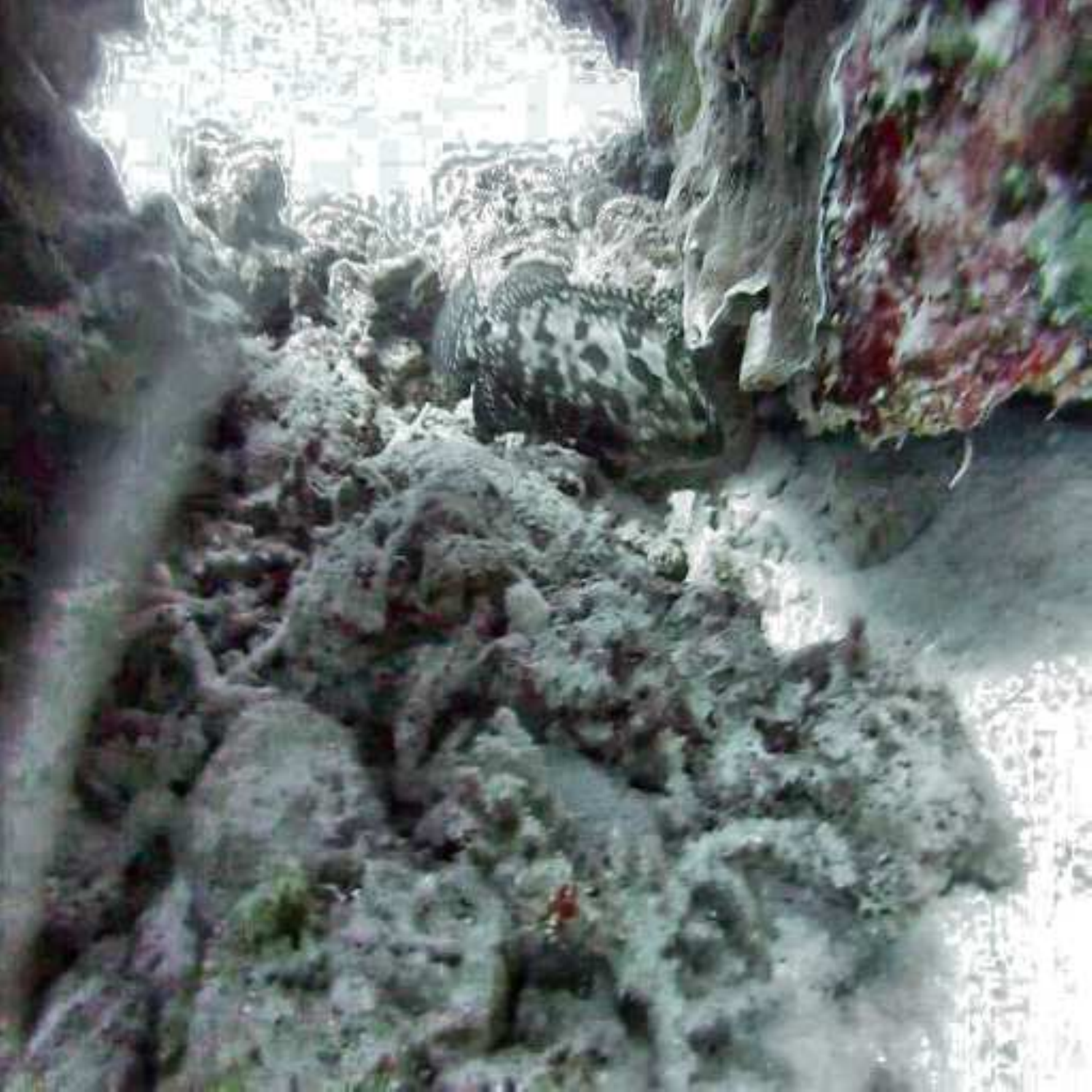}}
\hspace{-0.7mm}
\subfloat{\includegraphics[width=0.095\textwidth]{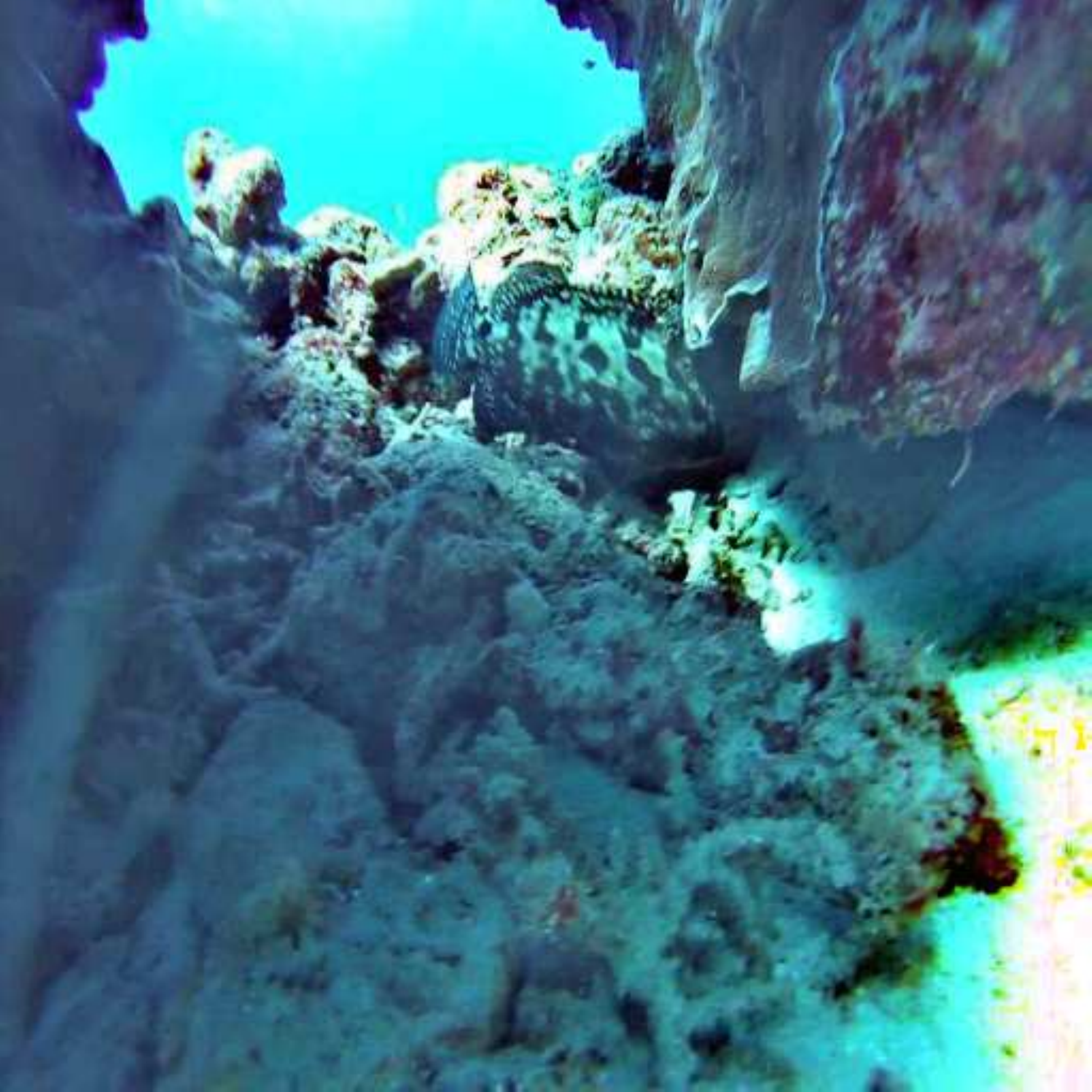}}
\hspace{-0.7mm}
\subfloat{\includegraphics[width=0.095\textwidth]{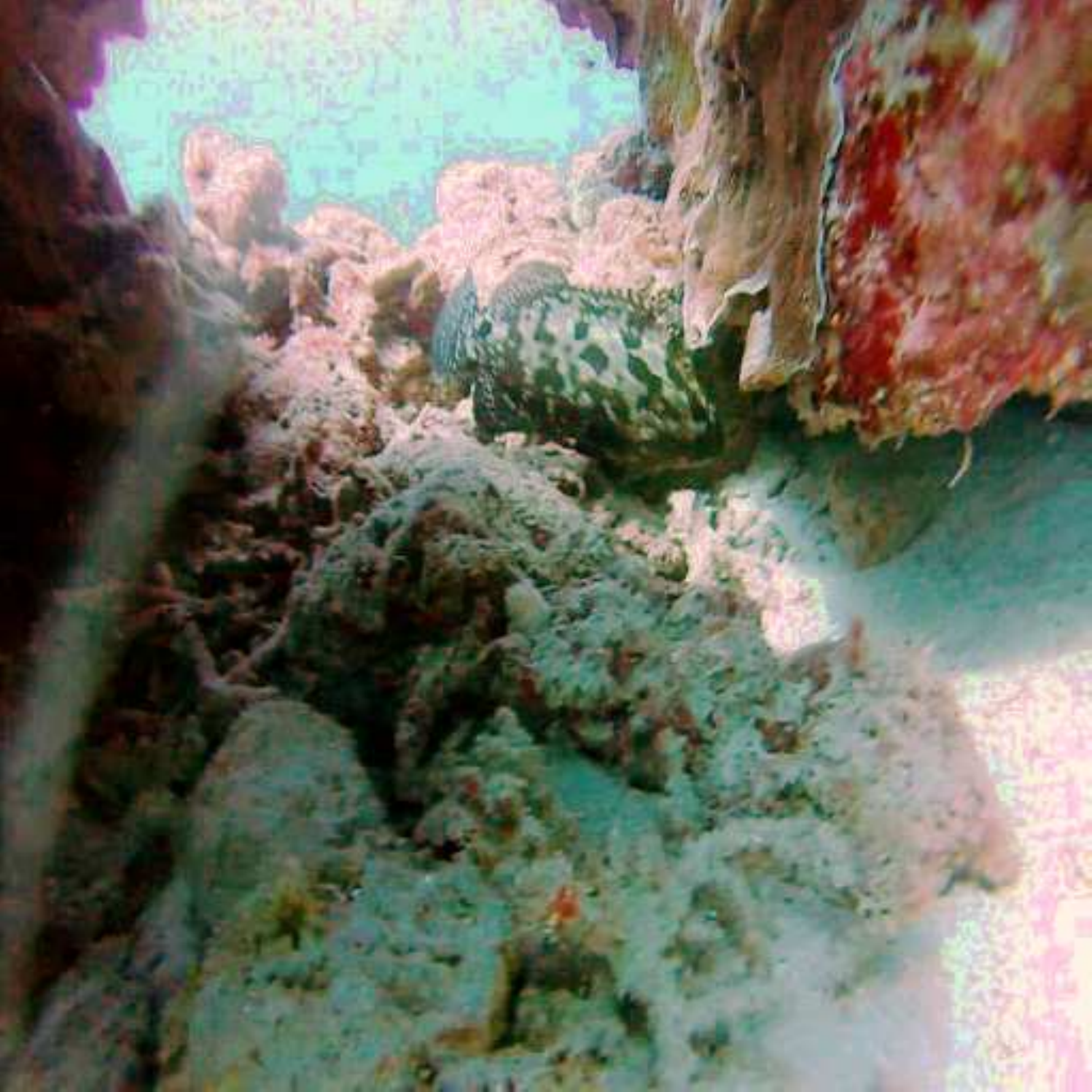}}
\hspace{-0.7mm}
\subfloat{\includegraphics[width=0.095\textwidth]{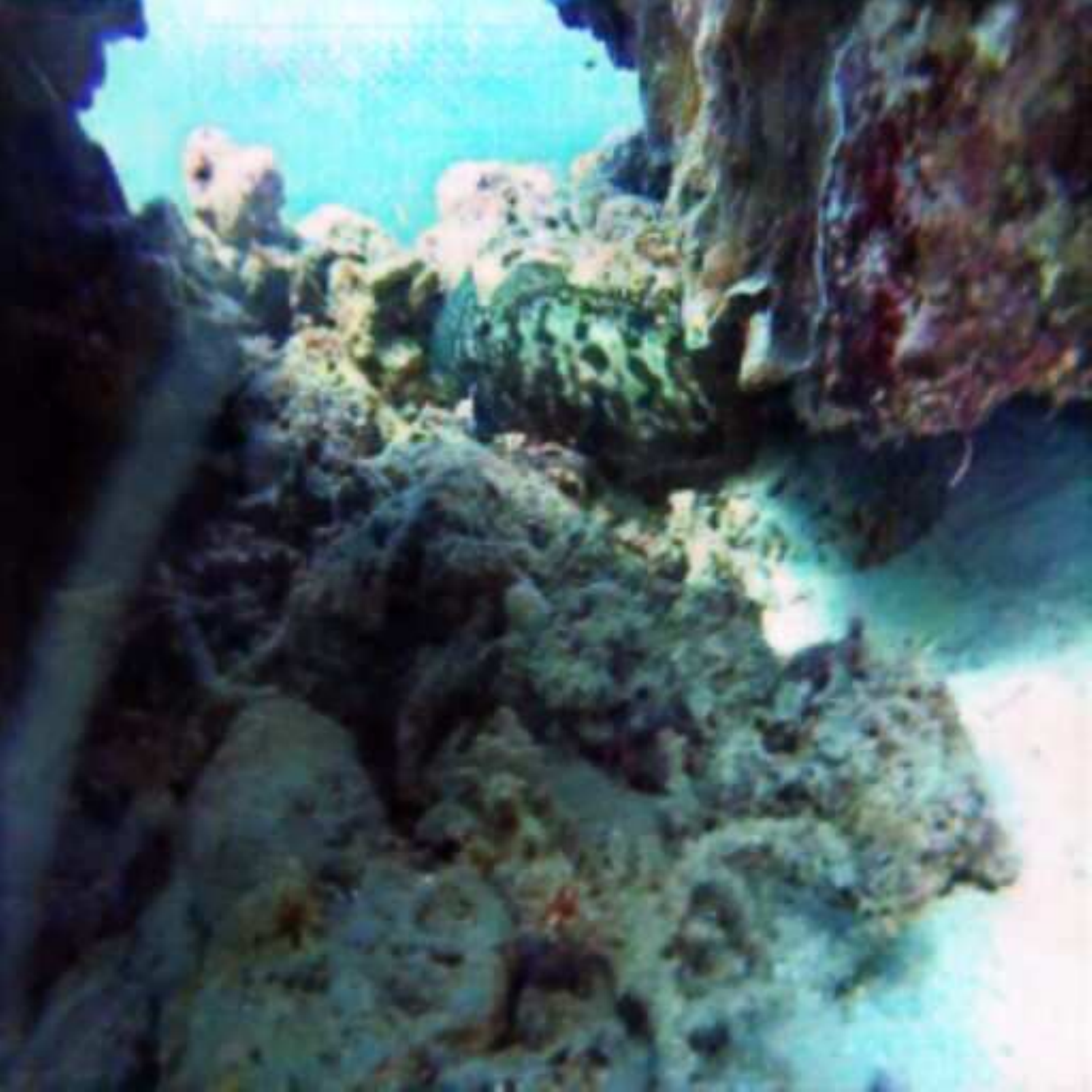}}
\hspace{-0.7mm}
\subfloat{\includegraphics[width=0.095\textwidth]{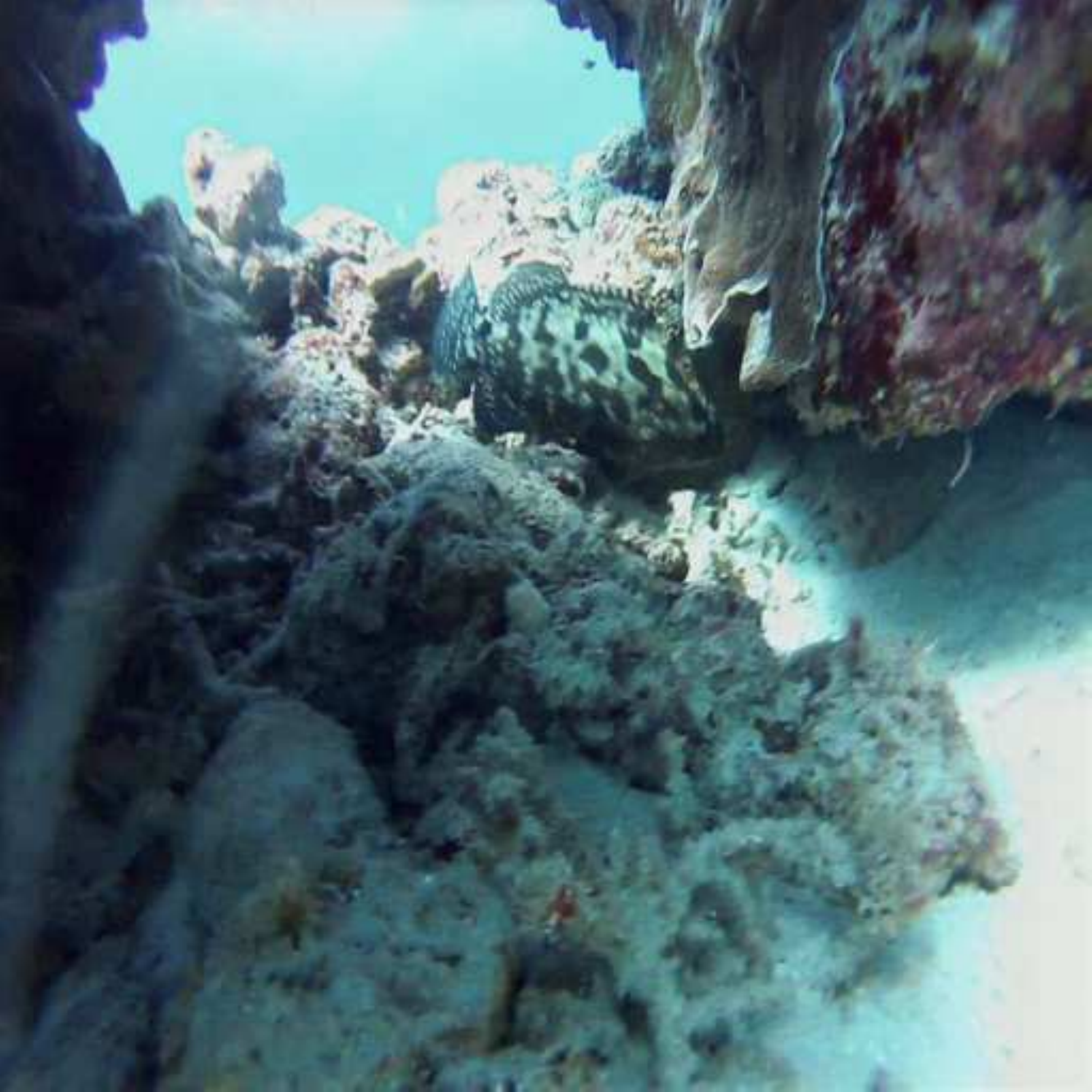}}
\hspace{-0.7mm}
\subfloat{\includegraphics[width=0.095\textwidth]{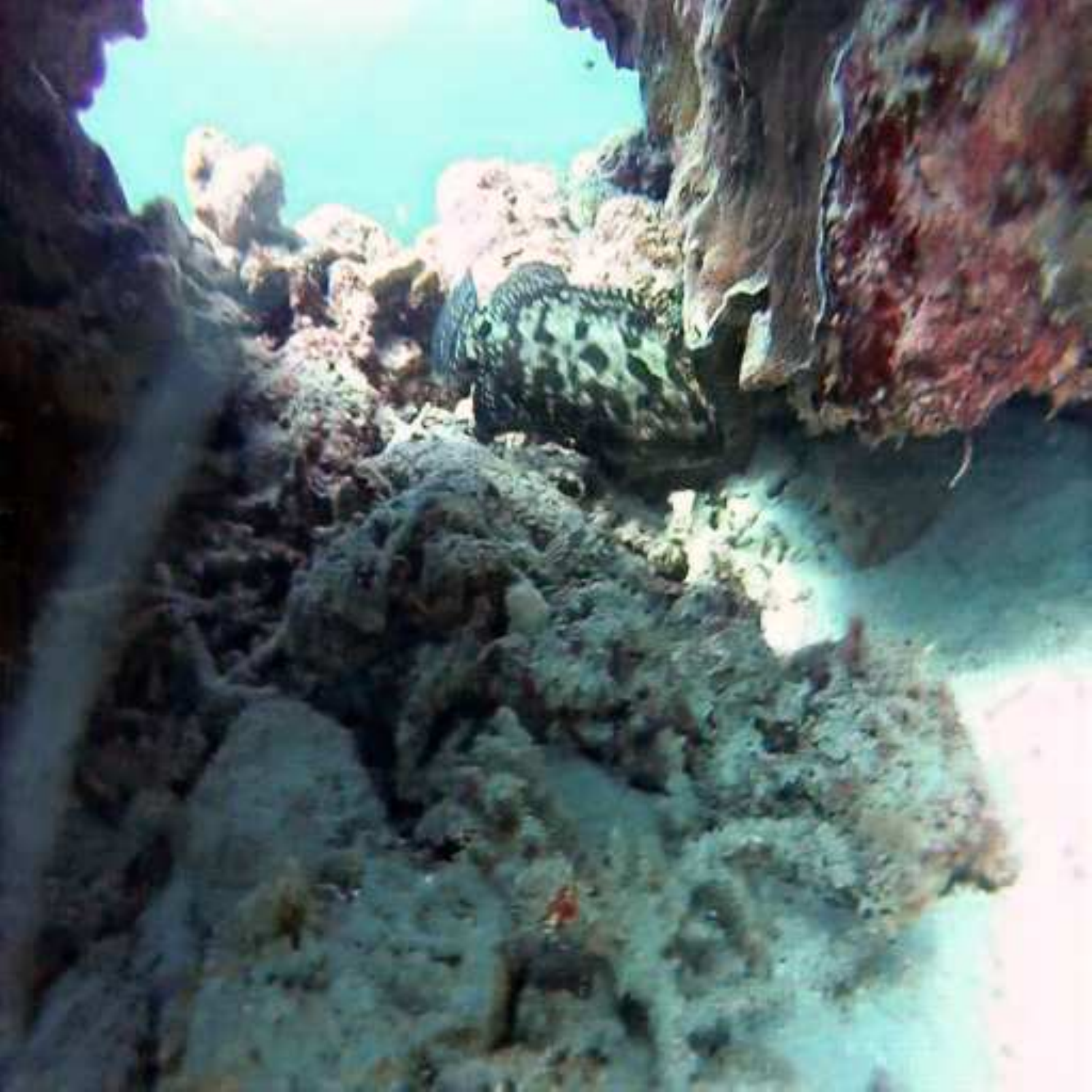}}
\hspace{-0.7mm}
\subfloat{\includegraphics[width=0.095\textwidth]{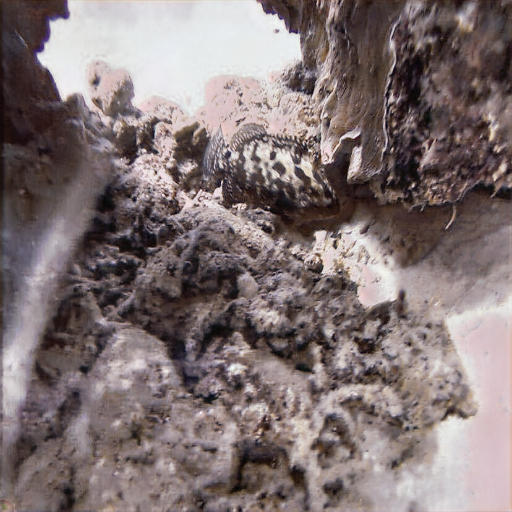}}
\hspace{-0.7mm}
\subfloat{\includegraphics[width=0.095\textwidth]{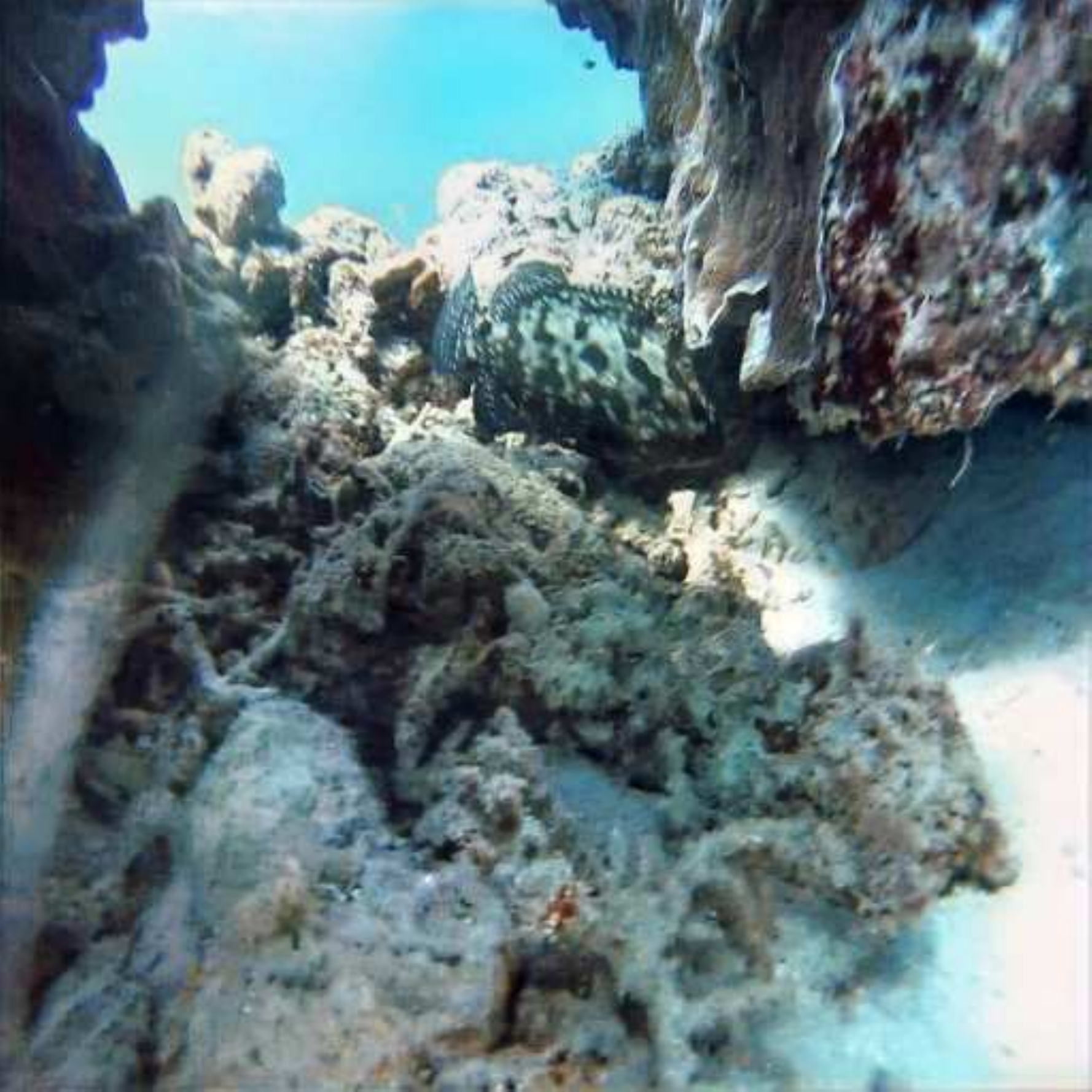}}

\vspace{-3mm}
\setcounter{subfigure}{0}
\subfloat[\footnotesize{Input}]{\includegraphics[width=0.095\textwidth]{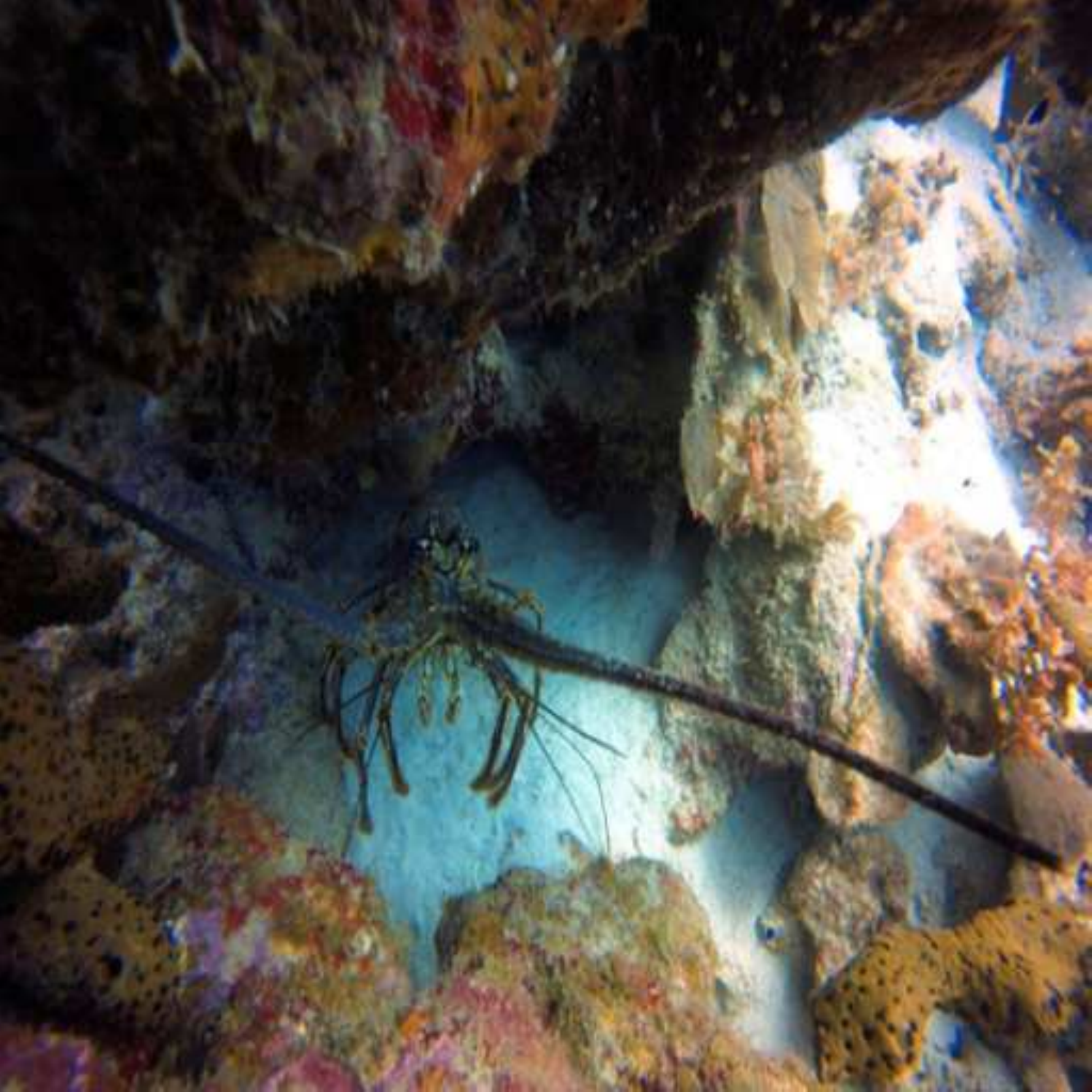}}
\hspace{-0.7mm}
\subfloat[\scriptsize{Reference}]{\includegraphics[width=0.095\textwidth]{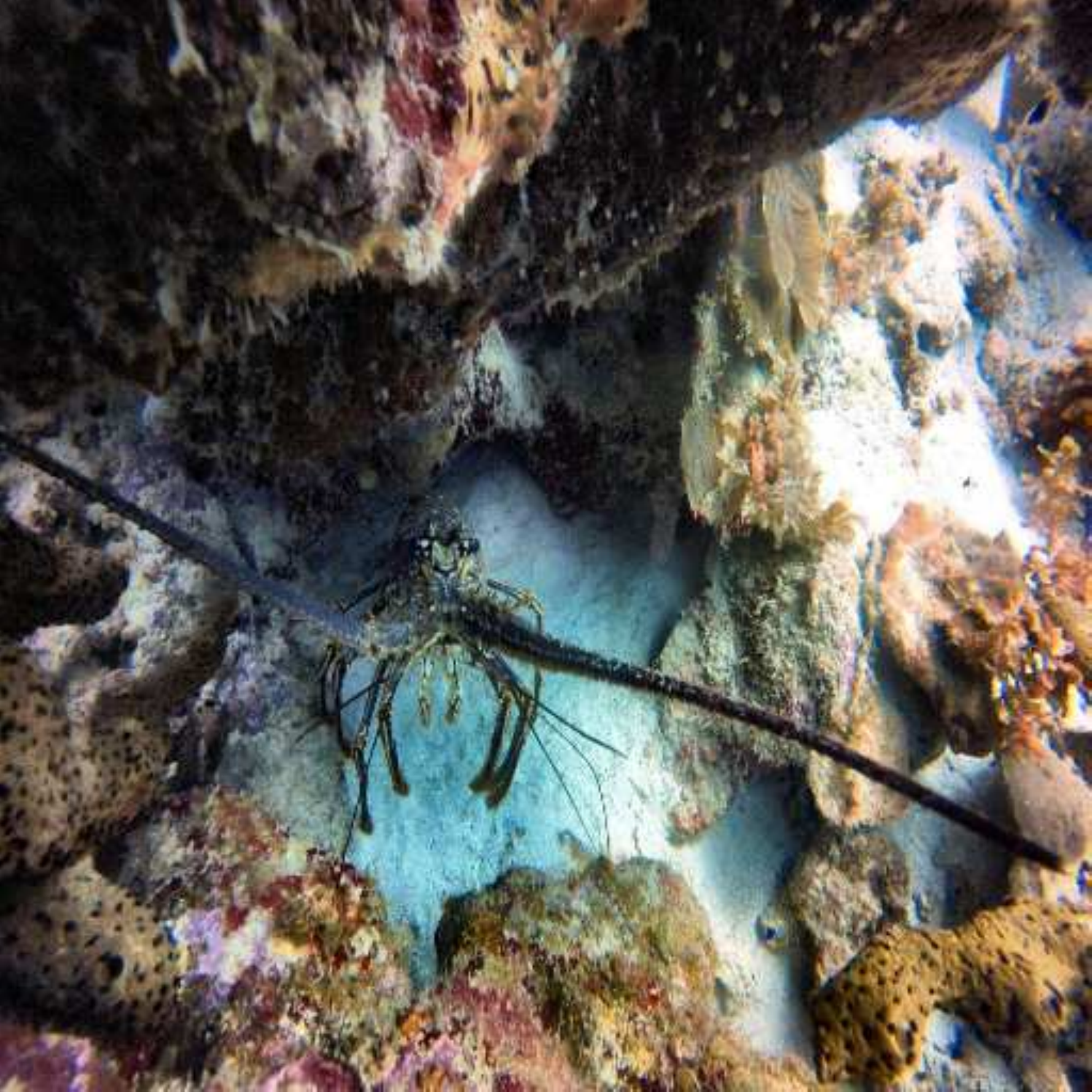}}
\hspace{-0.7mm}
\subfloat[\footnotesize{ACDC}]{\includegraphics[width=0.095\textwidth]{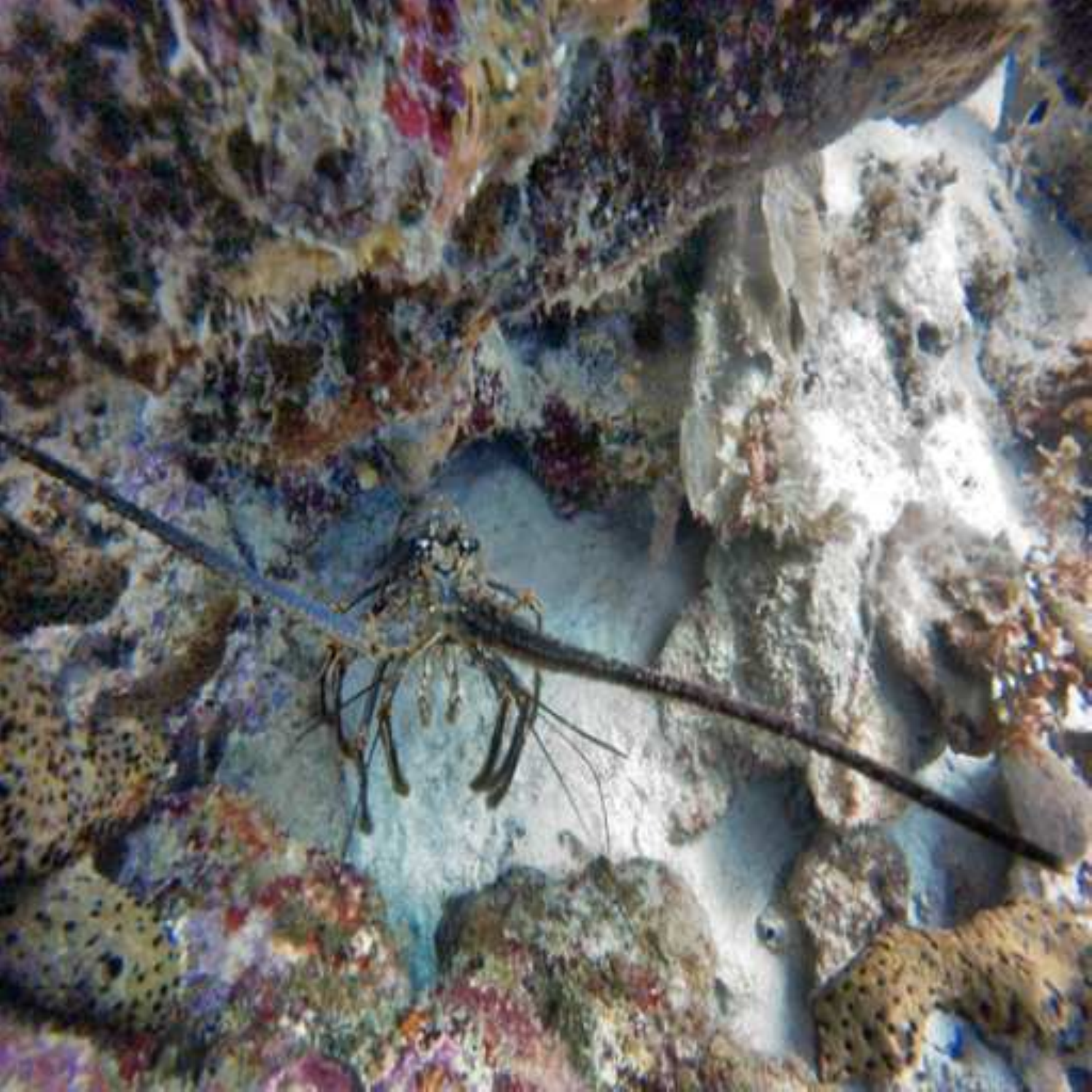}}
\hspace{-0.7mm}
\subfloat[\footnotesize{Rank1}]{\includegraphics[width=0.095\textwidth]{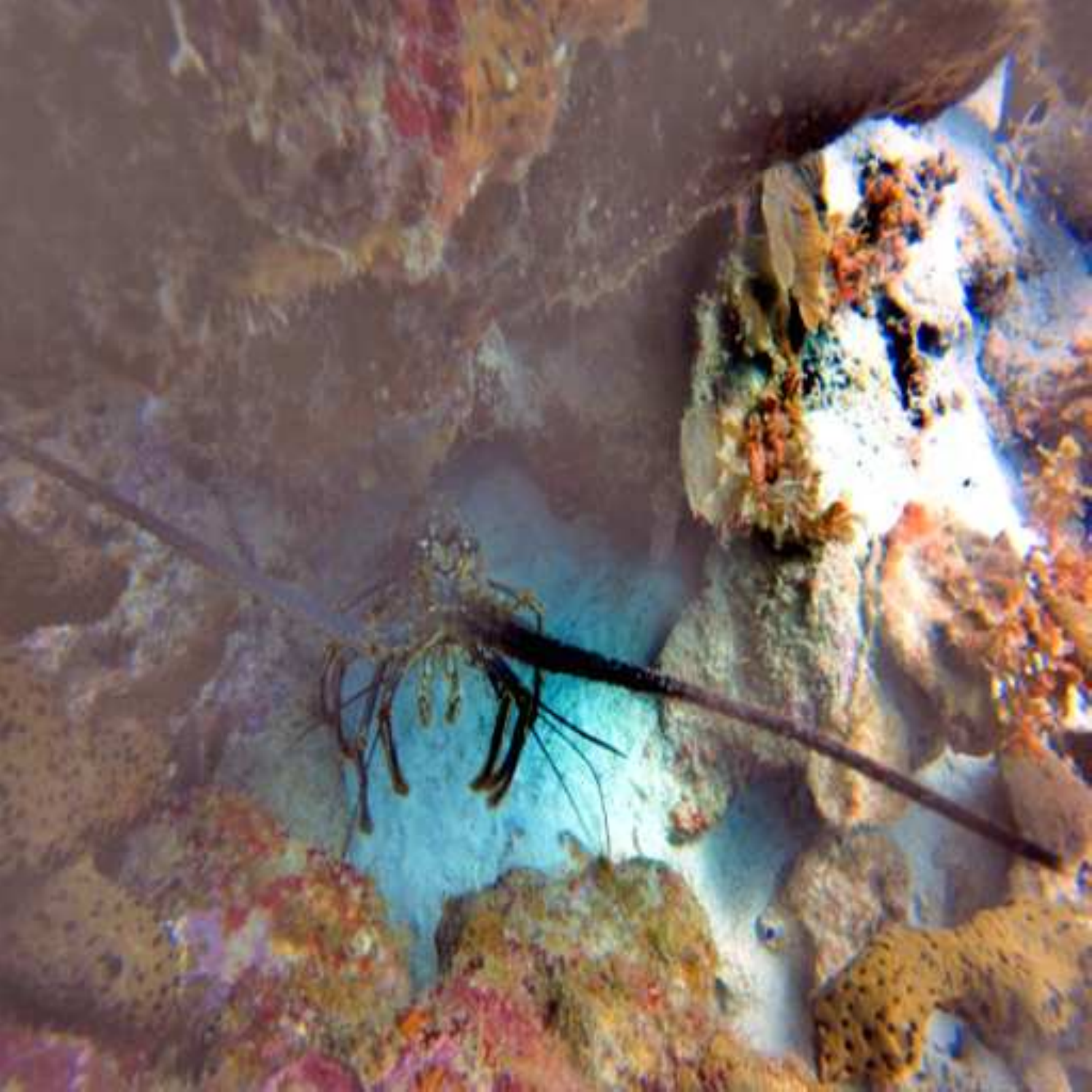}}
\hspace{-0.7mm}
\subfloat[\footnotesize{MMLE}]{\includegraphics[width=0.095\textwidth]{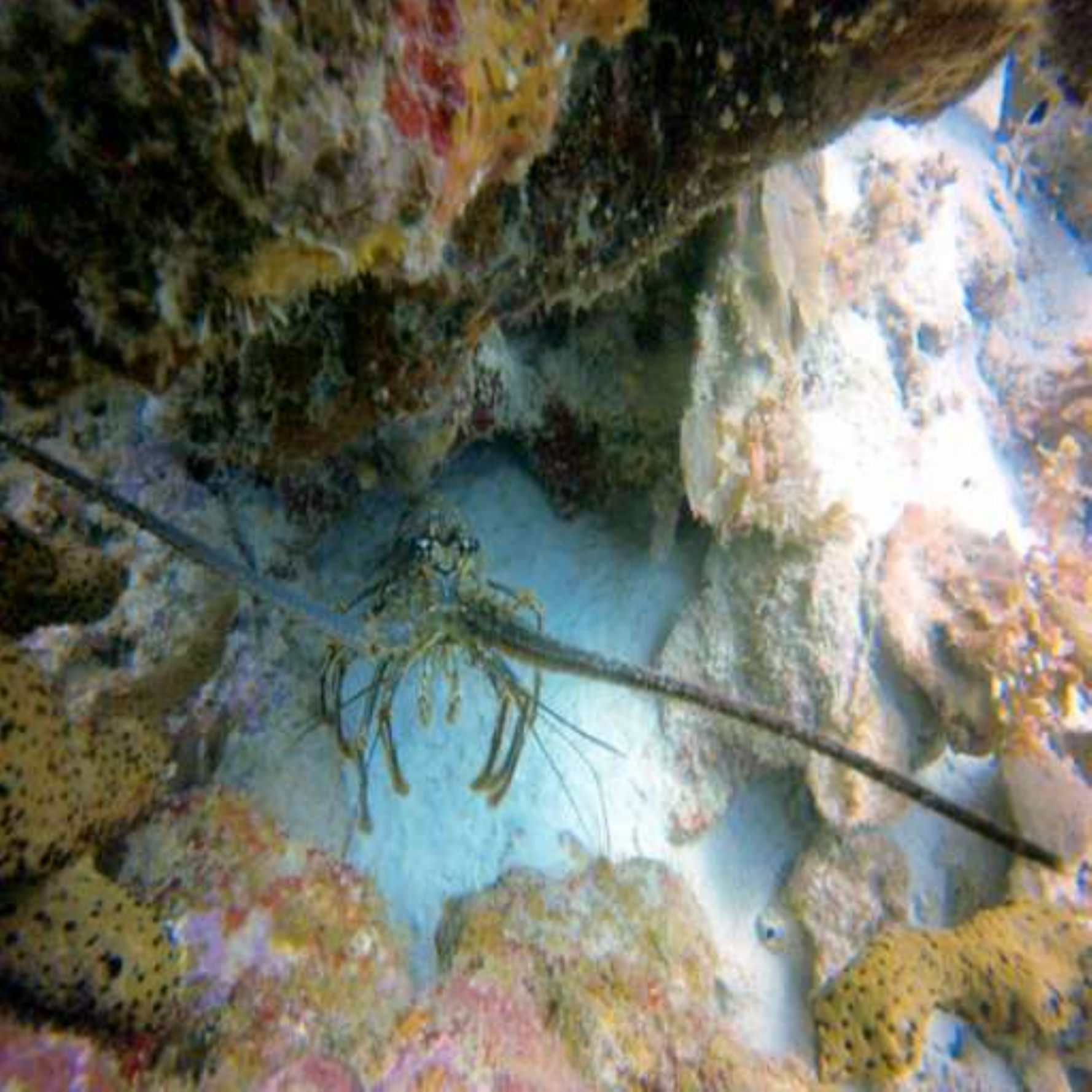}}
\hspace{-0.7mm}
\subfloat[\footnotesize{Funie}]{\includegraphics[width=0.095\textwidth]{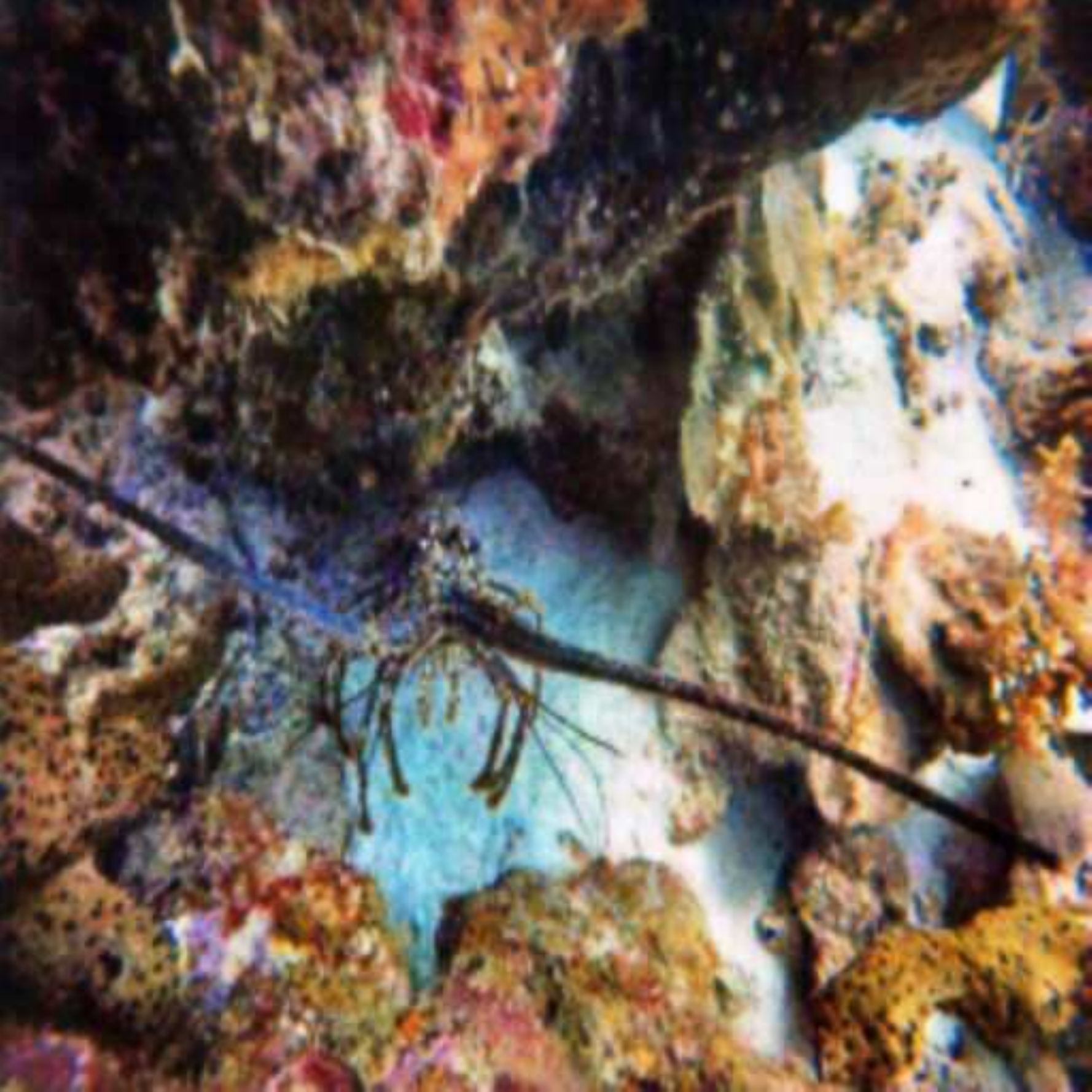}}
\hspace{-0.7mm}
\subfloat[\footnotesize{Ucolor}]{\includegraphics[width=0.095\textwidth]{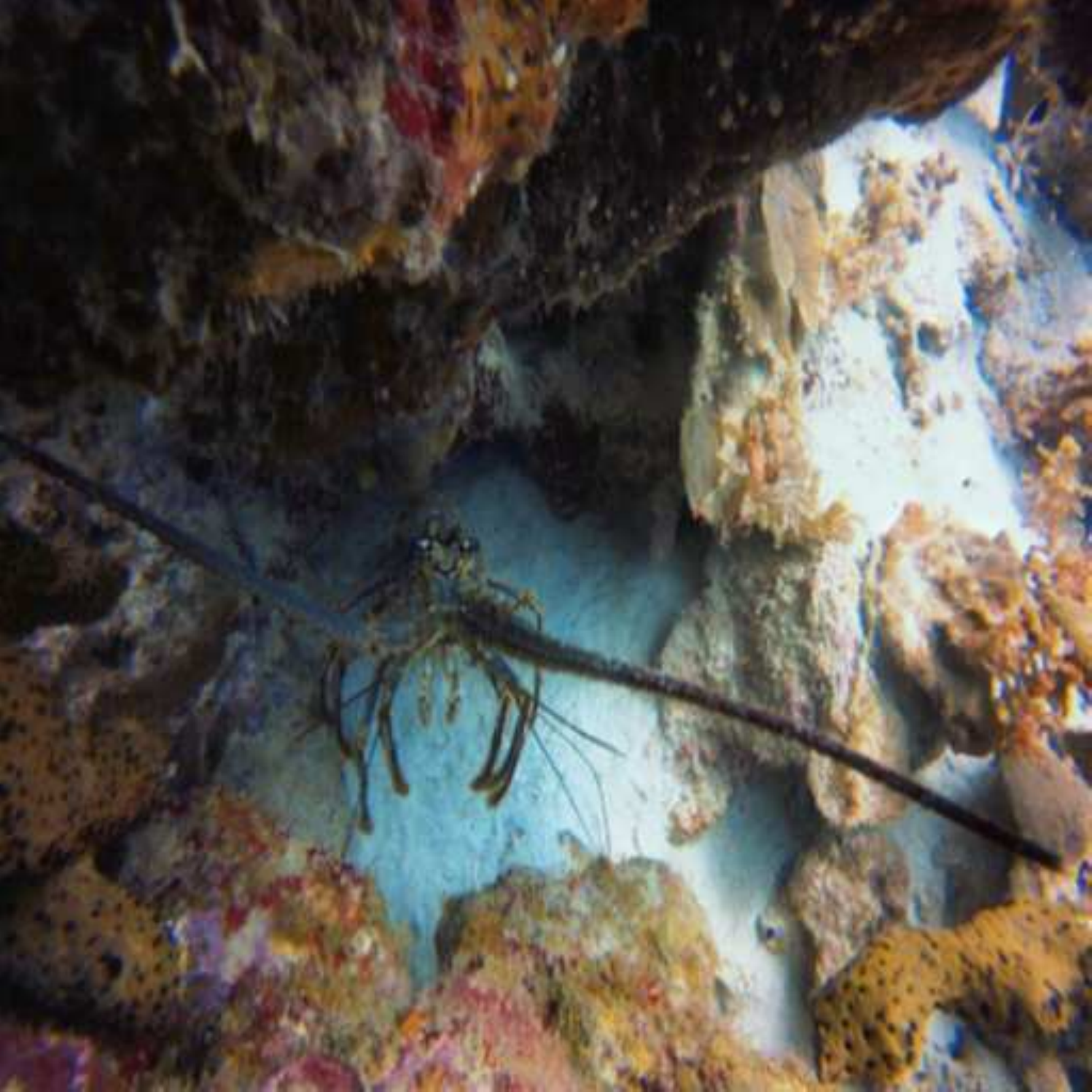}}
\hspace{-0.7mm}
\subfloat[\footnotesize{TACL}]{\includegraphics[width=0.095\textwidth]{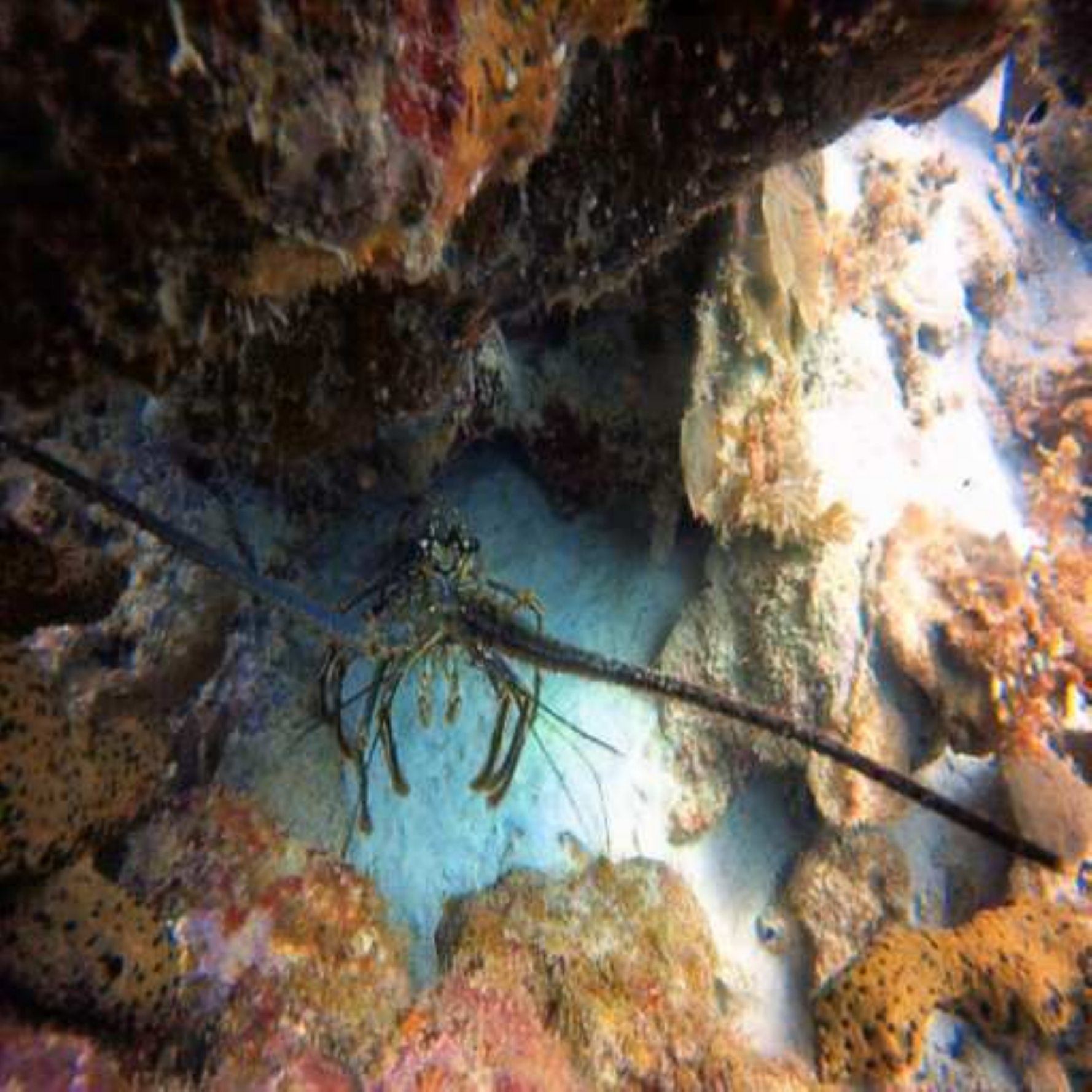}}
\hspace{-0.7mm}
\subfloat[\footnotesize{Pretrained}]{\includegraphics[width=0.095\textwidth]{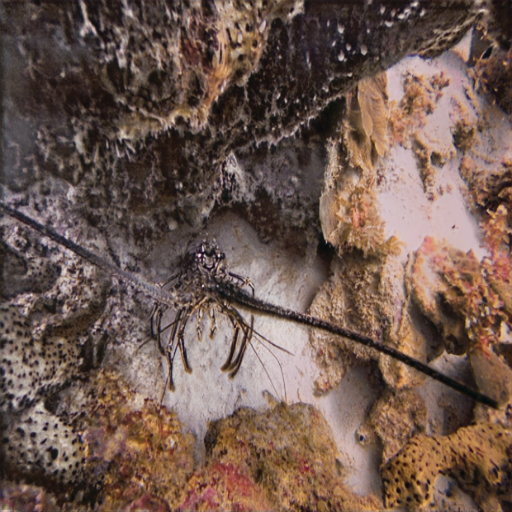}}
\hspace{-0.7mm}
\subfloat[\footnotesize{MetaUE}]{\includegraphics[width=0.095\textwidth]{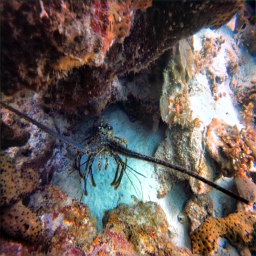}}

\caption{The visual comparison among the enhancement methods on the paired underwater images, where our MetaUE is fine-tuned on the UIEB dataset.}
\label{reference}
\end{figure*}

\begin{table*}[t]
\centering
 \footnotesize
	\tabcolsep=4pt
	\renewcommand\arraystretch{1}
  \centering
  \caption{The pre-trained model is fine-tuned on the UIEB and compared with other methods.  }
    \begin{tabular}{l|cccc|cccc|p{0.1pt}p{0.1pt}|p{0.1cm}p{0.1cm}|p{0.1cm}p{0.1cm}|p{0.05cm}p{0.05cm}}
    \hline
    \hline
      \multirow{2}{*}{\diagbox[width=2cm]{$\mathrm{Methods}$}{$\mathrm{Datasets}$}}   & \multicolumn{4}{c|}{UIEB}      & \multicolumn{4}{c|}{EUVP}      & \multicolumn{2}{c|}{U45} & \multicolumn{2}{c|}{UIQS} & \multicolumn{2}{c|}{UIEB-60} & \multicolumn{2}{c}{EUVPUN} \\
       \cline{2-17}
        & \multicolumn{1}{l}{\scriptsize{PSNR}} & \multicolumn{1}{l}{\scriptsize{SSIM}} & \multicolumn{1}{l}{\scriptsize{MSE}} & \multicolumn{1}{l|}{\scriptsize{UIQM}} & \multicolumn{1}{l}{\scriptsize{PSNR}} & \multicolumn{1}{l}{\scriptsize{SSIM}} & \multicolumn{1}{l}{\scriptsize{MSE}} & \multicolumn{1}{l|}{\scriptsize{UIQM}}  & \multicolumn{1}{l}{\scriptsize{UIQM}} & \multicolumn{1}{l|}{\scriptsize{UCIQE}} & \multicolumn{1}{l}{\scriptsize{UIQM}} & \multicolumn{1}{l|}{\scriptsize{UCIQE}}& \multicolumn{1}{l}{\scriptsize{UIQM}} & \multicolumn{1}{l|}{\scriptsize{UCIQE}}& \multicolumn{1}{l}{\scriptsize{UIQM}} & \multicolumn{1}{l}{\scriptsize{UCIQE}} \\
          \hline
    ACDC \cite{zhang2022underwaterACDC} &    18.04 & 0.8212 & 1.21  & 1.26  & 14.74 & 0.659 & 2.53  & 1.22  & 1.03  & 1.02  & 0.73  & 0.34  & 1.03  & 0.38  & 1.14  & 0.41 \\
    Rank1 \cite{liu2021rank} & 18.16 & 0.8443 & 1.21  & 1.63  & 15.16 & 0.7121 & 2.56  & 1.45  & 1.57  & 0.48  & 1.51  & 0.48  & 1.36  & 0.4   & 1.48  & 0.42 \\
    MMLE  \cite{zhang2022underwater}& 15.48 & 0.8242 & 2.34  & 1.48  & 14.84 & 0.6331 & 2.56  & 1.54  & 1.38  & \textbf{0.49} & 1.58  & 0.51 & 1.53  & 0.38  & 1.35  & 0.31 \\
    UWCNN \cite{li2020underwater} & 13.43 & 0.6253 & 3.88  & 0.89  & 14.84 & 0.7161 & 1.33  & 1.28  & 0.68  & 0.31  & 0.88  & 0.23  & 0.63  & 0.29  & 0.66  & 0.31 \\
    WaterNet \cite{UIEB} & 18.35 & 0.8397 & 1.34  & 1.73  & 12.24 & 0.5576 & 1.69  & 1.69  & 1.57  & 0.37  & 1.28  & 0.42  & 1.63  & 0.41  & 1.71  & 0.39 \\
    Funie \cite{islam2020fast}& 21.53 & 0.8512 & 0.55  & 1.69  & 18.31 & 0.7393 & 1.15  & 1.56  & 1.45  & 0.39  & 1.46  & 0.39  & 1.61  & 0.38  & 1.44  & 0.38 \\
    Ucolor \cite{li2021underwater} & 20.63 & 0.8495 & 1.08  & 1.74  & 15.54 & 0.7215 & 1.55  & 1.69  & 1.72  & 0.39  & 1.69 & 0.41  & 1.63  & 0.34  & 1.59  & 0.39 \\
    TACL\cite{liu2022twin}  & 20.41 & 0.8478 & 0.48  & 1.77  & 18.42 & 0.7501 & 1.24  & 2.24 & 1.85  & 0.42  & 1.59  & 0.39  & 1.58  & 0.41  & 1.91  & 0.43 \\
    \hline
    Pre-trained & 16.42 & 0.7354 & 1.54 & \textbf{1.95} & 15.57 & 0.7145 & 1.96 & \textbf{2.15}  & 1.88 & 0.46  & \textbf{1.72} & \textbf{0.55}  & 1.55 & \textbf{0.45} & 1.87 & \textbf{0.46}\\
    MetaUE & \textbf{22.01} & \textbf{0.8607} & \textbf{0.54} & 1.81 & \textbf{19.11} & \textbf{0.7804} & \textbf{1.09} & 2.12  & \textbf{1.93} & 0.42  & 1.61 & 0.41  & \textbf{1.65} & 0.42 & \textbf{1.98} & 0.39 \\
    \hline
    \hline
    \end{tabular}%
  \label{metau45}%
\end{table*}
\begin{table*}[t]
\centering
\footnotesize
	\tabcolsep=5pt
	\renewcommand\arraystretch{1}
  \centering
  \caption{The pre-trained model is fine-tuned on the EUVP and compared with other methods. }
    \begin{tabular}{l|p{0.1cm}p{0.1cm}p{0.1cm}p{0.1cm}|p{0.1pt}p{0.1pt}p{0.1pt}p{0.1pt}|p{0.1pt}p{0.1pt}|p{0.1cm}p{0.1cm}|p{0.1cm}p{0.1cm}|p{0.1cm}p{0.1cm}}
    \hline
    \hline
  \multirow{2}{*}{\diagbox[width=2cm]{$\mathrm{Methods}$}{$\mathrm{Datasets}$}}       & \multicolumn{4}{c|}{UIEB}      & \multicolumn{4}{c|}{EUVP}      & \multicolumn{2}{c|}{U45} & \multicolumn{2}{c|}{UIQS} & \multicolumn{2}{c|}{UIEB-60} & \multicolumn{2}{c}{EUVPUN} \\
     \cline{2-17}
     & \multicolumn{1}{l}{\scriptsize{PSNR}} & \multicolumn{1}{l}{\scriptsize{SSIM}} & \multicolumn{1}{l}{\scriptsize{MSE}} & \multicolumn{1}{l|}{\scriptsize{UIQM}} & \multicolumn{1}{l}{\scriptsize{PSNR}} & \multicolumn{1}{l}{\scriptsize{SSIM}} & \multicolumn{1}{l}{\scriptsize{MSE}} & \multicolumn{1}{l|}{\scriptsize{UIQM}}  & \multicolumn{1}{l}{\scriptsize{UIQM}} & \multicolumn{1}{l|}{\scriptsize{UCIQE}} & \multicolumn{1}{l}{\scriptsize{UIQM}} & \multicolumn{1}{l|}{\scriptsize{UCIQE}}& \multicolumn{1}{l}{\scriptsize{UIQM}} & \multicolumn{1}{l|}{\scriptsize{UCIQE}}& \multicolumn{1}{l}{\scriptsize{UIQM}} & \multicolumn{1}{l}{\scriptsize{UCIQE}} \\
          \hline
    ACDC \cite{zhang2022underwaterACDC} &    18.04 & 0.8212 & 1.21 & 1.26  & 14.74 & 0.659 & 2.53  & 1.22  & 1.03  & 1.02  & 0.73  & 0.34  & 1.03  & 0.38  & 1.14  & 0.41 \\
    Rank1 \cite{liu2021rank} &18.16 & 0.8443 & 1.21 & 1.63  & 15.16 & 0.7121 & 2.56  & 1.45  & 1.57  & 0.48  & 1.51  & 0.48  & 1.36  & 0.4   & 1.48  & 0.42 \\
    MMLE \cite{zhang2022underwater}&  15.48 & 0.8242 & 2.34  & 1.48  & 14.84 & 0.6331 & 2.56  & 1.54  & 1.38  & \textbf{0.49} & 1.58  & 0.51 & 1.53  & 0.38  & 1.35  & 0.31 \\
    WaterNet \cite{UIEB} &   15.42 & 0.6542 & 3.51  & 1.15  & 18.64 & 0.7215 & 2.15  & 1.78  & 1.39  & 0.31  & 1.41  & 0.33  & 1.31  & 0.35  & 1.45  & 0.33 \\
    Funie \cite{islam2020fast} &  16.91 & 0.7291 & 1.70   & 1.66  & 21.99 & 0.8003 & 0.45 & 2.05  & 1.45  & 0.35  & 1.31  & 0.31  & 1.33  & 0.37  & 1.74  & 0.37 \\
    Ucolor  \cite{li2021underwater}&  16.54 & 0.7464 & 1.82  & 1.16  & 20.08 & 0.7815 & 1.54  & 1.98  & 1.77  & 0.41  & 1.61  & 0.39  & 0.88  & 0.32  & 1.66  & 0.38 \\
    \hline
    Pre-trained & 16.42 & 0.7354 & 1.54 & \textbf{1.95} & 15.57 & 0.7145 & 1.96 & 2.15  & \textbf{1.88} & 0.46  & \textbf{1.72} & \textbf{0.55}  & 1.55 & \textbf{0.45} & \textbf{1.87} & \textbf{0.46}\\
     MetaUE&  \textbf{18.67} & \textbf{0.8494} &  \textbf{1.07}  & 1.71 & \textbf{22.72} & \textbf{0.8623} & \textbf{0.39}  & \textbf{2.19} &1.81 & 0.43  &1.61 & 0.42  & \textbf{1.68} & 0.41 & 1.75 & \textbf{0.46} \\
    \hline
    \hline
    \end{tabular}%
  \label{metaUE}%
\end{table*}%

\begin{figure*}[htbp]
\centering

\hspace{-5mm}
\subfloat{\includegraphics[width=0.095\textwidth]{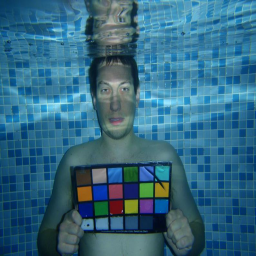}}
\hspace{-0.7mm}
\subfloat{\includegraphics[width=0.095\textwidth]{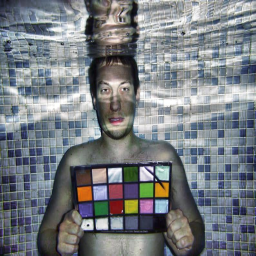}}
\hspace{-0.7mm}
\subfloat{\includegraphics[width=0.095\textwidth]{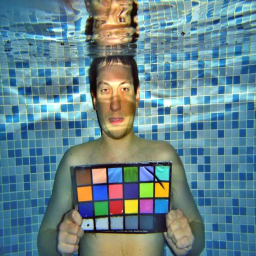}}
\hspace{-0.7mm}
\subfloat{\includegraphics[width=0.095\textwidth]{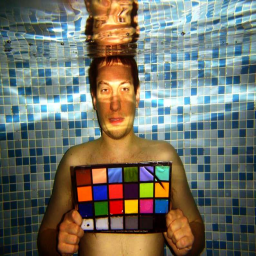}}
\hspace{-0.7mm}
\subfloat{\includegraphics[width=0.095\textwidth]{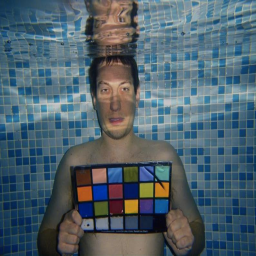}}
\hspace{-0.7mm}
\subfloat{\includegraphics[width=0.095\textwidth]{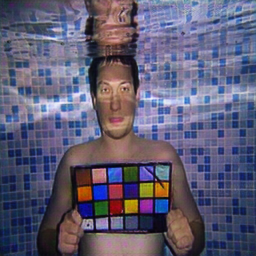}}
\hspace{-0.7mm}
\subfloat{\includegraphics[width=0.095\textwidth]{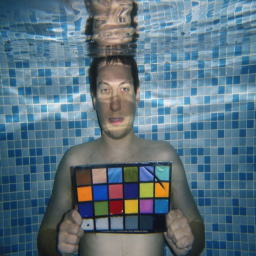}}
\hspace{-0.7mm}
\subfloat{\includegraphics[width=0.095\textwidth]{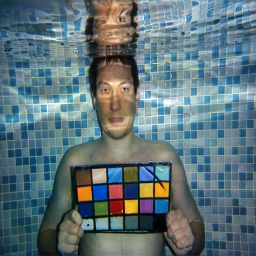}}
\hspace{-0.7mm}
\subfloat{\includegraphics[width=0.095\textwidth]{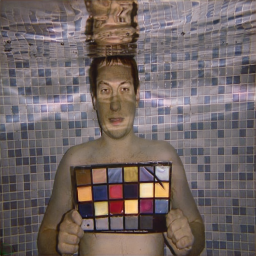}}
\hspace{-0.7mm}
\subfloat{\includegraphics[width=0.095\textwidth]{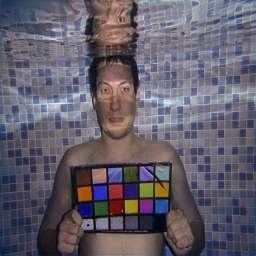}}

\vspace{0.5mm}
\hspace{-5mm}
\subfloat{\includegraphics[width=0.095\textwidth]{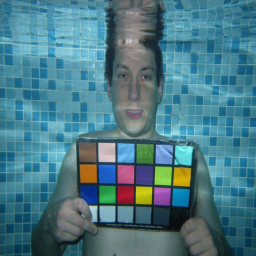}}
\hspace{-0.7mm}
\subfloat{\includegraphics[width=0.095\textwidth]{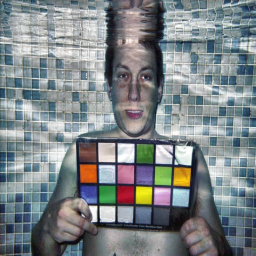}}
\hspace{-0.7mm}
\subfloat{\includegraphics[width=0.095\textwidth]{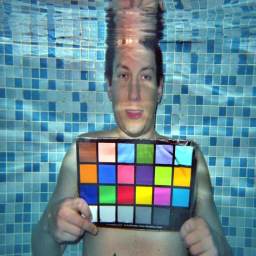}}
\hspace{-0.7mm}
\subfloat{\includegraphics[width=0.095\textwidth]{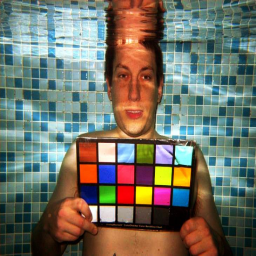}}
\hspace{-0.7mm}
\subfloat{\includegraphics[width=0.095\textwidth]{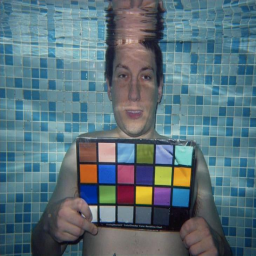}}
\hspace{-0.7mm}
\subfloat{\includegraphics[width=0.095\textwidth]{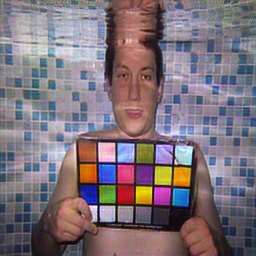}}
\hspace{-0.7mm}
\subfloat{\includegraphics[width=0.095\textwidth]{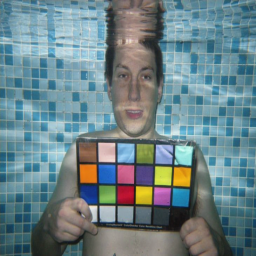}}
\hspace{-0.7mm}
\subfloat{\includegraphics[width=0.095\textwidth]{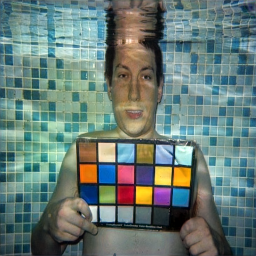}}
\hspace{-0.7mm}
\subfloat{\includegraphics[width=0.095\textwidth]{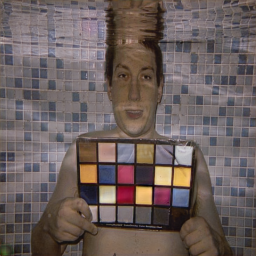}}
\hspace{-0.7mm}
\subfloat{\includegraphics[width=0.095\textwidth]{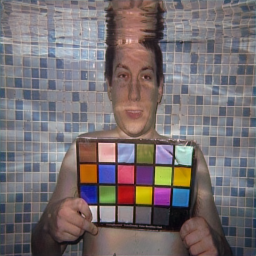}}

\vspace{0.5mm}
\hspace{-5mm}
\subfloat{\includegraphics[width=0.095\textwidth]{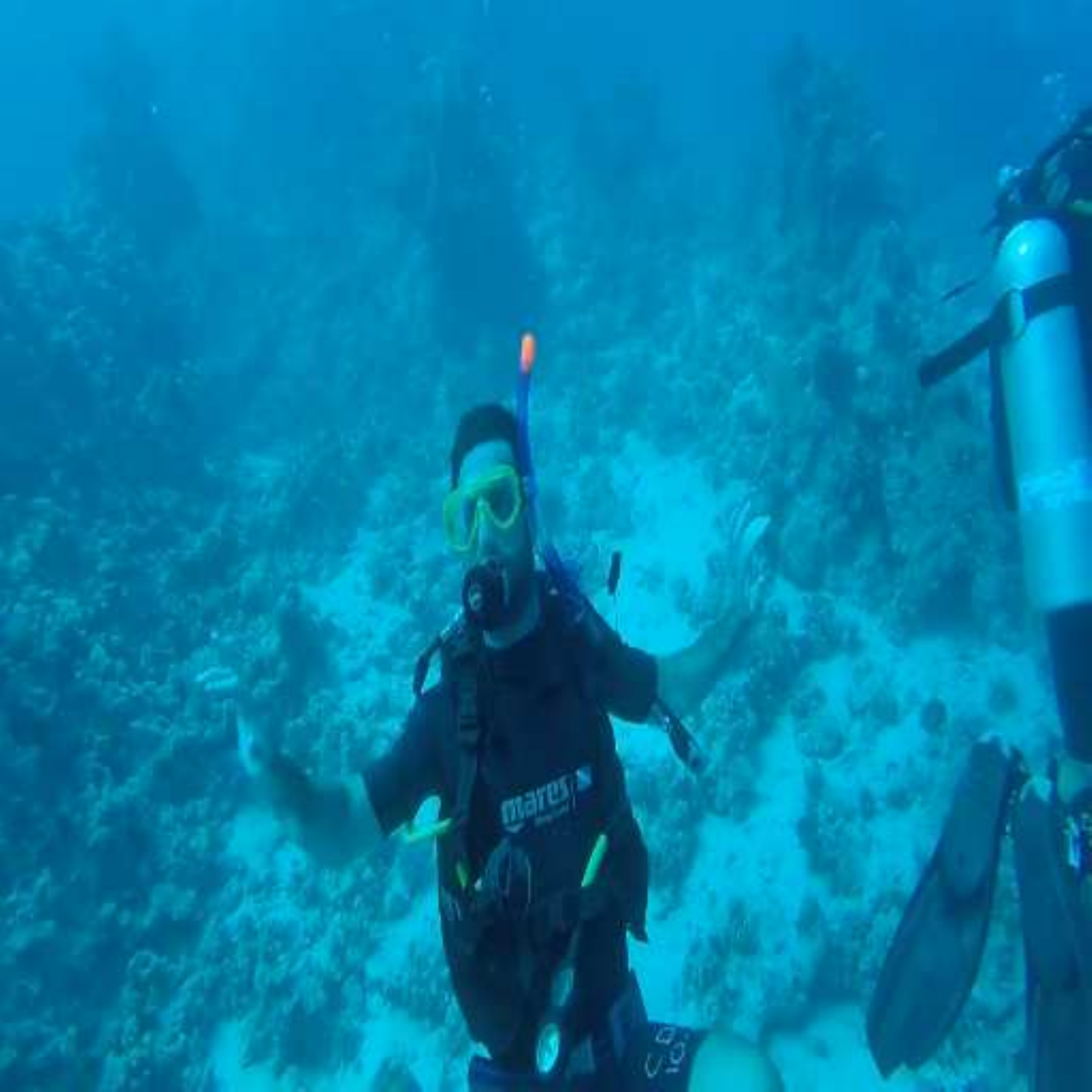}}
\hspace{-0.7mm}
\subfloat{\includegraphics[width=0.095\textwidth]{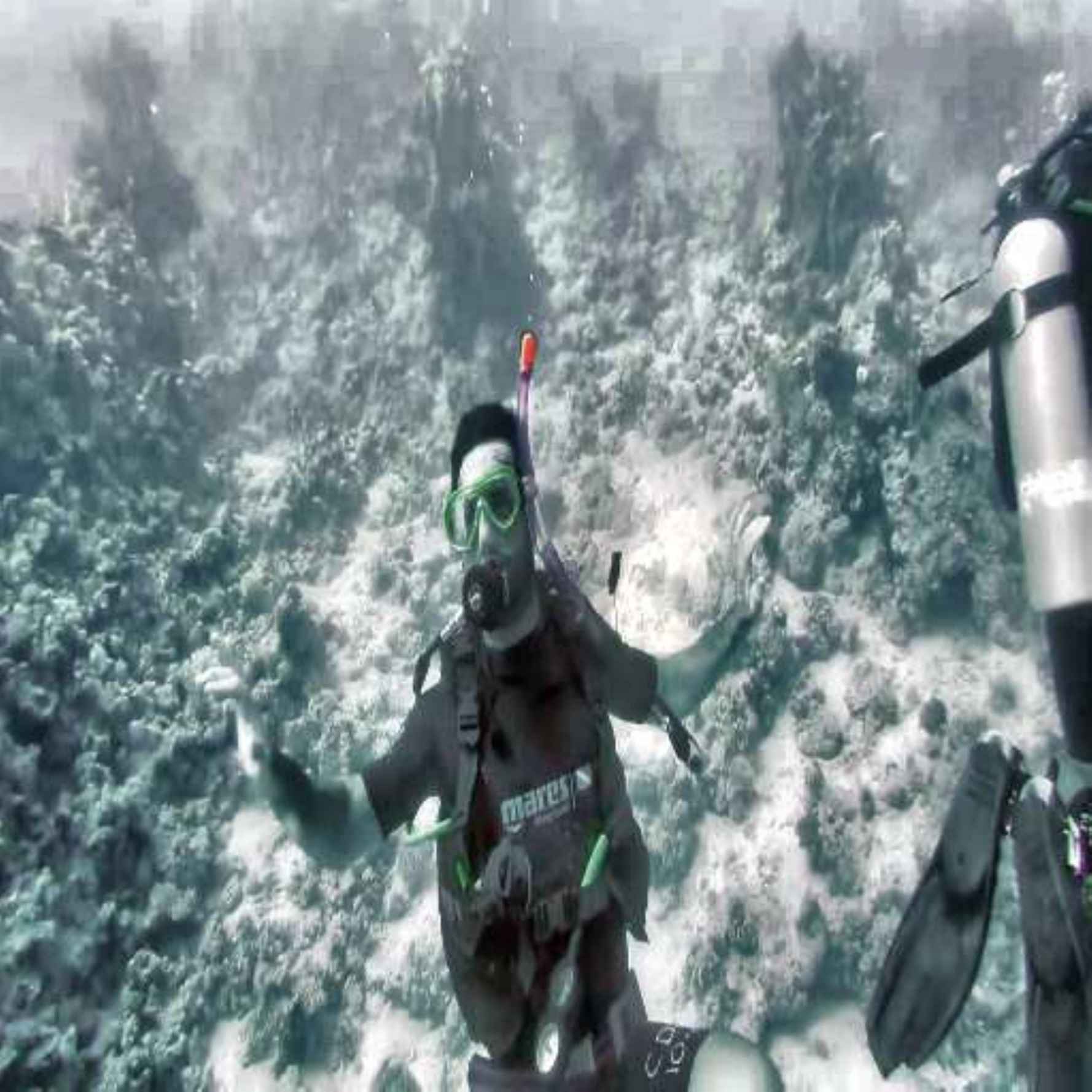}}
\hspace{-0.7mm}
\subfloat{\includegraphics[width=0.095\textwidth]{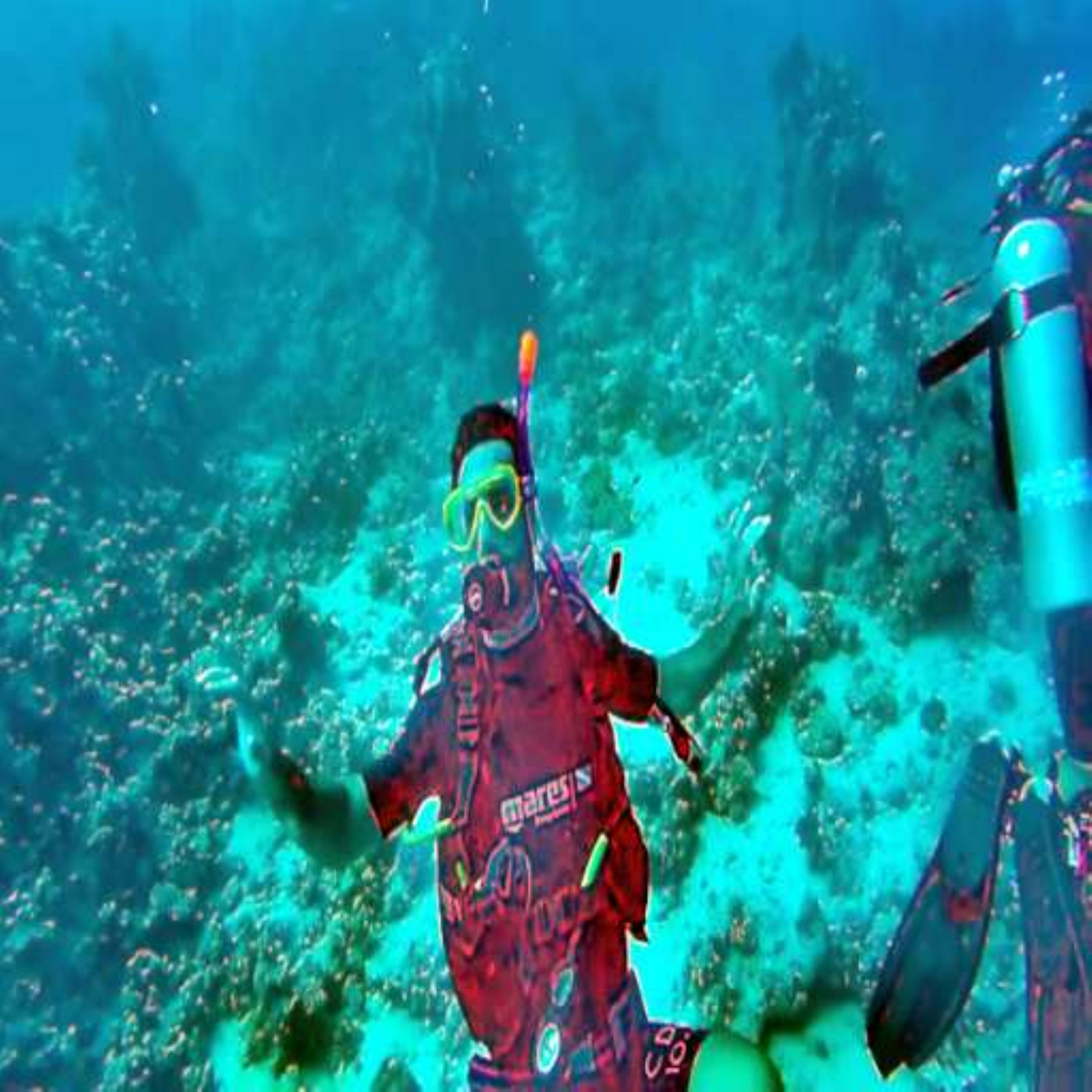}}
\hspace{-0.7mm}
\subfloat{\includegraphics[width=0.095\textwidth]{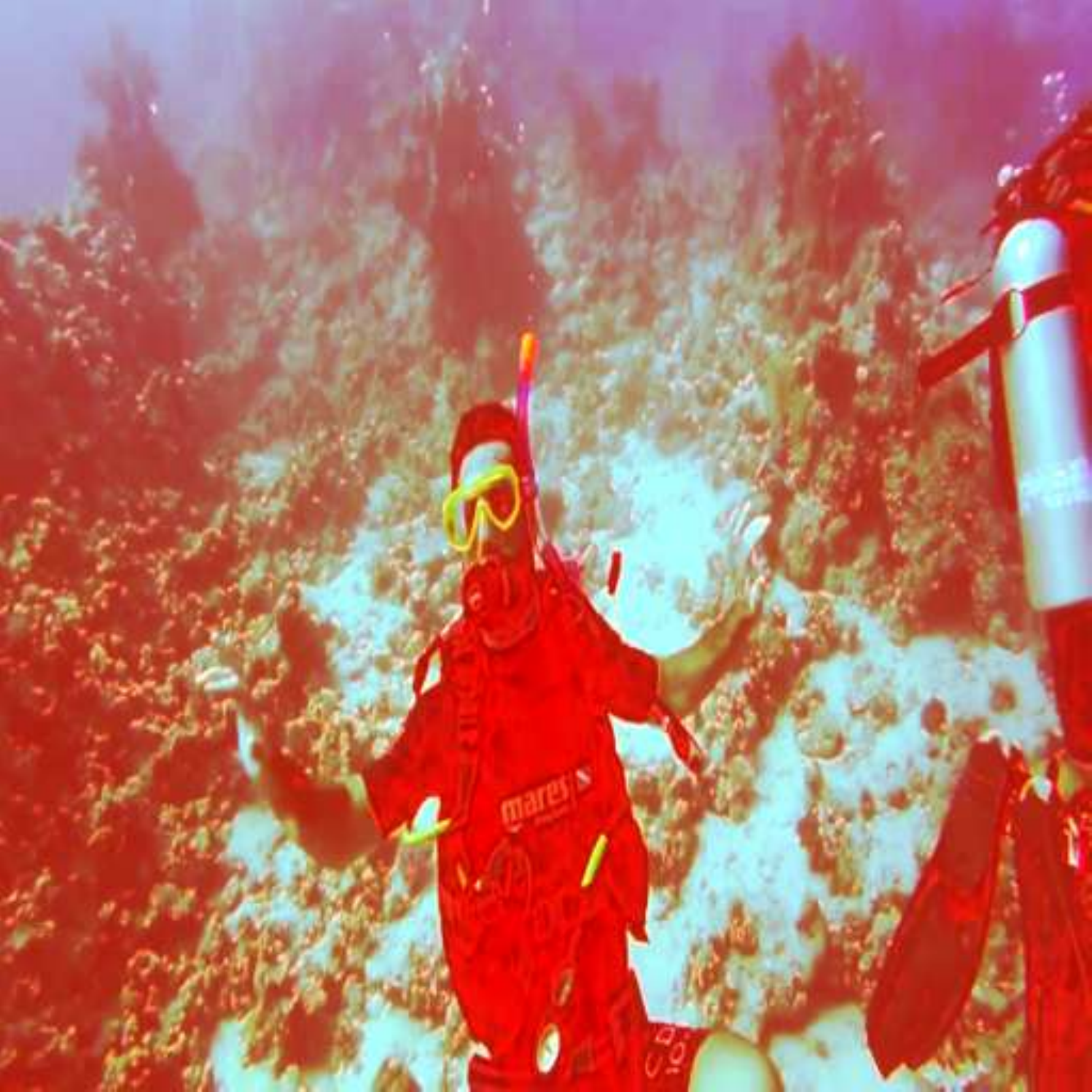}}
\hspace{-0.7mm}
\subfloat{\includegraphics[width=0.095\textwidth]{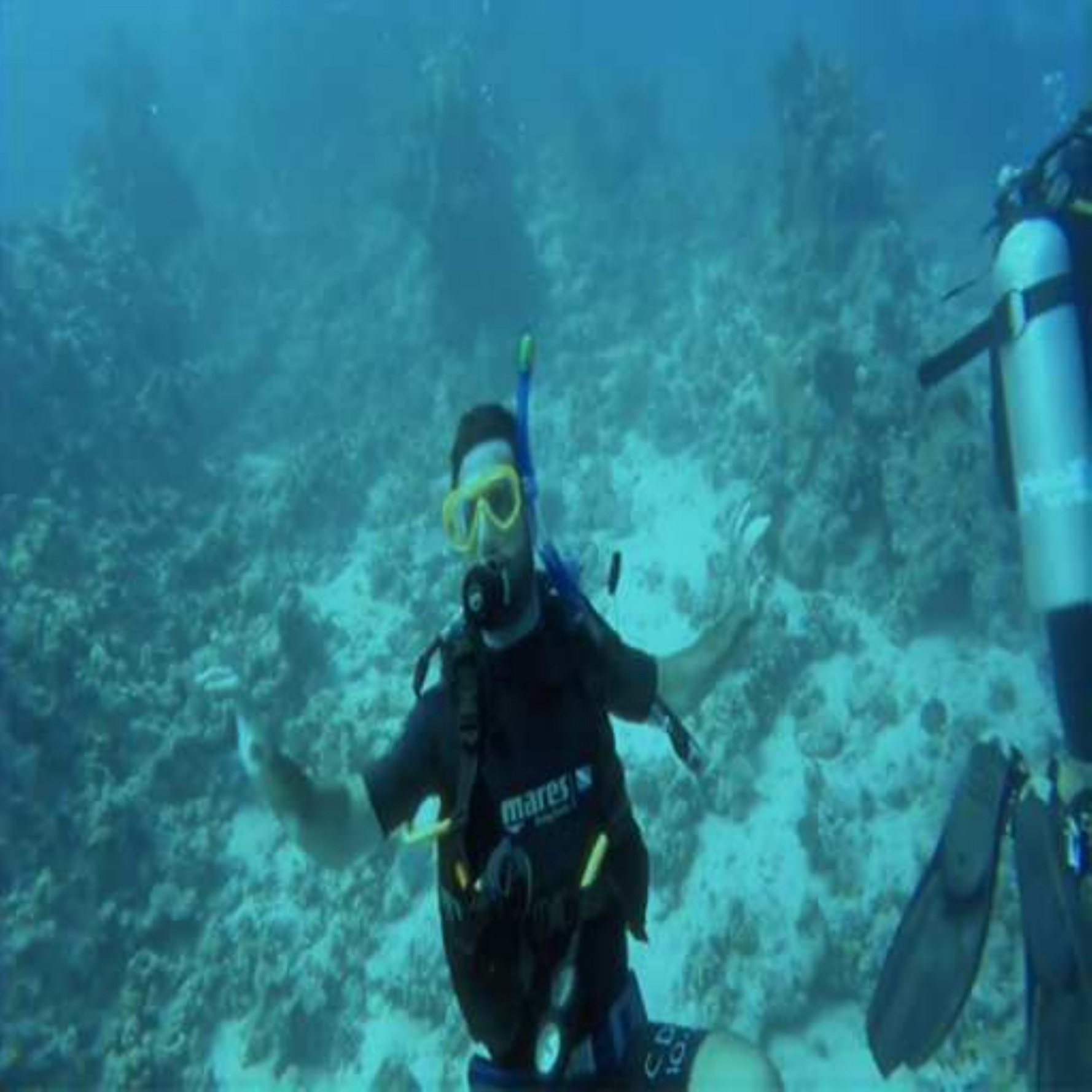}}
\hspace{-0.7mm}
\subfloat{\includegraphics[width=0.095\textwidth]{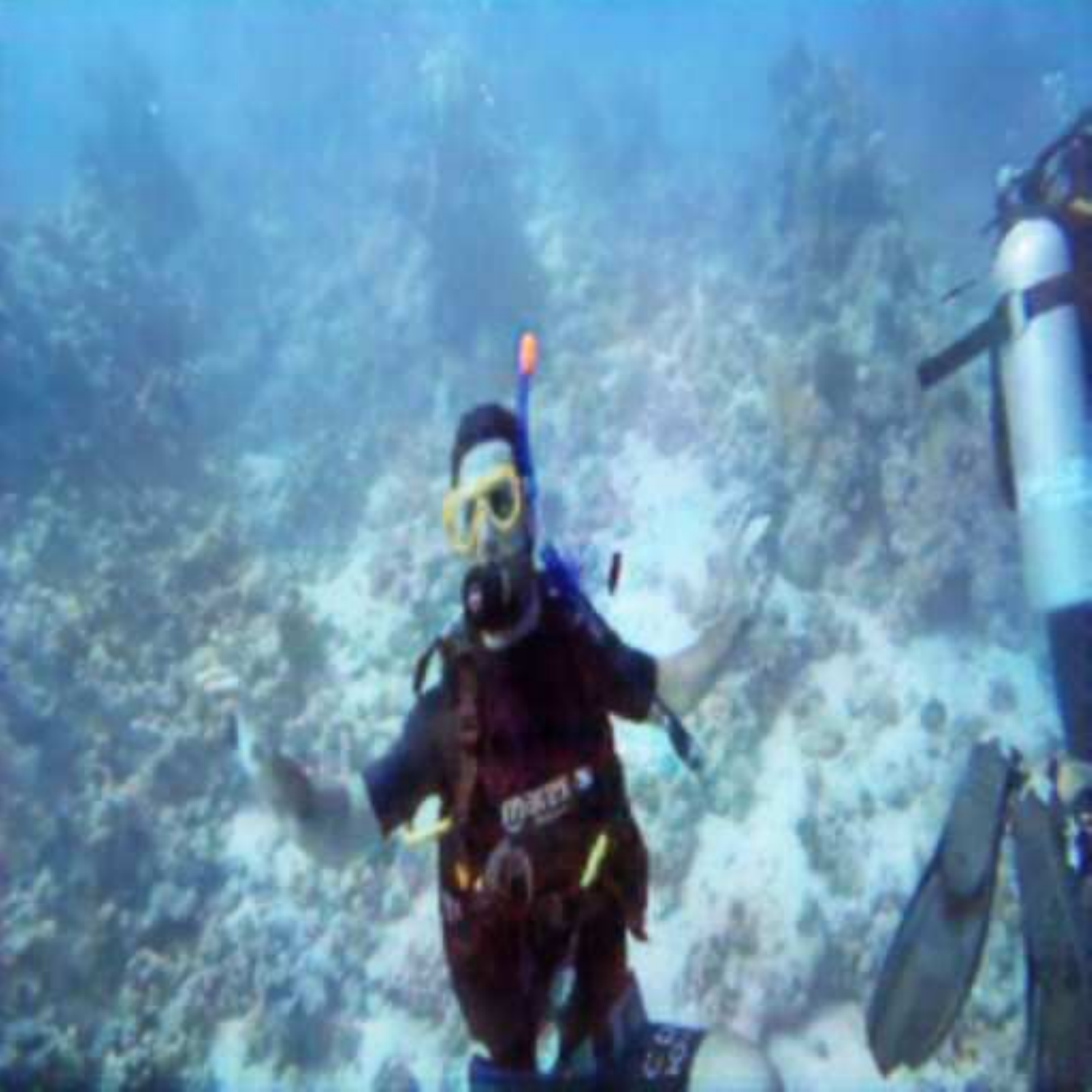}}
\hspace{-0.7mm}
\subfloat{\includegraphics[width=0.095\textwidth]{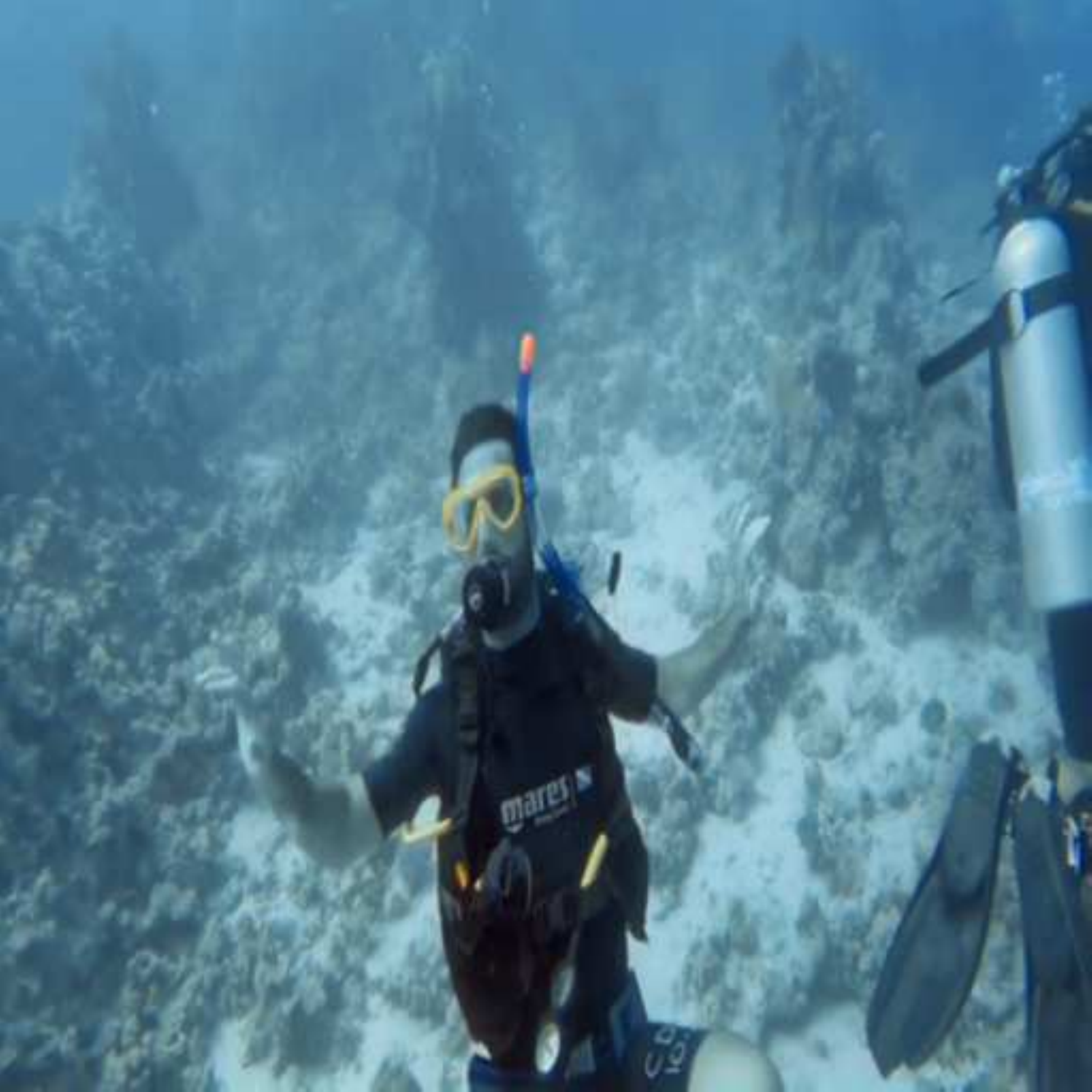}}
\hspace{-0.7mm}
\subfloat{\includegraphics[width=0.095\textwidth]{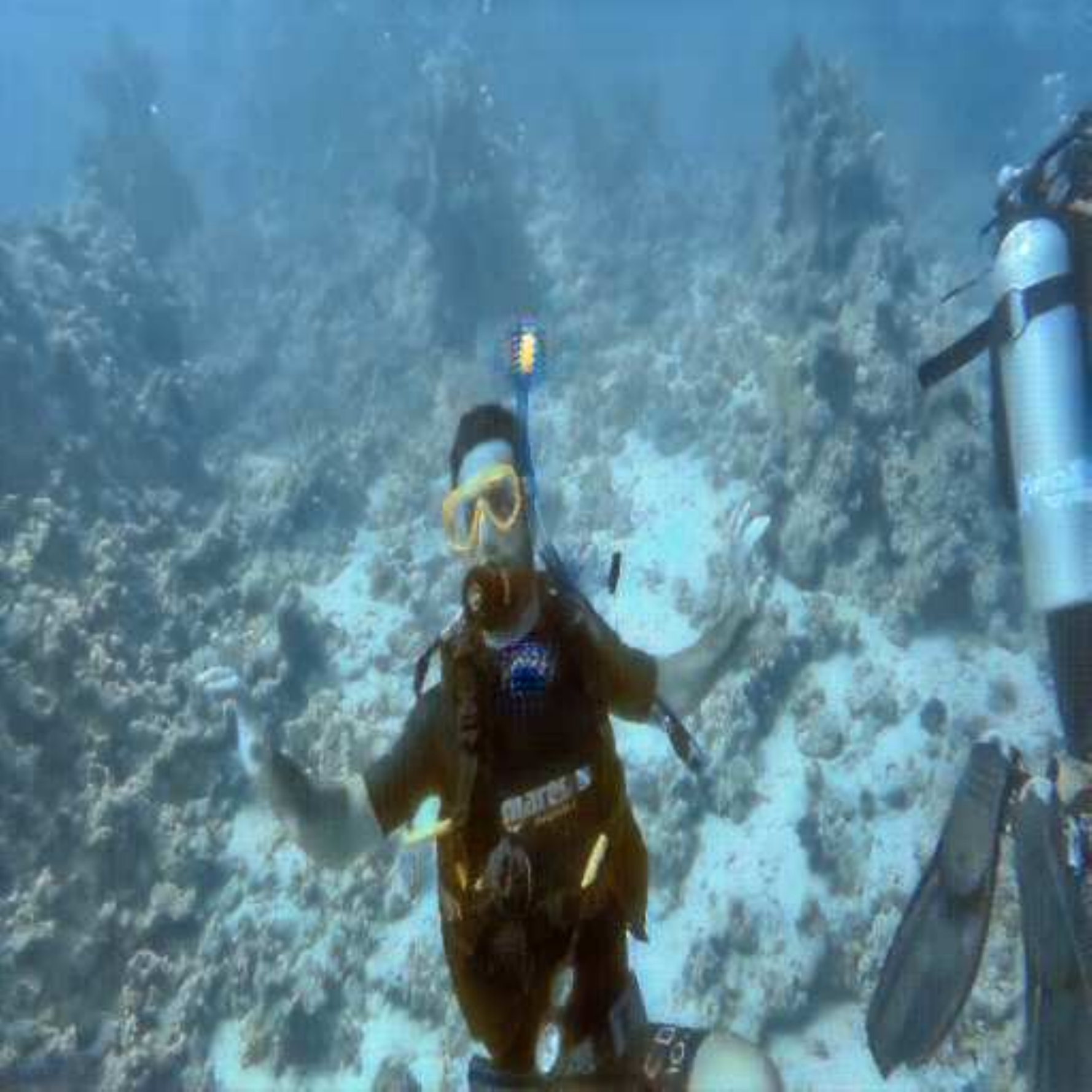}}
\hspace{-0.7mm}
\subfloat{\includegraphics[width=0.095\textwidth]{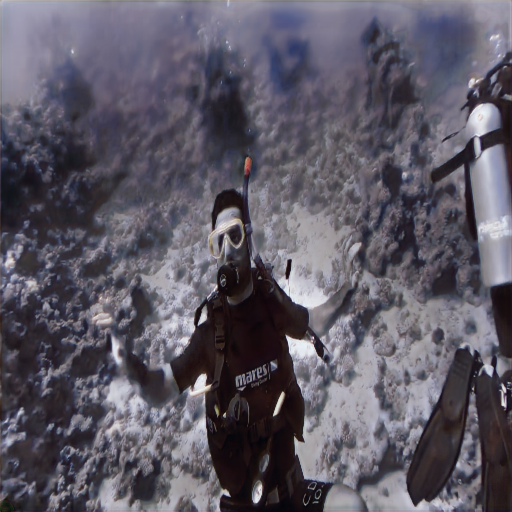}}
\hspace{-0.7mm}
\subfloat{\includegraphics[width=0.095\textwidth]{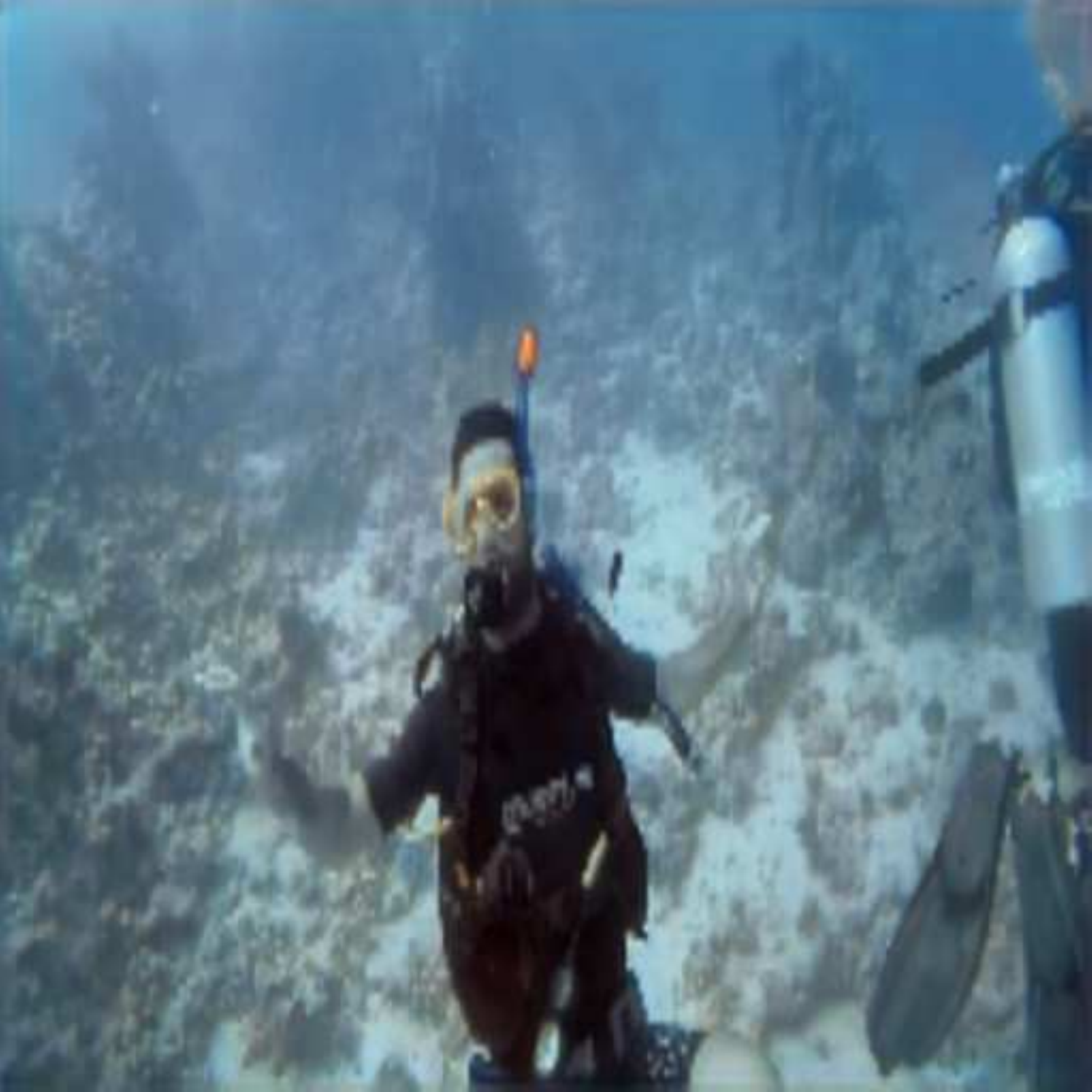}}

\vspace{0.5mm}
\hspace{-5mm}
\subfloat{\includegraphics[width=0.095\textwidth]{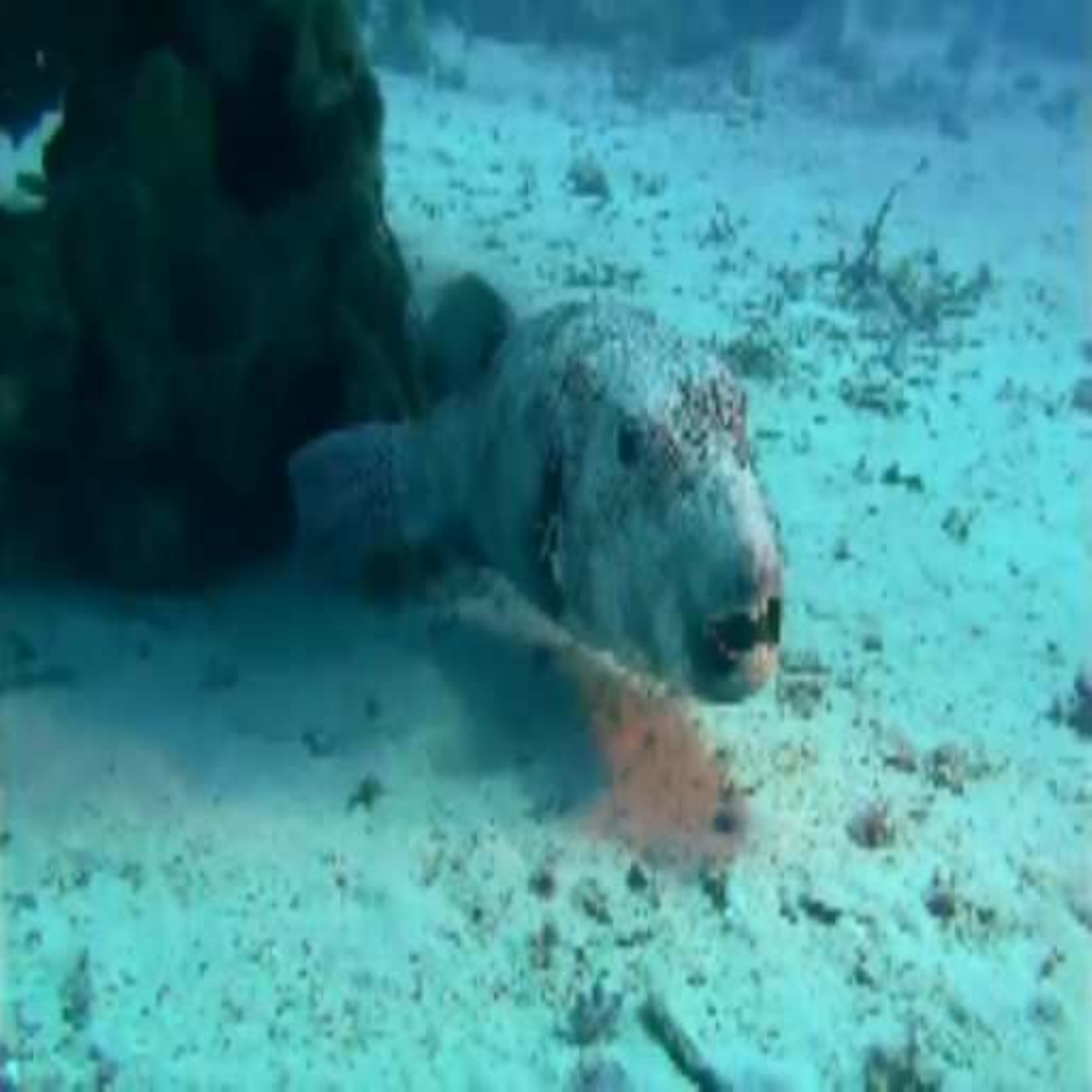}}
\hspace{-0.7mm}
\subfloat{\includegraphics[width=0.095\textwidth]{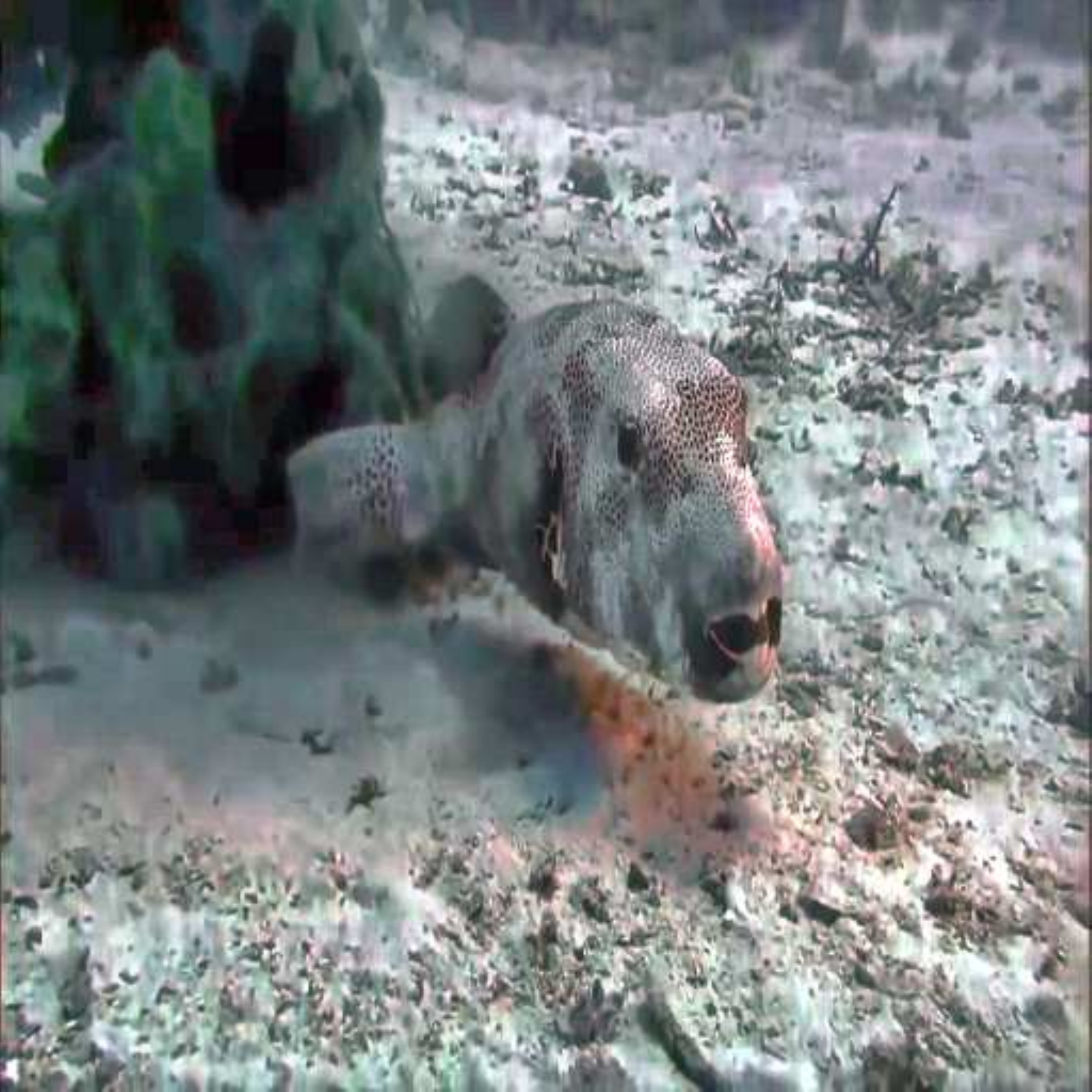}}
\hspace{-0.7mm}
\subfloat{\includegraphics[width=0.095\textwidth]{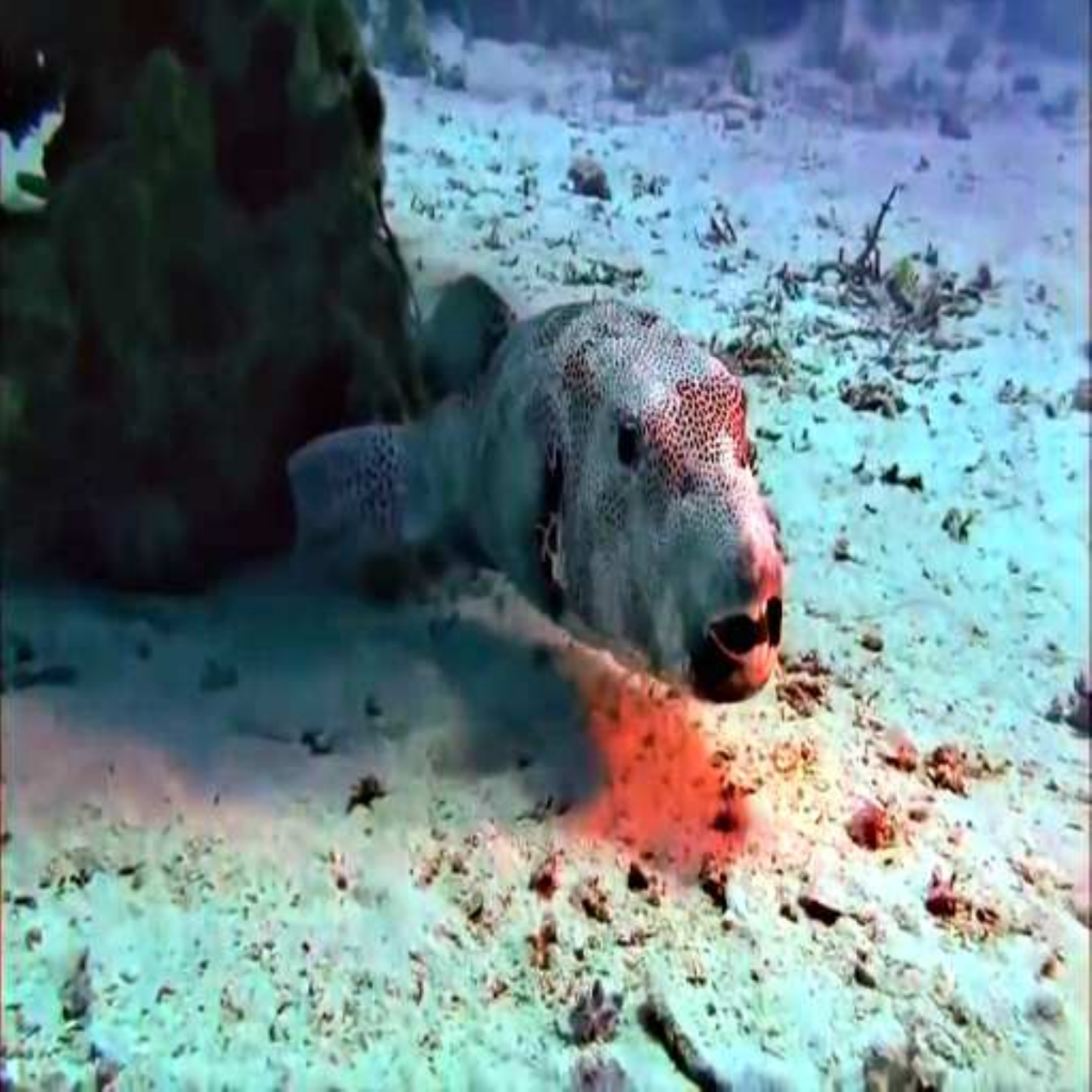}}
\hspace{-0.7mm}
\subfloat{\includegraphics[width=0.095\textwidth]{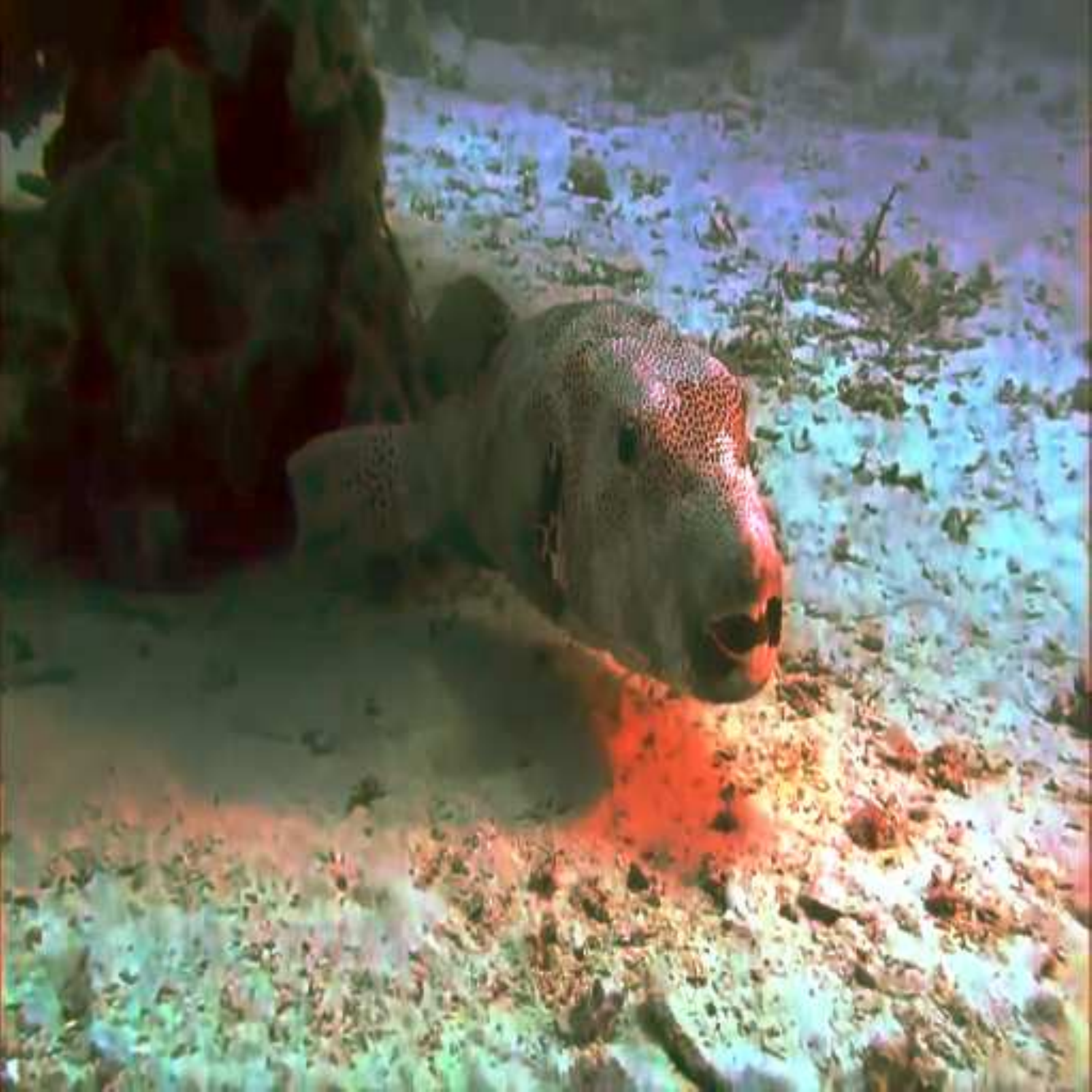}}
\hspace{-0.7mm}
\subfloat{\includegraphics[width=0.095\textwidth]{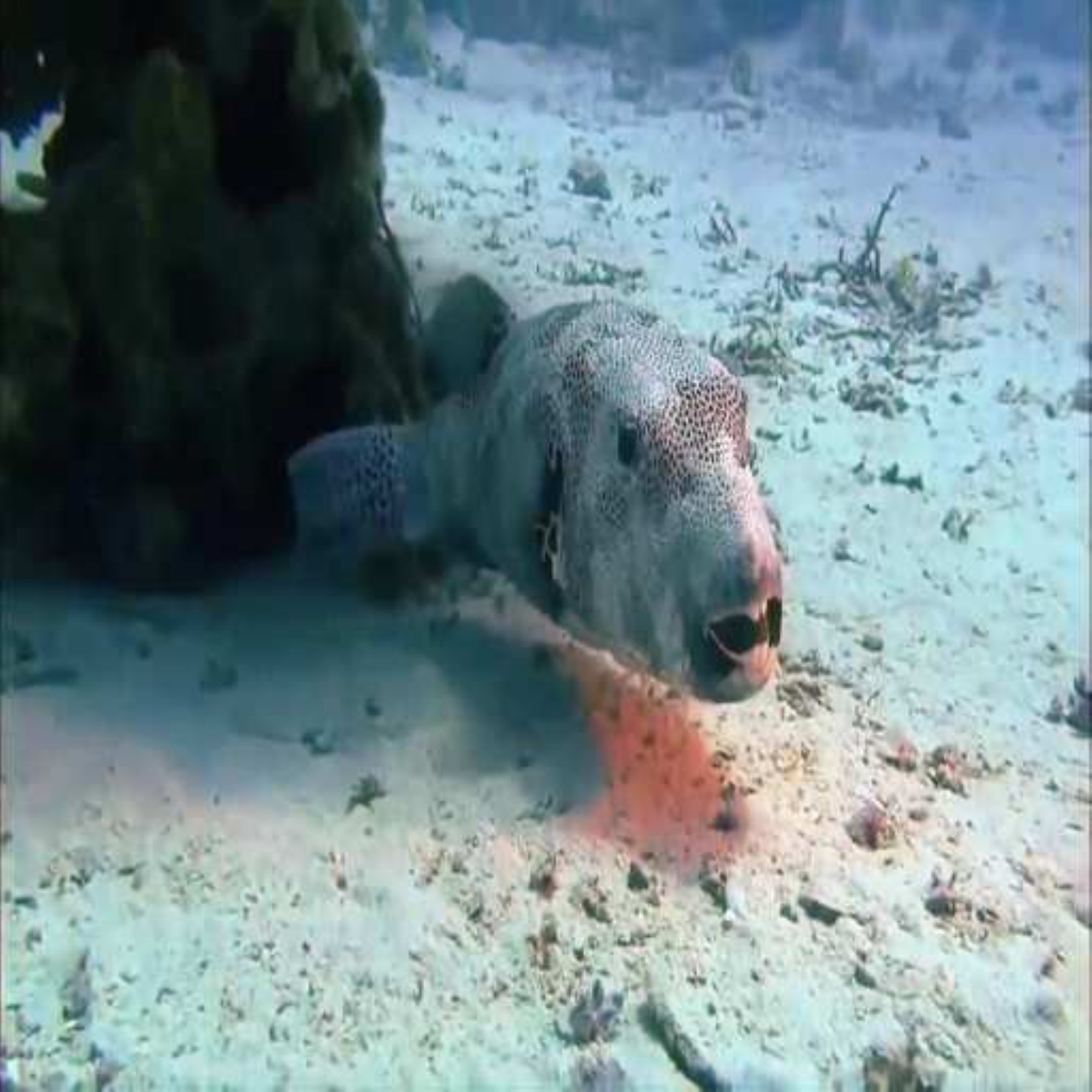}}
\hspace{-0.7mm}
\subfloat{\includegraphics[width=0.095\textwidth]{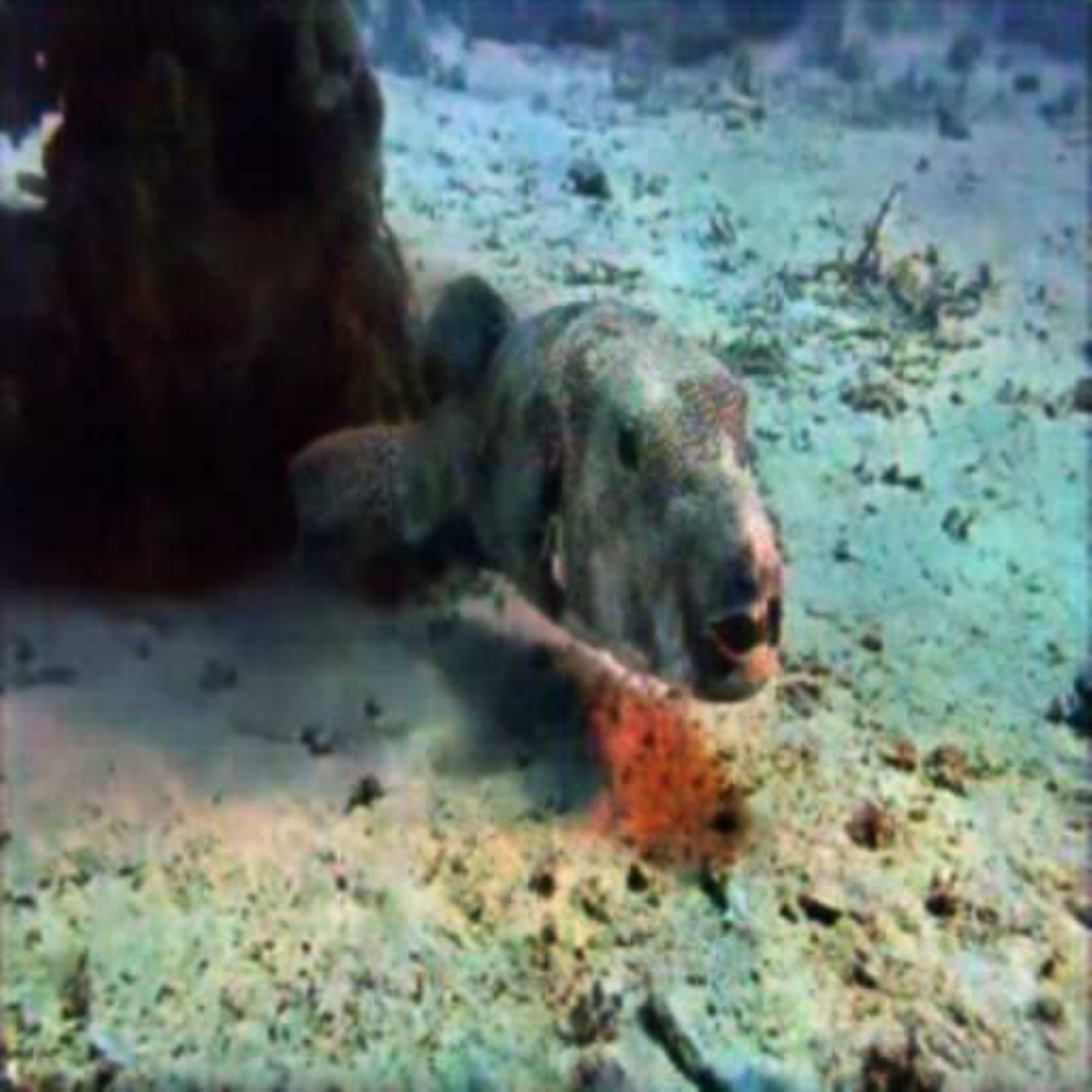}}
\hspace{-0.7mm}
\subfloat{\includegraphics[width=0.095\textwidth]{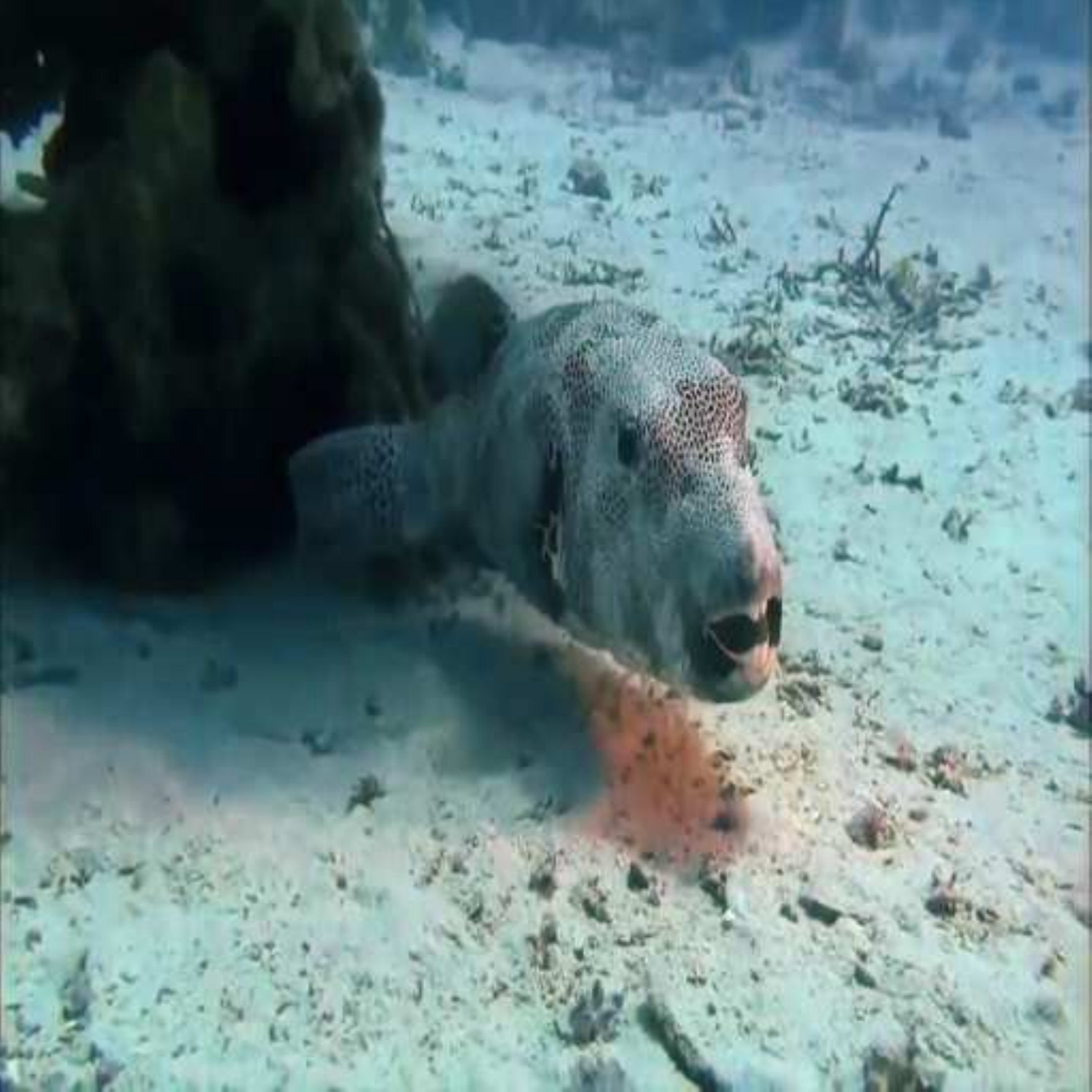}}
\hspace{-0.7mm}
\subfloat{\includegraphics[width=0.095\textwidth]{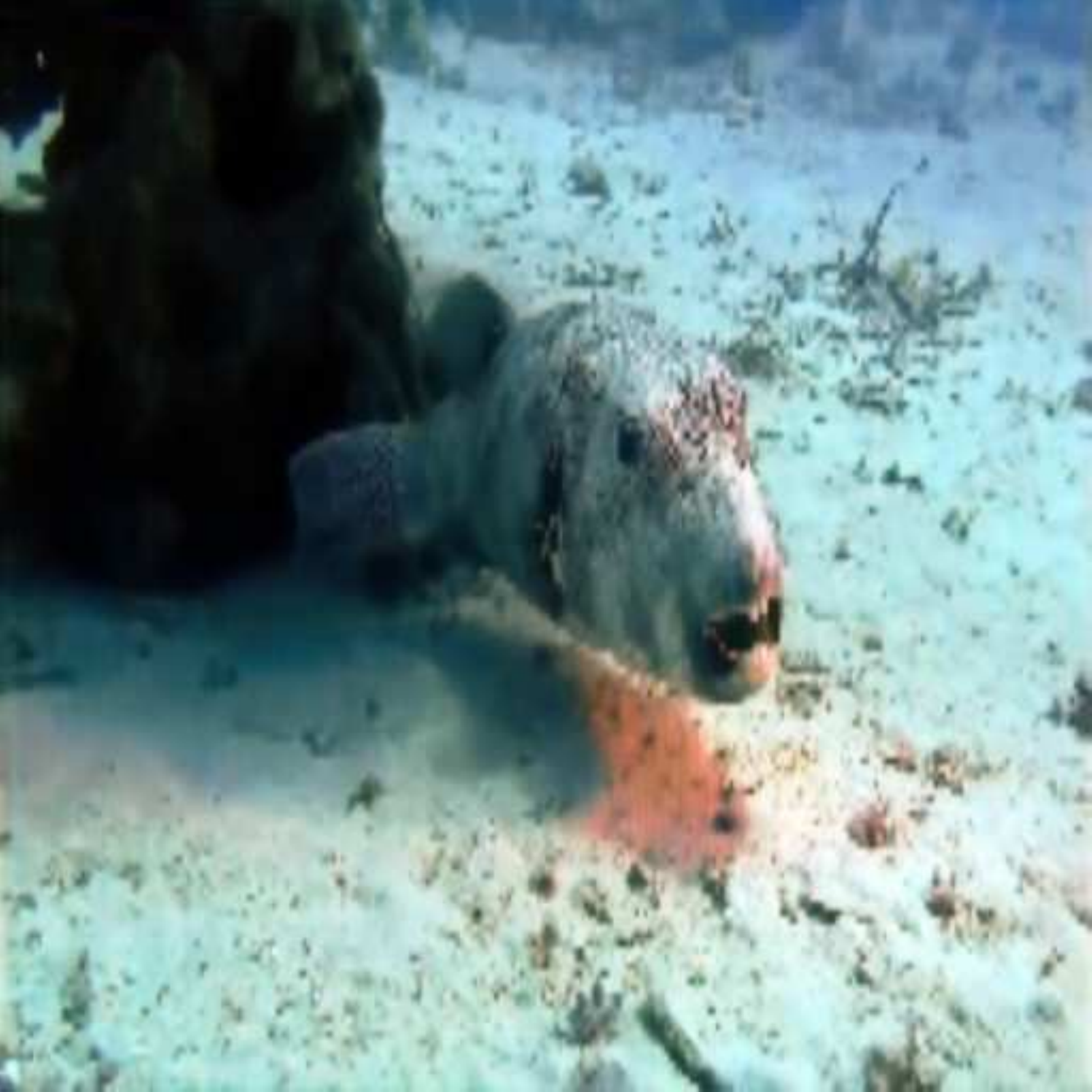}}
\hspace{-0.7mm}
\subfloat{\includegraphics[width=0.095\textwidth]{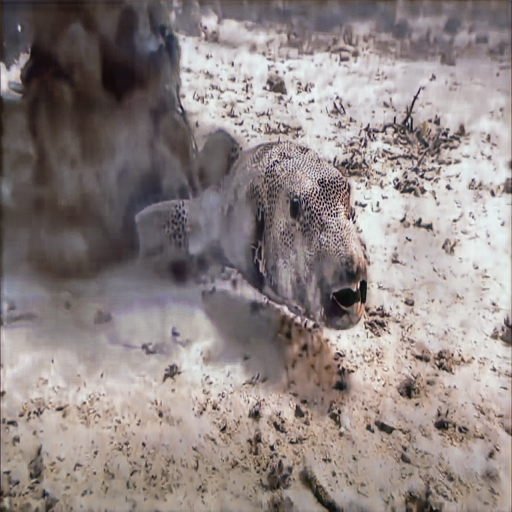}}
\hspace{-0.7mm}
\subfloat{\includegraphics[width=0.095\textwidth]{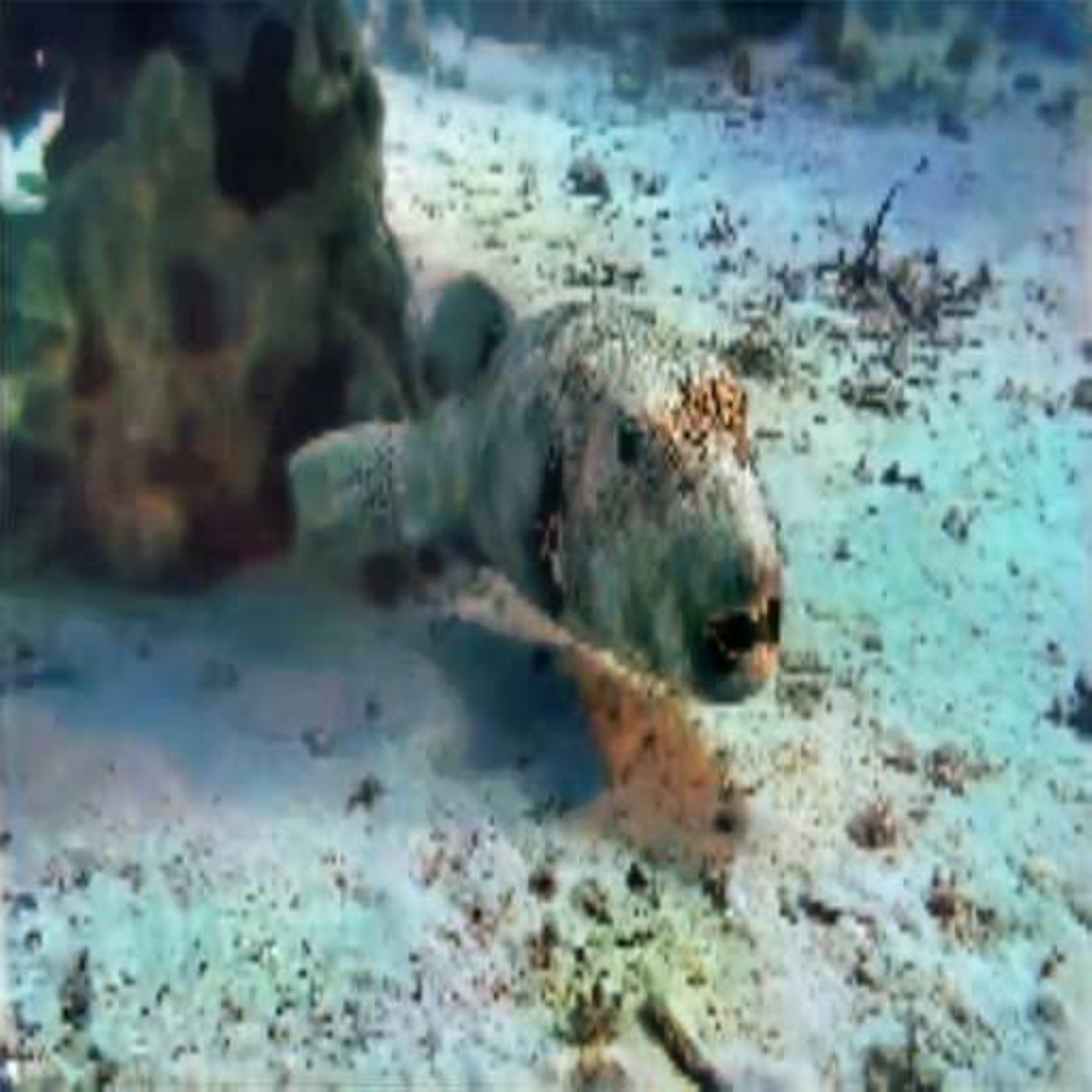}}

\vspace{0.5mm}
\hspace{-5mm}
\subfloat{\includegraphics[width=0.095\textwidth]{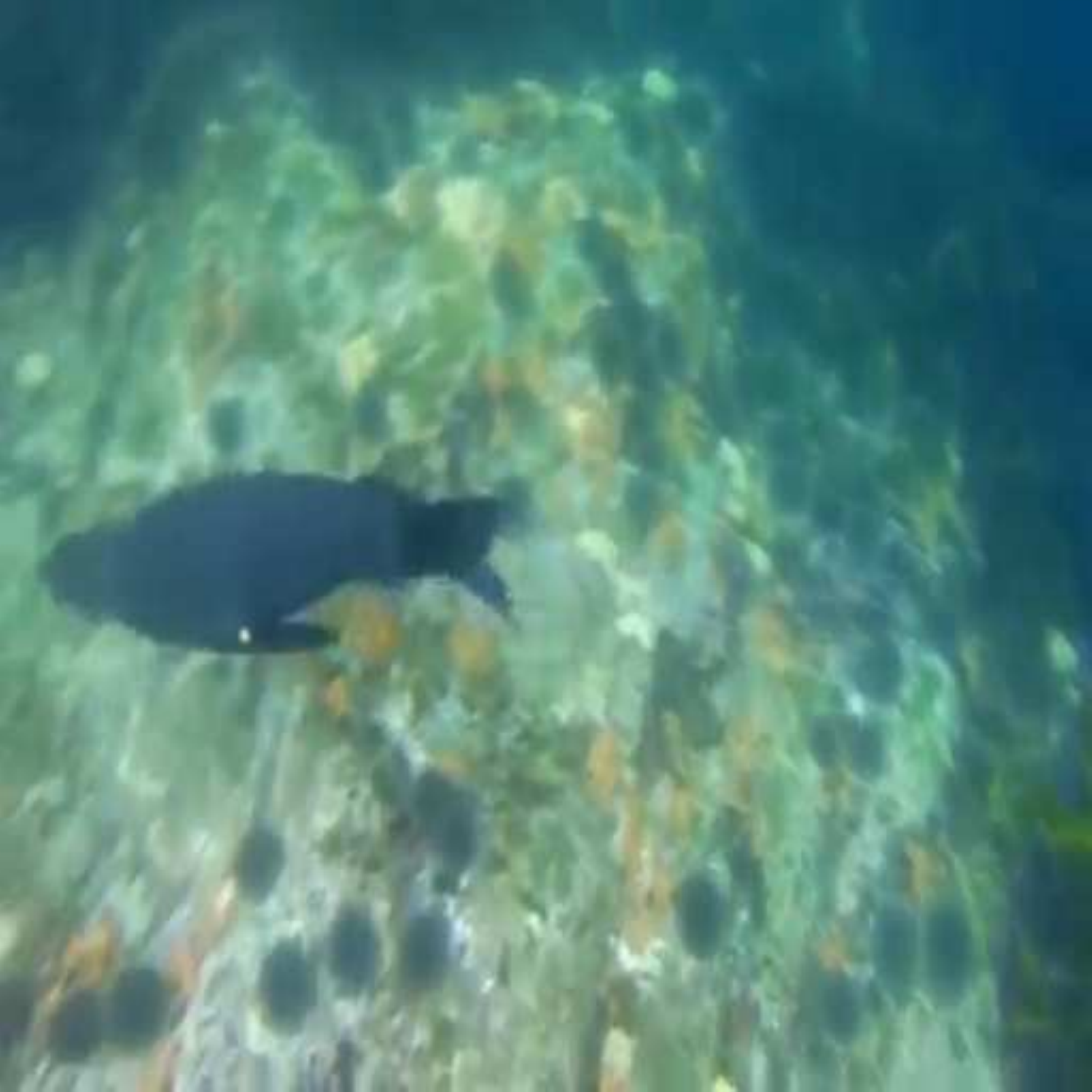}}
\hspace{-0.7mm}
\subfloat{\includegraphics[width=0.095\textwidth]{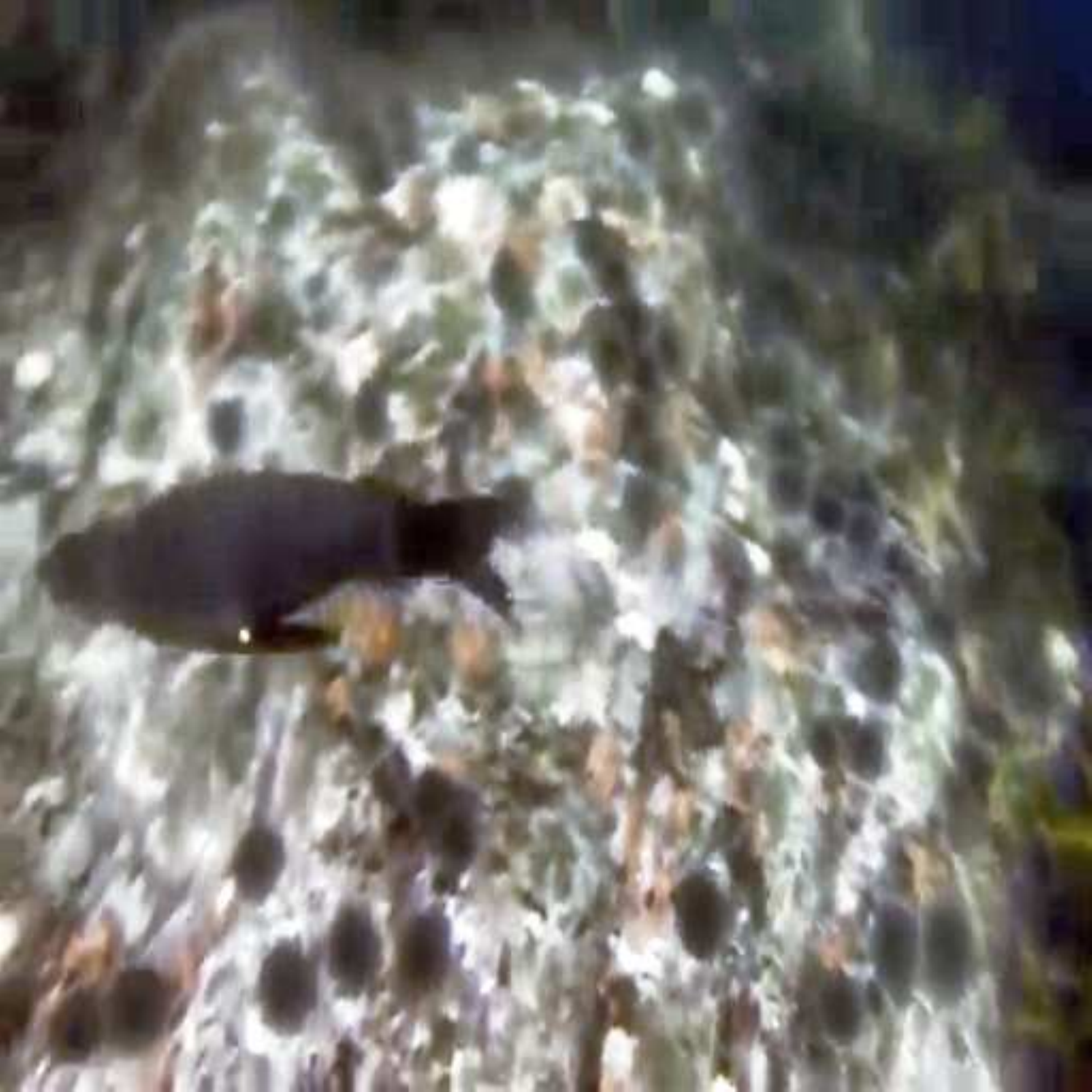}}
\hspace{-0.7mm}
\subfloat{\includegraphics[width=0.095\textwidth]{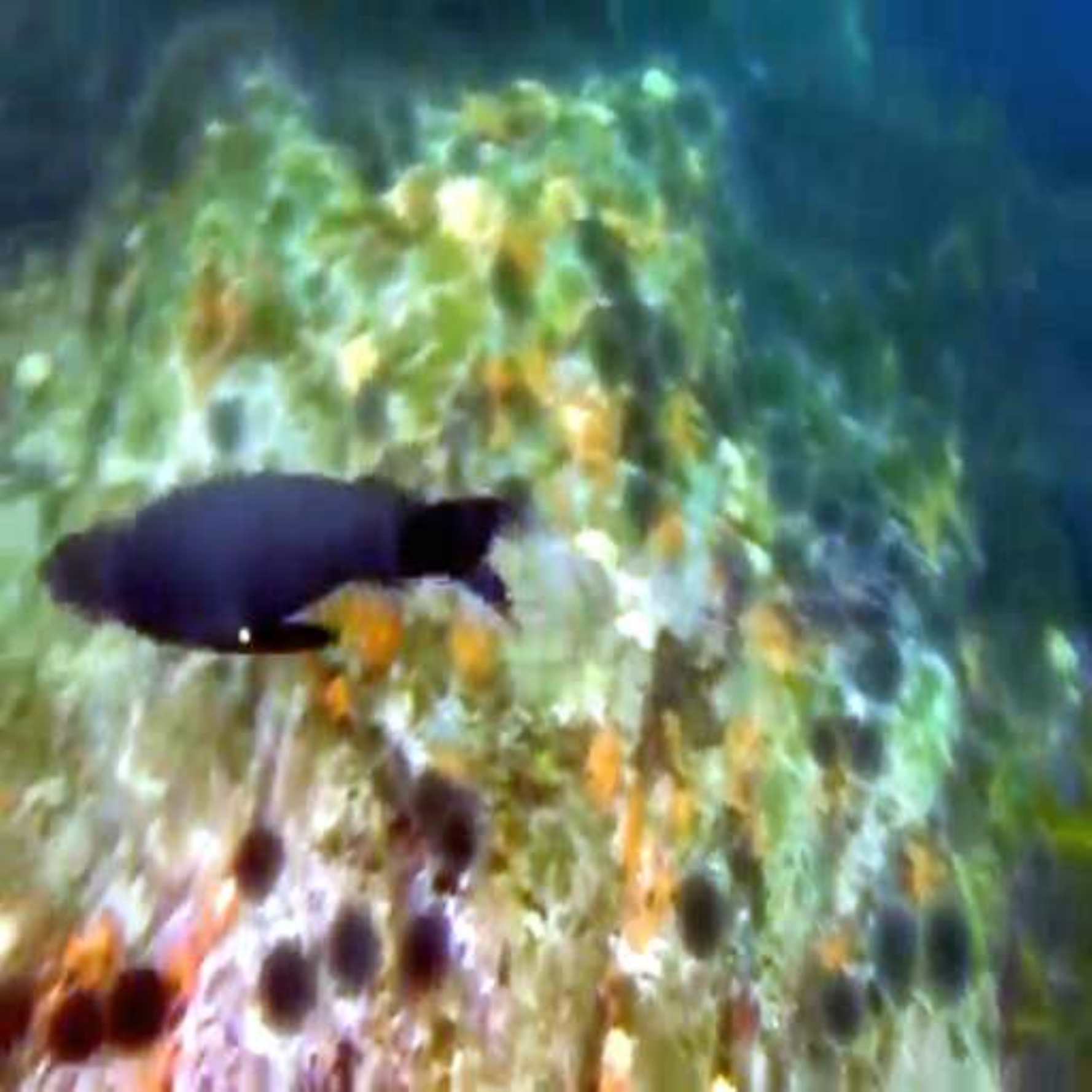}}
\hspace{-0.7mm}
\subfloat{\includegraphics[width=0.095\textwidth]{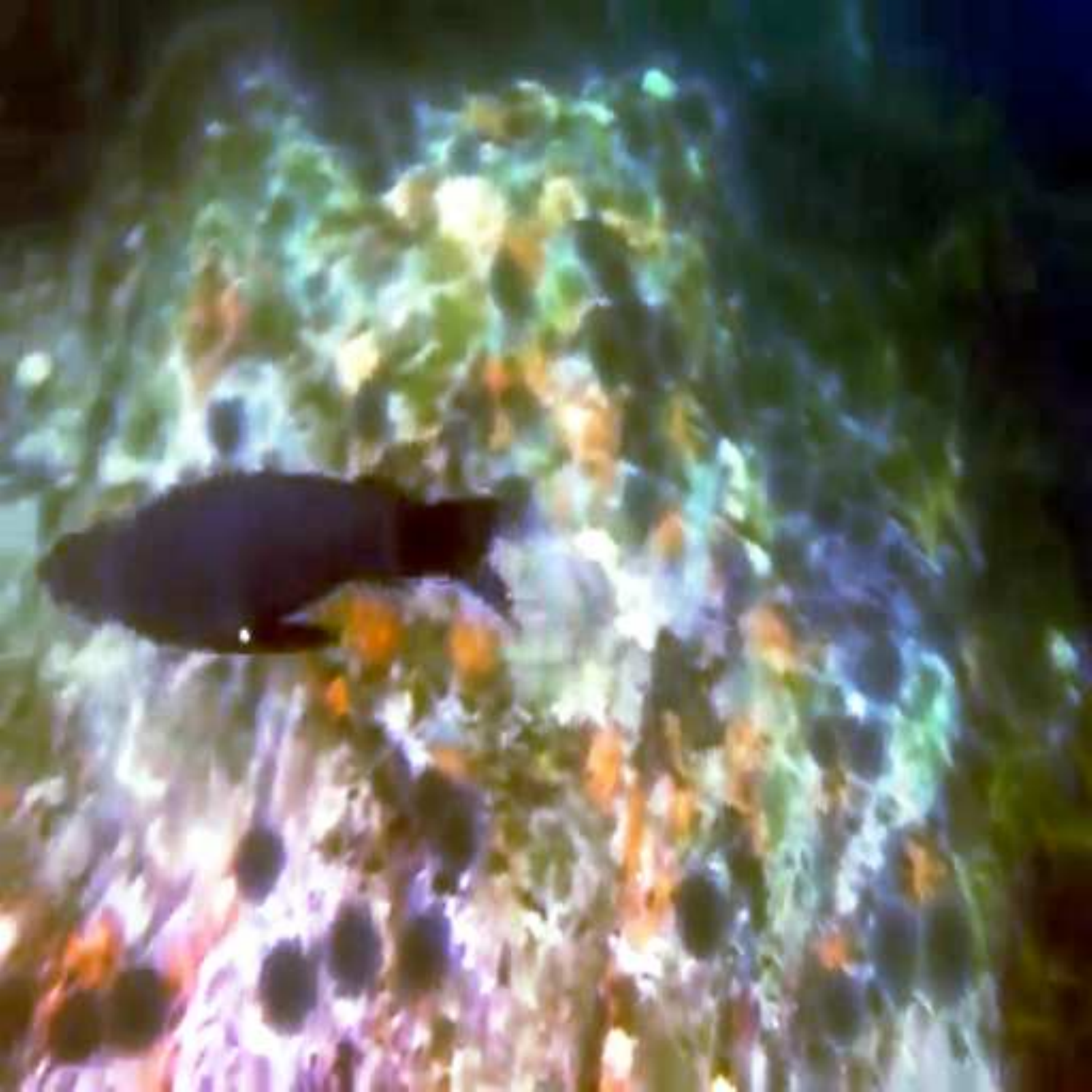}}
\hspace{-0.7mm}
\subfloat{\includegraphics[width=0.095\textwidth]{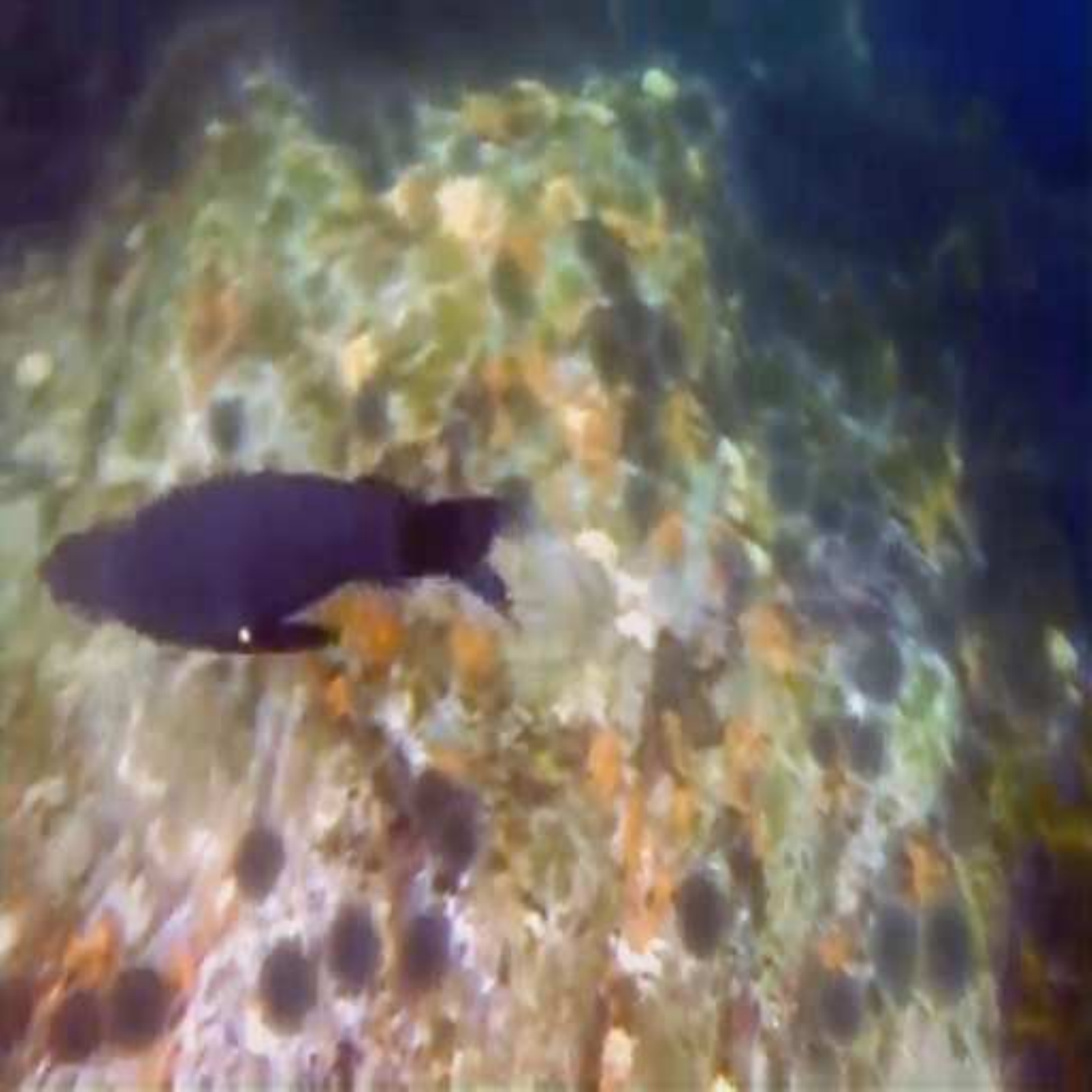}}
\hspace{-0.7mm}
\subfloat{\includegraphics[width=0.095\textwidth]{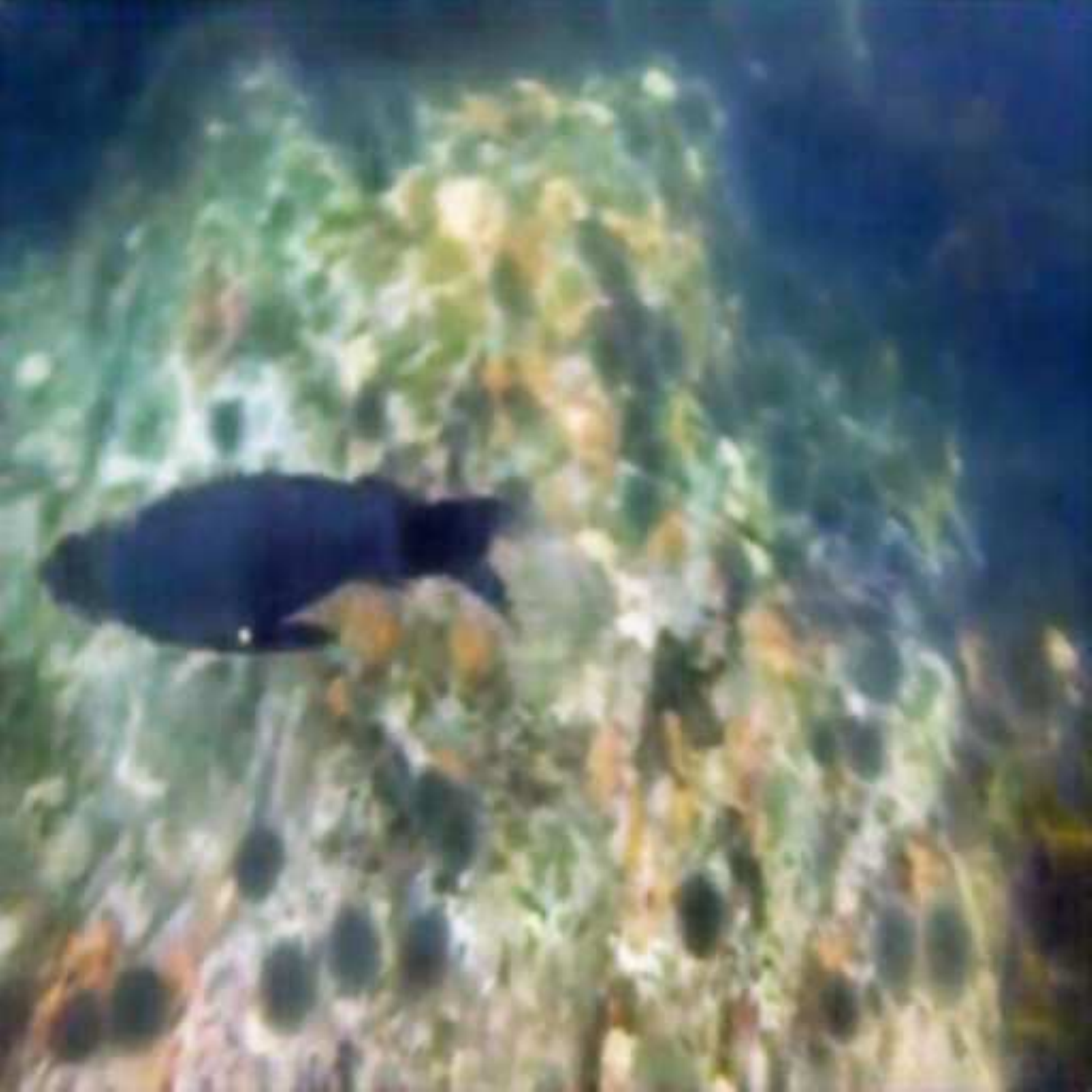}}
\hspace{-0.7mm}
\subfloat{\includegraphics[width=0.095\textwidth]{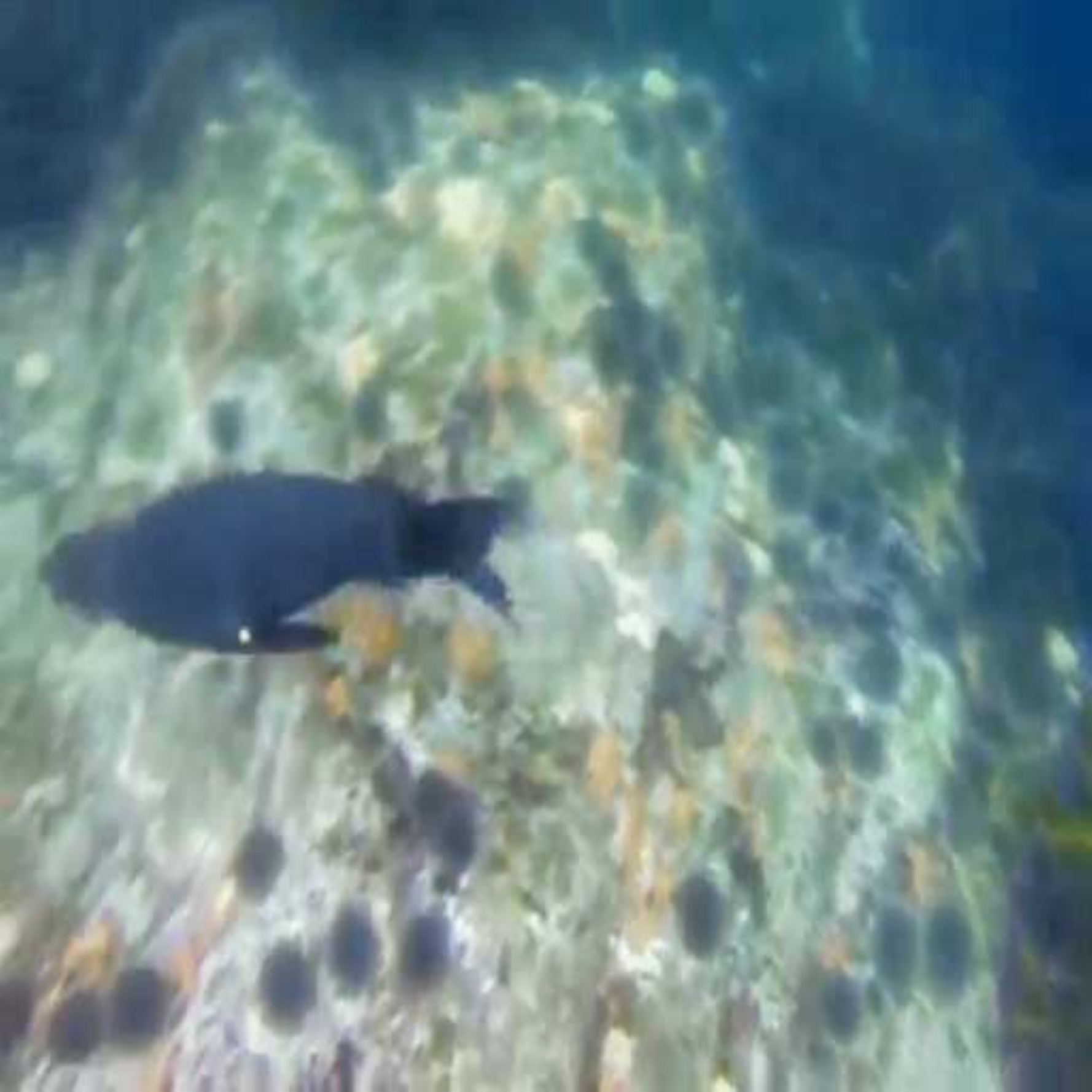}}
\hspace{-0.7mm}
\subfloat{\includegraphics[width=0.095\textwidth]{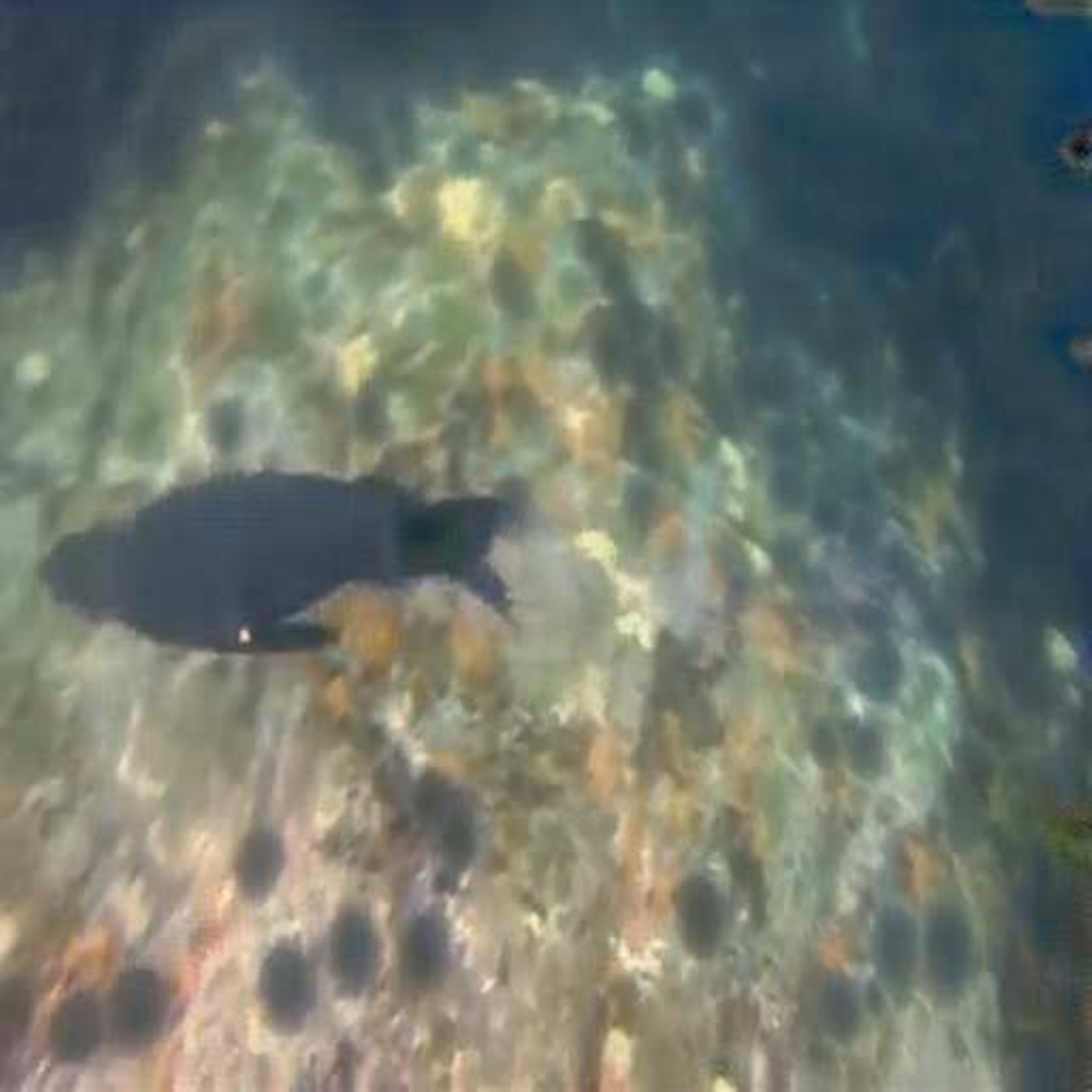}}
\hspace{-0.7mm}
\subfloat{\includegraphics[width=0.095\textwidth]{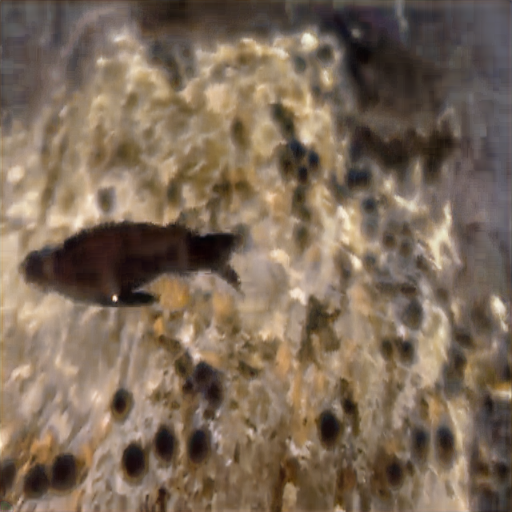}}
\hspace{-0.7mm}
\subfloat{\includegraphics[width=0.095\textwidth]{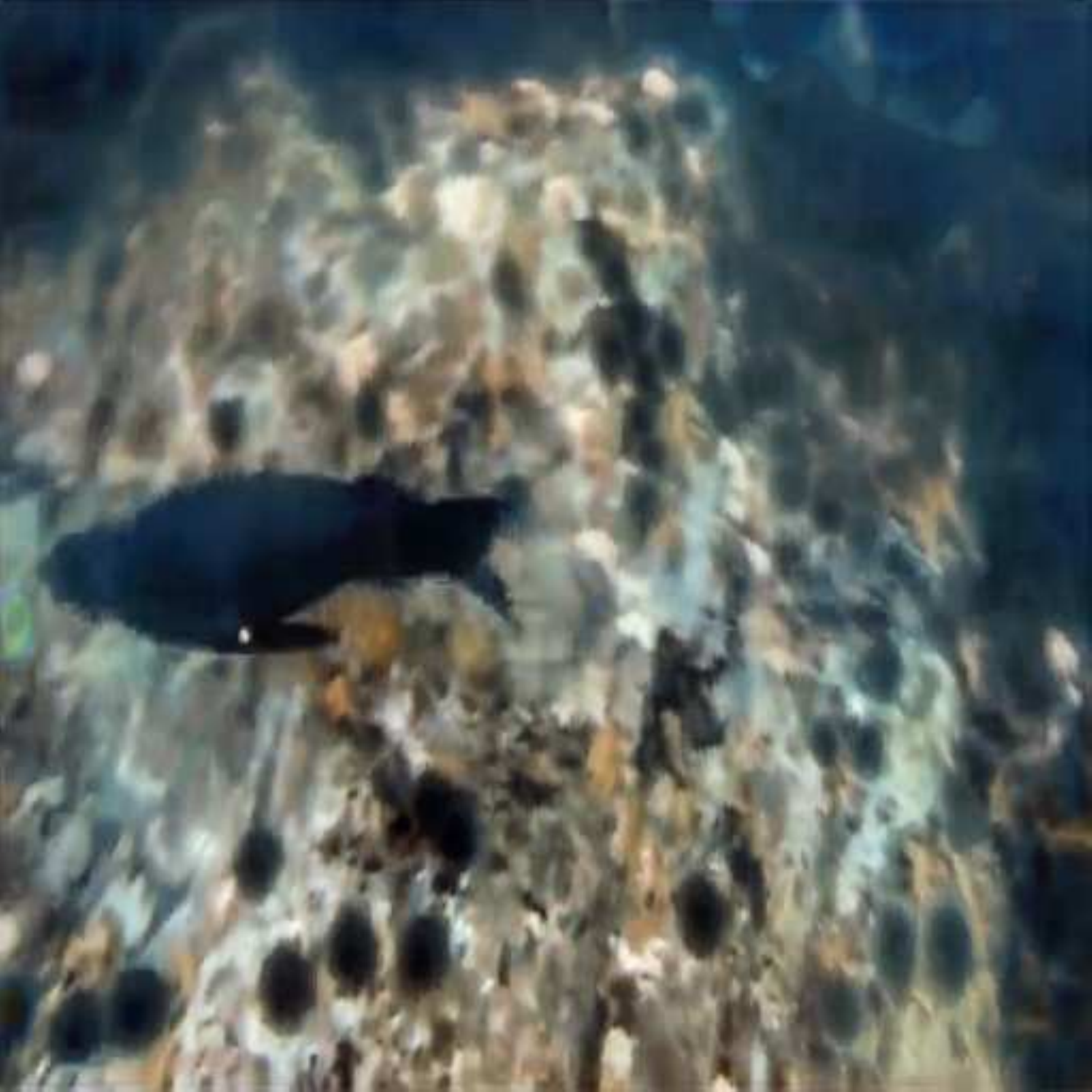}}

\vspace{0.5mm}
\hspace{-5mm}
\subfloat{\includegraphics[width=0.095\textwidth]{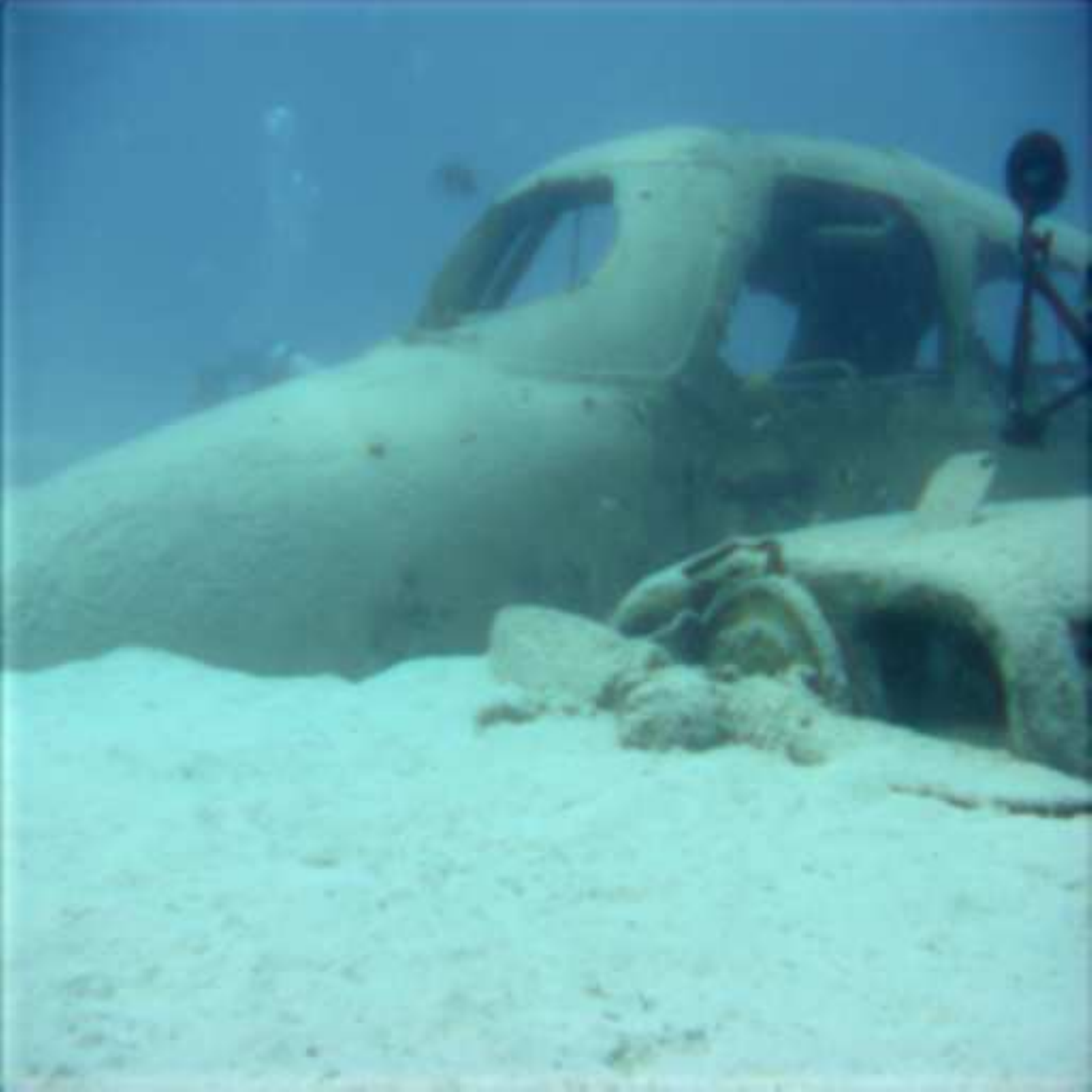}}
\hspace{-0.7mm}
\subfloat{\includegraphics[width=0.095\textwidth]{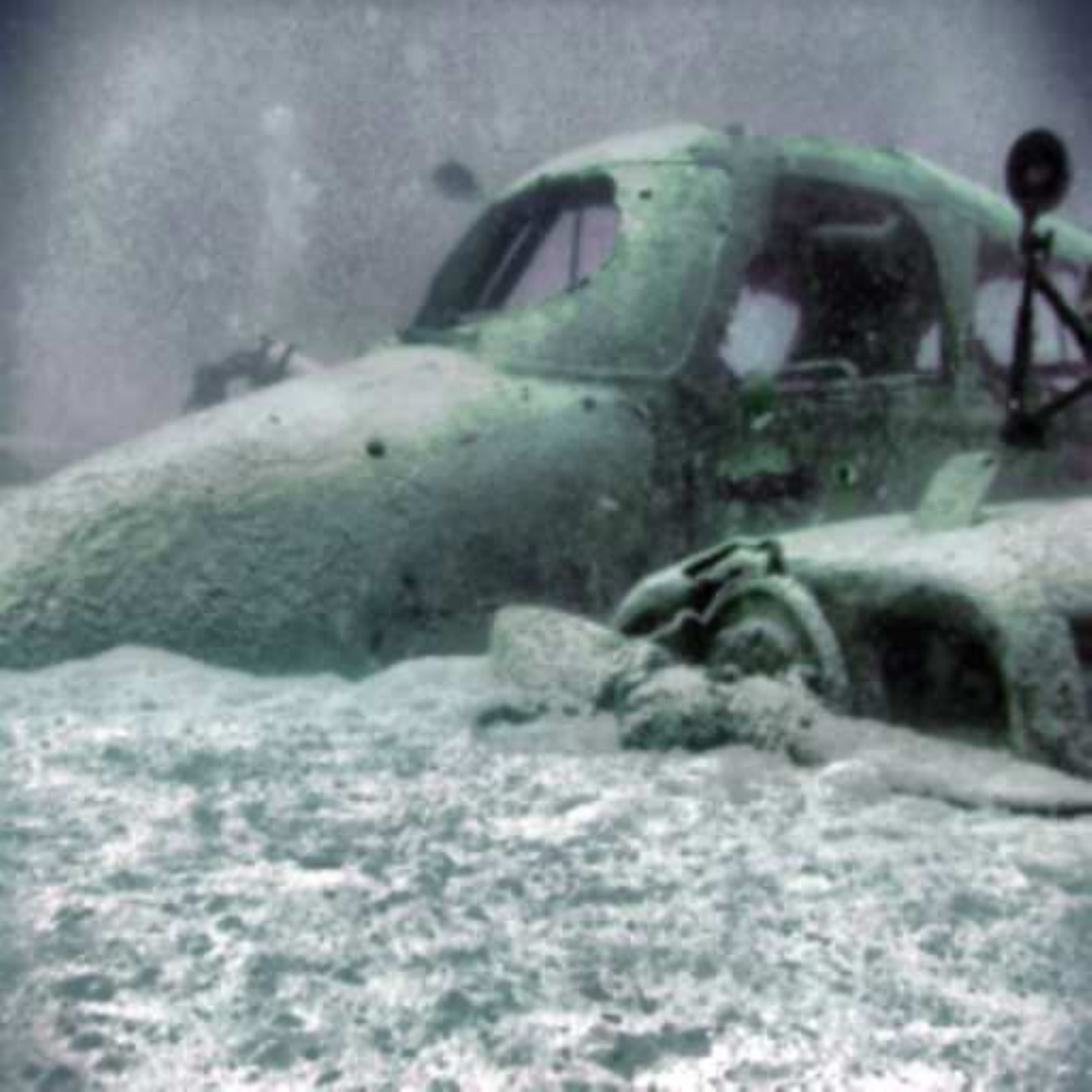}}
\hspace{-0.7mm}
\subfloat{\includegraphics[width=0.095\textwidth]{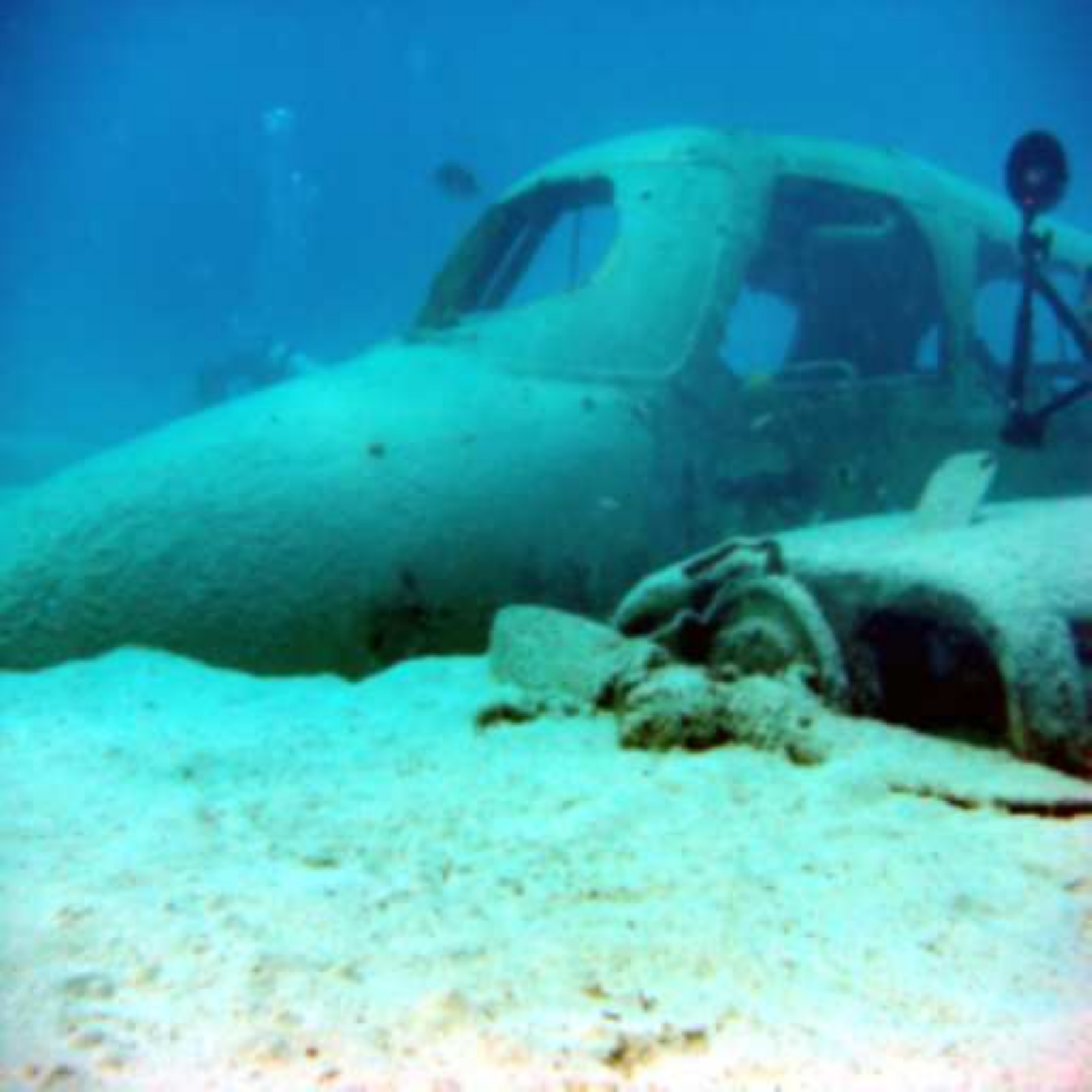}}
\hspace{-0.7mm}
\subfloat{\includegraphics[width=0.095\textwidth]{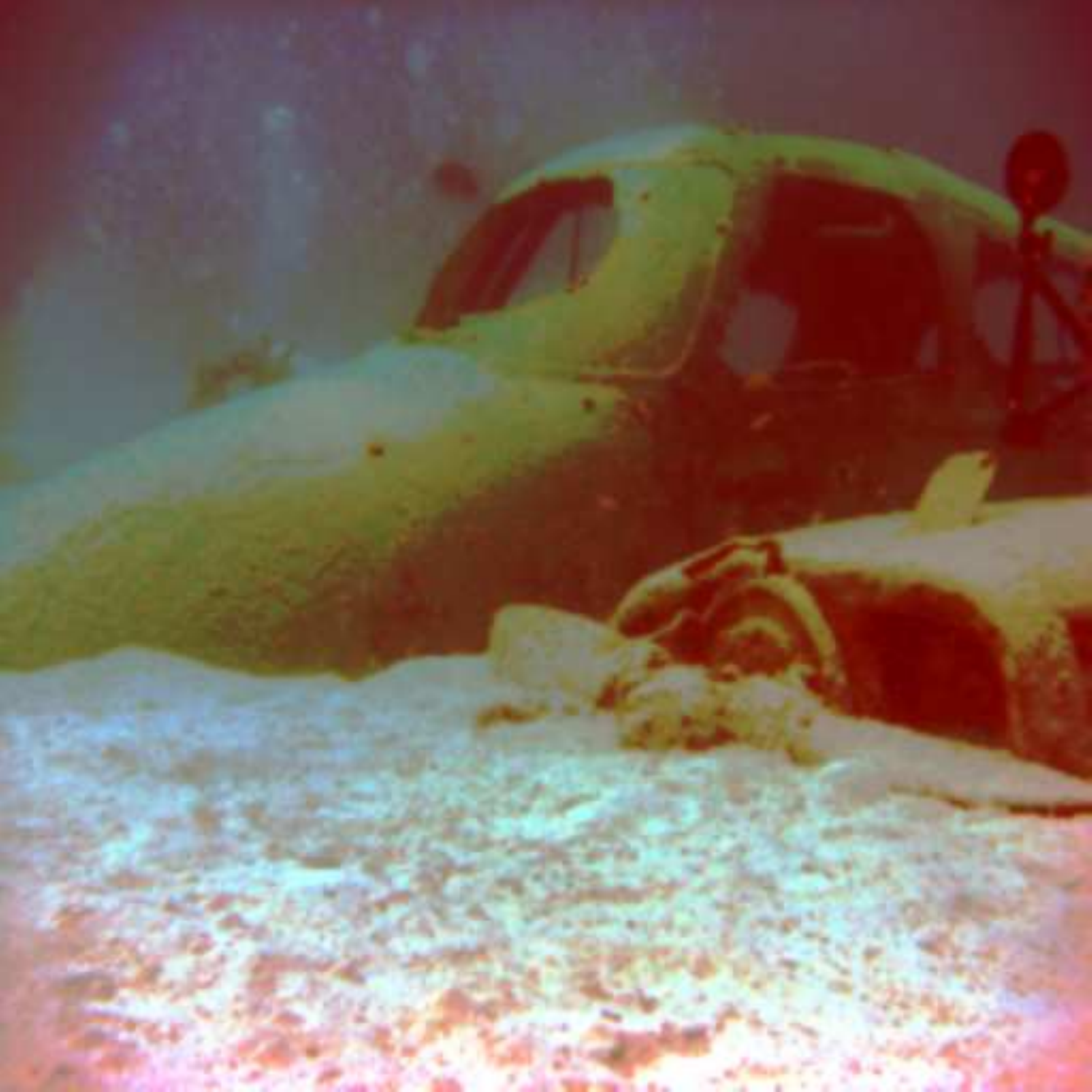}}
\hspace{-0.7mm}
\subfloat{\includegraphics[width=0.095\textwidth]{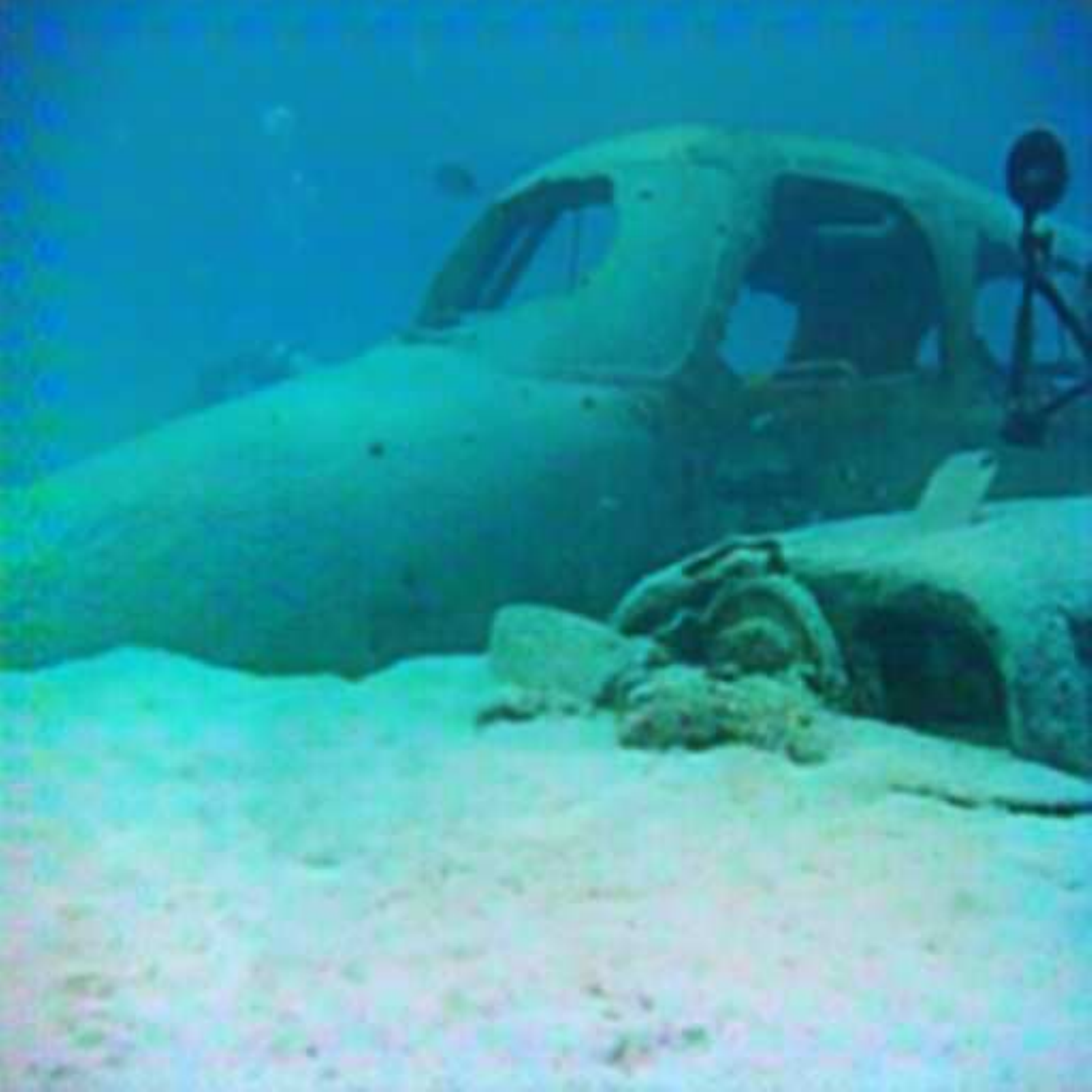}}
\hspace{-0.7mm}
\subfloat{\includegraphics[width=0.095\textwidth]{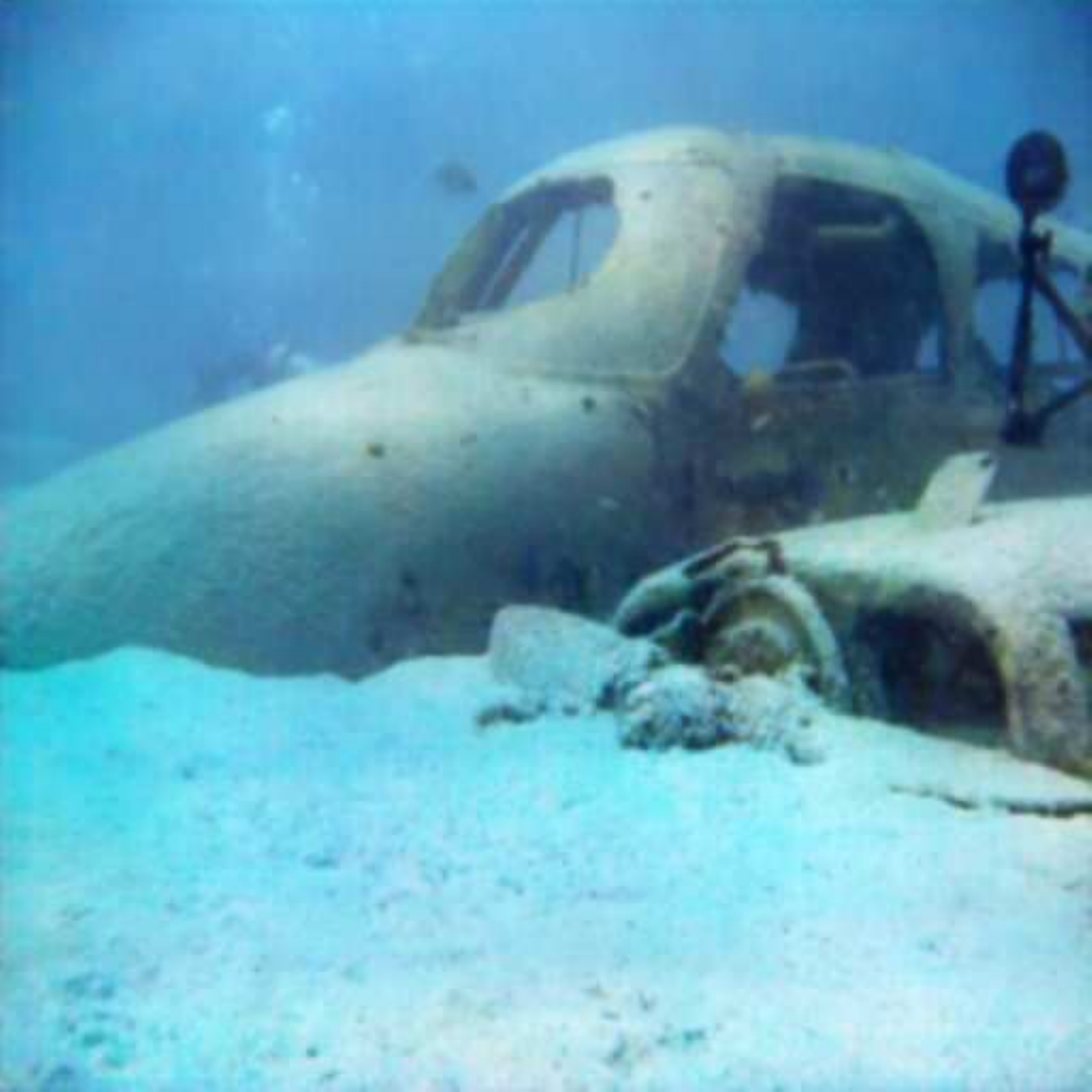}}
\hspace{-0.7mm}
\subfloat{\includegraphics[width=0.095\textwidth]{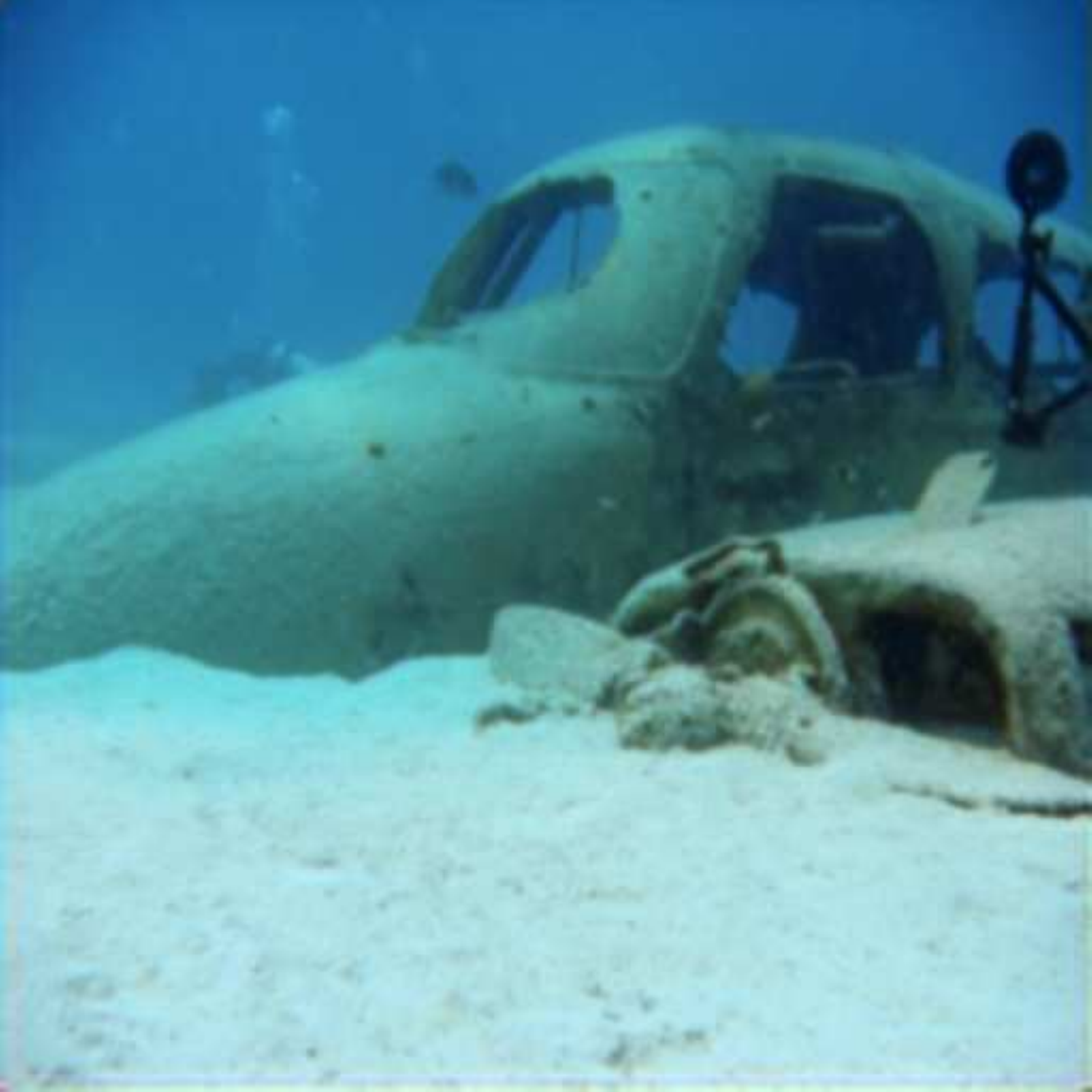}}
\hspace{-0.7mm}
\subfloat{\includegraphics[width=0.095\textwidth]{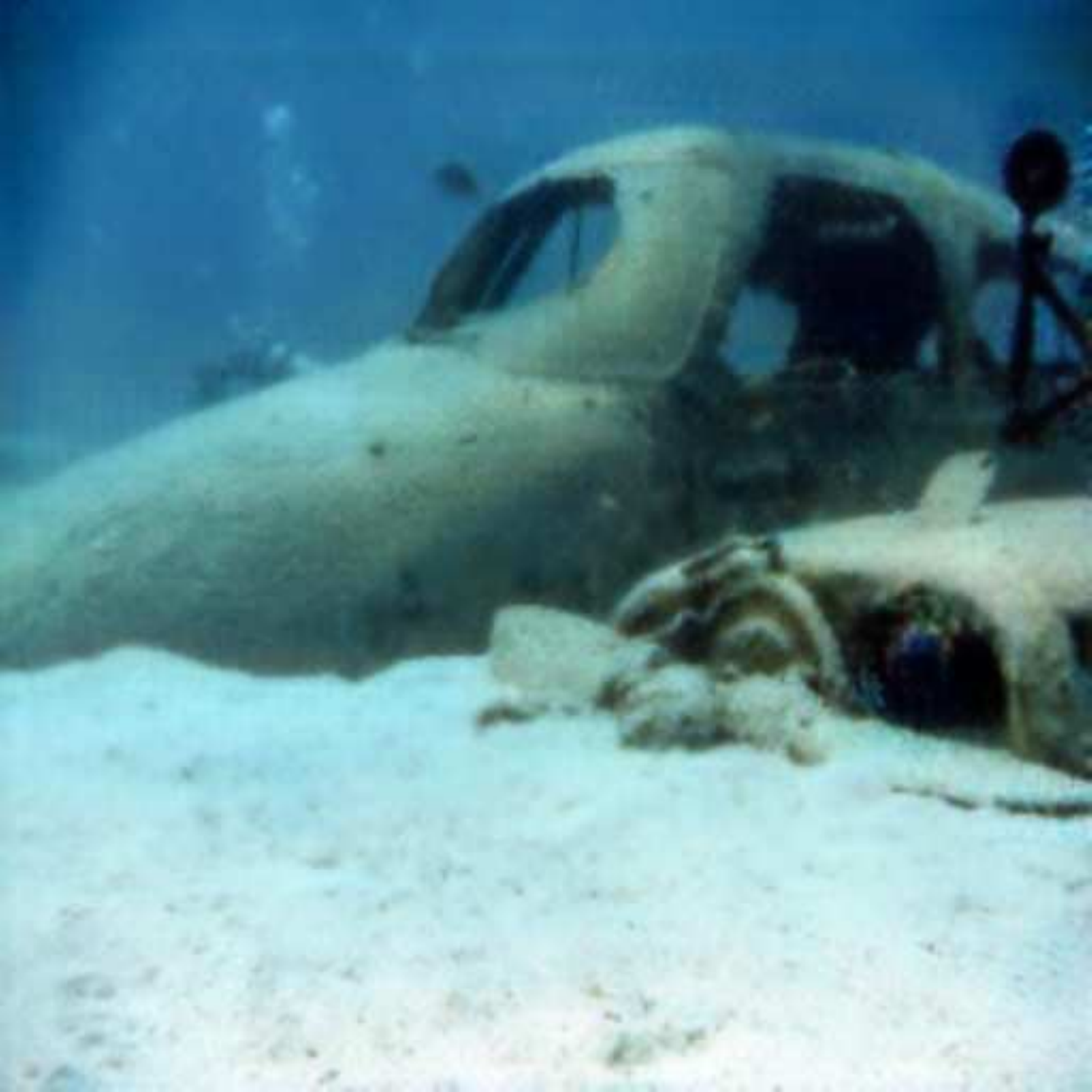}}
\hspace{-0.7mm}
\subfloat{\includegraphics[width=0.095\textwidth]{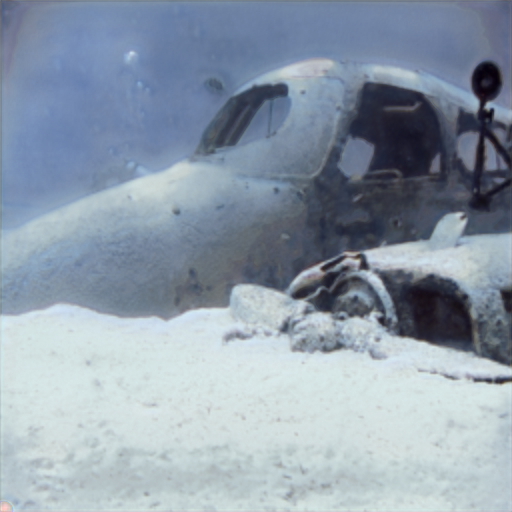}}
\hspace{-0.7mm}
\subfloat{\includegraphics[width=0.095\textwidth]{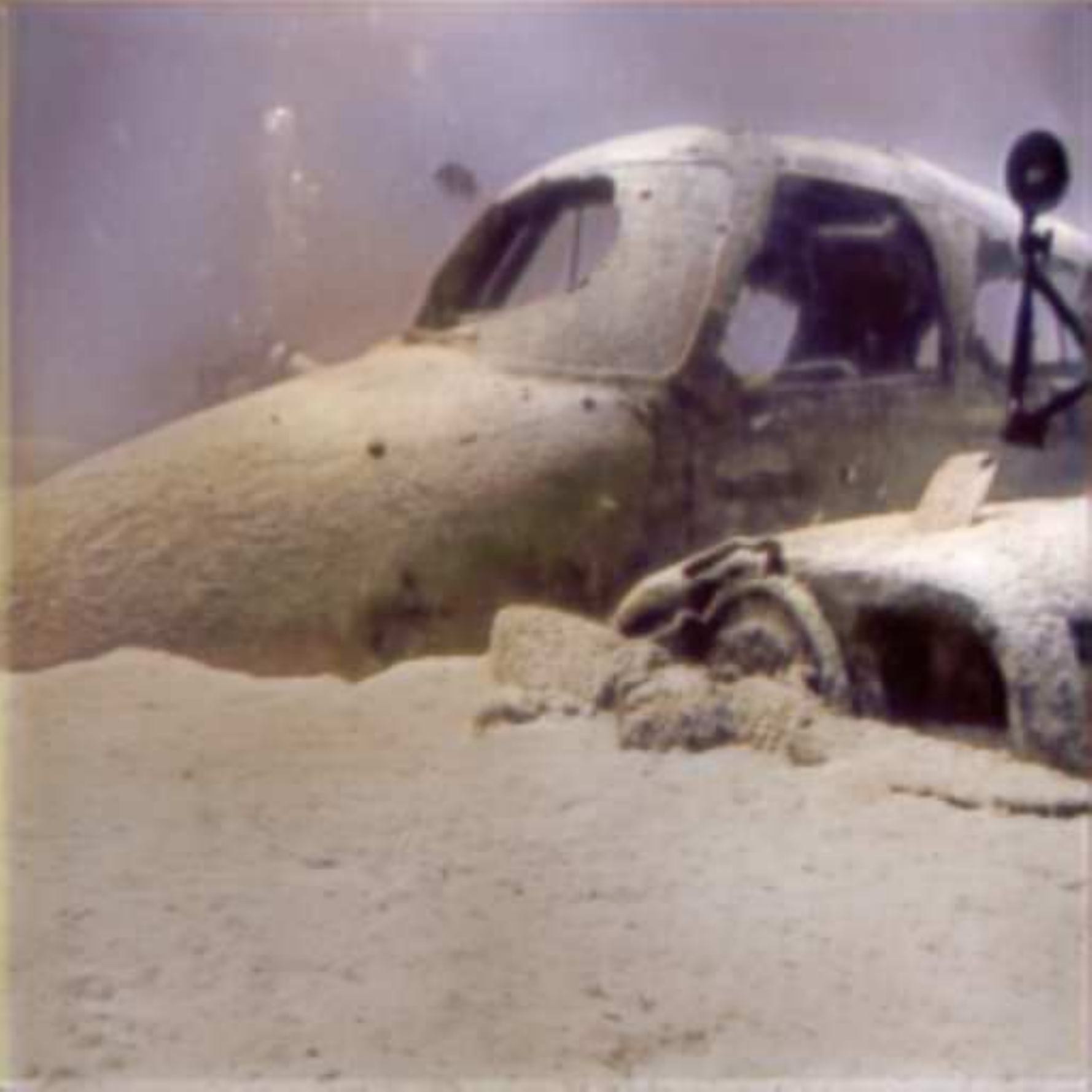}}

\vspace{0.5mm}
\hspace{-5.8mm}
\setcounter{subfigure}{0}
\subfloat[\small{Input}]{\includegraphics[width=0.095\textwidth]{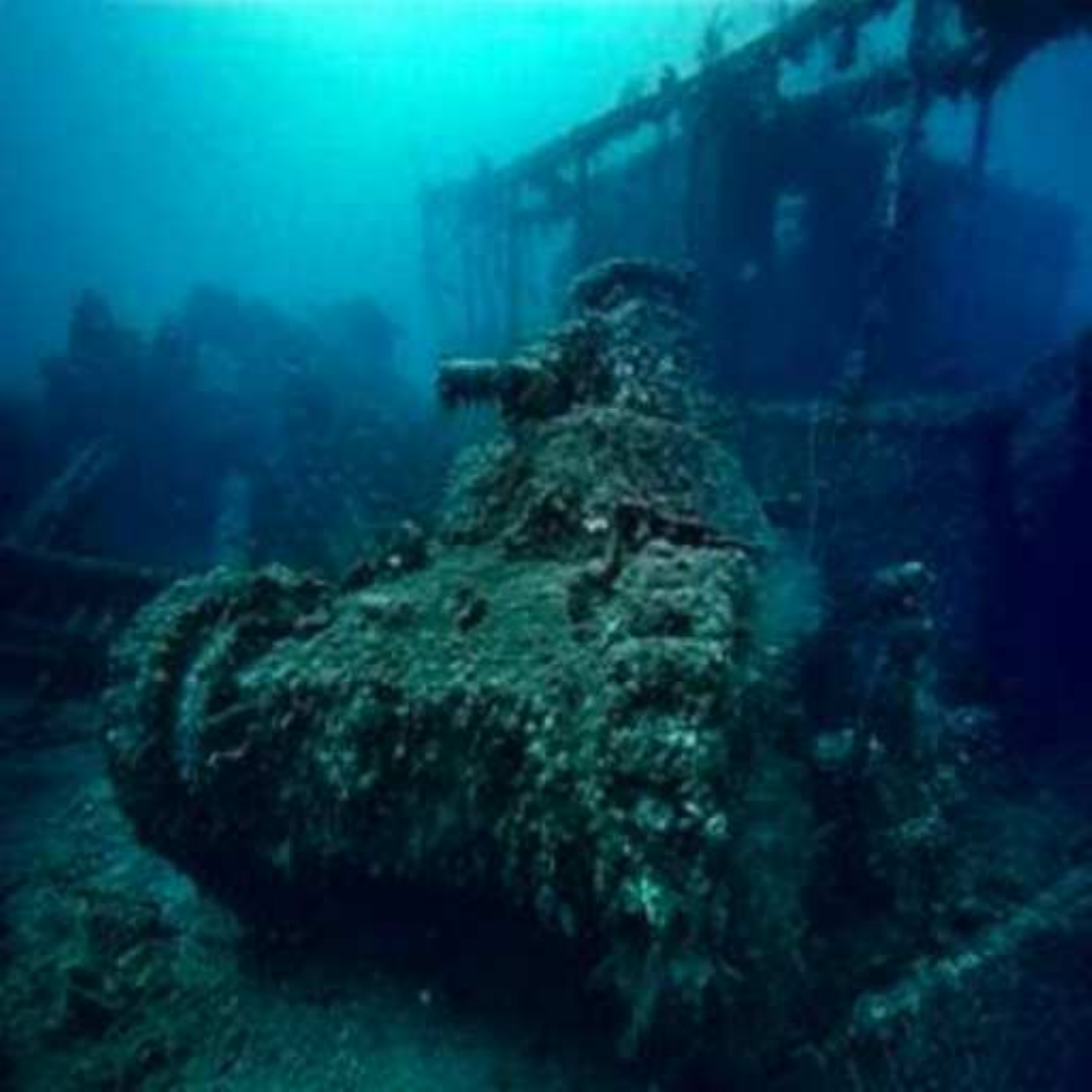}}
\hspace{-0.7mm}
\subfloat[\small{ACDC}]{\includegraphics[width=0.095\textwidth]{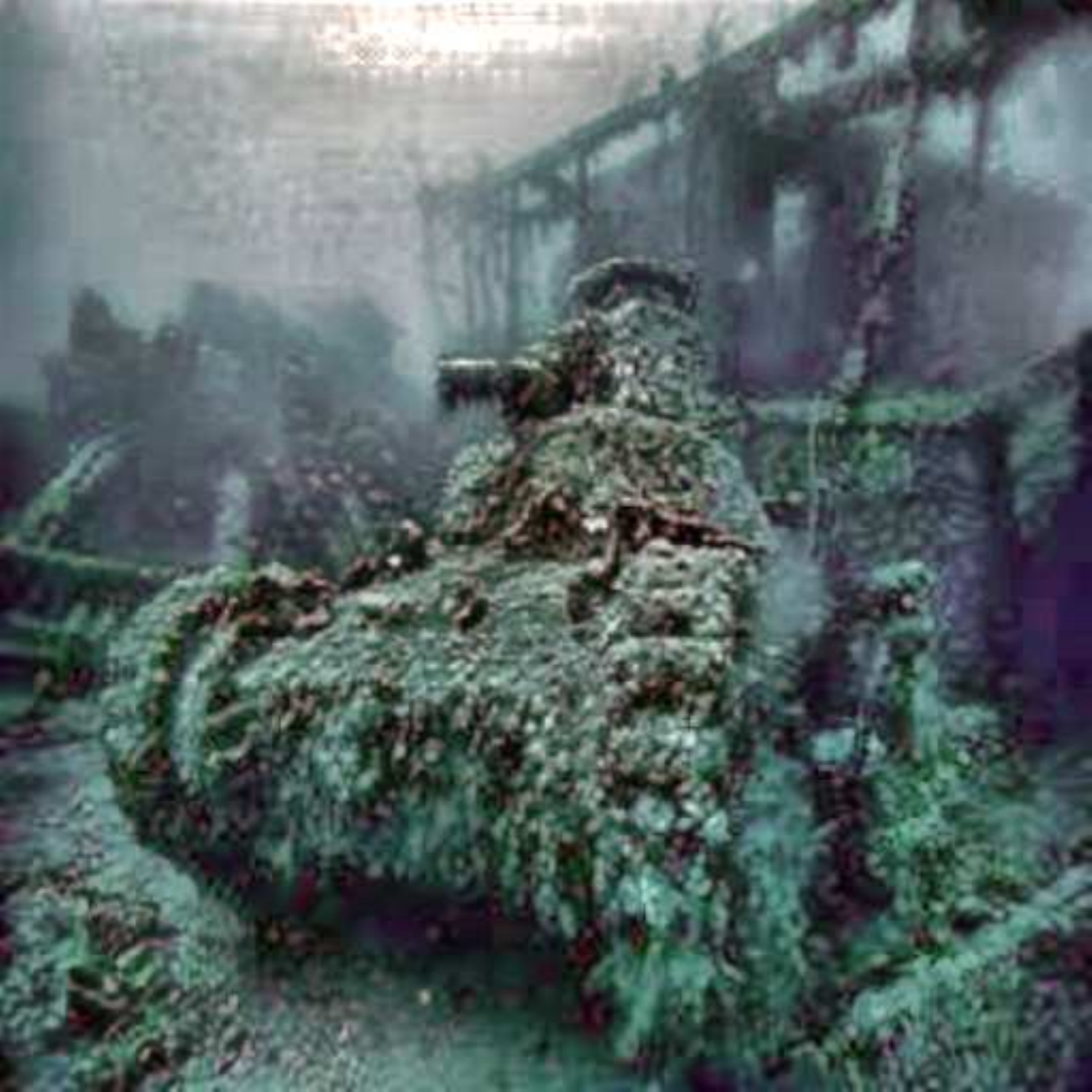}}
\hspace{-0.7mm}
\subfloat[\small{Rank1}]{\includegraphics[width=0.095\textwidth]{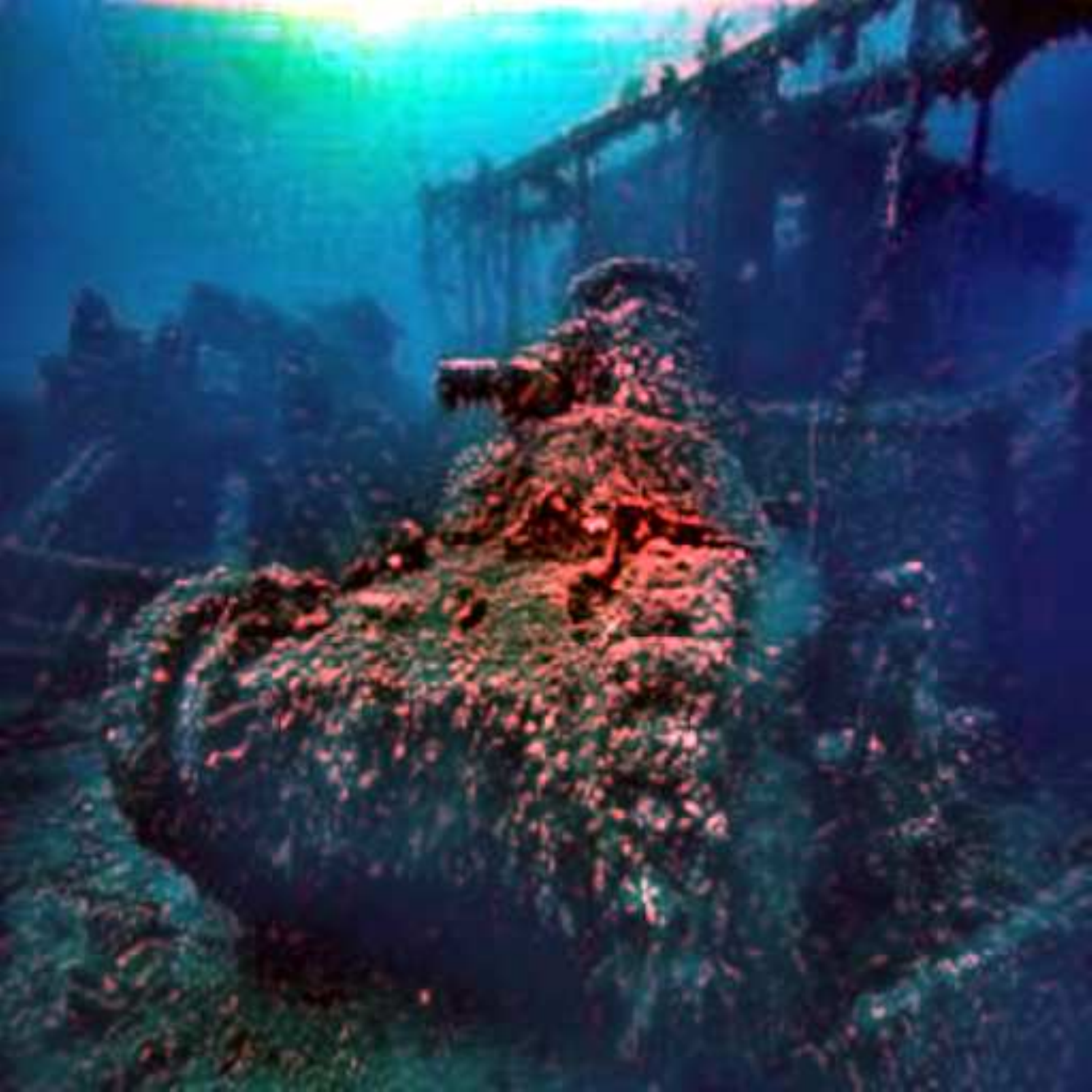}}
\hspace{-0.7mm}
\subfloat[\small{MMLE}]{\includegraphics[width=0.095\textwidth]{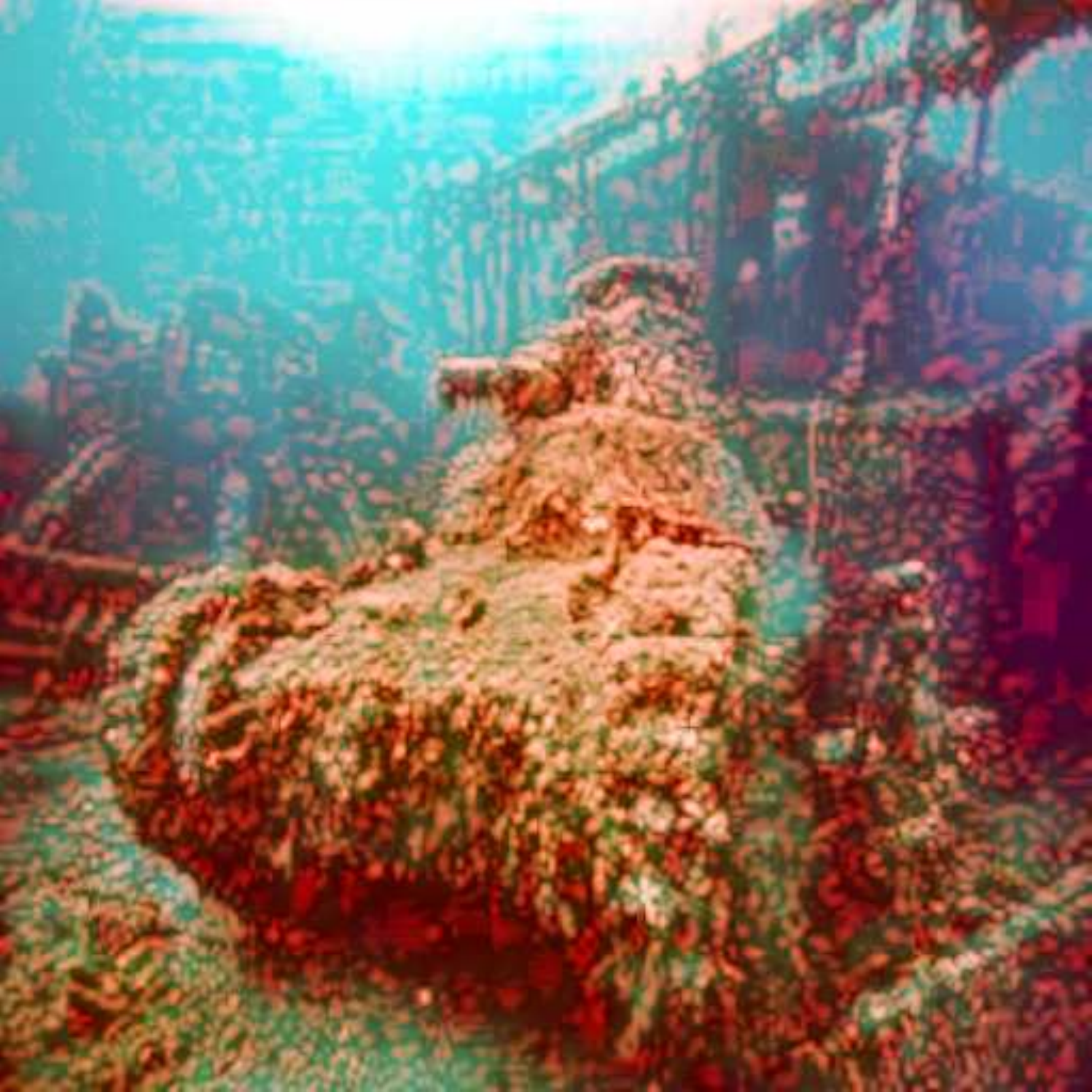}}
\hspace{-0.7mm}
\subfloat[\small{Water}]{\includegraphics[width=0.095\textwidth]{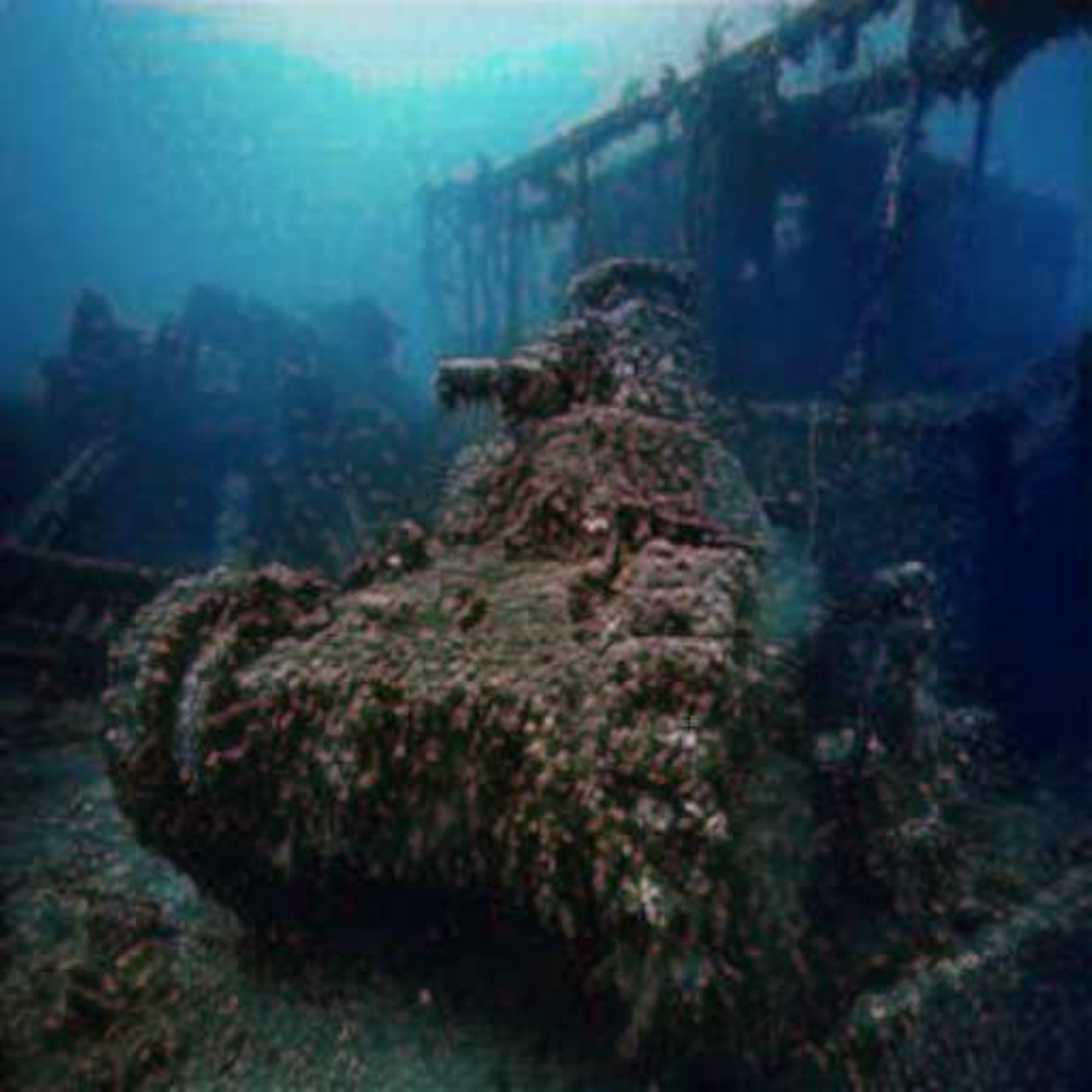}}
\hspace{-0.7mm}
\subfloat[\small{Funie}]{\includegraphics[width=0.095\textwidth]{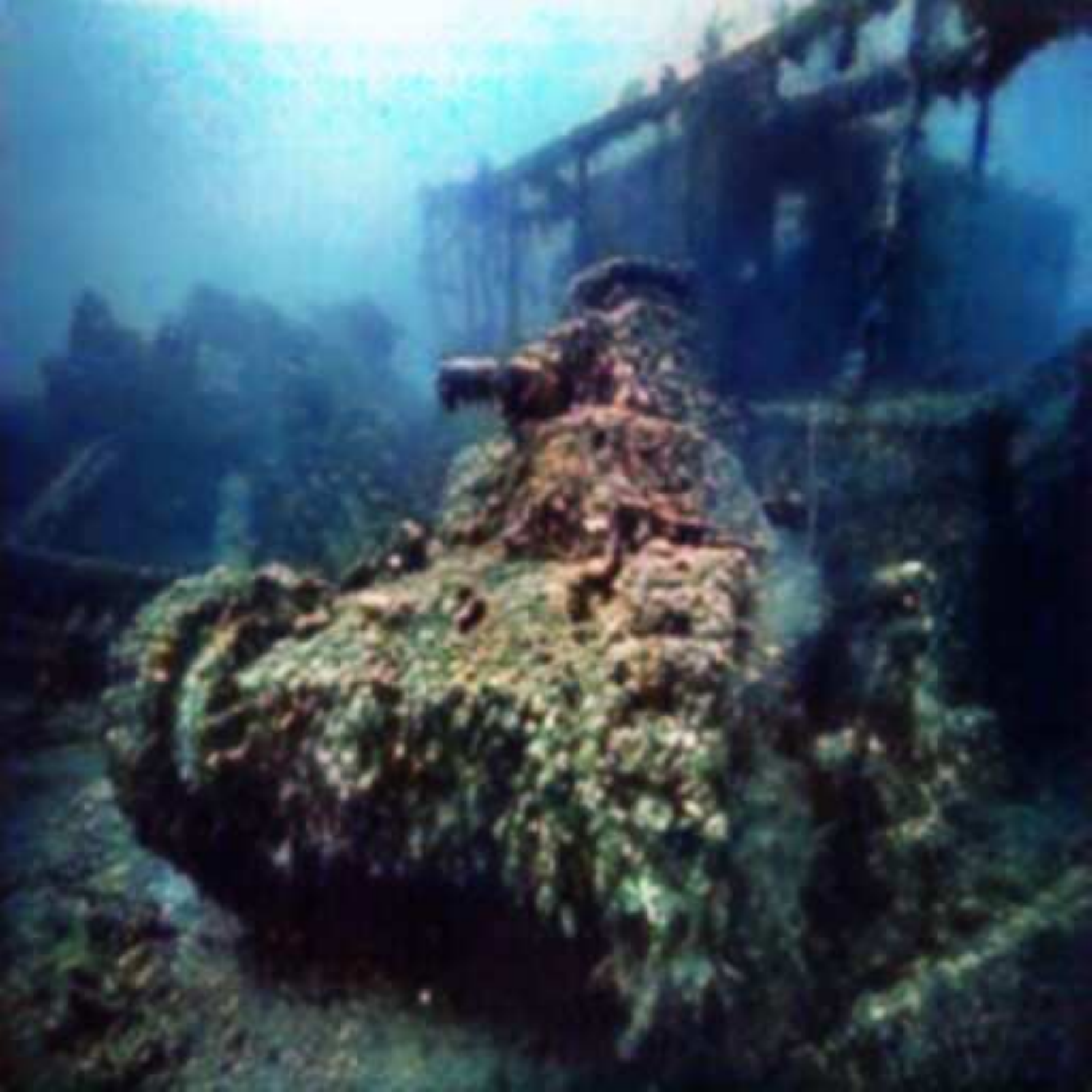}}
\hspace{-0.7mm}
\subfloat[\small{Ucolor}]{\includegraphics[width=0.095\textwidth]{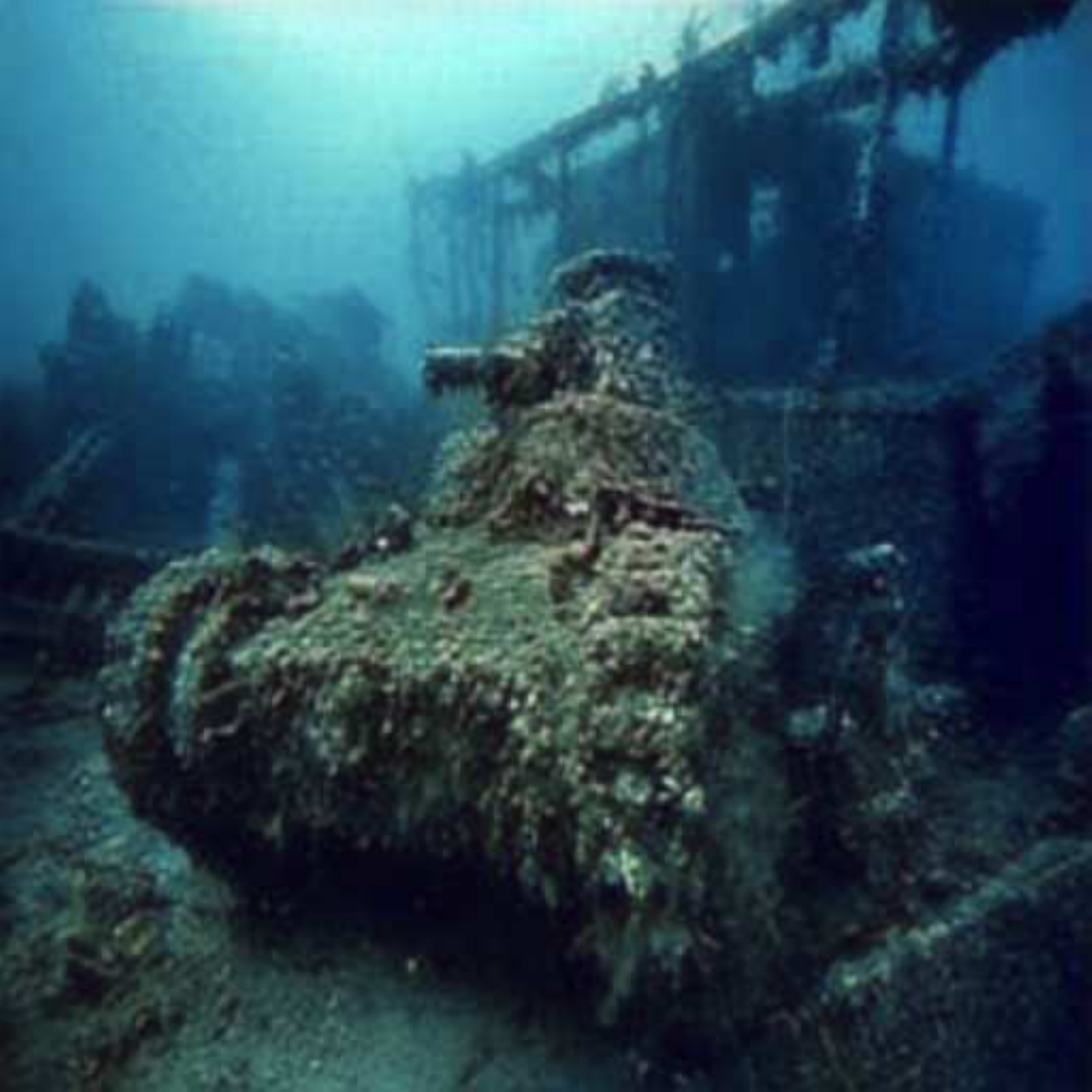}}
\hspace{-0.7mm}
\subfloat[\small{TACL}]{\includegraphics[width=0.095\textwidth]{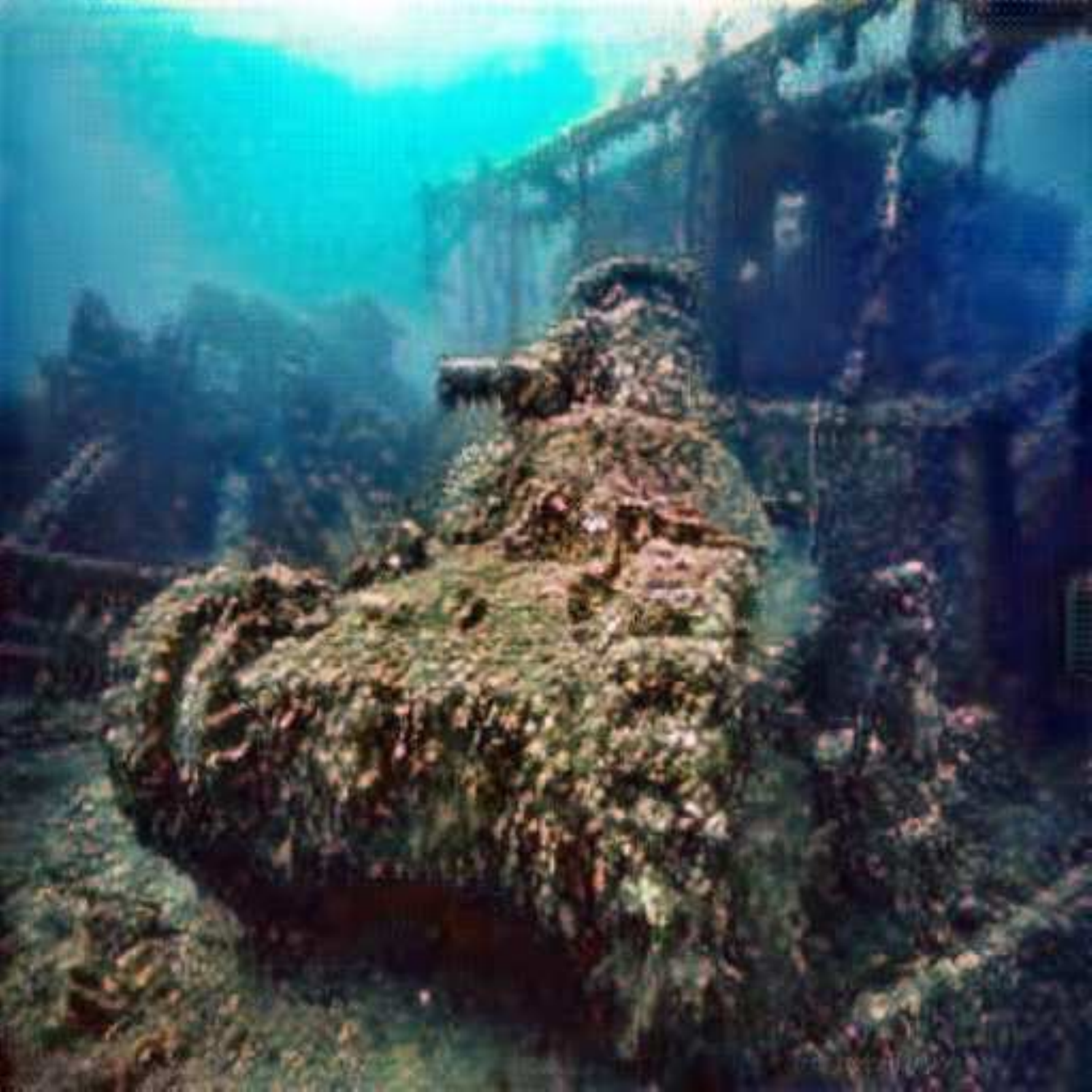}}
\hspace{-0.7mm}
\subfloat[\footnotesize{Pretrained}]{\includegraphics[width=0.095\textwidth]{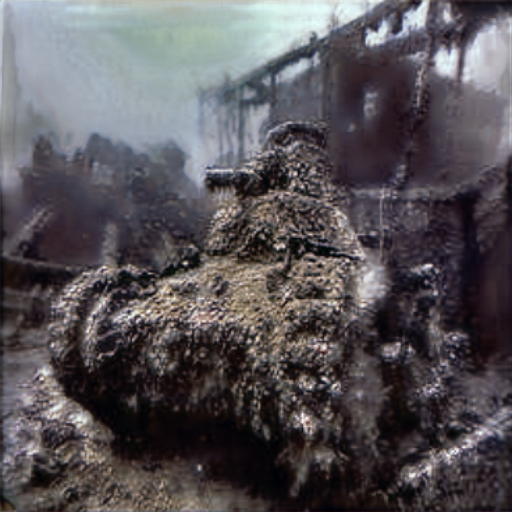}}
\hspace{-0.7mm}
\subfloat[\small{MetaUE}]{\includegraphics[width=0.095\textwidth]{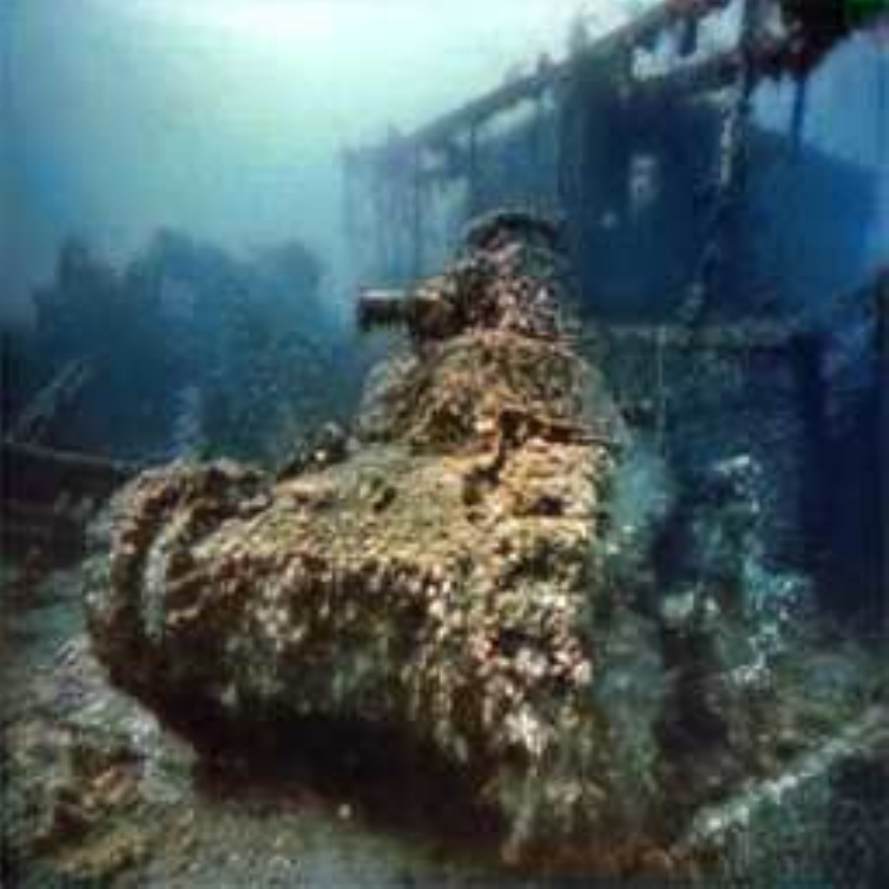}}

\caption{The visual comparison among the enhancement methods on the unpaired underwater images, where our MetaUE is fine-tuned on the EUVP dataset. }
\label{no reference}
\end{figure*}

\begin{figure*}[htbp]
\centering
\subfloat{\includegraphics[width=0.15\textwidth]{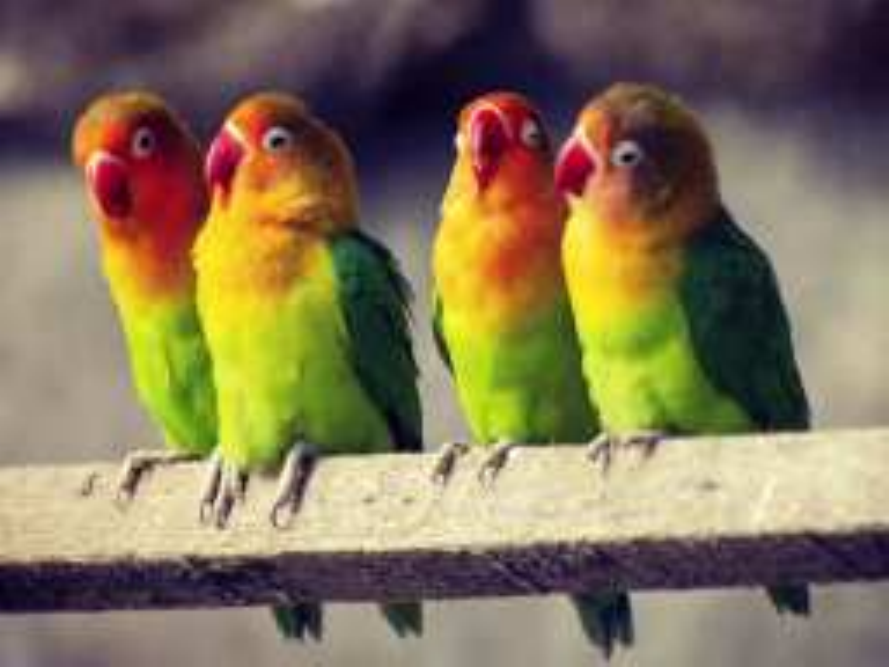}}
\hspace{-0.7mm}
\subfloat{\includegraphics[width=0.15\textwidth]{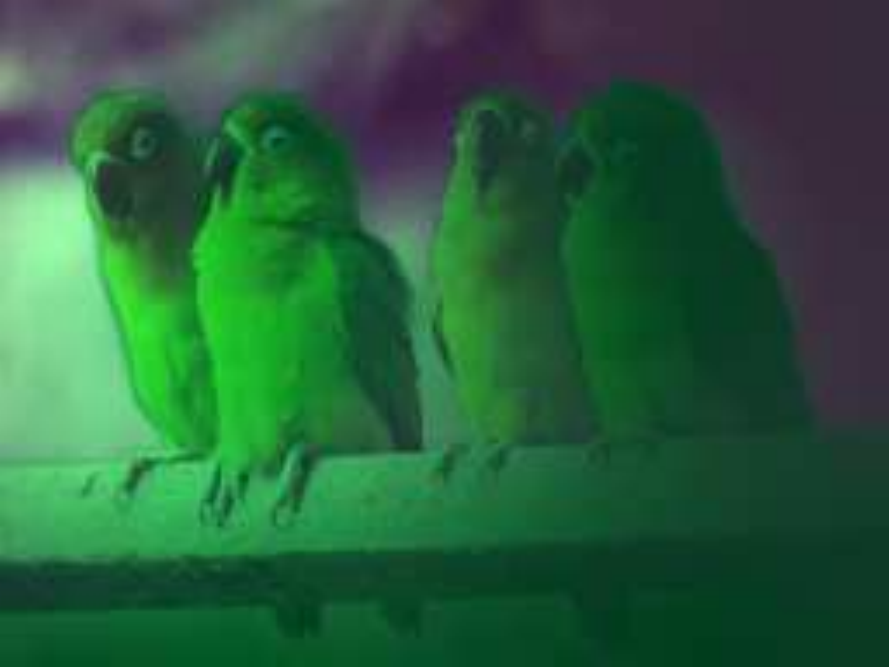}}
\hspace{-0.7mm}
\subfloat{\includegraphics[width=0.15\textwidth]{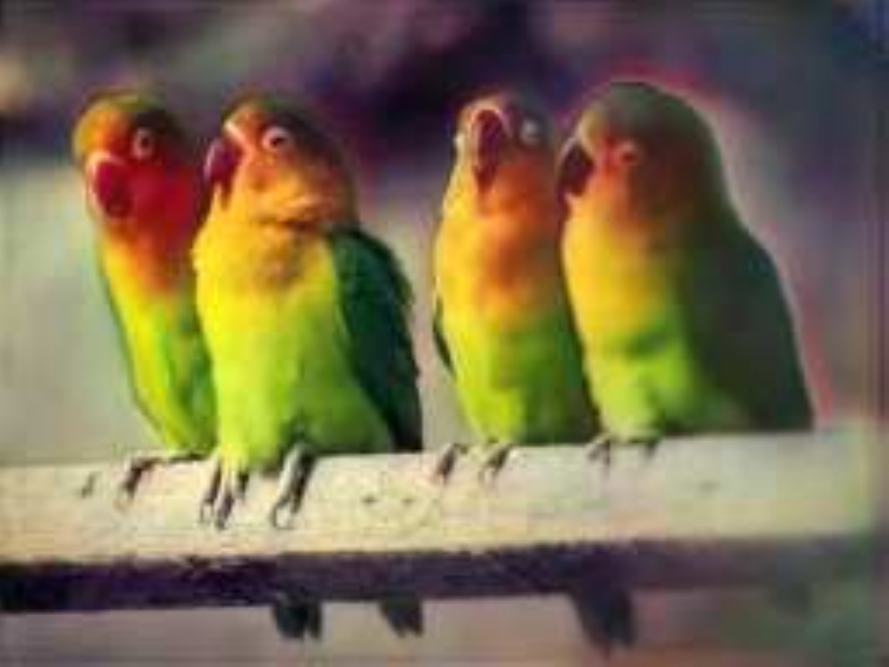}}
\hspace{-0.7mm}
\subfloat{\includegraphics[width=0.15\textwidth]{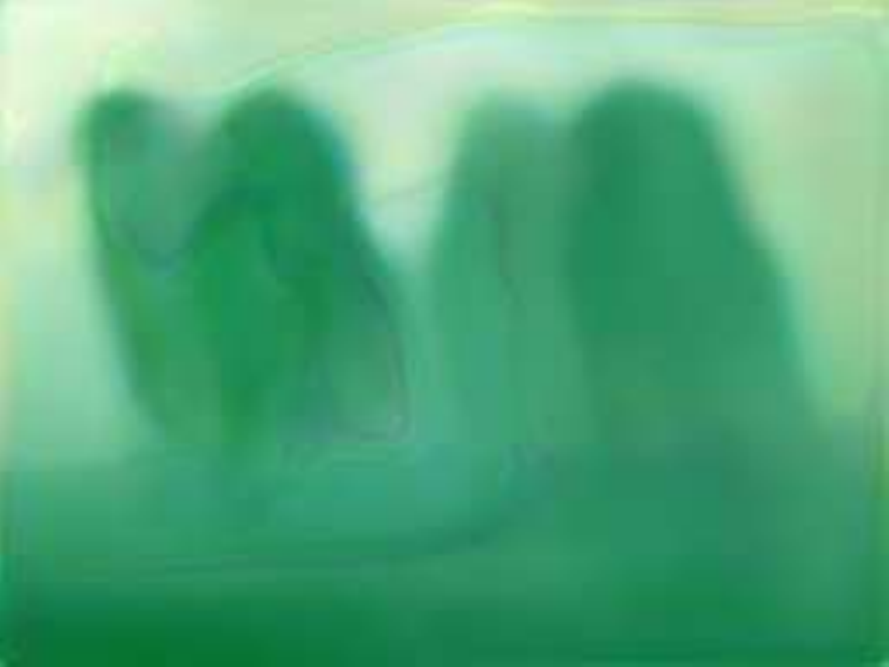}}
\hspace{-0.7mm}
\subfloat{\includegraphics[width=0.15\textwidth]{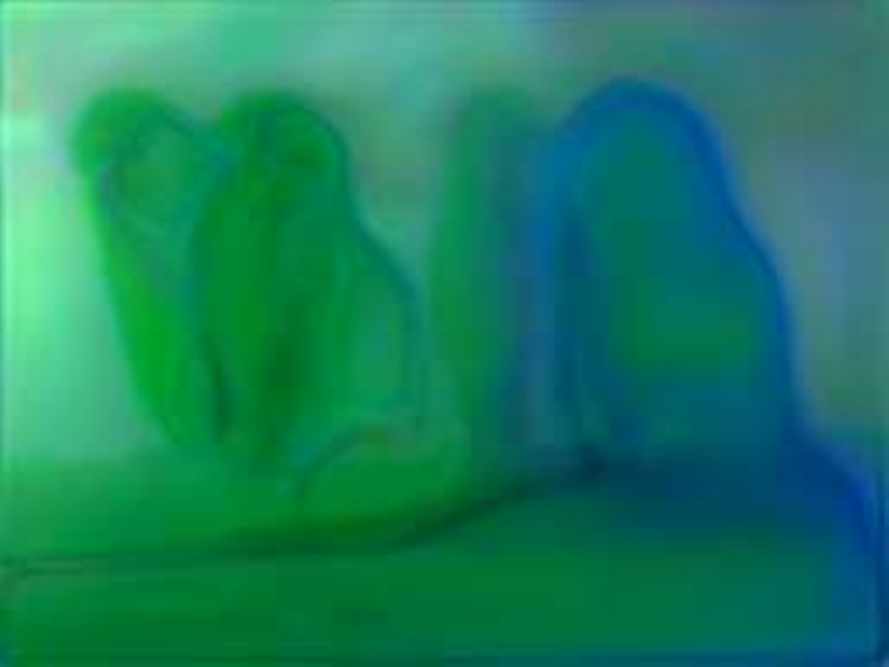}}
\hspace{-0.7mm}
\subfloat{\includegraphics[width=0.15\textwidth]{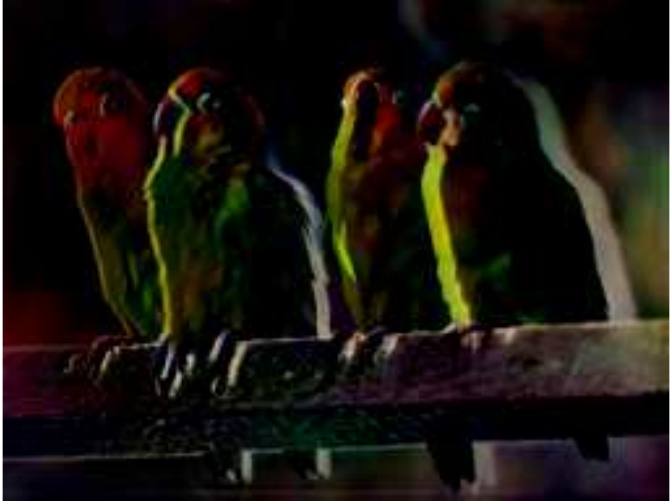}}

\vspace{-3mm}
\setcounter{subfigure}{0}
\subfloat[\footnotesize{Reference: $J_\lambda^{gt}$ }]{\includegraphics[width=0.15\textwidth]{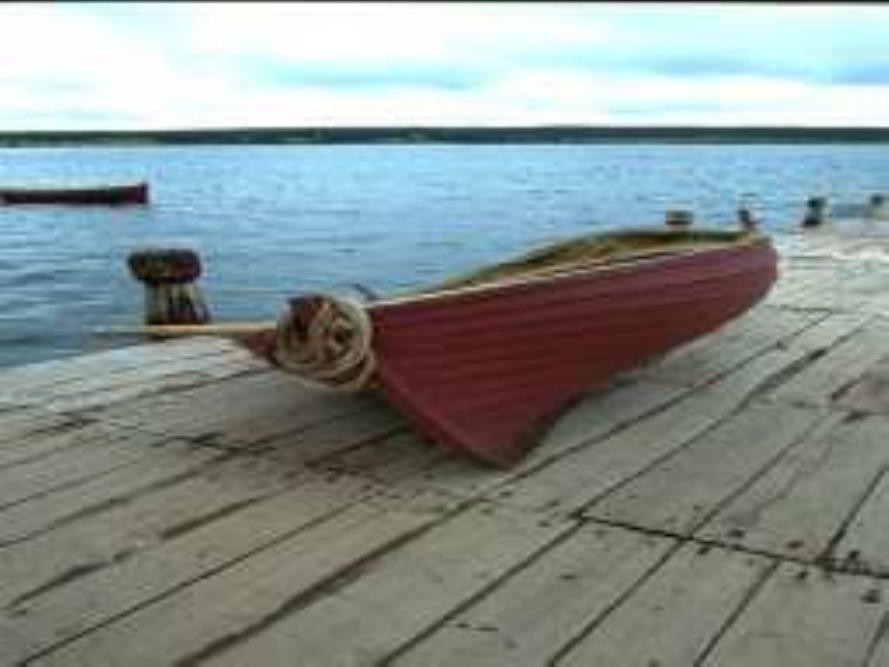}}
\hspace{-0.7mm}
\subfloat[\footnotesize{Input}]{\includegraphics[width=0.15\textwidth]{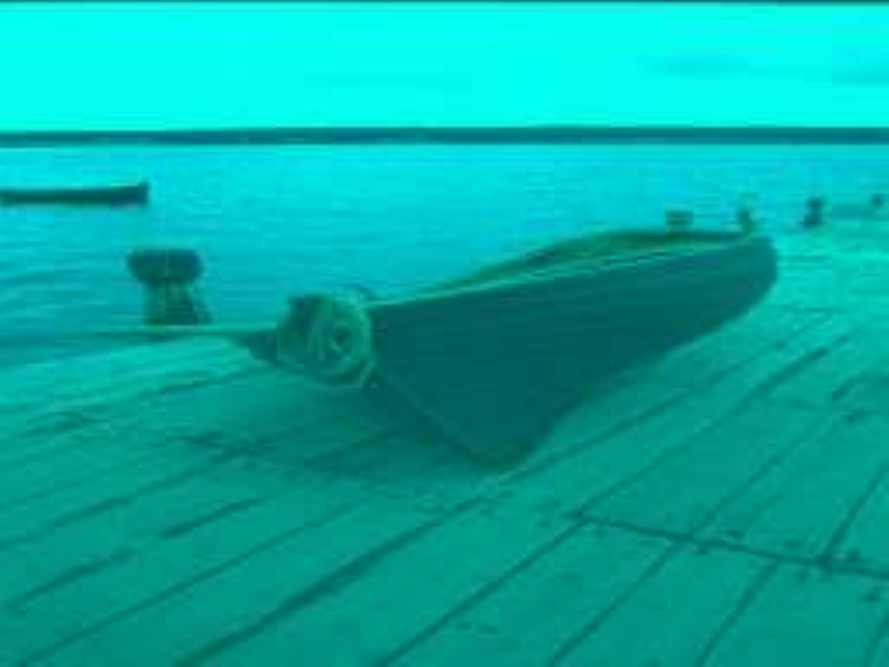}}
\hspace{-0.7mm}
\subfloat[\footnotesize{Clean image: $J_\lambda$}]{\includegraphics[width=0.15\textwidth]{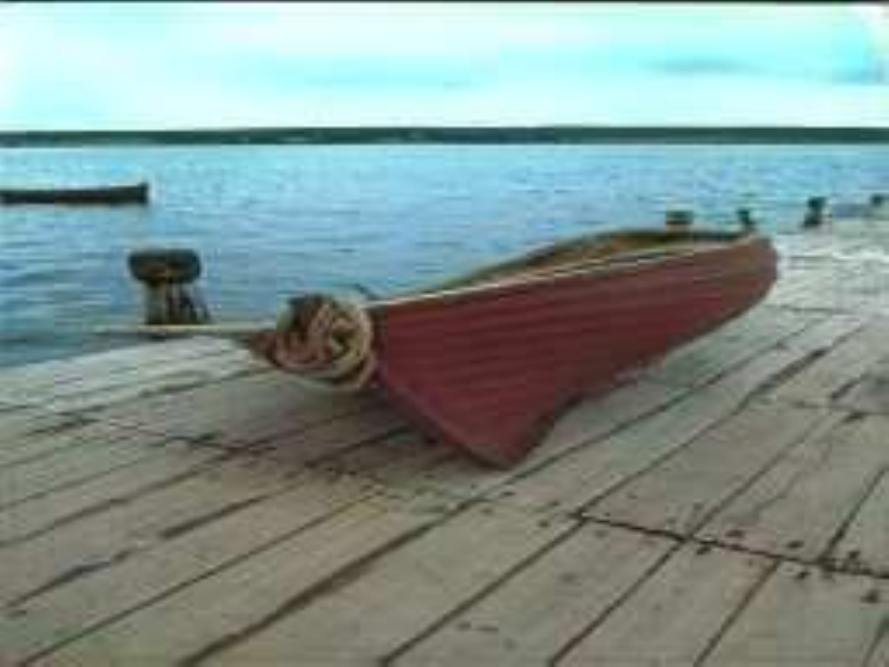}}
\hspace{-0.7mm}
\subfloat[\footnotesize{Transmission: $T_\lambda$}]{\includegraphics[width=0.15\textwidth]{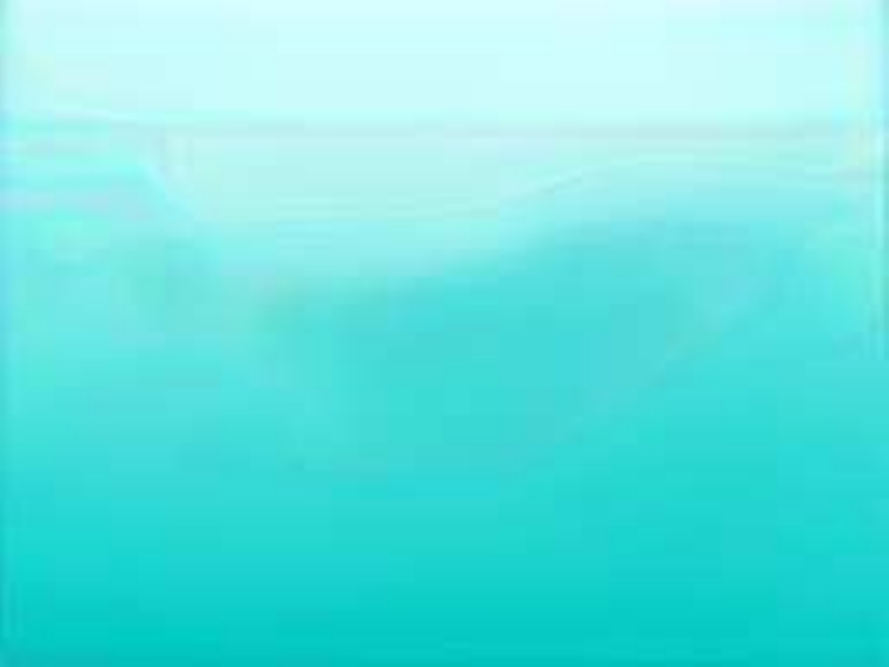}}
\hspace{-0.7mm}
\subfloat[\footnotesize{Background: $B_\lambda$}]{\includegraphics[width=0.15\textwidth]{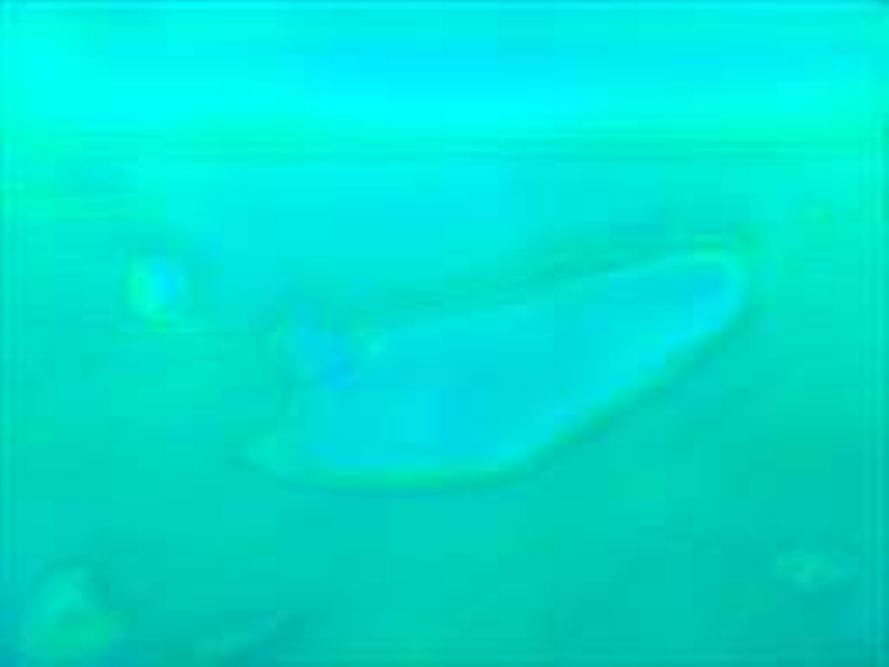}}
\hspace{-0.7mm}
\subfloat[\scriptsize{Difference: $J_\lambda-J_\lambda^{gt}$}]{\includegraphics[width=0.15\textwidth]{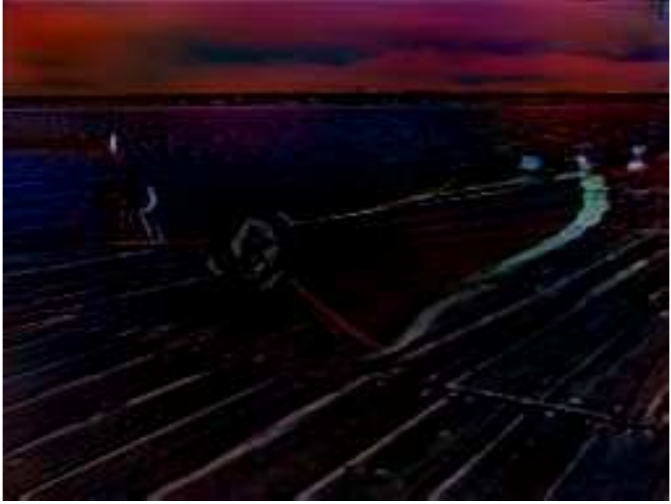}}
\caption{The visual results of the clean image $J_\lambda$, transmission map $T_\lambda$, background light $B_\lambda$ and the difference image, which is estimated as the difference between the reference image and the restored clean image.  }
\label{color}
\end{figure*}

\subsection{Comparison between the pre-trained and fine-tuned models}

We use meta-learning to obtain a pre-trained model by learning the shared knowledge among different types of distortions on the synthetic underwater image dataset.
Based on the pre-trained model, we then fine-tune two image enhancement models on the real underwater image dataset, i.e.,  UIEB and EUVP, where F-UIEB and F-EUVP are used to denote the fine-tuned models on UIEB and EUVP, respectively.
For simplicity,  we use P-UIEB and P-EUVP to represent the results of the pre-trained model on the UIEB and EUVP datasets, respectively.
As demonstrated in Fig. \ref{loss-psnr}, we monitor the training process by tracking the validation loss and averaged PSNR on the test datasets of UIEB and EUVP.
We can observe that both pre-trained models converge in around 30 epochs, while the fine-tuned models require about 20 epochs to reach convergence. More importantly, the values of loss function and PSNR have been significantly improved by the fine-tuning process on real underwater images, which convinces the merit of meta-learning strategy in dealing with such image restoration problems.

Generalization is an important capability for the underwater image enhancement model, which can equip it to handle diversified distortions.
As shown in Fig. \ref{enhance}, we evaluate the performance of the pre-trained model and fine-tuned model on typical underwater attenuation images including haze, blue, green, yellow, and turbid.
We notice that our model-driven pre-trained model can well remove the background color of the underwater images, while the fine-tuned models can adapt to the characteristics of real underwater datasets.
For quantitative comparison, Table \ref{Pre-training-uciqe} provides
the averaged UCIQE and UIQM as well as the component scores of UCIQE and UIQM for the test images in Fig. \ref{enhance}, where the pre-trained model gives higher UCIQE and UIQM than the two fine-tuned models. It confirms that the pre-trained model can provide enhancement results with better human visual perception.
By analyzing the composition of these indicators, the pre-trained model is superior to other models in contrast and sharpness, but the color saturation needs to be further improved.
Therefore, the model-based pre-trained model can remove the color degradation from various underwater scenarios. Since the fine-tuned models learned features from a specific real underwater dataset, they may have some performance degradation when the water environment changes.

\subsection{Comparison with other underwater image enhancement methods}
We further evaluate our MetaUE model by comparing it with state-of-the-art underwater image enhancement methods.
The quantitative comparison is displayed in Table \ref{metau45}, where our pre-trained model is fine-tuned on the UIEB dataset.
As shown in Table \ref{metau45}, our MetaUE model outperforms other comparison methods in terms of PSNR, SSIM, and MSE, demonstrating the advantages of our model-based meta-learning strategy.
The learning-based methods, Water-Net and Ucolor, perform poorly on other underwater image datasets except for UIEB, which convinces the advantages of building up the deep learning model based on the physical model for underwater images.
For the unpaired datasets, the UIQM and UCIQE are the main evaluation metrics. It can be observed that both the pre-trained model and MetaUE provide better results than other comparison methods.
Furthermore, it is shown that the pre-trained model presents its generalization ability on real underwater datasets.
When we fine-tune the pre-trained model on UIEB, both PSNR and SSIM on the test dataset are significantly improved on UIEB, but it loses generalization on the EUVP dataset. Similar results can be observed in Table \ref{metaUE}, for which we fine-tuned the learning-based models on the EUVP dataset.

\begin{figure*}[htbp]
\centering
\hspace{-5mm}
\subfloat{\includegraphics[width=0.12\textwidth]{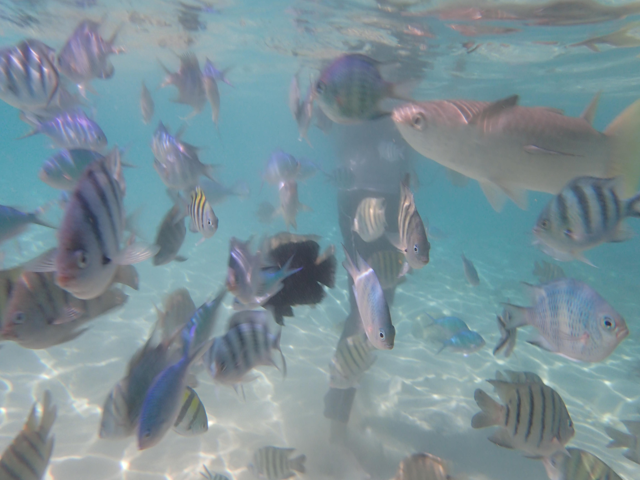}}
\hspace{-0.7mm}
\subfloat{\includegraphics[width=0.12\textwidth]{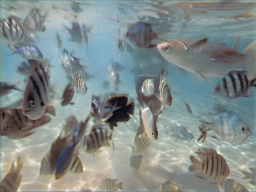}}
\hspace{-0.7mm}
\subfloat{\includegraphics[width=0.12\textwidth]{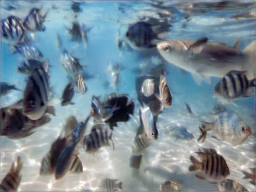}}
\hspace{-0.7mm}
\subfloat{\includegraphics[width=0.12\textwidth]{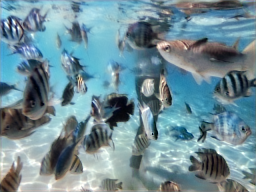}}
\hspace{-0.7mm}
\subfloat{\includegraphics[width=0.12\textwidth]{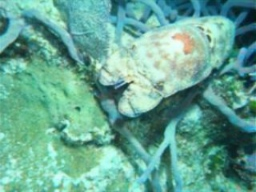}}
\hspace{-0.7mm}
\subfloat{\includegraphics[width=0.12\textwidth]{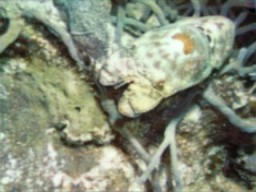}}
\hspace{-0.7mm}
\subfloat{\includegraphics[width=0.12\textwidth]{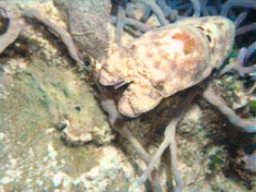}}
\hspace{-0.7mm}
\subfloat{\includegraphics[width=0.12\textwidth]{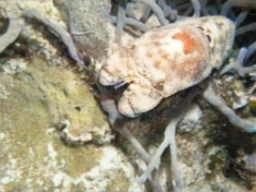}}

\vspace{0.5mm}
\hspace{-5mm}
\subfloat{\includegraphics[width=0.12\textwidth]{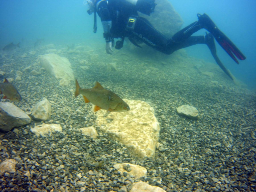}}
\hspace{-0.7mm}
\subfloat{\includegraphics[width=0.12\textwidth]{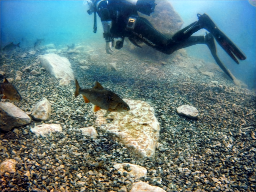}}
\hspace{-0.7mm}
\subfloat{\includegraphics[width=0.12\textwidth]{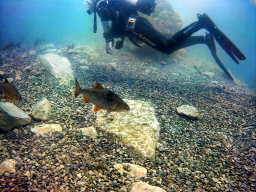}}
\hspace{-0.7mm}
\subfloat{\includegraphics[width=0.12\textwidth]{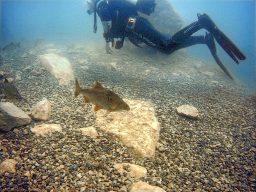}}
\hspace{-0.7mm}
\subfloat{\includegraphics[width=0.12\textwidth]{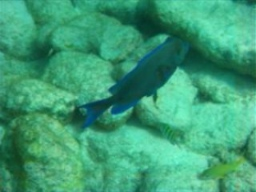}}
\hspace{-0.7mm}
\subfloat{\includegraphics[width=0.12\textwidth]{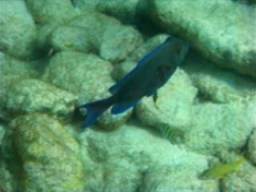}}
\hspace{-0.7mm}
\subfloat{\includegraphics[width=0.12\textwidth]{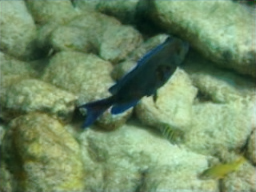}}
\hspace{-0.7mm}
\subfloat{\includegraphics[width=0.12\textwidth]{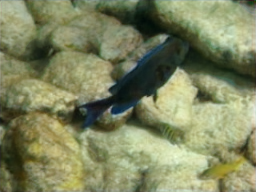}}

\vspace{0.5mm}
\hspace{-6mm}
\setcounter{subfigure}{0}
\subfloat[\scriptsize{Input}]{\includegraphics[width=0.12\textwidth]{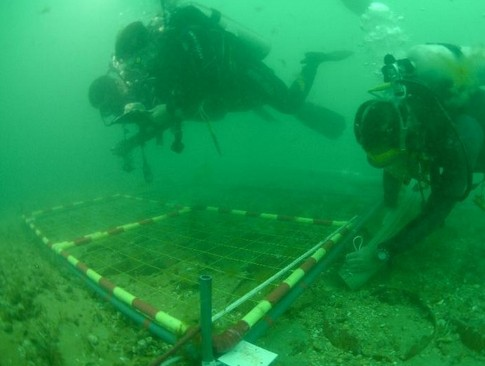}}
\hspace{-0.7mm}
\subfloat[\scriptsize{plain UNet}]{\includegraphics[width=0.12\textwidth]{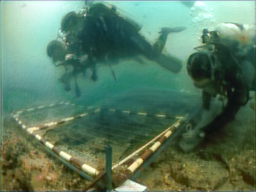}}
\hspace{-0.7mm}
\subfloat[\scriptsize{meta UNet}]{\includegraphics[width=0.12\textwidth]{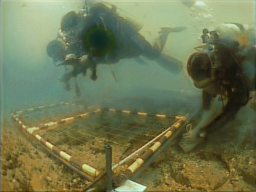}}
\hspace{-0.7mm}
\subfloat[\scriptsize{MetaUE}]{\includegraphics[width=0.12\textwidth]{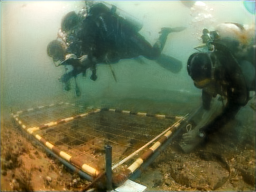}}
\setcounter{subfigure}{0}
\hspace{-1.8mm}
\subfloat[\scriptsize{Input}]{\includegraphics[width=0.12\textwidth]{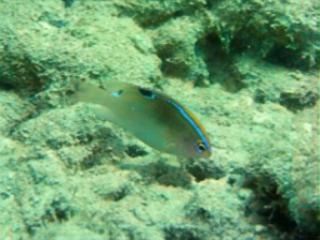}}
\hspace{-0.7mm}
\subfloat[\scriptsize{plain UNet}]{\includegraphics[width=0.12\textwidth]{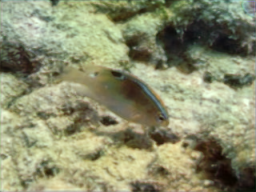}}
\hspace{-0.7mm}
\subfloat[\scriptsize{meta UNet}]{\includegraphics[width=0.12\textwidth]{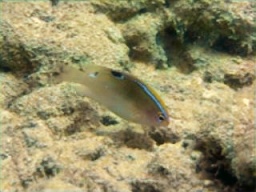}}
\hspace{-0.7mm}
\subfloat[\scriptsize{MetaUE}]{\includegraphics[width=0.12\textwidth]{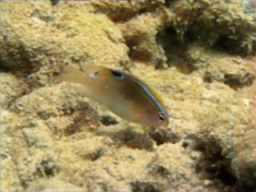}}

\caption{The visual comparison on ablation study, where the first to fourth columns are the results on the UIEB dataset and the fifth to eighth columns are the results of the EUVP.  }
\label{Ablation study}
\end{figure*}

In Fig. \ref{reference}, we exhibit the visual enhancement results among different methods on the degraded underwater images including scenes such as haze, green, blue, and low light. As can be observed, for the comparison methods, some work well on low-light images or yellow images, but none of them can provide satisfactory results on all applications. Our MetaUE presents good generalization ability on various underwater images, where the visual comparison is consistent with the quantitative results in Table \ref{metau45}.
Furthermore, the pre-trained model is shown to be able to remove the water to obtain clear images, which provides excellent results for low-brightness images by restoring the objects in the shadow. And the MetaUE produces similar restoration results as the reference images provided by the UIEB dataset, which reflects the effect of the fine-tuning process.
More specifically, the first two examples in Fig. \ref{no reference} reveal that the pre-trained model may lose color information due to the limited coverage of the synthetic dataset.
On the other hand, the fine-tuned model can learn the color features from real underwater images.
Similar results can be observed in Fig. \ref{no reference}, where the pre-trained model shows good generalization ability by removing the color degradation and repairing the insufficient illumination. The fine-tuning can effectively compensate for the underwater characteristic to make the enhanced images more realistic.

Some physical model-based methods \cite{galdran2015automatic,akkaynak2019sea,liu2021rank,zhang2022underwaterACDC} assume the background light to be a constant function, which leads to inaccurate estimation.
Actually, both background light and transmission map are smooth functions \cite{akkaynak2019sea}.
As shown in Fig. \ref{color}, we provide two examples to illustrate the solutions of the background light and transmission map in our model. As can be seen, the clean image, transmission map and background light are correctly estimated by our method. In particular, the residual image between the reference image $J_\lambda^{gt}$ and the estimated clean image $J_\lambda$ indicates that the pre-trained model can accurately remove the background color.

\begin{figure*}[t]
\centering
\hspace{-5mm}
\subfloat{\rotatebox{90}{\footnotesize{~~~~Input}}}
\hspace{-0.7mm}
\subfloat{\includegraphics[width=0.105\textwidth]{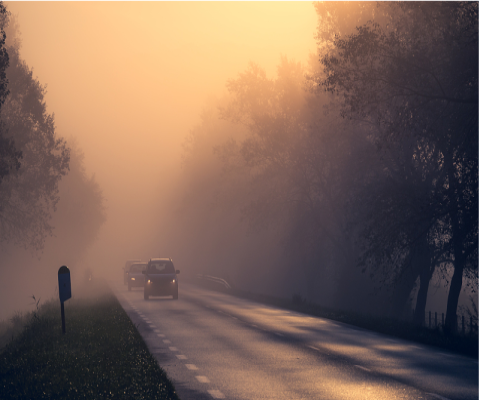}\hspace{-0.7mm}
\includegraphics[width=0.105\textwidth]{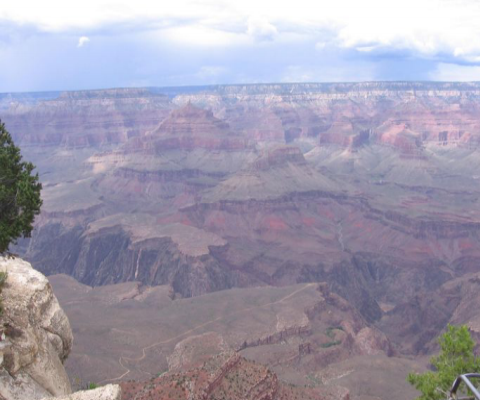}\hspace{-0.7mm}
\includegraphics[width=0.105\textwidth]{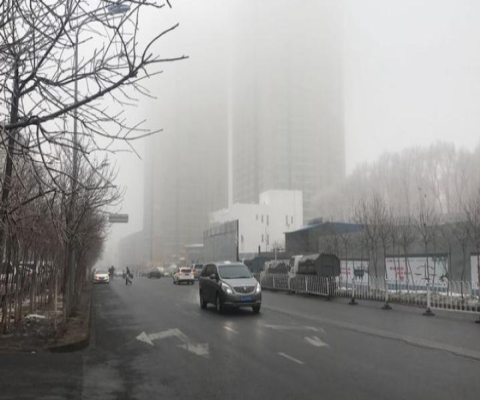}}
\hspace{0.2mm}
\subfloat{\includegraphics[width=0.105\textwidth]{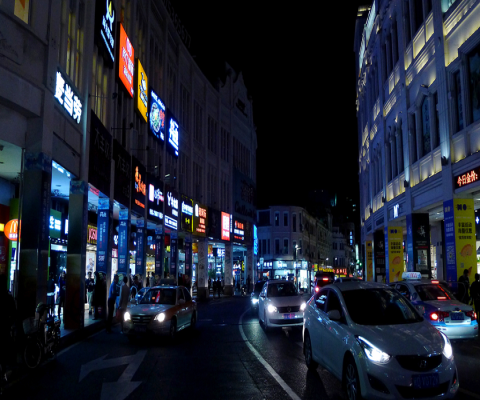}\hspace{-0.7mm}
\includegraphics[width=0.105\textwidth]{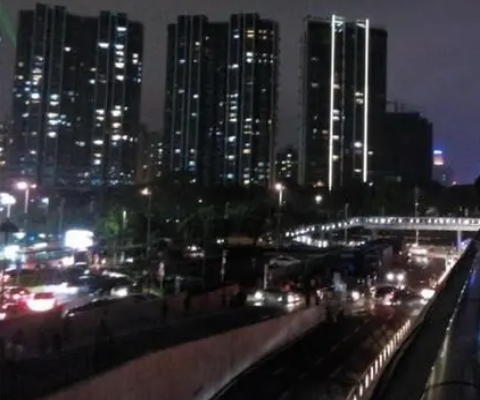}\hspace{-0.7mm}
\includegraphics[width=0.105\textwidth]{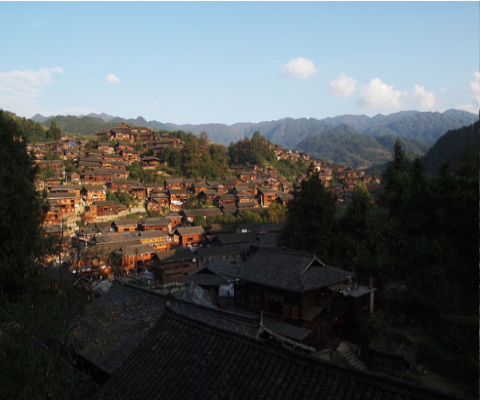}}
\hspace{0.2mm}
\subfloat{\includegraphics[width=0.105\textwidth]{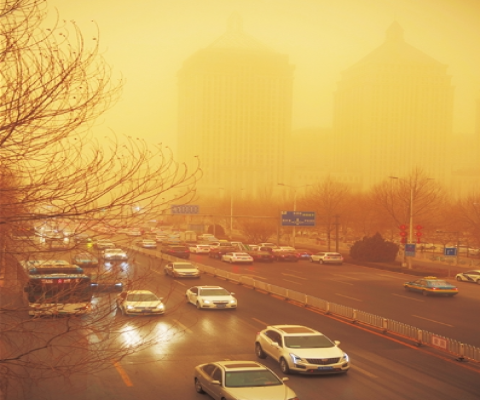}\hspace{-0.7mm}
\includegraphics[width=0.105\textwidth]{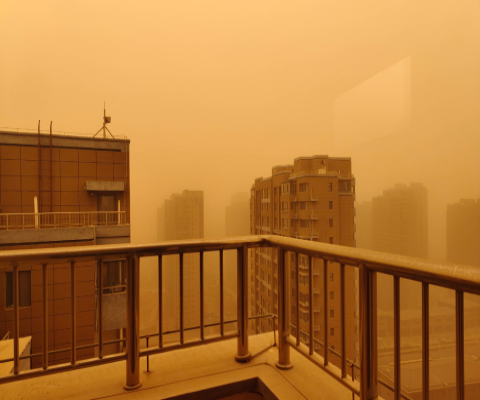}\hspace{-0.7mm}
\includegraphics[width=0.105\textwidth]{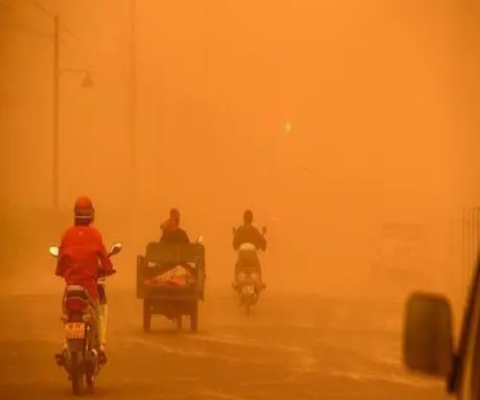}}

\vspace{0.5mm}
\hspace{-4.5mm}
\subfloat{\rotatebox{90}{\footnotesize{~Pre-trained}}}
\hspace{-0.7mm}
\subfloat{\includegraphics[width=0.105\textwidth]{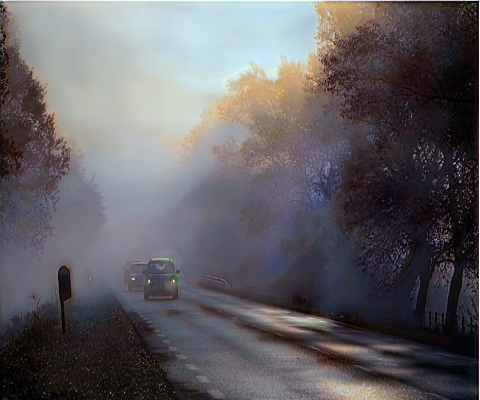}\hspace{-0.7mm}
\includegraphics[width=0.105\textwidth]{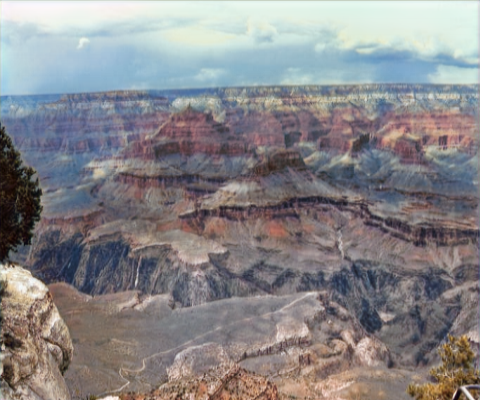}\hspace{-0.7mm}
\includegraphics[width=0.105\textwidth]{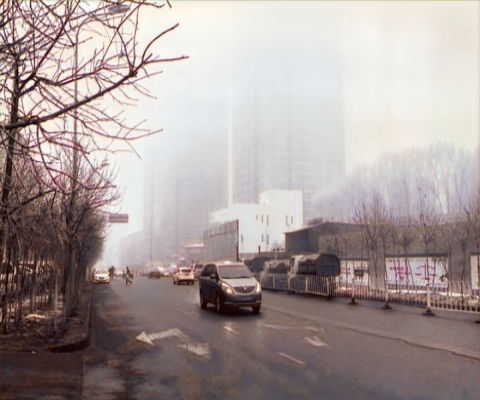}}
\hspace{0.2mm}
\subfloat{\includegraphics[width=0.105\textwidth]{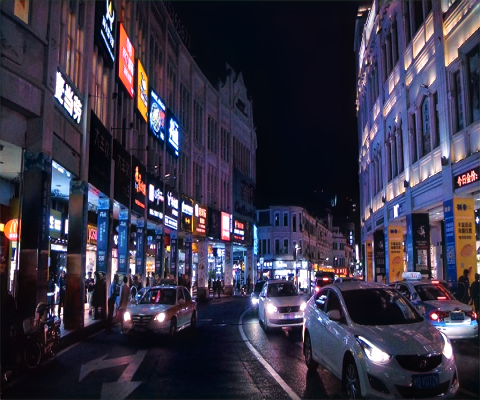}\hspace{-0.7mm}
\includegraphics[width=0.105\textwidth]{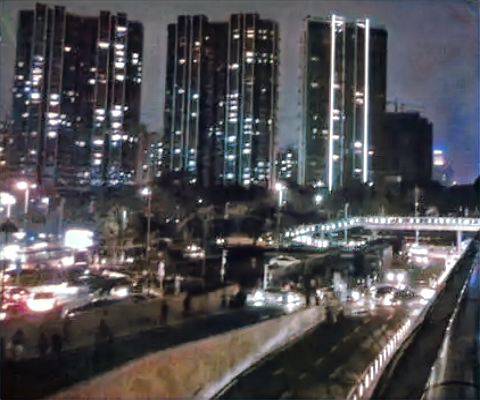}\hspace{-0.7mm}
\includegraphics[width=0.105\textwidth]{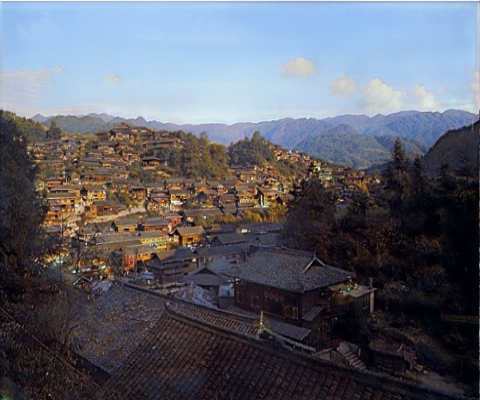}}
\hspace{0.2mm}
\subfloat{\includegraphics[width=0.105\textwidth]{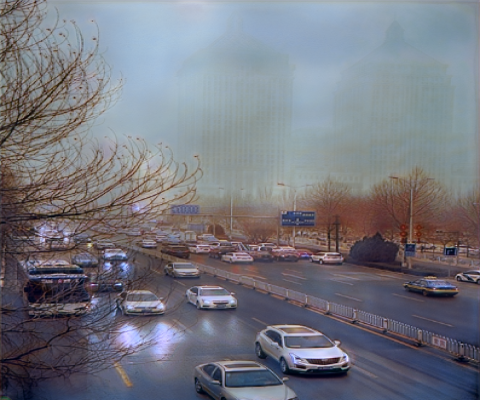}\hspace{-0.7mm}
\includegraphics[width=0.105\textwidth]{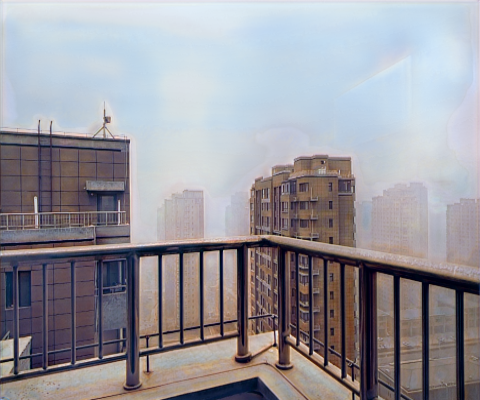}\hspace{-0.7mm}
\includegraphics[width=0.105\textwidth]{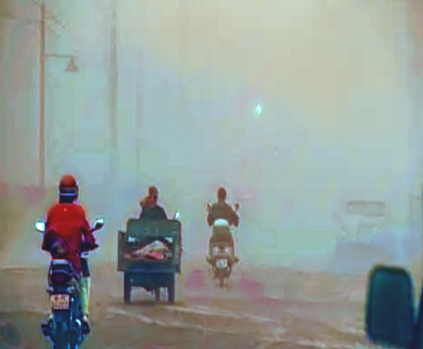}}

\vspace{0.5mm}
\hspace{-5mm}
\subfloat{\rotatebox{90}{\footnotesize{~~Rank1\cite{liu2021rank} }}}
\setcounter{subfigure}{0}
\hspace{-1.9mm}
\subfloat[\scriptsize{Hazing}]{\includegraphics[width=0.105\textwidth]{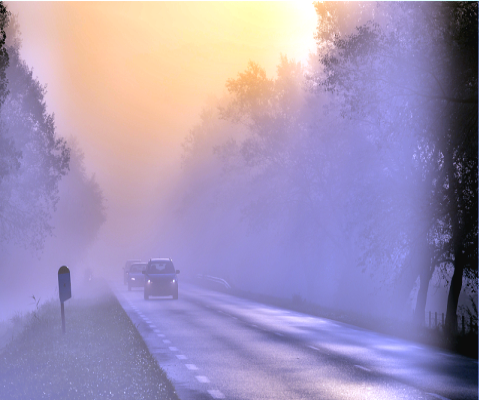}\hspace{-0.7mm}
\includegraphics[width=0.105\textwidth]{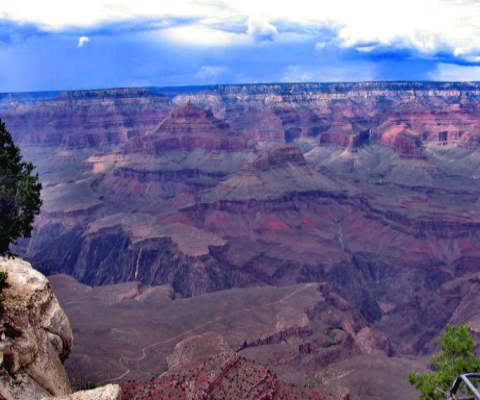}\hspace{-0.7mm}
\includegraphics[width=0.105\textwidth]{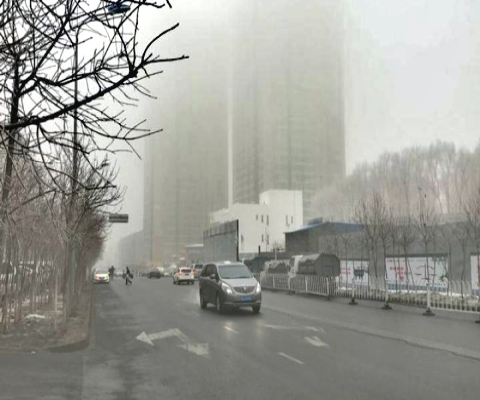}}\hspace{0.2mm}
\subfloat[\scriptsize{Low light}]{\includegraphics[width=0.105\textwidth]{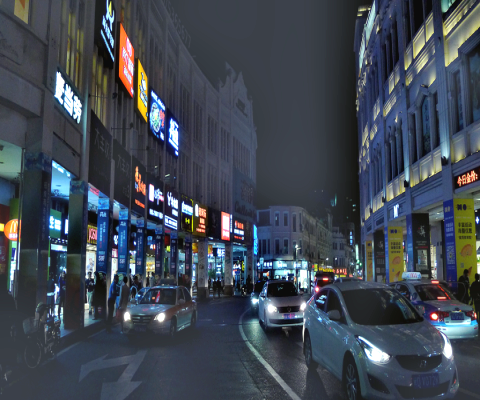}\hspace{-0.7mm}
\includegraphics[width=0.105\textwidth]{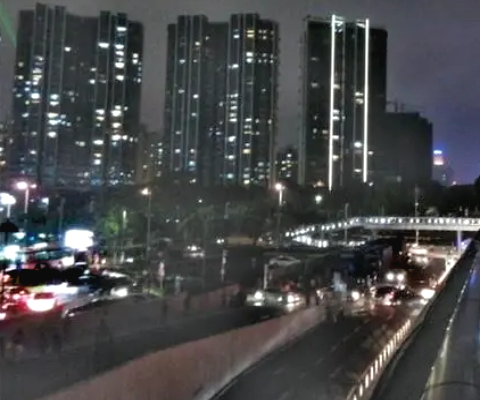}\hspace{-0.7mm}
\includegraphics[width=0.105\textwidth]{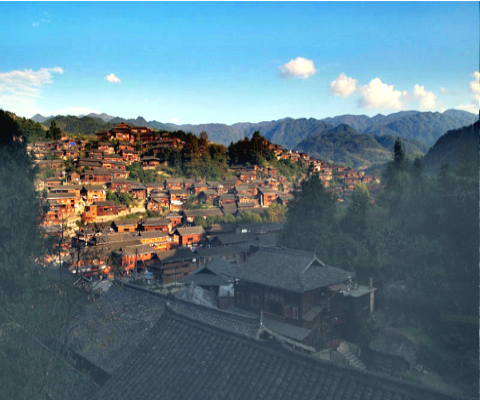}}\hspace{0.2mm}
\subfloat[\scriptsize{Sand storm}]{\includegraphics[width=0.105\textwidth]{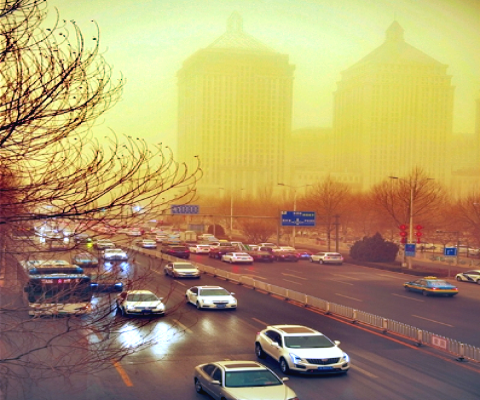}\hspace{-0.7mm}
\includegraphics[width=0.105\textwidth]{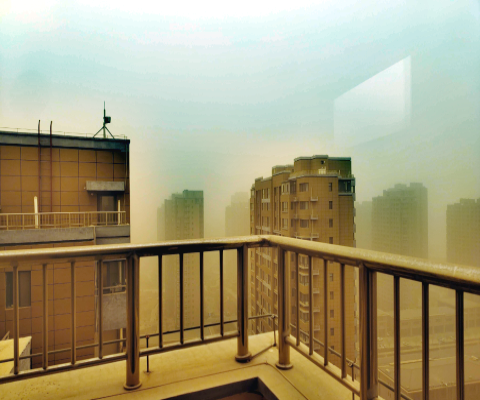}\hspace{-0.7mm}
\includegraphics[width=0.105\textwidth]{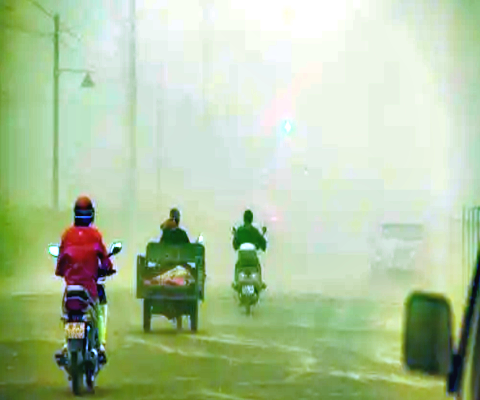}}
\caption{There are the application results of the pre-trained model for low quality images, including haze, low light, and sand storm, where Rank1 represent the results of \cite{liu2021rank}.}
\label{application}
\end{figure*}

\subsection{Ablation study}
The ablation studies are presented to explore the effectiveness of the model-driven meta learning method  for underwater image enhancement.
More specifically, we build up three deep learning models for comparison, namely the Plain UNet, Meta UNet, and MetaUE.
Note that the `Plain UNet' is realized by random initialization and trained on UIEB and EUVP datasets, respectively.
The `Meta UNet' model is introduced to evaluate the effectiveness of meta-learning, where only the clean image $N_J(I;\theta_J)$ was trained by meta-learning.
As shown in Table \ref{Ablation Study}, the pre-trained strategy plays a major role in improving underwater image quality, while the physical image model can guide the deep-learning model to recover clean images from the observations.
Fig. \ref{Ablation study} provides the visual comparison of the three methods, where the  first to fourth  columns are the results from the UIEB dataset and the fifth to eighth  columns are the results from the EUVP. We can observe that the fine-tuned models based on the physical model give the best results of the three models, confirming the effectiveness of our model.

\begin{table}[t]
  \centering
  \caption{The ablation studies on the UIEB and EUVP datasets, where three models are trained, the Plain UNet, Meta UNet and model-based Meta UNet. }
    \begin{tabular}{c|cc|cc}
    \hline
    \hline
    \multirow{2}{*}{\diagbox[width=2.5cm]{$\mathrm{Models}$}{$\mathrm{Datasets}$}}   & \multicolumn{2}{c|}{UIEB}   & \multicolumn{2}{c}{EUVP}  \\
     \cline{2-5}
     & \multicolumn{1}{l}{PSNR} & \multicolumn{1}{l|}{SSIM} & \multicolumn{1}{l}{PSNR} & \multicolumn{1}{l}{SSIM} \\
          \hline
    plain UNet  &  17.54 &0.8125       &  18.76     & 0.8239 \\
    meta UNet & 21.12 & 0.8054      &   21.24    &0.8211  \\
    MetaUE & \textbf{22.01} & \textbf{0.8609}       &  \textbf{22.12} & \textbf{0.8322}  \\
    \hline
    \hline
    \end{tabular}%
  \label{Ablation Study}%
\end{table}%

\section{Conclusion and Discussion}\label{sec6}

In this paper, we presented a model-based deep-learning model to improve the visual qualities of underwater images, which was demonstrated with better interpretability and generalization ability.
To be specific, we built up a multi-variable convolution model
to learn the composition of the underwater image, including the clean image, transmission map, and background light.
The meta-learning was used to first capture the shared prior knowledge of the degraded underwater images and then learn to adapt to a specific underwater environment by fine-tuning on the real underwater image dataset.
Numerical experiments showed that our MetaUE outperformed several SOTA underwater image enhancement methods in both quality and robustness.

In addition to underwater images, there exist other degraded images that could be regarded as a superimposition of a clean image with environmental conditions such as sand dust, haze, and low light, etc.
Various methods have been investigated to deal with such image restoration problems, e.g., low light enhancement \cite{ma2022toward,wang2022low} and image dehazing \cite{yang2022self,liu2022deep}. Indeed, these problems can be also modeled by the physical model \eqref{uw model} and recovered by our model-based deep learning model.
As illustrated in Fig. \ref{application}, our pre-trained model works well on these challenging applications, which outperforms the recent Rank1 prior method for scene recovery in \cite{liu2021rank}. Thus, it is beneficial to construct the network model based on a solid physical model, which can help to guide the network towards achieving desired outcomes.
Although our MetaUE has demonstrated the improved generalization capabilities, there is still an obvious disparity in the performance of the fine-tuned model across different datasets; see Table \ref{metau45} and Table \ref{metaUE}. Our future work includes developing more efficient underwater image enhancement models in dealing with various real underwater images.



\begin{thebibliography}{10}
\providecommand{\url}[1]{#1}
\csname url@samestyle\endcsname
\providecommand{\newblock}{\relax}
\providecommand{\bibinfo}[2]{#2}
\providecommand{\BIBentrySTDinterwordspacing}{\spaceskip=0pt\relax}
\providecommand{\BIBentryALTinterwordstretchfactor}{4}
\providecommand{\BIBentryALTinterwordspacing}{\spaceskip=\fontdimen2\font plus
\BIBentryALTinterwordstretchfactor\fontdimen3\font minus
  \fontdimen4\font\relax}
\providecommand{\BIBforeignlanguage}[2]{{%
\expandafter\ifx\csname l@#1\endcsname\relax
\typeout{** WARNING: IEEEtran.bst: No hyphenation pattern has been}%
\typeout{** loaded for the language `#1'. Using the pattern for}%
\typeout{** the default language instead.}%
\else
\language=\csname l@#1\endcsname
\fi
#2}}
\providecommand{\BIBdecl}{\relax}
\BIBdecl

\bibitem{jian2021underwater}
M.~Jian, X.~Liu, H.~Luo, X.~Lu, H.~Yu, and J.~Dong, ``Underwater image
  processing and analysis: A review,'' \emph{Signal Processing: Image
  Communication}, vol.~91, pp. 116--128, 2021.

\bibitem{akkaynak2017space}
D.~Akkaynak, T.~Treibitz, T.~Shlesinger, Y.~Loya, R.~Tamir, and D.~Iluz, ``What
  is the space of attenuation coefficients in underwater computer vision?'' in
  \emph{Proceedings of the IEEE Conference on Computer Vision and Pattern
  Recognition}, 2017, pp. 4931--4940.

\bibitem{akkaynak2019sea}
D.~Akkaynak and T.~Treibitz, ``Sea-{T}hru: A method for removing water from
  underwater images,'' in \emph{Proceedings of the IEEE/CVF Conference on
  Computer Vision and Pattern Recognition}, 2019, pp. 1682--1691.

\bibitem{liu2021rank}
J.~Liu, W.~Liu, J.~Sun, and T.~Zeng, ``Rank-one prior: Toward real-time scene
  recovery,'' in \emph{Proceedings of the IEEE/CVF Conference on Computer
  Vision and Pattern Recognition}, 2021, pp. 14\,802--14\,810.

\bibitem{liu2022rank}
J.~Liu, R.~W. Liu, J.~Sun, and T.~Zeng, ``Rank-one prior: Real-time scene
  recovery,'' \emph{IEEE Transactions on Pattern Analysis and Machine
  Intelligence}, 2022.

\bibitem{ding2021depth}
X.~Ding, Z.~Liang, Y.~Wang, and X.~Fu, ``Depth-aware total variation
  regularization for underwater image dehazing,'' \emph{Signal Processing:
  Image Communication}, vol.~98, pp. 116--128, 2021.

\bibitem{hou2020underwater}
G.~Hou, J.~Li, G.~Wang, Z.~Pan, and X.~Zhao, ``Underwater image dehazing and
  denoising via curvature variation regularization,'' \emph{Multimedia Tools
  and Applications}, vol.~79, no.~27, pp. 20\,199--20\,219, 2020.

\bibitem{jiao2021underwater}
Q.~Jiao, M.~Liu, P.~Li, L.~Dong, M.~Hui, L.~Kong, and Y.~Zhao, ``Underwater
  image restoration via non-convex non-smooth variation and thermal exchange
  optimization,'' \emph{Journal of Marine Science and Engineering}, vol.~9,
  no.~6, pp. 570--581, 2021.

\bibitem{li2018watergan}
J.~Li, K.~A. Skinner, R.~M. Eustice, and M.~Johnson-Roberson, ``Water {G}an:
  Unsupervised generative network to enable real-time color correction of
  monocular underwater images,'' \emph{IEEE Robotics and Automation Letters},
  vol.~3, no.~1, pp. 387--394, 2018.

\bibitem{li2018emerging}
C.~Li, J.~Guo, and C.~Guo, ``Emerging from water: Underwater image color
  correction based on weakly supervised color transfer,'' \emph{IEEE Signal
  Processing Letters}, vol.~25, no.~3, pp. 323--327, 2018.

\bibitem{ye2019deep}
X.~Ye, Z.~Li, B.~Sun, Z.~Wang, R.~Xu, H.~Li, and X.~Fan, ``Deep joint depth
  estimation and color correction from monocular underwater images based on
  unsupervised adaptation networks,'' \emph{IEEE Transactions on Circuits and
  Systems for Video Technology}, vol.~30, no.~11, pp. 3995--4008, 2019.

\bibitem{yang2020underwater}
M.~Yang, K.~Hu, Y.~Du, Z.~Wei, Z.~Sheng, and J.~Hu, ``Underwater image
  enhancement based on conditional generative adversarial network,''
  \emph{Signal Processing: Image Communication}, vol.~81, pp. 115--123, 2020.

\bibitem{liu2022twin}
R.~Liu, Z.~Jiang, S.~Yang, and X.~Fan, ``Twin adversarial contrastive learning
  for underwater image enhancement and beyond,'' \emph{IEEE Transactions on
  Image Processing}, vol.~31, pp. 4922--4936, 2022.

\bibitem{fabbi2018enhancing}
C.~Fabbri, M.~J. Islam, and J.~Sattar, ``Enhancing underwater imagery using
  generative adversarial networks,'' in \emph{2018 IEEE International
  Conference on Robotics and Automation (ICRA)}, vol.~5, no.~3, 2018, pp.
  7159--7165.

\bibitem{islam2020fast}
M.~J. Islam, Y.~Xia, and J.~Sattar, ``Fast underwater image enhancement for
  improved visual perception,'' \emph{IEEE Robotics and Automation Letters},
  vol.~5, no.~2, pp. 3227--3234, 2020.

\bibitem{li2020underwater}
C.~Li, S.~Anwar, and F.~Porikli, ``Underwater scene prior inspired deep
  underwater image and video enhancement,'' \emph{Pattern Recognition},
  vol.~98, pp. 107--118, 2020.

\bibitem{UIEB}
C.~Li, C.~Guo, W.~Ren, R.~Cong, J.~Hou, S.~Kwong, and D.~Tao, ``An underwater
  image enhancement benchmark dataset and beyond,'' \emph{IEEE Transactions on
  Image Processing}, vol.~29, no.~2, pp. 4376--4389, 2020.

\bibitem{xue2021joint}
X.~Xue, Z.~Hao, L.~Ma, Y.~Wang, and R.~Liu, ``Joint luminance and chrominance
  learning for underwater image enhancement,'' \emph{IEEE Signal Processing
  Letters}, vol.~28, no.~1, pp. 818--822, 2021.

\bibitem{li2021underwater}
C.~Li, S.~Anwar, J.~Hou, R.~Cong, C.~Guo, and W.~Ren, ``Underwater image
  enhancement via medium transmission-guided multi-color space embedding,''
  \emph{IEEE Transactions on Image Processing}, vol.~30, pp. 4985--5000, 2021.

\bibitem{zhang2019deep}
K.~Zhang, W.~Zuo, and L.~Zhang, ``Deep plug-and-play super-resolution for
  arbitrary blur kernels,'' in \emph{Proceedings of the IEEE/CVF Conference on
  Computer Vision and Pattern Recognition}, 2019, pp. 1671--1681.

\bibitem{zhang2021plug}
K.~Zhang, Y.~Li, W.~Zuo, L.~Zhang, L.~Van~Gool, and R.~Timofte, ``Plug-and-play
  image restoration with deep denoiser prior,'' \emph{IEEE Transactions on
  Pattern Analysis and Machine Intelligence}, vol.~44, no.~10, pp. 6360--6376,
  2021.

\bibitem{fang2022robust}
Y.~Fang, H.~Zhang, H.~S. Wong, and T.~Zeng, ``A robust non-blind deblurring
  method using deep denoiser prior,'' in \emph{Proceedings of the IEEE/CVF
  Conference on Computer Vision and Pattern Recognition}, 2022, pp. 735--744.

\bibitem{ho2020denoising}
J.~Ho, A.~Jain, and P.~Abbeel, ``Denoising diffusion probabilistic models,''
  \emph{Advances in Neural Information Processing Systems}, vol.~33, pp.
  6840--6851, 2020.

\bibitem{dhariwal2021diffusion}
P.~Dhariwal and A.~Nichol, ``Diffusion models beat {GAN}s on image synthesis,''
  \emph{Advances in Neural Information Processing Systems}, vol.~34, pp.
  8780--8794, 2021.

\bibitem{rombach2022high}
R.~Rombach, A.~Blattmann, D.~Lorenz, P.~Esser, and B.~Ommer, ``High-resolution
  image synthesis with latent diffusion models,'' in \emph{Proceedings of the
  IEEE/CVF Conference on Computer Vision and Pattern Recognition}, 2022, pp.
  10\,684--10\,695.

\bibitem{wyatt2022anoddpm}
J.~Wyatt, A.~Leach, S.~M. Schmon, and C.~G. Willcocks, ``Anoddpm: Anomaly
  detection with denoising diffusion probabilistic models using simplex
  noise,'' in \emph{Proceedings of the IEEE/CVF Conference on Computer Vision
  and Pattern Recognition}, 2022, pp. 650--656.

\bibitem{chung2022mr}
H.~Chung, E.~S. Lee, and J.~C. Ye, ``Mr image denoising and super-resolution
  using regularized reverse diffusion,'' \emph{IEEE Transactions on Medical
  Imaging}, 2022.

\bibitem{hoogeboom2021argmax}
E.~Hoogeboom, D.~Nielsen, P.~Jaini, P.~Forr{\'e}, and M.~Welling, ``Argmax
  flows and multinomial diffusion: Learning categorical distributions,''
  \emph{Advances in Neural Information Processing Systems}, vol.~34, pp.
  12\,454--12\,465, 2021.

\bibitem{santoro2016meta}
A.~Santoro, S.~Bartunov, M.~Botvinick, D.~Wierstra, and T.~Lillicrap,
  ``Meta-learning with memory-augmented neural networks,'' in
  \emph{International Conference on Machine Learning}.\hskip 1em plus 0.5em
  minus 0.4em\relax PMLR, 2016, pp. 1842--1850.

\bibitem{wang2019meta}
Y.-X. Wang, D.~Ramanan, and M.~Hebert, ``Meta-learning to detect rare
  objects,'' in \emph{Proceedings of the IEEE/CVF International Conference on
  Computer Vision}, 2019, pp. 9925--9934.

\bibitem{hospedales2021meta}
T.~Hospedales, A.~Antoniou, P.~Micaelli, and A.~Storkey, ``Meta-learning in
  neural networks: A survey,'' \emph{IEEE Transactions on Pattern Analysis and
  Machine Intelligence}, vol.~44, no.~9, pp. 5149--5169, 2021.

\bibitem{zhu2020metaiqa}
H.~Zhu, L.~Li, J.~Wu, W.~Dong, and G.~Shi, ``Meta{IQA}: Deep meta-learning for
  no-reference image quality assessment,'' in \emph{Proceedings of the IEEE/CVF
  Conference on Computer Vision and Pattern Recognition}, 2020, pp.
  14\,143--14\,152.

\bibitem{zhao2022federated}
H.~Zhao, F.~Ji, Q.~Li, Q.~Guan, S.~Wang, and M.~Wen, ``Federated
  meta-{L}earning enhanced acoustic radio cooperative framework for ocean of
  things,'' \emph{IEEE Journal of Selected Topics in Signal Processing},
  vol.~16, no.~3, pp. 474--486, 2022.

\bibitem{wang2022remember}
W.~Wang, L.~Duan, Y.~Wang, Q.~En, J.~Fan, and Z.~Zhang, ``Remember the
  difference: Cross-domain few-shot semantic segmentation via meta-memory
  transfer,'' in \emph{Proceedings of the IEEE/CVF Conference on Computer
  Vision and Pattern Recognition}, 2022, pp. 7065--7074.

\bibitem{hong2021wsuie}
L.~Hong, X.~Wang, Z.~Xiao, G.~Zhang, and J.~Liu, ``Wsuie: Weakly supervised
  underwater image enhancement for improved visual perception,'' \emph{IEEE
  Robotics and Automation Letters}, vol.~6, no.~4, pp. 8237--8244, 2021.

\bibitem{laradji2021weakly}
I.~H. Laradji, A.~Saleh, P.~Rodriguez, D.~Nowrouzezahrai, M.~R. Azghadi, and
  D.~Vazquez, ``Weakly supervised underwater fish segmentation using affinity
  lcfcn,'' \emph{Scientific reports}, vol.~11, no.~1, pp. 173--179, 2021.

\bibitem{saleh2022transformer}
A.~Saleh, M.~Sheaves, D.~Jerry, and M.~R. Azghadi, ``Transformer-based
  self-supervised fish segmentation in underwater videos,'' \emph{arXiv
  preprint arXiv:2206.05390}, 2022.

\bibitem{cai2022underwater}
S.~Cai, G.~Li, and Y.~Shan, ``Underwater object detection using collaborative
  weakly supervision,'' \emph{Computers and Electrical Engineering}, vol. 102,
  p. 108159, 2022.

\bibitem{uplavikar2019all}
P.~M. Uplavikar, Z.~Wu, and Z.~Wang, ``All-in-{O}ne underwater image
  enhancement using domain-adversarial learning,'' in \emph{CVPR Workshops},
  2019, pp. 1--8.

\bibitem{info13040187}
\BIBentryALTinterwordspacing
Y.~Liu, H.~Xu, B.~Zhang, K.~Sun, J.~Yang, B.~Li, C.~Li, and X.~Quan,
  ``Model-based underwater image simulation and learning-based underwater image
  enhancement method,'' \emph{Information}, vol.~13, no.~4, pp. 2176--2289,
  2022. [Online]. Available: \url{https://www.mdpi.com/2078-2489/13/4/187}
\BIBentrySTDinterwordspacing

\bibitem{wang2023domain}
Z.~Wang, L.~Shen, M.~Xu, M.~Yu, K.~Wang, and Y.~Lin, ``Domain adaptation for
  underwater image enhancement,'' \emph{IEEE Transactions on Image Processing},
  2023.

\bibitem{wang2021uiec}
Y.~Wang, J.~Guo, H.~Gao, and H.~Yue, ``{UIEC}$^2$-{N}et: {CNN}-based underwater
  image enhancement using two color space,'' \emph{Signal Processing: Image
  Communication}, vol.~96, pp. 116--130, 2021.

\bibitem{liu2022adaptive}
S.~Liu, H.~Fan, S.~Lin, Q.~Wang, N.~Ding, and Y.~Tang, ``Adaptive learning
  attention network for underwater image enhancement,'' \emph{IEEE Robotics and
  Automation Letters}, vol.~7, no.~2, pp. 5326--5333, 2022.

\bibitem{zhou2022deep}
W.-H. Zhou, D.-M. Zhu, M.~Shi, Z.-X. Li, M.~Duan, Z.-Q. Wang, G.-L. Zhao, and
  C.-D. Zheng, ``Deep images enhancement for turbid underwater images based on
  unsupervised learning,'' \emph{Computers and Electronics in Agriculture},
  vol. 202, pp. 107--122, 2022.

\bibitem{chai2022unsupervised}
S.~Chai, Z.~Fu, Y.~Huang, X.~Tu, and X.~Ding, ``Unsupervised and untrained
  underwater image restoration based on physical image formation model,'' in
  \emph{ICASSP 2022-2022 IEEE International Conference on Acoustics, Speech and
  Signal Processing (ICASSP)}.\hskip 1em plus 0.5em minus 0.4em\relax IEEE,
  2022, pp. 2774--2778.

\bibitem{guo2022unsupervised}
Z.~Guo, D.~Guo, Z.~Gu, H.~Zheng, B.~Zheng, and G.~Wang, ``Unsupervised
  underwater image clearness via transformer,'' in \emph{OCEANS
  2022-Chennai}.\hskip 1em plus 0.5em minus 0.4em\relax IEEE, 2022, pp. 1--4.

\bibitem{jiang2022two}
Q.~Jiang, Y.~Zhang, F.~Bao, X.~Zhao, C.~Zhang, and P.~Liu, ``Two-step domain
  adaptation for underwater image enhancement,'' \emph{Pattern Recognition},
  vol. 122, pp. 108--124, 2022.

\bibitem{espinosa2023efficient}
A.~R. Espinosa, D.~McIntosh, and A.~B. Albu, ``An efficient approach for
  underwater image improvement: Deblurring, dehazing, and color correction,''
  in \emph{Proceedings of the IEEE/CVF Winter Conference on Applications of
  Computer Vision}, 2023, pp. 206--215.

\bibitem{he2016deep}
K.~He, X.~Zhang, S.~Ren, and J.~Sun, ``Deep residual learning for image
  recognition,'' in \emph{Proceedings of the IEEE Conference on Computer Vision
  and Pattern Recognition}, 2016, pp. 770--778.

\bibitem{silberman2011indoor}
N.~Silberman and R.~Fergus, ``Indoor scene segmentation using a structured
  light sensor,'' in \emph{2011 IEEE International Conference on Computer
  Vision Workshops (ICCV workshops)}.\hskip 1em plus 0.5em minus 0.4em\relax
  IEEE, 2011, pp. 601--608.

\bibitem{zhao2015deriving}
X.~Zhao, T.~Jin, and S.~Qu, ``Deriving inherent optical properties from
  background color and underwater image enhancement,'' \emph{Ocean
  Engineering}, vol.~94, pp. 163--172, 2015.

\bibitem{mobley1994light}
C.~D. Mobley and C.~D. Mobley, \emph{Light and water: radiative transfer in
  natural waters}.\hskip 1em plus 0.5em minus 0.4em\relax Academic Press, 1994.

\bibitem{paszke2017automatic}
A.~Paszke, S.~Gross, S.~Chintala, G.~Chanan, E.~Yang, Z.~DeVito, Z.~Lin,
  A.~Desmaison, L.~Antiga, and A.~Lerer, ``Automatic differentiation in
  pytorch,'' 2017.

\bibitem{zhang2022underwaterACDC}
W.~Zhang, Y.~Wang, and C.~Li, ``Underwater image enhancement by attenuated
  color channel correction and detail preserved contrast enhancement,''
  \emph{IEEE Journal of Oceanic Engineering}, 2022.

\bibitem{zhang2022underwater}
W.~Zhang, P.~Zhuang, H.-H. Sun, G.~Li, S.~Kwong, and C.~Li, ``Underwater image
  enhancement via minimal color loss and locally adaptive contrast
  enhancement,'' \emph{IEEE Transactions on Image Processing}, vol.~31, pp.
  3997--4010, 2022.

\bibitem{li2019fusion}
H.~Li, J.~Li, and W.~Wang, ``A fusion adversarial underwater image enhancement
  network with a public test dataset,'' \emph{arXiv preprint arXiv:1906.06819},
  2019.

\bibitem{duarte2016dataset}
A.~Duarte, F.~Codevilla, J.~D.~O. Gaya, and S.~S. Botelho, ``A dataset to
  evaluate underwater image restoration methods,'' in \emph{OCEANS
  2016-Shanghai}.\hskip 1em plus 0.5em minus 0.4em\relax IEEE, 2016, pp. 1--6.

\bibitem{yang2015underwater}
M.~Yang and A.~Sowmya, ``An underwater color image quality evaluation metric,''
  \emph{IEEE Transactions on Image Processing}, vol.~24, no.~12, pp.
  6062--6071, 2015.

\bibitem{panetta2015human}
K.~Panetta, C.~Gao, and S.~Agaian, ``Human-visual-system-inspired underwater
  image quality measures,'' \emph{IEEE Journal of Oceanic Engineering},
  vol.~41, no.~3, pp. 541--551, 2015.

\bibitem{galdran2015automatic}
A.~Galdran, D.~Pardo, A.~Pic{\'o}n, and A.~Alvarez-Gila, ``Automatic
  red-channel underwater image restoration,'' \emph{Journal of Visual
  Communication and Image Representation}, vol.~26, pp. 132--145, 2015.

\bibitem{ma2022toward}
L.~Ma, T.~Ma, R.~Liu, X.~Fan, and Z.~Luo, ``Toward fast, flexible, and robust
  low-light image enhancement,'' in \emph{Proceedings of the IEEE/CVF
  Conference on Computer Vision and Pattern Recognition}, 2022, pp. 5637--5646.

\bibitem{wang2022low}
M.~Wang, Y.~Huang, J.~Xiong, and W.~Xie, ``Low-light images in-the-wild: A
  novel visibility perception-guided blind quality indicator,'' \emph{IEEE
  Transactions on Industrial Informatics}, 2022.

\bibitem{yang2022self}
Y.~Yang, C.~Wang, R.~Liu, L.~Zhang, X.~Guo, and D.~Tao, ``Self-augmented
  unpaired image dehazing via density and depth decomposition,'' in
  \emph{Proceedings of the IEEE/CVF Conference on Computer Vision and Pattern
  Recognition}, 2022, pp. 2037--2046.

\bibitem{liu2022deep}
R.~W. Liu, Y.~Guo, Y.~Lu, K.~T. Chui, and B.~B. Gupta, ``Deep network-enabled
  haze visibility enhancement for visual iot-driven intelligent transportation
  systems,'' \emph{IEEE Transactions on Industrial Informatics}, vol.~19,
  no.~2, pp. 1581--1591, 2022.

\end{thebibliography}
\end{document}